\pdfoutput=1
\documentclass[11pt]{article}

\usepackage[final]{acl}

\usepackage{times}
\usepackage{latexsym}
\usepackage[T1]{fontenc}
\usepackage[utf8]{inputenc}
\usepackage{amsmath} %
\usepackage{amssymb} %
\usepackage{microtype}
\usepackage{inconsolata}
\usepackage{graphicx}
\usepackage{multirow}
\usepackage{booktabs}
\usepackage{array}
\usepackage{longtable} %
\usepackage{xcolor}
\usepackage{url}
\usepackage{tcolorbox}
\usepackage{makecell}
\usepackage{pgfplots}
\usepackage{needspace}
\usepgfplotslibrary{groupplots}
\pgfplotsset{compat=1.18}

\tcbuselibrary{skins,breakable}
\definecolor{promptframe}{HTML}{5A2A6A}
\definecolor{promptback}{HTML}{FBFAF7}
\definecolor{promptph}{HTML}{8C2A2A}
\newcommand{\ph}[1]{\textcolor{promptph}{\{#1\}}}
\newcommand{\promptmono}{\fontfamily{lmtt}\selectfont}
\newcommand{\promptdivider}{}
\newcommand{\promptlang}[1]{{\sffamily\bfseries\footnotesize%
  \textcolor{promptframe}{\MakeUppercase{#1}}}}
\tcbset{promptbox/.style={
  enhanced, breakable, arc=2pt,
  colback=promptback, colframe=promptframe, boxrule=0.6pt,
  left=7pt, right=7pt, top=5pt, bottom=6pt,
  fonttitle=\sffamily\bfseries\small, coltitle=white,
  colbacktitle=promptframe}}

\definecolor{newgold}{HTML}{C79214}
\newcommand{\newdsmark}{{\color{newgold}\ensuremath{\bigstar}}}
\newcommand{\newds}{\nobreak\hspace{0.2em}\newdsmark}

\definecolor{tickgreen}{HTML}{1F8A3A}
\newcommand{\yes}{\textcolor{tickgreen}{\checkmark}}

\title{sk-bench: A Native-First Benchmark for Evaluating\\ Large Language Models in Slovak}

\author{\bf Marek \v{S}uppa$^{\alpha,~\beta}$\thanks{\,Correspondence: \href{mailto:marek@suppa.sk}{marek@suppa.sk}} \quad Ivan Vykopal$^{\delta,~\varepsilon}$ \quad Andrej Ridzik$^{\delta}$ \quad Kristi\'{a}n Sopkovi\v{c}$^{\gamma}$\\
\bf Nat\'{a}lia K\v{n}a\v{z}ekov\'{a}$^{\alpha,~\delta}$ \quad Jaroslav Kop\v{c}an$^{\delta}$ \quad Miroslav Bl\v{s}t\'{a}k$^{\delta}$ \quad Vikt\'{o}ria Ondrejov\'{a}$^{\beta}$\\
\bf Daniel Hl\'{a}dek$^{\gamma}$ \quad Michal Gregor$^{\delta}$ \quad Martin Tamajka$^{\delta}$ \quad Mari\'{a}n \v{S}imko$^{\delta}$\\
$^{\alpha}$Comenius University in Bratislava, Slovakia,
$^{\beta}$Cisco Systems, \\
$^{\gamma}$Technical University of Ko\v{s}ice, Slovakia,
$^{\varepsilon}$Brno University of Technology, Czechia \\
$^{\delta}$Kempelen Institute of Intelligent Technologies, Bratislava, Slovakia \\[1pt]
{\small \href{https://github.com/slovak-nlp/sk-bench}{\raisebox{-0.2em}{\includegraphics[height=1em]{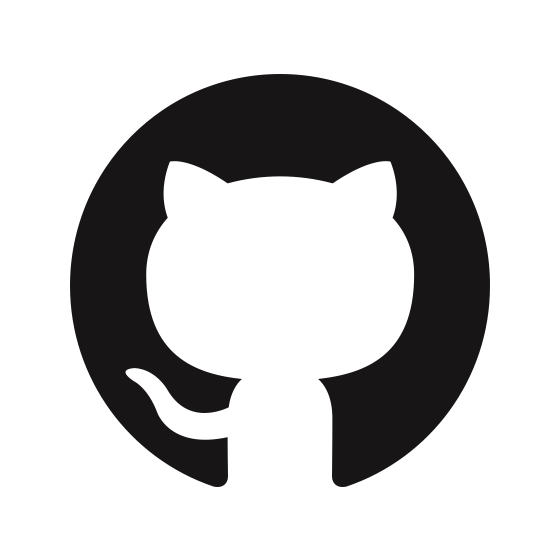}}~\texttt{github.com/slovak-nlp/sk-bench}}\hspace{2em}\href{https://huggingface.co/collections/slovak-nlp/sk-bench}{\raisebox{-0.2em}{\includegraphics[height=1em]{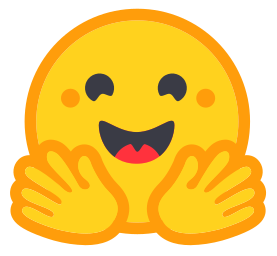}}~\texttt{huggingface.co/collections/slovak-nlp/sk-bench}}}\\[15pt]
}

\begin{document}
\maketitle

\begin{abstract}
Multilingual LLM benchmarks omit Slovak, a morphologically rich West
Slavic language of five million speakers, or cover it only by machine
translation. We present sk-bench, a native-first Slovak benchmark with 30
datasets (33 scored task variants) across ten skill categories. Eleven
resources are introduced or first packaged for generative-LLM evaluation,
including IFEval-SK with Slovak-adapted instruction checkers and native
Chiby/SKJ1 resources for Slovak grammar and morphology. We
evaluate 55 open- and closed-weights models under one harness. The best open model trails
proprietary APIs by 12.6 points. Model rankings are similar for native
and translated closed-form data ($\rho\geq0.98$), though translation
separates the strongest models less well. By contrast,
human-authored and LLM-generated QA questions rank models differently
($\rho=0.72$). For Qwen3-14B,
continued Slovak pretraining lowers the overall score by
13.9 points. A small instruction set restores three quarters of that
loss. Test-time reasoning improves scores by 8.5 to 12.5 points for
models of 9B and above. Together, these findings suggest four design
lessons for other under-resourced languages: use native data where translation fails, plan instruction
repair after language adaptation, enable test-time reasoning before
scaling up, and avoid overinvesting in target-language prompts.
We release the data and code at
\url{github.com/slovak-nlp/sk-bench}.
\end{abstract}

\section{Introduction}
\label{sec:intro}

\begin{figure*}[t]
  \centering
  \includegraphics[width=\linewidth]{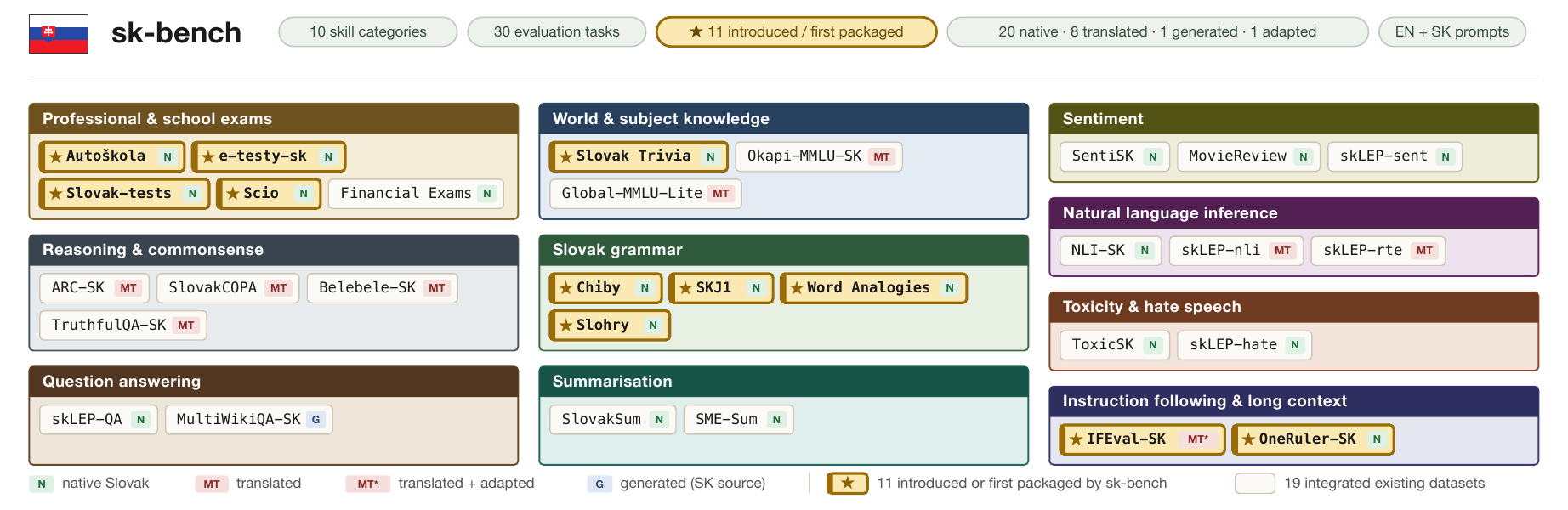}
  \caption{Overview of the sk-bench design: ten skill categories, 30
  evaluation tasks. Chips with a gold fill, heavy gold border and a
  leading star ($\star$) are the 11 resources introduced or first
  packaged for generative-LLM evaluation by sk-bench, listed first
  within each panel. Plain chips with a hairline border are the 19
  pre-existing datasets we integrate into the harness. Novelty (the
  chip itself) and provenance (the N~/~MT~/~MT$^{*}$~/~G badge at the
  right: native Slovak, translated from English, translated then
  adapted with a custom Slovak registry, LLM-generated over a Slovak
  source) are independent of each other---MultiWikiQA-SK is generated but
  pre-existing, Autoškola is native and new. Tasks with bilingual
templates are run with both English and Slovak prompts. IFEval-SK and
Slohry are Slovak-only. These ten skill categories
organise the datasets. The results tables instead use six reporting
groups defined by answer format (Table~\ref{tab:results-main},
\S\ref{sec:results}).}
  \label{fig:overview}
\end{figure*}

Slovak, a fusional, morphologically rich West Slavic language spoken
by roughly five million people, is not short of NLP resources: the
contemporary Slovak National Corpus passes one billion tokens
\cite{snk2025}, a Slovak-centric encoder \citep{slovakbert2022} has
existed since 2021, and the skLEP benchmark \citep{sklep2025} covers
nine encoder-style NLU tasks (POS, two NER datasets, NLI, RTE, QA,
sentiment, hate-speech, and STS). Yet Slovak remains under-served in two respects: (a)~\emph{generative-LLM evaluation
under a unified harness}: an LLM practitioner picking between, say,
a 70\,B-class open model and a closed API for a Slovak deployment
cannot get zero-shot, identically prompted scores for both across the skills LLMs are used for; and
(b)~\emph{Slovak coverage in the multilingual benchmarks that shape
model development}: XNLI \citep{xnli2018} and XCOPA \citep{xcopa2020} omit Slovak, and
HELM \citep{helm2023} and BIG-bench \citep{bigbench2023} provide
minimal Slovak-specific signal. Multilingual suites that do include
Slovak rely mostly on machine translation
\citep{okapi2023, globalmmlu2024}, which inherits source-language bias
and culturally Western-centric content \citep{globalmmlu2024} and
leaves exploitable translation artefacts in the data
\citep{artetxe2020translation}. We
address (a) and (b) by combining native Slovak data with selected
translated proxies under a single generative-LLM harness.

\paragraph{Research questions.} We use sk-bench to address three
empirical questions about Slovak LLMs (\S\ref{sec:results}). Their
answers inform the cross-language lessons in \S\ref{sec:conclusion}
and Appendix~\ref{app:lessons}.

\textbf{RQ1 (data provenance).} For the \emph{same skill}, how much do
native, machine-translated, and LLM-generated evaluation data disagree
about which LLM is best in Slovak? We separate translation from question generation (\S\ref{sec:rq1}).

\textbf{RQ2 (language adaptation).} What does adapting an
instruction-tuned model to Slovak \emph{cost}, and can the loss be
\emph{repaired}? We compare a Slovak continued-pretraining checkpoint
with its base and add a Slovak instruction-tuning step across three
bases (\S\ref{sec:rq2}).

\begin{samepage}
\textbf{RQ3 (test-time reasoning).} Does test-time reasoning transfer
to Slovak, and which skills does it move? We evaluate nine models in
both standard and reasoning modes (\S\ref{sec:rq3}).

\end{samepage}

We make four contributions:
(1)~\emph{sk-bench}, a \emph{native-first} benchmark for
generative-LLM evaluation in Slovak (\S\ref{sec:task-selection}): 30
tasks across ten skill categories, run under one harness against 55
contemporary open- and closed-weights models, with a unified Slovak
leaderboard (\S\ref{sec:results}).
(2)~\emph{Eleven new Slovak evaluation resources}: These did not exist as LLM evaluations before sk-bench: five we built from scratch (authored, annotated, or constructed by us), four we curated from public institutional sources whose items existed but where no dataset did, and two are existing expert resources first packaged for generative-LLM evaluation. All
eleven are marked with a star \mbox{(\newdsmark{})} in Figure~\ref{fig:overview},
which separates them from the nineteen datasets we integrate, and
itemised by tier in Table~\ref{tab:dataset-catalogue}. Two could not be built by translation: IFEval-SK, whose
Slovak instruction-checker registry we re-implemented from scratch
(Appendix~\ref{app:ifevalsk}), and the paired Chiby/SKJ1
grammaticality probe~\citep{skj1}. The native exam sets (driving law,
admission tests, school subjects) are topically selective, but their
professionally validated gold labels make them the native counterpart
to translated MMLU in RQ1 (\S\ref{sec:rq1}).
(3)~\emph{A reproducible harness and open release.} The harness
(\S\ref{sec:harness}) scores open-weights and closed (API) models
through one generate-and-match pipeline. We release the prompts, run
scripts, and per-task breakdowns for reproducible
follow-up work (\S\ref{sec:reproducibility}).
(4)~\emph{Four transferable design lessons} (L1--L4) for
under-resourced-language benchmarks, each backed by a Slovak result:
translation preserves closed-form rankings but blurs the frontier and
diverges on QA, adaptation
costs Qwen3-14B 13.9 points that instruction tuning largely repairs, and
reasoning adds 8.5--12.5 points at $\geq$9B (\S\ref{sec:conclusion},
Appendix~\ref{app:lessons}).

\section{Related Work}
\label{sec:related}

\paragraph{Slovak NLP benchmarks.}
The skLEP benchmark~\citep{sklep2025} bundled nine NLU tasks (POS,
two NER datasets, NLI, RTE, QA, sentiment, hate-speech, STS) into
a single GLUE-style suite and is the closest precedent to sk-bench.
We reuse several skLEP subtasks under the LLM harness and extend its encoder-style scope to multi-choice reasoning, summarisation, instruction-following, and Slovak professional and school exams.
SK-QuAD~\citep{skquad2023} provides
$\sim$91\,k manually annotated extractive QA pairs.
WikiGoldSK~\citep{wikigoldsk2023} contributes a gold-standard NER
set. Slovak summarisation has been studied on SME-Sum
\citep{smesum2020} and SlovakSum~\citep{slovaksum2024}. Sentiment
classification has a longer history~\citep{pecar-simko-2019-sentiment},
and the SentiSK and ToxicSK datasets
\citep{toxic_sk} underpin the corresponding tasks in our suite.
The recent Trojsten benchmark~\citep{trojsten2025} evaluates LLMs on
Slovak high-school STEM competition problems but is single-domain.
SkMTEB~\citep{skmteb2026}, a Slovak text-embedding benchmark rather
than a generative evaluation, introduced the natively annotated
NLI-SK-annotated dataset that we reuse in our suite.
Table~\ref{tab:benchmark-comparison} (App.~\ref{app:lessons}) summarises how sk-bench differs from skLEP and the Slavic, Central-European, and multilingual benchmarks discussed below.

\paragraph{Slavic and Central European benchmarks.}
The closest analogues are BenCzechMark~\citep{benczechmark2025} for
Czech; KLEJ~\citep{klej2020} and LEPISZCZE~\citep{lepiszcze2022} for
Polish; RussianSuperGLUE~\citep{rsglue2020} and the LLM-oriented
MERA~\citep{mera2024} for Russian; HuLU~\citep{hulu2024} for
Hungarian; and SI-NLI~\citep{sinli2024} together with the SloBench
platform~\citep{slobench} for Slovene. sk-bench borrows the
multi-skill, multi-metric design of BenCzechMark and the
GLUE-lineage benchmark design of KLEJ/HuLU but specialises the
tasks, prompts, and metrics for Slovak.

\paragraph{Single-language native benchmarks.}
sk-bench belongs to a family of single-language, natively-sourced LLM
benchmarks that pair a curated task suite with a unified harness:
C-Eval for Chinese~\citep{ceval2023}, IndoMMLU for
Indonesian~\citep{indommlu2023}, ArabicMMLU for
Arabic~\citep{arabicmmlu2024}, and BenCzechMark
\citep{benczechmark2025} for the fellow West Slavic Czech. What sk-bench adds is the \emph{paired provenance contrast} (\S\ref{sec:design-principles}): measuring a skill from natively authored, machine-translated, and LLM-generated data lets us test whether a translated proxy ranks models reliably for that skill (\S\ref{sec:rq1}). No native-first generative-LLM suite of this
breadth previously existed for Slovak.

\paragraph{Multilingual benchmarks containing Slovak.}
Slovak is one of the 122 variants in Belebele~\citep{belebele2024},
one of the 26 languages in Okapi's translated MMLU/ARC/HellaSwag
\citep{okapi2023}, and one of 300+ in
MultiWikiQA~\citep{multiwikiqa2025}. INCLUDE~\citep{include2024}
is a \emph{non-translated} multilingual benchmark; sk-bench shares its preference for native data. Global MMLU~\citep{globalmmlu2024} documents cultural bias in purely translated MMLU.
MEGAVERSE~\citep{megaverse2024} (22 datasets, 83 languages) is a
precedent for our open/closed side-by-side comparison.
MultiBLiMP~\citep{multiblimp2026} covers Slovak grammaticality
natively rather than by translation, but is restricted to
subject--verb agreement minimal pairs generated automatically from
Universal Dependencies and UniMorph. Chiby and SKJ1 instead pair
naturalistic Slovak error data with expert-authored items across nine
error types. Closest to our RQ1 in design is
MultiLoKo~\citep{multiloko2025}, which contrasts locally-sourced with
translated data across 31 languages. sk-bench makes that contrast per skill within one language.

\paragraph{Evaluation methodology.}
sk-bench builds on LightEval~\citep{lighteval} and the
evaluation-harness lineage of HELM~\citep{helm2023}, the EleutherAI
\texttt{lm-evaluation-harness}~\citep{lmevalharness}, the Open LLM
Leaderboard~\citep{openllmleaderboard}, GLUE/SuperGLUE
\citep{glue2018, superglue2019}, BIG-bench~\citep{bigbench2023}, and
XTREME~\citep{xtreme2020}. For instruction-following, we adapt
IFEval~\citep{ifeval2023} to Slovak via the multilingual M-IFEval
\citep{mifeval2025}.

\section{The sk-bench Task Suite}
\label{sec:tasks}

Figure~\ref{fig:overview} summarises the suite. The
full per-dataset catalogue (Table~\ref{tab:dataset-catalogue}) appears in
Appendix~\ref{app:dataset-catalogue}. The suite combines three kinds of data:
native Slovak data that tests language-specific competence,
Slovak domain data whose labels come from professional or educational
settings, and translated or generated proxies that cover skills for
which no native Slovak benchmark exists. We keep these provenance types explicit because RQ1 (\S\ref{sec:rq1}) tests whether a model's Slovak ranking is stable across them.

\subsection{Design Principles}
\label{sec:design-principles}

sk-bench is organised around a
\emph{paired provenance contrast per skill}: wherever possible the same
skill is measured from two differently-sourced datasets, so that
provenance can be varied while the skill is held fixed and a model's
Slovak ranking can be compared across the two
(Table~\ref{tab:paired-contrasts}, Appendix~\ref{app:tasks-detail}). The new native datasets
fill the native side where none existed, which enables the analysis in \S\ref{sec:rq1}. \S\ref{sec:limitations} lists what the suite does not cover (dialogue, agentic tool use, code, multimodality, and dialects).

\subsection{Inclusion Criteria}
\label{sec:task-selection}

We selected datasets according to four criteria, in priority order:
(i)~\emph{native Slovak data preferred over translation},
(ii)~\emph{publicly available redistribution-friendly licence},
(iii)~\emph{usable test split of at least 100 examples},
(iv)~\emph{automatic, reproducible scoring}. We call the result \emph{native-first}: we prioritise native Slovak evaluation data and admit translated datasets only when no native equivalent exists for
the skill of interest (e.g.\ MMLU-style multi-choice reasoning). Datasets failing (iii)--(iv) are noted but excluded from the main results.

\subsection{The Skill Categories}
The full per-dataset catalogue (Table~\ref{tab:dataset-catalogue},
Appendix~\ref{app:dataset-catalogue}) lists every task with its size,
provenance, skill, and licence. Per-task descriptions and metrics are
in Appendix~\ref{app:tasks-detail}. We group the ten skill categories
by provenance.

\paragraph{Native Slovak data.}
Most of the suite is native. Classification uses SentiSK and
MovieReview (sentiment), ToxicSK and skLEP-hate (toxicity and hate
speech), and the native NLI-SK-annotated set. Question answering uses
the human-authored skLEP-QA task from skLEP~\citep{sklep2025}, built on
SK-QuAD~\citep{skquad2023}. Professional and school knowledge comes from
five native exam sets (Financial Exams, Autoškola, e-testy-sk,
Slovak-tests, and Scio admission tests) plus the Slovak Trivia set. Grammar and morphology, which translation cannot produce,
are tested by Chiby and SKJ1~\citep{skj1} (grammaticality), 
Slohry (inflectional morphology), 
and Word Analogies derived from the Slovak Conceptual Dictionary \citep{blstak2025slovakconceptualdictionary}. 
Summarisation uses SlovakSum and SME-Sum, each down-sampled to 1{,}500
(Appendix~\ref{app:downsampling}) and scored with ROUGE, METEOR, and
chrF/chrF++ against single human references. IFEval-SK measures
verifiable instruction following: we translated the M-IFEval
\citep{mifeval2025} templates
to Slovak and re-implemented the programmatic checkers against Slovak
word boundaries, casing, and punctuation. OneRuler-SK probes
long-context retrieval over Slovak text.

\paragraph{Translated and generated proxies.}
Where no native resource of comparable breadth exists, we admit
translated proxies: Okapi-MMLU-SK and Global-MMLU-Lite (subject
knowledge), and ARC-SK, SlovakCOPA, Belebele-SK, and TruthfulQA-SK
(reasoning and commonsense). For QA we add
MultiWikiQA-SK~\citep{multiwikiqa2025}, an existing published dataset
whose questions were generated by Gemini~1.5~Pro from the November~2023
Slovak Wikipedia and then rephrased by the same model in a second pass
to blunt surface word-matching. It is released under CC~BY-NC-SA~4.0.
Because its generator belongs to a proprietary model family that we also evaluate, we
check for generator self-preference in
Appendix~\ref{app:self-preference} and find none. These form the translated and generated sides of the RQ1 contrasts (\S\ref{sec:rq1}).

\section{Evaluation Harness}
\label{sec:harness}

sk-bench wraps LightEval's \texttt{Pipeline}~\citep{lighteval} so the
same prompt, few-shot policy, and scoring logic apply to every model:
open weights are served with vLLM~\citep{vllm2023} and closed models
queried through LiteLLM~\citep{litellm}. Every task is scored by
\emph{text generation followed by string matching} against the gold
answer (or the allowed labels): we generate up to a per-task token
budget (32 tokens for classification and multi-choice, 512 for QA and
summarisation, 4096 for IFEval), normalise, and compute the metric.
Open and closed models are therefore scored identically. The only remaining difference is decoding compliance (some closed APIs ignore
\texttt{seed} or apply post-training that \texttt{temperature=0} does
not override). We hand-write Slovak instruction templates per task
category and parallel English templates as a robustness check.

\paragraph{Interpreting free-form output.}
Because every task is scored from generated text, variation in output
format and length is a source of measurement error. A fixed-order
parser strips chat-template artefacts
before matching, per-task token budgets are identical across providers,
and refusals and unresolvable outputs are scored incorrect rather than
dropped. This rarely matters for standard models (a multiple-choice answer is a median of two tokens), only for small models in reasoning mode, which exhaust the budget inside the trace before
emitting a parseable answer (\S\ref{sec:rq3}).
Appendices~\ref{app:scoring} and~\ref{app:prompts} give the parsing
rules, the language-mismatch policy, per-task output-length statistics,
and the templates.

\paragraph{Metrics and aggregation.}
We use accuracy for classification and multi-choice tasks;
multilingual quasi-exact-match (QEM) and quasi-F1 (QF1)
\citep{lighteval} for QA; ROUGE~\citep{rouge2004},
chrF/chrF++~\citep{chrf2015, chrfpp2017}, and
METEOR~\citep{banerjee2005meteor} for summarisation; and the
prompt- and instruction-level strict/loose accuracies of
\citet{ifeval2023} for IFEval-SK. The \emph{main sk-bench score}
is the unweighted mean of the reporting-group averages, in the
spirit of GLUE/SuperGLUE \citep{glue2018, superglue2019}.
\emph{Within-group} aggregation is by task macro-average, not by
example count, to prevent the 13{,}062-item Okapi-MMLU-SK from
dominating its group against the 500-item SlovakCOPA.

\begin{table*}[ht]
\centering\small
\setlength{\tabcolsep}{5pt}
\resizebox{\linewidth}{!}{%
\begin{tabular}{lrrrrrr|r}
\toprule
\textbf{Model} & \makecell{\textbf{Knowledge \&}\\\textbf{reasoning}} & \makecell{\textbf{Reading \&}\\\textbf{QA}} & \textbf{Classification} & \textbf{Generation} & \makecell{\textbf{Instruction}\\\textbf{following}} & \makecell{\textbf{Grammar \&}\\\textbf{morphology}} & \textbf{Overall} \\
\midrule
\multicolumn{8}{l}{\textit{$<$2B}} \\
Granite 4.0-H 1B & 39.8\,{\tiny\color{gray}$\pm$0.6} & 28.2\,{\tiny\color{gray}$\pm$0.6} & 50.1\,{\tiny\color{gray}$\pm$7.8} & 14.8\,{\tiny\color{gray}$\pm$0.9} & 14.9 & 6.0\,{\tiny\color{gray}$\pm$0.8} & 25.6\,{\tiny\color{gray}$\pm$1.8} \\
Qwen3 1.7B & 35.0\,{\tiny\color{gray}$\pm$5.5} & 30.9\,{\tiny\color{gray}$\pm$2.8} & 52.4\,{\tiny\color{gray}$\pm$7.8} & 12.8\,{\tiny\color{gray}$\pm$0.9} & 23.5 & 12.3\,{\tiny\color{gray}$\pm$1.9} & 27.8\,{\tiny\color{gray}$\pm$2.4} \\
\midrule
\multicolumn{8}{l}{\textit{2--9B}} \\
Ministral 3 8B & 60.4\,{\tiny\color{gray}$\pm$1.3} & 39.2\,{\tiny\color{gray}$\pm$6.2} & 70.6\,{\tiny\color{gray}$\pm$4.4} & 13.5\,{\tiny\color{gray}$\pm$0.8} & 24.0 & 37.1\,{\tiny\color{gray}$\pm$1.0} & 40.8\,{\tiny\color{gray}$\pm$2.3} \\
Ministral 3 8B (R) & 54.9\,{\tiny\color{gray}$\pm$1.9} & 37.8\,{\tiny\color{gray}$\pm$4.0} & 61.9\,{\tiny\color{gray}$\pm$7.2} & 13.1\,{\tiny\color{gray}$\pm$1.1} & 39.9 & 32.1\,{\tiny\color{gray}$\pm$2.8} & 39.9\,{\tiny\color{gray}$\pm$2.4} \\
\midrule
\multicolumn{8}{l}{\textit{9--32B}} \\
Qwen3.5 27B & 81.2\,{\tiny\color{gray}$\pm$0.2} & 55.7\,{\tiny\color{gray}$\pm$6.3} & 79.5\,{\tiny\color{gray}$\pm$3.4} & 14.9\,{\tiny\color{gray}$\pm$0.4} & 36.1 & 64.6\,{\tiny\color{gray}$\pm$0.8} & 55.3\,{\tiny\color{gray}$\pm$1.8} \\
Gemma 4 31B & 78.8\,{\tiny\color{gray}$\pm$0.7} & 53.1\,{\tiny\color{gray}$\pm$8.7} & 82.4\,{\tiny\color{gray}$\pm$0.7} & 14.4\,{\tiny\color{gray}$\pm$0.5} & 30.2 & 66.7\,{\tiny\color{gray}$\pm$2.1} & 54.3\,{\tiny\color{gray}$\pm$1.3} \\
\midrule
\multicolumn{8}{l}{\textit{32--200B}} \\
Llama 3.3 70B & 72.1\,{\tiny\color{gray}$\pm$0.5} & 53.9\,{\tiny\color{gray}$\pm$4.0} & 79.4\,{\tiny\color{gray}$\pm$0.2} & \textbf{21.4}\,{\tiny\color{gray}$\pm$0.5} & 35.4 & 56.4\,{\tiny\color{gray}$\pm$1.0} & 53.1\,{\tiny\color{gray}$\pm$1.0} \\
gpt-oss 120B & 73.7\,{\tiny\color{gray}$\pm$1.0} & 53.2\,{\tiny\color{gray}$\pm$6.3} & 79.0\,{\tiny\color{gray}$\pm$0.5} & 13.5\,{\tiny\color{gray}$\pm$0.4} & 74.9 & 65.3\,{\tiny\color{gray}$\pm$1.0} & 60.0\,{\tiny\color{gray}$\pm$1.1} \\
\midrule
\multicolumn{8}{l}{\textit{$>$200B}} \\
Qwen3.5 397B-A17B & 83.4\,{\tiny\color{gray}$\pm$0.5} & 55.5\,{\tiny\color{gray}$\pm$8.2} & 80.6\,{\tiny\color{gray}$\pm$0.8} & 14.6\,{\tiny\color{gray}$\pm$0.2} & 37.8 & 72.3\,{\tiny\color{gray}$\pm$3.0} & 57.4\,{\tiny\color{gray}$\pm$1.7} \\
GLM-5.1 & 80.6\,{\tiny\color{gray}$\pm$0.4} & 61.2\,{\tiny\color{gray}$\pm$6.9} & 82.6\,{\tiny\color{gray}$\pm$1.0} & 14.2\,{\tiny\color{gray}$\pm$0.2} & 35.9 & 70.8\,{\tiny\color{gray}$\pm$1.3} & 57.6\,{\tiny\color{gray}$\pm$1.2} \\
\midrule
\multicolumn{8}{l}{\textit{Proprietary (size undisclosed)}} \\
Gemini 3.1 Pro & \textbf{92.4}\,{\tiny\color{gray}$\pm$0.2} & \textbf{71.3}\,{\tiny\color{gray}$\pm$4.5} & \underline{82.6}\,{\tiny\color{gray}$\pm$0.4} & 14.8\,{\tiny\color{gray}$\pm$0.0} & \textbf{87.2} & \textbf{87.6}\,{\tiny\color{gray}$\pm$0.4} & \textbf{72.6}\,{\tiny\color{gray}$\pm$0.9} \\
GPT-5.5 & \underline{89.2}\,{\tiny\color{gray}$\pm$0.3} & \underline{67.3}\,{\tiny\color{gray}$\pm$6.8} & 81.6\,{\tiny\color{gray}$\pm$0.3} & 15.5\,{\tiny\color{gray}$\pm$0.3} & \underline{86.4} & \underline{86.1}\,{\tiny\color{gray}$\pm$0.3} & \underline{71.0}\,{\tiny\color{gray}$\pm$1.2} \\
Claude Opus 4.6 & 87.2\,{\tiny\color{gray}$\pm$0.5} & 58.6\,{\tiny\color{gray}$\pm$10.2} & \textbf{83.4}\,{\tiny\color{gray}$\pm$0.4} & 15.4\,{\tiny\color{gray}$\pm$0.2} & 80.6 & 83.6\,{\tiny\color{gray}$\pm$0.5} & 68.1\,{\tiny\color{gray}$\pm$2.0} \\
\bottomrule
\end{tabular}
}
\caption{No open-weights model reaches the proprietary APIs on Slovak. Main sk-bench leaderboard (Slovak, 0-shot): top two open-weights models per size tier and top three proprietary APIs. Cells are the unweighted mean of the group's per-task primary metrics ($\times100$). \textbf{Best} and \underline{second-best} are over \emph{all 55} models, so a mark appears only where a shown model holds the record. \textbf{Overall} averages the six groups. Grey $\pm$ is the standard deviation across the three prompt templates. Full leaderboard and per-task scores: Tables~\ref{tab:results-full} and~\ref{tab:skbench-comprehensive}.}
\label{tab:results-main}
\end{table*}

\section{Experimental Setup}
\label{sec:setup}

\subsection{Models Evaluated}
We evaluate 55 models: 49 open-weight (46 base checkpoints plus three
we instruction-tune on Slovak data, below) and 6 proprietary, grouped
into four parameter tiers ($\leq$2\,B, 2--10\,B, 10--70\,B,
$\geq$70\,B) spanning 350\,M to 1.1\,T parameters. Per-tier membership
is in the leaderboard (Table~\ref{tab:results-main}). The full
catalogue with sizes, licences, and revisions is in
Appendix~\ref{app:models}. The open weights cover the IBM Granite,
Qwen3/3.5/3.6, Gemma, Llama, and Mistral/Ministral families; the
European-centric TildeOpen~\citep{tildeopen2025},
EuroLLM~\citep{eurollm2024}, Teuken~\citep{teuken2025}, Bielik, and
Apertus models; the Aya, GPT-OSS, Command-A, GLM, MiniMax, and Kimi
checkpoints; and two Slovak-adapted
bases (below).
The proprietary models are two each from the
GPT-5~\citep{gpt5} (GPT-5.4~Mini, GPT-5.5),
Gemini~3.1~\citep{gemini3} (Pro, Flash~Lite), and
Claude~4.6~\citep{claude4} (Opus, Sonnet).

\paragraph{Slovak instruction tuning.} We fine-tuned the base models \texttt{slovak-nlp/mistral-sk-7b}, \texttt{slovak-nlp/Qwen3-14B-sk}, and \texttt{TildeAI/TildeOpen-30b-64k} (Mistral-SK, Qwen3-SK, and TildeOpen-SK below) on the Slovak split of the TaCo Alpaca data~\citep{taco2024} (machine-translated Stanford Alpaca~\citep{alpaca2023}) to create instruction-tuned counterparts.

\subsection{Inference, Decoding, and Prompt Language}
Open-weights models run on H200 141\,GB and A100 40\,GB nodes (served
as described in \S\ref{sec:harness}). The default
protocol is \emph{zero-shot} with greedy decoding
(\texttt{temperature=0}, fixed seed=1234 where supported).  Per-task token budgets are kept identical
across providers. The one exception is reasoning mode, where the budget is 8192 tokens. The released results use the Slovak instruction templates.
The harness also ships parallel English templates for each task.

\subsection{Few-Shot Policy}
The main results are zero-shot. We additionally ran every task with
a few-shot counterpart at 3-shot (random exemplars from each task's
training or validation split) under both Slovak and English prompts.
The effect is analysed in \S\ref{sec:oneruler}
(Table~\ref{tab:fewshot}).

\subsection{Reproducibility}
\label{sec:reproducibility}
The harness, prompts, model configuration, decoding settings, and
per-example raw model outputs are released with the benchmark.
Software dependencies are pinned in the released harness
(\texttt{pyproject.toml}), and each Hugging Face dataset is loaded at a
fixed revision. Closed-weights API calls record the model snapshot and
the call date. Generation is greedy (\texttt{temperature=0}), so the
main results carry no generation stochasticity for open-weights
models.

\paragraph{Crediting the component datasets.}
sk-bench integrates existing Slovak resources alongside the datasets we
introduce, and aggregation of this kind can redirect citations
from the original authors to the wrapper. To preserve credit for the
original authors, we follow the practice established by Universal
Dependencies, where the collection as a whole is citable but every
component keeps its own authorship and its own citation. The release ships a per-component BibTeX entry for each integrated
dataset, so a user of any subset can cite its original authors. Every dataset card names the original authors first, ahead of
any sk-bench attribution. The release README asks users to cite
\emph{the datasets they actually use}, not only sk-bench. In the paper
itself, every integrated dataset is cited in its task description
(\S\ref{app:tasks-detail}), with its licence and provenance recorded in
the dataset catalogue (Table~\ref{tab:dataset-catalogue}).

\paragraph{Closed-API non-determinism.}
We query closed-API models at \texttt{temperature=0}. We did not run
repeated calls per (model, task). The three Slovak and three English
prompt variants instead give a per-model spread of scores that bounds
prompt-induced variation.

\section{Results and Analysis}
\label{sec:results}

Proprietary APIs lead the Slovak leaderboard and the best open-weights
model trails them by 12.6 points. Among open models, scale is a weak
predictor of Slovak ability.
Table~\ref{tab:results-main} is the main sk-bench leaderboard:
the top two models per size tier (and the top three proprietary
APIs) scored on six \emph{reporting groups} (classification,
knowledge \& reasoning, reading \& QA, generation, grammar \&
morphology, and instruction following), plus an overall macro-average
over the six. The full 55-model
leaderboard and per-task scores are in Appendix~\ref{app:detailed}
(Tables~\ref{tab:results-full} and~\ref{tab:skbench-comprehensive}), with a thinking-mode breakdown
alongside. Reporting groups are defined by answer format and cut across the ten skill categories of Figure~\ref{fig:overview}: grammar
\& morphology comprises the two \emph{production} tasks (Word
Analogies, Slohry), while the Chiby and SKJ1 grammaticality judgements
are selection tasks and are scored under reading \& QA and knowledge \&
reasoning respectively.

\begin{figure}[t]
  \centering
  \includegraphics[width=\linewidth]{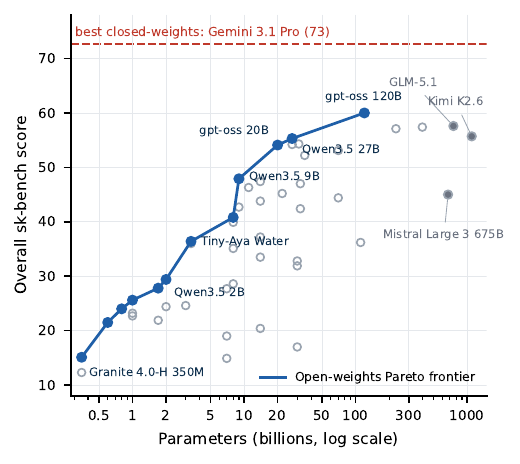}
  \caption{Open-weights Slovak scores plateau past $\sim$20B:
  overall sk-bench score vs.\ parameter count (log scale). The frontier
  (blue) flattens past gpt-oss~20B/Qwen3.5~27B---no larger open model,
  up to the 1.1T Kimi~K2.6, beats gpt-oss~120B (60.0). The dashed line
  marks the best closed model (Gemini~3.1~Pro, 72.6). A fully-labelled
  version is in Appendix~\ref{app:detailed}
  (Figure~\ref{fig:pareto-full}).}
  \label{fig:pareto}
\end{figure}

Gemini~3.1~Pro leads at 72.6 overall, ahead of GPT-5.5 (71.0) and
Claude~Opus~4.6 (68.1). The strongest open-weights model is
gpt-oss~120B at 60.0. Mid-sized open models are nonetheless
competitive: Qwen3.5~27B (55.3), Gemma~4~31B (54.3), and gpt-oss~20B
(54.1) come within a few points of the largest ($>$200B) open models
(45.0--57.6), and the Pareto frontier flattens past gpt-oss~20B with no
larger model beating gpt-oss~120B (Figure~\ref{fig:pareto}). The
open--closed gap
is also uneven across skills: it is widest on instruction following
(best closed 87.2 vs.\ best open 74.9) and smallest, even reversed, on summarisation, but the reversal comes from the metric. Under single-reference ROUGE-L, Llama~3.3~70B (21.4)
tops the group ahead of GPT-5.5 (15.5) and Gemini~3.1~Pro (14.8), yet
Llama falls to 13th under chrF (28.6, behind both at 29.0), and under
METEOR the best summariser is Claude~Opus~4.6. The generation column therefore ranks models under one metric and says little about summarisation quality (\S\ref{sec:limitations}).

\subsection{RQ1: Where Does Data Provenance Reorder Rankings?}
\label{sec:rq1}

\begin{table}[t]
\centering\small
\setlength{\tabcolsep}{3pt}
\resizebox{\columnwidth}{!}{%
\begin{tabular}{l r rrrrr r c}
\toprule
& & \multicolumn{5}{c}{\textbf{Within parameter band}} & & \\
\cmidrule(lr){3-7}
\textbf{Contrast} & \textbf{Overall} & \textbf{$<$2B} & \textbf{2--9B} & \textbf{9--32B} & \textbf{32--200B} & \textbf{$>$200B} & \textbf{Prop.} & \textbf{top-10}\\
\midrule
NLI {\tiny(N\,vs\,MT)} & 0.99 & 0.93 & 0.99 & 0.98 & 0.96$^{\dagger}$ & 0.90$^{\dagger}$ & 0.94$^{\dagger}$ & 10/10\\
MCQ {\tiny(N\,vs\,MT)} & 0.98 & 0.93 & 0.96 & 0.98 & 0.93$^{\dagger}$ & 0.40$^{\dagger}$ & 0.71$^{\dagger}$ & 7/10\\
QA {\tiny(N\,vs\,G)}   & 0.72 & 0.17 & 0.43 & 0.74 & 0.64$^{\dagger}$ & 0.80$^{\dagger}$ & 0.94$^{\dagger}$ & 6/10\\
\bottomrule
\end{tabular}}
\caption{Translation is a safe ranking proxy for closed-form Slovak
skills, but question provenance is not. Spearman $\rho$ between the native
and translated/generated sk-bench leaderboards---overall, within
parameter-size bands, and as top-10 rank overlap---over all 55 models
(zero-shot, Slovak prompts, accuracy for NLI/MCQ, QEM for QA). Overall
95\% bootstrap CIs: NLI~$[0.97,1.00]$, MCQ~$[0.94,0.99]$,
QA~$[0.57,0.83]$. $^{\dagger}$~band has $n\le7$, so its
within-band $\rho$ is high-variance (wide CIs) and read only
qualitatively. N=native, MT=translated, G=LLM-generated.}
\label{tab:provenance-divergence}
\end{table}

Translation preserves the coarse ordering
of models on closed-form tasks, but how the QA \emph{questions} are produced reorders models substantially, so native data pays off on some skills more than others, and a paired contrast shows which. We test this with three skill-level dataset contrasts that compare provenance while holding the broad skill fixed
(Table~\ref{tab:provenance-divergence}): along the \emph{translation} contrast, native NLI-SK-annotated vs.\ translated skLEP-nli, and native
\emph{e-testy-sk} vs.\ translated Okapi-MMLU-SK; and along the
\emph{question-generation} contrast, human-authored skLEP-QA vs.\
LLM-generated MultiWikiQA-SK. For each we report the Spearman rank
correlation between the two leaderboards, overall and \emph{within}
parameter-size bands, plus top-10 rank overlap. The per-model
native-vs-translated scatter is in Appendix~\ref{app:lessons}
(Figure~\ref{fig:provenance-scatter}).
A single correlation over a pool spanning 350\,M to 1.1\,T parameters is
dominated by the between-tier ordering, so the within-band values and the top-10 overlap are more informative.

\paragraph{Translation is a safe ranking proxy for closed-form skills.}
For NLI the native and translated leaderboards are near-identical
(overall $\rho=0.99$, 95\% bootstrap CI $[0.97,1.00]$, \emph{identical}
top-10) and stay highly correlated in every band we can estimate. For
multiple choice, translation is likewise safe overall
($\rho=0.98$ $[0.94,0.99]$), but the two top-10s share only 7 of 10 models, suggesting translated MMLU separates the
\emph{strongest} models less reliably. Our frontier
bands ($n\le6$) are too small to estimate this divergence reliably.

\paragraph{Limits of the dataset-level contrast.}
The two translation-contrast rows pair differently-sourced datasets covering
the same skill rather than the same items in two languages, so they mix
content differences with provenance. We report them at the
dataset level because that is the choice a benchmark builder faces: build a native set or adopt a translated suite such as Okapi-MMLU (if a native set exists, one uses it). An item-matched control would isolate translation
\emph{per se}, and is constructible from data we release, since
Belebele-SK and Global-MMLU-Lite are parallel across
Slovak and English. We leave it to future work.

\paragraph{Question provenance reorders QA rankings.}
Human-authored skLEP-QA and LLM-generated MultiWikiQA-SK agree only at
$\rho=0.72$ $[0.57,0.83]$ (top-10 overlap 6/10), and the disagreement
is largest below the frontier ($\rho=0.17$ and $0.43$ at $<$2B and
2--9B, rising to $0.74$ at 9--32B). For Slovak QA, \emph{how the
questions are produced} moves model rankings substantially more than
translation of multiple-choice items does. Two mechanisms plausibly
separate QA from the closed-form skills. First, the formats differ: NLI is format-robust closed classification over a fixed label
set, whereas open QA is free-text generation scored by
quasi-exact-match and token F1, and is far more exposed to format and
surface variation. Second, MultiWikiQA-SK's questions were deliberately
rephrased to reduce surface word overlap, which fits the divergence being largest below the frontier, where models lean most on word-overlap shortcuts. The
per-task scores for both sides of every contrast are in
Appendix~\ref{app:detailed}
(Table~\ref{tab:skbench-comprehensive}).

\subsection{RQ2: What Does Slovak Adaptation Cost, and Can It Be
Repaired?}
\label{sec:rq2}
By ``cost'' we mean capability lost relative to the base model, not training compute (App.~\ref{app:cost}).

\paragraph{The cost of adaptation.}
Adaptation is expensive and the loss is largely repairable: continued
Slovak pretraining trades broad capability for a narrow grammar gain,
and a small Slovak instruction set then recovers about three quarters
of it. We evaluate one adapted/base pair whose members are both in the sweep,
\texttt{slovak-nlp/Qwen3-14B-sk} (Qwen3-SK) against its \texttt{Qwen/Qwen3-14B} base.
A single pair cannot support a population-level claim. We read it as a
case study isolating the adaptation step from base architecture (though
not from other post-training differences). Per-group and per-task
scores for both models are in Appendix~\ref{app:detailed}. For this pair, Slovak adaptation
\emph{lowers} the overall score from 47.4 to 33.5: it helps only the
grammar group ($+6.8$) while degrading knowledge ($-23.7$), reading
($-25.2$), and classification ($-30.6$). Instruction following falls by only $-6.4$,
far less than knowledge, reading, or classification, so adaptation mainly costs content and comprehension, not instruction following.
Figure~\ref{fig:adaptation-trajectory} (App.~\ref{app:lessons}) plots
the full base-adapted-repaired arc per group.

\paragraph{Repairing it with instruction tuning.}
\label{sec:alpaca}
Three of our open models are Slovak base LMs that we further tuned on a
small Slovak Alpaca-style instruction set, giving a base/tuned pair at
7B, 14B, and 30B (\S\ref{sec:setup}, Table~\ref{tab:alpaca-delta},
App.~\ref{app:lessons}). The
effect of this single tuning step is strongly base-dependent and rises
monotonically with the capability of the base: the weakest base
(Mistral-SK, 19.0 overall) \emph{loses} 4.0
points, the mid base (TildeOpen-SK, 31.9) is essentially flat
($+0.8$), and the strongest (Qwen3-SK, 33.5) gains $+10.3$. The only consistent effect is on instruction following, which the data targets: instruction
following improves for two bases ($+13.2$, $+11.5$) and is unchanged
for the third, and free-form generation moves little. Every other group
swings by up to 30 points in \emph{both} directions depending on the
base (Figure~\ref{fig:adaptation}, App.~\ref{app:lessons}): Qwen3-SK
trades 19.8 grammar points
for large gains in knowledge
($+26.6$), reading, and classification, whereas TildeOpen-SK does the
reverse ($+30.7$ grammar, $-29.2$ classification). A small Alpaca set
thus reliably teaches the instruction format but does not add knowledge
or reasoning, so whether it nets a gain depends on how much capability
the base already has to expose.

For the Qwen lineage, Alpaca instruction-tuning recovers most of the
adaptation loss, back to 43.8 (about three-quarters of the 13.9-point
deficit, though still 3.6 below the original Qwen3-14B).
Instruction tuning therefore mostly re-exposes capability the base already has instead of adding new Slovak capability, which is also why it helps least on the bases with the least capability to recover.

\subsection{RQ3: Does Test-Time Reasoning Transfer to Slovak?}
\label{sec:rq3}
Test-time reasoning gives large, consistent gains in Slovak for models of 9B or more, mostly in instruction following, and it backfires below that size, so its benefit depends on capacity.
Nine models expose an optional reasoning (``thinking'') mode and were
evaluated in both settings under an 8{,}192-token generation cap:
eight Qwen variants from 0.6B to the 35B-A3B mixture-of-experts (MoE),
plus Gemma~4~31B
(Table~\ref{tab:thinking-delta}, App.~\ref{app:lessons}, with full
per-task scores in Appendix~\ref{app:detailed},
Table~\ref{tab:skbench-thinking}).
Every model of 9B or more gains $+8.5$ to $+12.5$ overall, and the gain
is highly \emph{concentrated}: instruction following (IFEval-SK)
improves for \emph{every} model in the set---by $+7$ to $+54$ points,
and by at least $+21$ above 9B, the largest and most consistent effect---while grammar~\&~morphology and knowledge rise at scale and
free-form generation (summarisation ROUGE-L) is flat or slightly
negative everywhere ($\leq +0.2$). The gains therefore carry over to a morphologically rich Slavic language.

Reasoning can also reduce scores. The two smallest Qwen3.5 models (0.8B,
2B) \emph{regress} sharply ($-16.8$ and $-9.7$ overall), while the
older Qwen3~0.6B/1.7B at the same sizes improve, so capacity, not size
alone, decides the outcome. The regression is a format failure: these models exhaust the budget inside the trace without emitting a parseable answer, so reasoning helps only once a model can close its trace within budget
(App.~\ref{app:lessons}).

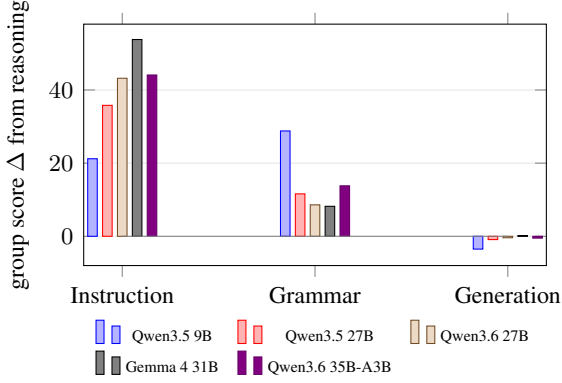
\begin{figure}[t]
\centering
\begin{tikzpicture}
\begin{axis}[
  ybar, width=\columnwidth, height=0.62\columnwidth,
  bar width=3.6pt,
  symbolic x coords={Instruction, Grammar, Generation},
  xtick=data, ymin=-8, ymax=58,
  ylabel={group score $\Delta$ from reasoning}, ylabel style={font=\footnotesize},
  tick label style={font=\footnotesize},
  ymajorgrids, grid style={gray!20},
  legend style={font=\tiny, at={(0.5,-0.20)}, anchor=north,
                legend columns=3, draw=none, fill=none,
                /tikz/every even column/.append style={column sep=5pt}},
]
\addplot coordinates {(Instruction,21.2) (Grammar,28.8) (Generation,-3.5)};
  \addlegendentry{Qwen3.5 9B}
\addplot coordinates {(Instruction,35.8) (Grammar,11.6) (Generation,-0.9)};
  \addlegendentry{Qwen3.5 27B}
\addplot coordinates {(Instruction,43.2) (Grammar,8.6) (Generation,-0.4)};
  \addlegendentry{Qwen3.6 27B}
\addplot coordinates {(Instruction,53.8) (Grammar,8.2) (Generation,0.2)};
  \addlegendentry{Gemma 4 31B}
\addplot coordinates {(Instruction,44.1) (Grammar,13.8) (Generation,-0.5)};
  \addlegendentry{Qwen3.6 35B-A3B}
\draw[gray, thin] (axis cs:Instruction,0) -- (axis cs:Generation,0);
\end{axis}
\end{tikzpicture}
\caption{Test-time reasoning transfers to Slovak, and concentrates on
instruction following. Per-group sk-bench change (Slovak, 0-shot) from
thinking mode for the five $\geq$9B models, each gaining $+8.5$ to
$+12.5$ overall. The effect is flat on free-form generation.}
\label{fig:thinking}
\end{figure}

\subsection{Robustness: Prompt Language, Few-Shot, Context Length}
\label{sec:oneruler}

Three perturbations leave both the overall score and the model ordering
largely intact. The per-group tables and their discussion are in
Appendix~\ref{app:lessons}. \emph{Prompt language (L4):} re-running the
sweep with English rather than Slovak prompts, on every task that has an
English template, moves the overall score by just $+0.4$
(Table~\ref{tab:lang-comparison}). Grammar \& morphology is the only
sizable shift ($-2.6$, where Slovak helps). \emph{Few-shot exemplars:}
3-shot raises the overall score only modestly across the 52 models with
few-shot results (Slovak $+2.0$, English $+1.6$,
Table~\ref{tab:fewshot}), and unevenly---it helps classification and
reading \& QA, where exemplars pin down the answer format, but slightly
\emph{lowers} knowledge and reasoning, echoing the instruction-tuning
result of \S\ref{sec:rq2}: demonstrations teach the answer format but add no knowledge.
\emph{Long context:} on OneRuler-SK's matched Slovak and English
haystacks, Slovak retrieval declines only mildly from 4k to 16k
($58.1\to54.4$) and tracks English throughout, so there is no
Slovak-specific length cliff up to 16k. Scores vary across models far more than across lengths (Table~\ref{tab:oneruler-length}).

\section{Conclusion}
\label{sec:conclusion}

sk-bench is a native-first benchmark for Slovak. It covers ten skills
with 30 datasets, eleven of which are introduced or first packaged here.
Several of them, such as IFEval-SK, Chiby, and SKJ1, could not have been
built by translation. We evaluate 55 open- and closed-weights models
under one harness and release all per-task scores (\S\ref{sec:results},
Appendix~\ref{app:detailed}). The best open model still trails
proprietary APIs by 12.6 points.

Building sk-bench taught us four design lessons for other under-resourced
languages (Appendix~\ref{app:lessons}).

\par\smallskip\needspace{3\baselineskip}\noindent\textbf{L1: Spend native data where translation fails.}
Translation preserves the overall ranking on closed-form skills
($\rho\geq0.98$) but separates the strongest models less well (top-10
overlap 7 of 10 for multiple choice). Native
data matters most for grammar and morphology, which translation cannot
supply, and for QA, where LLM-generated questions rank models
differently from human-authored ones ($\rho=0.72$).

\par\smallskip\needspace{3\baselineskip}\noindent\textbf{L2: Plan an instruction-repair step after language
adaptation.} Continued Slovak pretraining raised Qwen3-14B's grammar
score by 6.8 points but cut its overall score from 47.4 to 33.5, a
familiar case of catastrophic forgetting
\citep{french1999forgetting, luo2023forgetting}. A small instruction
set recovers three quarters of the loss, but only for bases with
capability left to recover.

\par\smallskip\needspace{3\baselineskip}\noindent\textbf{L3: Enable test-time reasoning before scaling up.} It
adds 8.5 to 12.5 points for models of 9B and above, mostly in
instruction following, but backfires on the smallest models (Qwen3.5
0.8B and 2B lose 16.8 and 9.7 points), which exhaust their budget
before answering.

\par\smallskip\needspace{3\baselineskip}\noindent\textbf{L4: Do not over-invest in target-language prompts.}
They barely move scores in our case: switching every template to
English changes the overall score by only $+0.4$ points, and grammar,
the one group that shifts noticeably, by $-2.6$.

\par\smallskip\noindent We therefore recommend that benchmarks for under-resourced
languages include at least one native paired contrast per skill and
re-implement instruction checkers natively instead of translating them.

\section*{Limitations}
\label{sec:limitations}

\paragraph{Coverage.}
sk-bench focuses on textual single-turn evaluation in standard
written Slovak. It does not currently cover dialogue, multi-turn
agentic tool use, multimodality, code generation, or
dialectal/colloquial variants of Slovak.

\paragraph{Translated tasks.}
Several tasks (Okapi-MMLU-SK, ARC-SK, TruthfulQA-SK, Belebele-SK,
SlovakCOPA, IFEval-SK) are translations from English source data.
They inherit cultural and topical bias from English-speaking
contexts~\citep{globalmmlu2024}. We retain them because no native
Slovak counterpart of equivalent breadth exists, but caution against
treating them as measurements of Slovak knowledge per se.

\paragraph{Single-reference summarisation.}
SlovakSum and SME-Sum each provide one human reference per article.
ROUGE and chrF under-penalise paraphrastic-but-correct summaries and
over-penalise stylistic divergence. We treat the resulting numbers
as relative rankings of summarisation quality rather than absolute
measures.

\paragraph{Test-set contamination.}
We have not measured or certified the absence of any sk-bench test
item from the training corpora of the models we evaluate. A new dataset does not imply unseen content, so we distinguish three layers at which leakage can occur.
\emph{Benchmark-format} leakage is exposure to the exact prompt and
gold answer in sk-bench form. This format is new for all eleven resources
introduced or first packaged here. \emph{Answer-key} leakage is exposure to the
question--answer pairs of the underlying item banks. These have been
public for years for the exam sets (Autoškola, Scio, e-testy-sk,
Slovak-tests). \emph{Source-document} leakage is exposure to the raw
text behind the task, which is public in most cases: Chiby is built over Slovak Wikipedia revisions and Word Analogies over the Slovak
Conceptual Dictionary. The tasks least exposed on all three layers
are therefore those whose gold signal we authored ourselves: the
Chiby annotations and the IFEval-SK checkers. Assembling a fresh
benchmark reduces exposure of the exact test split but not exposure to the underlying public material, so scores on the
previously released datasets (NLI-SK-annotated, skLEP, SK-QuAD,
SlovakSum, SME-Sum, ARC, MMLU, TruthfulQA), particularly for closed
models whose training data is undisclosed, should not be read as
contamination-free measurements of Slovak ability.
Appendix~\ref{app:contamination} gives the per-dataset breakdown.

\paragraph{Aggregation.}
The sk-bench score weights all reporting groups equally and uses task
macro-averaging within each group. This prevents large translated tasks (Okapi-MMLU-SK) from dominating their group, but it
remains contestable: an example-weighted score would yield different
rankings. We release per-task and per-group breakdowns
(Appendix~\ref{app:detailed}) so readers can re-aggregate under their
own weighting.

\paragraph{Reproducibility caveats.}
Closed-weights APIs can change without notice, so we cannot
guarantee bit-identical reproducibility of those numbers later.
Open-weights results are reproducible given the released harness, the
released checkpoints, and the same vLLM version.

\section*{Ethical Considerations}

The datasets aggregated by sk-bench were originally released under
their respective licences (catalogued in
Appendix~\ref{app:dataset-catalogue}). We re-distribute only what those
licences allow. Tasks whose licence forbids redistribution are run from
the original source. We use every artefact solely for non-commercial
research evaluation, consistent with the intended use of the benchmark
and evaluation resources it derives from. The sk-bench artefacts we
release (the new datasets, the harness, and the per-task model
outputs) are likewise intended for research evaluation. Components
derived from non-commercial (CC~BY-NC$^{*}$) or research-only sources
inherit those restrictions and must not be used commercially.
Hate-speech and
toxicity tasks contain material that may be offensive. We mark these
tasks as such in the released metadata and recommend appropriate
IRB-style review for any follow-up annotation work. The Slovak
Wikipedia, news, and exam content underlying several tasks is
publicly available. We add no personally-identifying information
beyond what is already public. The user-generated sentiment and
toxicity tasks (SentiSK, ToxicSK) consist of public social-media
comments used as released by their original authors. We perform no re-identification and add no annotations
linking text to individuals.

\paragraph{Use of AI assistants.}
AI assistants were used to help generate the result tables and
figures, to script parts of the evaluation harness and analysis, and,
as the authors are not native English speakers, to improve the
grammar and language of the manuscript. The benchmark design,
experiments, analyses, and conclusions are the authors' own and were
verified by the authors. LLM use within the benchmark itself
(LLM-generated MultiWikiQA-SK and the LLM-based annotation of
Autoškola) is marked by the dataset provenance labels and documented
in the dataset appendices.

\section*{Acknowledgements}

This work was partially funded by the Ministry of Education, Research, Development and Youth of the Slovak Republic under the project KEGA 049TUKE-4/2024; by VEGA 1/0685/26; by the Slovak Research and Development Agency under the project APVV-22-0414; by the European Union under the project lorAI -- Low Resource Artificial Intelligence, GA No.~\href{https://doi.org/10.3030/101136646}{101136646}; and by CLARIN-SK, the Slovak national consortium of the CLARIN ERIC research infrastructure.  It was also partially funded by the EU NextGenerationEU through the Recovery and Resilience Plan for Slovakia under the project No. 09I02-03-V01-00029.

Part of the research results was obtained using the computational resources procured in the national project National competence centre for high performance computing (project code: 311070AKF2), funded by the European Regional Development Fund, EU Structural Funds Informatization of Society, Operational Program Integrated Infrastructure, as well as the computational resources of the supercomputer PERUN, operated by the Supercomputing Centre at the Technical University of Košice (TUKE), Slovakia, with the support of the European Union from the funds of the Recovery and Resilience Plan of the Slovak Republic within the framework of project No.~17I03-04-P03-00001, Development and design of a supercomputer for the National Supercomputing Center.

\bibliography{custom}

\begin{thebibliography}{85}
\providecommand{\natexlab}[1]{#1}

\bibitem[{Ahuja et~al.(2024)Ahuja, Aggarwal, Gumma, Watts, Sathe, Ochieng,
  Hada, Jain, Ahmed, Bali, and Sitaram}]{megaverse2024}
Sanchit Ahuja, Divyanshu Aggarwal, Varun Gumma, Ishaan Watts, Ashutosh Sathe,
  Millicent Ochieng, Rishav Hada, Prachi Jain, Mohamed Ahmed, Kalika Bali, and
  Sunayana Sitaram. 2024.
\newblock \href {https://doi.org/10.18653/V1/2024.NAACL-LONG.143} {{MEGAVERSE:}
  benchmarking large language models across languages, modalities, models and
  tasks}.
\newblock In \emph{Proceedings of the 2024 Conference of the North American
  Chapter of the Association for Computational Linguistics: Human Language
  Technologies (Volume 1: Long Papers), {NAACL} 2024, Mexico City, Mexico, June
  16-21, 2024}, pages 2598--2637. Association for Computational Linguistics.

\bibitem[{Ali et~al.(2024)Ali, Fromm, Thellmann, Ebert, Weber, Rutmann, Jain,
  Lübbering, Steinigen, Leveling, Klug, Buschhoff, Jurkschat, Abdelwahab,
  Stein, Sylla, Denisov, Brandizzi, Saleem, Bhowmick, Helmer, John, Suarez,
  Ostendorff, Jude, Manjunath, Weinbach, Penke, Filatov, Barth, Mirza, Weber,
  Wendler, Sifa, Küch, Herten, Jäkel, Rehm, Kesselheim, Köhler, and
  Flores-Herr}]{teuken2025}
Mehdi Ali, Michael Fromm, Klaudia Thellmann, Jan Ebert, Alexander~Arno Weber,
  Richard Rutmann, Charvi Jain, Max Lübbering, Daniel Steinigen, Johannes
  Leveling, Katrin Klug, Jasper~Schulze Buschhoff, Lena Jurkschat, Hammam
  Abdelwahab, Benny~Jörg Stein, Karl-Heinz Sylla, Pavel Denisov, Nicolo'
  Brandizzi, Qasid Saleem, and 22 others. 2024.
\newblock \href {https://arxiv.org/abs/2410.03730} {Teuken-7b-base \&
  teuken-7b-instruct: Towards european llms}.
\newblock \emph{Preprint}, arXiv:2410.03730.

\bibitem[{{Anthropic}(2026)}]{claude4}
{Anthropic}. 2026.
\newblock {System Card: Claude Sonnet 4.6}.
\newblock
  \url{https://www-cdn.anthropic.com/bbd8ef16d70b7a1665f14f306ee88b53f686aa75.pdf}.
\newblock Accessed: 2026-05-25.

\bibitem[{Apertus et~al.(2025)Apertus, Hernández-Cano, Hägele, Huang,
  Romanou, Solergibert, Pasztor, Messmer, Garbaya, Ďurech, Hakimi, Giraldo,
  Ismayilzada, Foroutan, Moalla, Chen, Sabolčec, Xu, Aerni, AlKhamissi,
  Mariñas, Amani, Ansaripour, Badanin, Benoit, Boros, Browning, Bösch,
  Böther, Canova, Challier, Charmillot, Coles, Deriu, Devos, Drescher,
  Dzenhaliou, Ehrmann, Fan, Fan, Gao, Gila, Grandury, Hashemi, Hoyle, Jiang,
  Klein, Kucharavy, Kucherenko, Lübeck, Machacek, Manitaras, Marfurt, Matoba,
  Matrenok, Mendonça, Mohamed, Montariol, Mouchel, Najem-Meyer, Ni, Oliva,
  Pagliardini, Palme, Panferov, Paoletti, Passerini, Pavlov, Poiroux, Ponkshe,
  Ranchin, Rando, Sauser, Saydaliev, Sayfiddinov, Schneider, Schuppli,
  Scialanga, Semenov, Shridhar, Singhal, Sotnikova, Sternfeld, Tarun,
  Teiletche, Vamvas, Yao, Zhao, Ilic, Klimovic, Krause, Gulcehre, Rosenthal,
  Ash, Tramèr, VandeVondele, Veraldi, Rajman, Schulthess, Hoefler, Bosselut,
  Jaggi, and Schlag}]{apertus}
Project Apertus, Alejandro Hernández-Cano, Alexander Hägele, Allen~Hao Huang,
  Angelika Romanou, Antoni-Joan Solergibert, Barna Pasztor, Bettina Messmer,
  Dhia Garbaya, Eduard~Frank Ďurech, Ido Hakimi, Juan~García Giraldo, Mete
  Ismayilzada, Negar Foroutan, Skander Moalla, Tiancheng Chen, Vinko Sabolčec,
  Yixuan Xu, Michael Aerni, and 84 others. 2025.
\newblock \href {https://arxiv.org/abs/2509.14233} {Apertus: Democratizing open
  and compliant llms for global language environments}.
\newblock \emph{Preprint}, arXiv:2509.14233.

\bibitem[{Artetxe et~al.(2020)Artetxe, Labaka, and
  Agirre}]{artetxe2020translation}
Mikel Artetxe, Gorka Labaka, and Eneko Agirre. 2020.
\newblock \href {https://doi.org/10.18653/v1/2020.emnlp-main.618} {Translation
  artifacts in cross-lingual transfer learning}.
\newblock In \emph{Proceedings of the 2020 Conference on Empirical Methods in
  Natural Language Processing (EMNLP)}, pages 7674--7684. Association for
  Computational Linguistics.

\bibitem[{Augustyniak et~al.(2022)Augustyniak, Tagowski, Sawczyn, Janiak,
  Bartusiak, Szymczak, Janz, Szyma{\'{n}}ski, W{\k{a}}troba, Morzy,
  Kajdanowicz, and Piasecki}]{lepiszcze2022}
{\L}ukasz Augustyniak, Kamil Tagowski, Albert Sawczyn, Denis Janiak, Roman
  Bartusiak, Adrian Szymczak, Arkadiusz Janz, Piotr Szyma{\'{n}}ski, Marcin
  W{\k{a}}troba, Miko{\l}aj Morzy, Tomasz~Jan Kajdanowicz, and Maciej Piasecki.
  2022.
\newblock \href
  {http://papers.nips.cc/paper\_files/paper/2022/hash/890b206ebb79e550f3988cb8db936f42-Abstract-Datasets\_and\_Benchmarks.html}
  {This is the way: designing and compiling {LEPISZCZE}, a comprehensive {NLP}
  benchmark for {P}olish}.
\newblock In \emph{Advances in Neural Information Processing Systems 35: Annual
  Conference on Neural Information Processing Systems 2022, NeurIPS 2022, New
  Orleans, LA, USA, November 28 - December 9, 2022}.

\bibitem[{Bandarkar et~al.(2024)Bandarkar, Liang, Muller, Artetxe, Shukla,
  Husa, Goyal, Krishnan, Zettlemoyer, and Khabsa}]{belebele2024}
Lucas Bandarkar, Davis Liang, Benjamin Muller, Mikel Artetxe, Satya~Narayan
  Shukla, Donald Husa, Naman Goyal, Abhinandan Krishnan, Luke Zettlemoyer, and
  Madian Khabsa. 2024.
\newblock \href {https://doi.org/10.18653/V1/2024.ACL-LONG.44} {The {B}elebele
  benchmark: a parallel reading comprehension dataset in 122 language
  variants}.
\newblock In \emph{Proceedings of the 62nd Annual Meeting of the Association
  for Computational Linguistics (Volume 1: Long Papers), {ACL} 2024, Bangkok,
  Thailand, August 11-16, 2024}, pages 749--775. Association for Computational
  Linguistics.

\bibitem[{Banerjee and Lavie(2005)}]{banerjee2005meteor}
Satanjeev Banerjee and Alon Lavie. 2005.
\newblock \href {https://aclanthology.org/W05-0909/} {{METEOR}: An automatic
  metric for {MT} evaluation with improved correlation with human judgments}.
\newblock In \emph{Proceedings of the {ACL} Workshop on Intrinsic and Extrinsic
  Evaluation Measures for Machine Translation and/or Summarization}, pages
  65--72, Ann Arbor, Michigan. Association for Computational Linguistics.

\bibitem[{Bergmanis et~al.(2026)Bergmanis, Kronis, Pretkalniņš, Nicmanis,
  Jelinska, Rozis, Vīksna, and Pinnis}]{tildeopen2025}
Toms Bergmanis, Martins Kronis, Ingus~Jānis Pretkalniņš, Dāvis Nicmanis,
  Jeļizaveta Jelinska, Roberts Rozis, Rinalds Vīksna, and Mārcis Pinnis.
  2026.
\newblock \href {https://arxiv.org/abs/2603.08182} {Tildeopen llm: Leveraging
  curriculum learning to achieve equitable language representation}.
\newblock \emph{Preprint}, arXiv:2603.08182.

\bibitem[{{BerriAI}(2024)}]{litellm}
{BerriAI}. 2024.
\newblock {LiteLLM}: A unified interface for {LLM} apis.
\newblock \url{https://github.com/BerriAI/litellm}.
\newblock Accessed: 2026-05-25.

\bibitem[{Blšták(2025)}]{blstak2025slovakconceptualdictionary}
Miroslav Blšták. 2025.
\newblock \href {https://arxiv.org/abs/2512.00579} {Slovak conceptual
  dictionary}.
\newblock \emph{Preprint}, arXiv:2512.00579.

\bibitem[{{CJVT and University of Ljubljana}(2024)}]{slobench}
{CJVT and University of Ljubljana}. 2024.
\newblock {SloBench}: The {S}lovene language technology evaluation platform.
\newblock \url{https://slobench.cjvt.si}.
\newblock Accessed: 2026-04-16.

\bibitem[{Clark et~al.(2018)Clark, Cowhey, Etzioni, Khot, Sabharwal, Schoenick,
  and Tafjord}]{arc2018}
Peter Clark, Isaac Cowhey, Oren Etzioni, Tushar Khot, Ashish Sabharwal, Carissa
  Schoenick, and Oyvind Tafjord. 2018.
\newblock \href {https://arxiv.org/abs/1803.05457} {Think you have solved
  question answering? try arc, the ai2 reasoning challenge}.
\newblock \emph{Preprint}, arXiv:1803.05457.

\bibitem[{Cohere et~al.(2025)Cohere, :, Aakanksha, Ahmadian, Ahmed, Alammar,
  Alizadeh, Alnumay, Althammer, Arkhangorodsky, Aryabumi, Aumiller, Avalos,
  Aviv, Bae, Baji, Barbet, Bartolo, Bebensee, Beladia, Beller-Morales, Bérard,
  Berneshawi, Bialas, Blunsom, Bobkin, Bongale, Braun, Brunet, Cahyawijaya,
  Cairuz, Campos, Cao, Cao, Castagné, Cendrero, Currie, Chandak, Chang,
  Chatziveroglou, Chen, Cheng, Chevalier, Chiu, Cho, Choi, Choi, Chung, Cirik,
  Cismaru, Clavier, Conklin, Crawhall-Stein, Crouse, Cruz-Salinas, Cyrus,
  D'souza, Dalla-Torre, Dang, Darling, Domingues, Dash, Debugne, Dehaze, Desai,
  Devassy, Dholakia, Duffy, Edalati, Eldeib, Elkady, Elsharkawy, Ergün, Ermis,
  Fadaee, Fan, Fayoux, Flet-Berliac, Frosst, Gallé, Galuba, Garg, Geist, Azar,
  Gilsenan-McMahon, Goldfarb-Tarrant, Goldsack, Gomez, Gonzaga, Govindarajan,
  Govindassamy, Grinsztajn, Gritsch, Gu, Guo, Haefeli, Hajjar, Hawes, He,
  Hofstätter, Hong, Hooker, Hosking, Howe, Hu, Huang, Jain, Jain, Jakobi,
  Jenkins, Jordan, Joshi, Jung, Kalyanpur, Kamalakara, Kedrzycki, Keskin, Kim,
  Kim, Ko, Kocmi, Kozakov, Kryściński, Jain, Teru, Land, Lasby, Lasche, Lee,
  Lewis, Li, Li, Lin, Locatelli, Luong, Ma, Mach, Machado, Magbitang, Lopez,
  Mann, Marchisio, Markham, Matton, McKinney, McLoughlin, Mokry, Morisot,
  Moulder, Moynehan, Mozes, Muppalla, Murakhovska, Nagarajan, Nandula, Nasir,
  Nehra, Netto-Rosen, Ohashi, Owers-Bardsley, Ozuzu, Padilla, Park, Passaglia,
  Pekmez, Penstone, Piktus, Ploeg, Poulton, Qi, Raghvendra, Ramos, Ranjan,
  Richemond, Robert-Michon, Rodriguez, Roy, Ruder, Ruis, Rust, Sachan,
  Salamanca, Saravanakumar, Satyakam, Sebag, Sen, Sepehri, Seshadri, Shen,
  Sherborne, Shi, Shivaprasad, Shmyhlo, Shrinivason, Shteinbuk, Shukayev,
  Simard, Snyder, Spataru, Spooner, Starostina, Strub, Su, Sun, Talupuru,
  Tarassov, Tommasone, Tracey, Trend, Tumer, Üstün, Venkitesh, Venuto, Verga,
  Voisin, Wang, Wang, Wang, Wen, White, Willman, Winkels, Xia, Xie, Xu, Yang,
  Yi-Chern, Zhang, Zhao, and Zhao}]{commanda}
Team Cohere, :, Aakanksha, Arash Ahmadian, Marwan Ahmed, Jay Alammar, Milad
  Alizadeh, Yazeed Alnumay, Sophia Althammer, Arkady Arkhangorodsky, Viraat
  Aryabumi, Dennis Aumiller, Raphaël Avalos, Zahara Aviv, Sammie Bae, Saurabh
  Baji, Alexandre Barbet, Max Bartolo, Björn Bebensee, and 211 others. 2025.
\newblock \href {https://arxiv.org/abs/2504.00698} {Command a: An
  enterprise-ready large language model}.
\newblock \emph{Preprint}, arXiv:2504.00698.

\bibitem[{Conneau et~al.(2018)Conneau, Rinott, Lample, Williams, Bowman,
  Schwenk, and Stoyanov}]{xnli2018}
Alexis Conneau, Ruty Rinott, Guillaume Lample, Adina Williams, Samuel~R.
  Bowman, Holger Schwenk, and Veselin Stoyanov. 2018.
\newblock \href {https://doi.org/10.18653/V1/D18-1269} {{XNLI:} evaluating
  cross-lingual sentence representations}.
\newblock In \emph{Proceedings of the 2018 Conference on Empirical Methods in
  Natural Language Processing, Brussels, Belgium, October 31 - November 4,
  2018}, pages 2475--2485. Association for Computational Linguistics.

\bibitem[{Csaki et~al.(2024)Csaki, Li, Li, Xu, Pawakapan, Zhang, Du, Zhao, Hu,
  and Thakker}]{sambalingo2024}
Zoltan Csaki, Bo~Li, Jonathan~Lingjie Li, Qiantong Xu, Pian Pawakapan, Leon
  Zhang, Yun Du, Hengyu Zhao, Changran Hu, and Urmish Thakker. 2024.
\newblock \href {https://doi.org/10.18653/v1/2024.mrl-1.1} {{S}amba{L}ingo:
  Teaching large language models new languages}.
\newblock In \emph{Proceedings of the Fourth Workshop on Multilingual
  Representation Learning (MRL)}, pages 1--21, Miami, Florida, USA. Association
  for Computational Linguistics.

\bibitem[{Dang et~al.(2024)Dang, Singh, D'souza, Ahmadian, Salamanca, Smith,
  Peppin, Hong, Govindassamy, Zhao, Kublik, Amer, Aryabumi, Campos, Tan, Kocmi,
  Strub, Grinsztajn, Flet-Berliac, Locatelli, Lin, Talupuru, Venkitesh, Cairuz,
  Yang, Chung, Ko, Shi, Shukayev, Bae, Piktus, Castagné, Cruz-Salinas, Kim,
  Crawhall-Stein, Morisot, Roy, Blunsom, Zhang, Gomez, Frosst, Fadaee, Ermis,
  Üstün, and Hooker}]{ayaexpanse}
John Dang, Shivalika Singh, Daniel D'souza, Arash Ahmadian, Alejandro
  Salamanca, Madeline Smith, Aidan Peppin, Sungjin Hong, Manoj Govindassamy,
  Terrence Zhao, Sandra Kublik, Meor Amer, Viraat Aryabumi, Jon~Ander Campos,
  Yi-Chern Tan, Tom Kocmi, Florian Strub, Nathan Grinsztajn, Yannis
  Flet-Berliac, and 26 others. 2024.
\newblock \href {https://arxiv.org/abs/2412.04261} {Aya expanse: Combining
  research breakthroughs for a new multilingual frontier}.
\newblock \emph{Preprint}, arXiv:2412.04261.

\bibitem[{Dobe\v{s} and Bedn\'{a}r(2026)}]{skj1}
Marek Dobe\v{s} and Peter Bedn\'{a}r. 2026.
\newblock \href {https://huggingface.co/datasets/slovak-nlp/SKJ1} {{SKJ1}:
  {S}lovak grammar correction multi-choice set}.
\newblock Hugging Face dataset.
\newblock To appear as ``SKJ1 - Benchmark for Slovak Language Capabilities of
  Large Language Models'', TCL 2026. Accessed: 2026-04-20.

\bibitem[{Dussolle et~al.(2025)Dussolle, Carde{\~n}a~D{\'i}az, Sato, and
  Devine}]{mifeval2025}
Antoine Dussolle, Andrea Carde{\~n}a~D{\'i}az, Shota Sato, and Peter Devine.
  2025.
\newblock \href {https://doi.org/10.18653/v1/2025.findings-naacl.344}
  {{M}-{IFE}val: Multilingual instruction-following evaluation}.
\newblock In \emph{Findings of the Association for Computational Linguistics:
  NAACL 2025}, pages 6176--6191, Albuquerque, New Mexico. Association for
  Computational Linguistics.

\bibitem[{Fajcik et~al.(2025)Fajcik, Docekal, Dolezal, Ondrej, Beneš, Kapsa,
  Smrz, Polok, Hradis, Neverilova, Horak, Sabol, Stefanik, Jirkovsky, Adamczyk,
  Hyner, Hula, and Kydlicek}]{benczechmark2025}
Martin Fajcik, Martin Docekal, Jan Dolezal, Karel Ondrej, Karel Beneš, Jan
  Kapsa, Pavel Smrz, Alexander Polok, Michal Hradis, Zuzana Neverilova, Ales
  Horak, Radoslav Sabol, Michal Stefanik, Adam Jirkovsky, David Adamczyk, Petr
  Hyner, Jan Hula, and Hynek Kydlicek. 2025.
\newblock \href {https://doi.org/10.1162/TACL.a.32} {{BenCzechMark : A
  Czech-Centric Multitask and Multimetric Benchmark for Large Language Models
  with Duel Scoring Mechanism}}.
\newblock \emph{Transactions of the Association for Computational Linguistics},
  13:1068--1095.

\bibitem[{Fenogenova et~al.(2024)Fenogenova, Chervyakov, Martynov, Kozlova,
  Tikhonova, Akhmetgareeva, Emelyanov, Shevelev, Lebedev, Sinev, Isaeva,
  Kolomeytseva, Moskovskiy, Goncharova, Savushkin, Mikhailova, Minaeva,
  Dimitrov, Panchenko, and Markov}]{mera2024}
Alena Fenogenova, Artem Chervyakov, Nikita Martynov, Anastasia Kozlova, Maria
  Tikhonova, Albina Akhmetgareeva, Anton~A. Emelyanov, Denis Shevelev, Pavel
  Lebedev, Leonid Sinev, Ulyana Isaeva, Katerina Kolomeytseva, Daniil
  Moskovskiy, Elizaveta Goncharova, Nikita Savushkin, Polina Mikhailova,
  Anastasia Minaeva, Denis Dimitrov, Alexander Panchenko, and Sergey Markov.
  2024.
\newblock \href {https://doi.org/10.18653/V1/2024.ACL-LONG.534} {{MERA:} {A}
  comprehensive {LLM} evaluation in {R}ussian}.
\newblock In \emph{Proceedings of the 62nd Annual Meeting of the Association
  for Computational Linguistics (Volume 1: Long Papers), {ACL} 2024, Bangkok,
  Thailand, August 11-16, 2024}, pages 9920--9948. Association for
  Computational Linguistics.

\bibitem[{Fourrier et~al.(2024)Fourrier, Habib, Lozovskaya, Szafer, and
  Wolf}]{openllmleaderboard}
Cl{\'e}mentine Fourrier, Nathan Habib, Alina Lozovskaya, Konrad Szafer, and
  Thomas Wolf. 2024.
\newblock Open {LLM} leaderboard v2.
\newblock Hugging Face Space
  \url{https://huggingface.co/spaces/open-llm-leaderboard/open_llm_leaderboard}.
\newblock Accessed: 2026-04-28.

\bibitem[{French(1999)}]{french1999forgetting}
Robert~M. French. 1999.
\newblock \href {https://doi.org/10.1016/S1364-6613(99)01294-2} {Catastrophic
  forgetting in connectionist networks}.
\newblock \emph{Trends in Cognitive Sciences}, 3(4):128--135.

\bibitem[{Gao et~al.(2023)Gao, Tow, Abbasi, Biderman, Black, DiPofi, Foster,
  Golding, Hsu, Le~Noac'h, Li, McDonell, Muennighoff, Ociepa, Phang, Reynolds,
  Schoelkopf, Skowron, Sutawika, Tang, Thite, Wang, Wang, and
  Zou}]{lmevalharness}
Leo Gao, Jonathan Tow, Baber Abbasi, Stella Biderman, Sid Black, Anthony
  DiPofi, Charles Foster, Laurence Golding, Jeffrey Hsu, Alain Le~Noac'h,
  Haonan Li, Kyle McDonell, Niklas Muennighoff, Chris Ociepa, Jason Phang,
  Laria Reynolds, Hailey Schoelkopf, Aviya Skowron, Lintang Sutawika, and 5
  others. 2023.
\newblock \href {https://doi.org/10.5281/zenodo.10256836} {A framework for
  few-shot language model evaluation}.

\bibitem[{Garabík(2026)}]{Garabik2026}
Radovan Garabík. 2026.
\newblock \href
  {https://www.sav.sk/journals/uploads/0518105173-79-Slovenska-Wikipedia-ako-zdroj-dat-pre-korpus-chyb.pdf}
  {Slovenská wikipédia ako zdroj dát pre korpus chýb}.
\newblock \emph{Kultúra slova}.
\newblock In print.

\bibitem[{{Google DeepMind}(2026)}]{gemini3}
{Google DeepMind}. 2026.
\newblock {Gemini 3.1 Pro Model Card}.
\newblock \url{https://deepmind.google/models/model-cards/gemini-3-1-pro/}.
\newblock Accessed: 2026-05-07.

\bibitem[{{Granite Team, IBM}(2024)}]{granite2024}
{Granite Team, IBM}. 2024.
\newblock \href
  {https://github.com/ibm-granite/granite-3.0-language-models/blob/main/paper.pdf}
  {Granite 3.0 language models}.
\newblock Technical report, IBM.

\bibitem[{Grattafiori et~al.(2024)Grattafiori, Dubey, Jauhri, Pandey, Kadian,
  Al-Dahle, Letman, Mathur, Schelten, Vaughan, Yang, Fan, Goyal, Hartshorn,
  Yang, Mitra, Sravankumar, Korenev, Hinsvark, Rao, Zhang, Rodriguez,
  Gregerson, Spataru, Roziere, Biron, Tang, Chern, Caucheteux, Nayak, Bi,
  Marra, McConnell, Keller, Touret, Wu, Wong, Ferrer, Nikolaidis, Allonsius,
  Song, Pintz, Livshits, Wyatt, Esiobu, Choudhary, Mahajan, Garcia-Olano,
  Perino, Hupkes, Lakomkin, AlBadawy, Lobanova, Dinan, Smith, Radenovic,
  Guzmán, Zhang, Synnaeve, Lee, Anderson, Thattai, Nail, Mialon, Pang,
  Cucurell, Nguyen, Korevaar, Xu, Touvron, Zarov, Ibarra, Kloumann, Misra,
  Evtimov, Zhang, Copet, Lee, Geffert, Vranes, Park, Mahadeokar, Shah, van~der
  Linde, Billock, Hong, Lee, Fu, Chi, Huang, Liu, Wang, Yu, Bitton, Spisak,
  Park, Rocca, Johnstun, Saxe, Jia, Alwala, Prasad, Upasani, Plawiak, Li,
  Heafield, Stone, El-Arini, Iyer, Malik, Chiu, Bhalla, Lakhotia,
  Rantala-Yeary, van~der Maaten, Chen, Tan, Jenkins, Martin, Madaan, Malo,
  Blecher, Landzaat, de~Oliveira, Muzzi, Pasupuleti, Singh, Paluri, Kardas,
  Tsimpoukelli, Oldham, Rita, Pavlova, Kambadur, Lewis, Si, Singh, Hassan,
  Goyal, Torabi, Bashlykov, Bogoychev, Chatterji, Zhang, Duchenne, Çelebi,
  Alrassy, Zhang, Li, Vasic, Weng, Bhargava, Dubal, Krishnan, Koura, Xu, He,
  Dong, Srinivasan, Ganapathy, Calderer, Cabral, Stojnic, Raileanu, Maheswari,
  Girdhar, Patel, Sauvestre, Polidoro, Sumbaly, Taylor, Silva, Hou, Wang,
  Hosseini, Chennabasappa, Singh, Bell, Kim, Edunov, Nie, Narang, Raparthy,
  Shen, Wan, Bhosale, Zhang, Vandenhende, Batra, Whitman, Sootla, Collot,
  Gururangan, Borodinsky, Herman, Fowler, Sheasha, Georgiou, Scialom,
  Speckbacher, Mihaylov, Xiao, Karn, Goswami, Gupta, Ramanathan, Kerkez,
  Gonguet, Do, Vogeti, Albiero, Petrovic, Chu, Xiong, Fu, Meers, Martinet,
  Wang, Wang, Tan, Xia, Xie, Jia, Wang, Goldschlag, Gaur, Babaei, Wen, Song,
  Zhang, Li, Mao, Coudert, Yan, Chen, Papakipos, Singh, Srivastava, Jain,
  Kelsey, Shajnfeld, Gangidi, Victoria, Goldstand, Menon, Sharma, Boesenberg,
  Baevski, Feinstein, Kallet, Sangani, Teo, Yunus, Lupu, Alvarado, Caples, Gu,
  Ho, Poulton, Ryan, Ramchandani, Dong, Franco, Goyal, Saraf, Chowdhury,
  Gabriel, Bharambe, Eisenman, Yazdan, James, Maurer, Leonhardi, Huang, Loyd,
  Paola, Paranjape, Liu, Wu, Ni, Hancock, Wasti, Spence, Stojkovic, Gamido,
  Montalvo, Parker, Burton, Mejia, Liu, Wang, Kim, Zhou, Hu, Chu, Cai, Tindal,
  Feichtenhofer, Gao, Civin, Beaty, Kreymer, Li, Adkins, Xu, Testuggine, David,
  Parikh, Liskovich, Foss, Wang, Le, Holland, Dowling, Jamil, Montgomery,
  Presani, Hahn, Wood, Le, Brinkman, Arcaute, Dunbar, Smothers, Sun, Kreuk,
  Tian, Kokkinos, Ozgenel, Caggioni, Kanayet, Seide, Florez, Schwarz, Badeer,
  Swee, Halpern, Herman, Sizov, Guangyi, Zhang, Lakshminarayanan, Inan,
  Shojanazeri, Zou, Wang, Zha, Habeeb, Rudolph, Suk, Aspegren, Goldman, Zhan,
  Damlaj, Molybog, Tufanov, Leontiadis, Veliche, Gat, Weissman, Geboski, Kohli,
  Lam, Asher, Gaya, Marcus, Tang, Chan, Zhen, Reizenstein, Teboul, Zhong, Jin,
  Yang, Cummings, Carvill, Shepard, McPhie, Torres, Ginsburg, Wang, Wu, U,
  Saxena, Khandelwal, Zand, Matosich, Veeraraghavan, Michelena, Li, Jagadeesh,
  Huang, Chawla, Huang, Chen, Garg, A, Silva, Bell, Zhang, Guo, Yu, Moshkovich,
  Wehrstedt, Khabsa, Avalani, Bhatt, Mankus, Hasson, Lennie, Reso, Groshev,
  Naumov, Lathi, Keneally, Liu, Seltzer, Valko, Restrepo, Patel, Vyatskov,
  Samvelyan, Clark, Macey, Wang, Hermoso, Metanat, Rastegari, Bansal,
  Santhanam, Parks, White, Bawa, Singhal, Egebo, Usunier, Mehta, Laptev, Dong,
  Cheng, Chernoguz, Hart, Salpekar, Kalinli, Kent, Parekh, Saab, Balaji,
  Rittner, Bontrager, Roux, Dollar, Zvyagina, Ratanchandani, Yuvraj, Liang,
  Alao, Rodriguez, Ayub, Murthy, Nayani, Mitra, Parthasarathy, Li, Hogan,
  Battey, Wang, Howes, Rinott, Mehta, Siby, Bondu, Datta, Chugh, Hunt, Dhillon,
  Sidorov, Pan, Mahajan, Verma, Yamamoto, Ramaswamy, Lindsay, Lindsay, Feng,
  Lin, Zha, Patil, Shankar, Zhang, Zhang, Wang, Agarwal, Sajuyigbe, Chintala,
  Max, Chen, Kehoe, Satterfield, Govindaprasad, Gupta, Deng, Cho, Virk,
  Subramanian, Choudhury, Goldman, Remez, Glaser, Best, Koehler, Robinson, Li,
  Zhang, Matthews, Chou, Shaked, Vontimitta, Ajayi, Montanez, Mohan, Kumar,
  Mangla, Ionescu, Poenaru, Mihailescu, Ivanov, Li, Wang, Jiang, Bouaziz,
  Constable, Tang, Wu, Wang, Wu, Gao, Kleinman, Chen, Hu, Jia, Qi, Li, Zhang,
  Zhang, Adi, Nam, Yu, Wang, Zhao, Hao, Qian, Li, He, Rait, DeVito, Rosnbrick,
  Wen, Yang, Zhao, and Ma}]{llama3}
Aaron Grattafiori, Abhimanyu Dubey, Abhinav Jauhri, Abhinav Pandey, Abhishek
  Kadian, Ahmad Al-Dahle, Aiesha Letman, Akhil Mathur, Alan Schelten, Alex
  Vaughan, Amy Yang, Angela Fan, Anirudh Goyal, Anthony Hartshorn, Aobo Yang,
  Archi Mitra, Archie Sravankumar, Artem Korenev, Arthur Hinsvark, and 542
  others. 2024.
\newblock \href {https://arxiv.org/abs/2407.21783} {{The Llama 3 Herd of
  Models}}.
\newblock \emph{Preprint}, arXiv:2407.21783.

\bibitem[{Habib et~al.(2023)Habib, Fourrier, Kydlíček, Wolf, and
  Tunstall}]{lighteval}
Nathan Habib, Clémentine Fourrier, Hynek Kydlíček, Thomas Wolf, and Lewis
  Tunstall. 2023.
\newblock \href {https://github.com/huggingface/lighteval} {Lighteval: A
  lightweight framework for llm evaluation}.
\newblock Accessed: 2026-02-15.

\bibitem[{Hendrycks et~al.(2020)Hendrycks, Burns, Basart, Zou, Mazeika, Song,
  and Steinhardt}]{mmlu2021}
Dan Hendrycks, Collin Burns, Steven Basart, Andy Zou, Mantas Mazeika, Dawn
  Song, and Jacob Steinhardt. 2020.
\newblock \href {https://arxiv.org/abs/2009.03300} {Measuring massive multitask
  language understanding}.
\newblock \emph{Preprint}, arXiv:2009.03300.

\bibitem[{Hl{\'a}dek et~al.(2025)Hl{\'a}dek, Sopkovi{\v{c}}, Sta{\v{s}},
  Sokolov{\'a}, and Pleva}]{financial_exams}
Daniel Hl{\'a}dek, Kristi{\'a}n Sopkovi{\v{c}}, J{\'a}n Sta{\v{s}}, Zuzana
  Sokolov{\'a}, and Mat{\'u}{\v{s}} Pleva. 2025.
\newblock \href {https://doi.org/10.2478/jazcas-2025-0022} {Financial
  question-answering dataset for slovak language model evaluation}.
\newblock \emph{Journal of Linguistics/Jazykovedn\'{y} \v{c}asopis},
  76(1):247--257.

\bibitem[{Hl{\'{a}}dek et~al.(2023)Hl{\'{a}}dek, Sta{\v{s}}, Juh{\'{a}}r, and
  Koct{\'{u}}r}]{skquad2023}
Daniel Hl{\'{a}}dek, J{\'{a}}n Sta{\v{s}}, Jozef Juh{\'{a}}r, and
  Tom{\'{a}}{\v{s}} Koct{\'{u}}r. 2023.
\newblock \href {https://doi.org/10.1109/ACCESS.2023.3262308} {{S}lovak dataset
  for multilingual question answering}.
\newblock \emph{{IEEE} Access}, 11:32869--32881.

\bibitem[{Hu et~al.(2020)Hu, Ruder, Siddhant, Neubig, Firat, and
  Johnson}]{xtreme2020}
Junjie Hu, Sebastian Ruder, Aditya Siddhant, Graham Neubig, Orhan Firat, and
  Melvin Johnson. 2020.
\newblock \href {http://proceedings.mlr.press/v119/hu20b.html} {{XTREME:} {A}
  massively multilingual multi-task benchmark for evaluating cross-lingual
  generalisation}.
\newblock In \emph{Proceedings of the 37th International Conference on Machine
  Learning, {ICML} 2020, 13-18 July 2020, Virtual Event}, Proceedings of
  Machine Learning Research, pages 4411--4421. {PMLR}.

\bibitem[{Huang et~al.(2023)Huang, Bai, Zhu, Zhang, Zhang, Su, Liu, Lv, Zhang,
  Lei, Fu, Sun, and He}]{ceval2023}
Yuzhen Huang, Yuzhuo Bai, Zhihao Zhu, Junlei Zhang, Jinghan Zhang, Tangjun Su,
  Junteng Liu, Chuancheng Lv, Yikai Zhang, Jiayi Lei, Yao Fu, Maosong Sun, and
  Junxian He. 2023.
\newblock \href {https://doi.org/10.52202/075280-2749} {C-eval: A multi-level
  multi-discipline chinese evaluation suite for foundation models}.
\newblock In \emph{Advances in Neural Information Processing Systems (NeurIPS):
  Datasets and Benchmarks Track}, volume~36, pages 62991--63010.

\bibitem[{Hupkes and Bogoychev(2025)}]{multiloko2025}
Dieuwke Hupkes and Nikolay Bogoychev. 2025.
\newblock \href {https://arxiv.org/abs/2504.10356} {Multiloko: a multilingual
  local knowledge benchmark for llms spanning 31 languages}.
\newblock \emph{Preprint}, arXiv:2504.10356.

\bibitem[{Jiang et~al.(2023)Jiang, Sablayrolles, Mensch, Bamford, Chaplot,
  de~las Casas, Bressand, Lengyel, Lample, Saulnier, Lavaud, Lachaux, Stock,
  Scao, Lavril, Wang, Lacroix, and Sayed}]{mistral7b}
Albert~Q. Jiang, Alexandre Sablayrolles, Arthur Mensch, Chris Bamford,
  Devendra~Singh Chaplot, Diego de~las Casas, Florian Bressand, Gianna Lengyel,
  Guillaume Lample, Lucile Saulnier, Lélio~Renard Lavaud, Marie-Anne Lachaux,
  Pierre Stock, Teven~Le Scao, Thibaut Lavril, Thomas Wang, Timothée Lacroix,
  and William~El Sayed. 2023.
\newblock \href {https://arxiv.org/abs/2310.06825} {Mistral 7b}.
\newblock \emph{Preprint}, arXiv:2310.06825.

\bibitem[{Jumelet et~al.(2026)Jumelet, Weissweiler, Nivre, and
  Bisazza}]{multiblimp2026}
Jaap Jumelet, Leonie Weissweiler, Joakim Nivre, and Arianna Bisazza. 2026.
\newblock \href {https://doi.org/10.1162/tacl.a.600} {{MultiBLiMP} 1.0: A
  massively multilingual benchmark of linguistic minimal pairs}.
\newblock \emph{Transactions of the Association for Computational Linguistics},
  14:193--216.

\bibitem[{Klemen et~al.(2024)Klemen, {\v{Z}}agar, {\v{C}}ibej, and
  Robnik{-}{\v{S}}ikonja}]{sinli2024}
Matej Klemen, Ale{\v{s}} {\v{Z}}agar, Jaka {\v{C}}ibej, and Marko
  Robnik{-}{\v{S}}ikonja. 2024.
\newblock \href {https://aclanthology.org/2024.lrec-main.1294} {{SI-NLI:} {A}
  {S}lovene natural language inference dataset and its evaluation}.
\newblock In \emph{Proceedings of the 2024 Joint International Conference on
  Computational Linguistics, Language Resources and Evaluation, {LREC/COLING}
  2024, 20-25 May, 2024, Torino, Italy}, pages 14859--14870. {ELRA} and {ICCL}.

\bibitem[{Koto et~al.(2023)Koto, Aisyah, Li, and Baldwin}]{indommlu2023}
Fajri Koto, Nurul Aisyah, Haonan Li, and Timothy Baldwin. 2023.
\newblock \href {https://doi.org/10.18653/v1/2023.emnlp-main.760} {Large
  language models only pass primary school exams in indonesia: A comprehensive
  test on {IndoMMLU}}.
\newblock In \emph{Proceedings of the 2023 Conference on Empirical Methods in
  Natural Language Processing (EMNLP)}, pages 12359--12374.

\bibitem[{Koto et~al.(2024)Koto, Li, Shatnawi, Doughman, Sadallah, Alraeesi,
  Almubarak, Alyafeai, Sengupta, Shehata, Habash, Nakov, and
  Baldwin}]{arabicmmlu2024}
Fajri Koto, Haonan Li, Sara Shatnawi, Jad Doughman, Abdelrahman Sadallah, Aisha
  Alraeesi, Khalid Almubarak, Zaid Alyafeai, Neha Sengupta, Shady Shehata,
  Nizar Habash, Preslav Nakov, and Timothy Baldwin. 2024.
\newblock \href {https://doi.org/10.18653/v1/2024.findings-acl.334}
  {{ArabicMMLU}: Assessing massive multitask language understanding in arabic}.
\newblock In \emph{Findings of the Association for Computational Linguistics:
  ACL 2024}, pages 5622--5640.

\bibitem[{Kwon et~al.(2023)Kwon, Li, Zhuang, Sheng, Zheng, Yu, Gonzalez, Zhang,
  and Stoica}]{vllm2023}
Woosuk Kwon, Zhuohan Li, Siyuan Zhuang, Ying Sheng, Lianmin Zheng, Cody~Hao Yu,
  Joseph Gonzalez, Hao Zhang, and Ion Stoica. 2023.
\newblock \href {https://doi.org/10.1145/3600006.3613165} {Efficient memory
  management for large language model serving with pagedattention}.
\newblock In \emph{Proceedings of the 29th Symposium on Operating Systems
  Principles, {SOSP} 2023, Koblenz, Germany, October 23-26, 2023}, pages
  611--626. {ACM}.

\bibitem[{Kňažeková and Šuppa(2025)}]{nli_sk}
Natália Kňažeková and Marek Šuppa. 2025.
\newblock \href {https://huggingface.co/datasets/natalia-nk/NLI-SK-annotated}
  {{NLI-SK-annotated}: A native {S}lovak natural language inference dataset}.
\newblock Hugging Face dataset.
\newblock Accessed: 2026-02-20.

\bibitem[{Lai et~al.(2023)Lai, Nguyen, Ngo, Nguyen, Dernoncourt, Rossi, and
  Nguyen}]{okapi2023}
Viet~Dac Lai, Chien~Van Nguyen, Nghia~Trung Ngo, Thuat Nguyen, Franck
  Dernoncourt, Ryan~A. Rossi, and Thien~Huu Nguyen. 2023.
\newblock \href {https://doi.org/10.18653/V1/2023.EMNLP-DEMO.28} {Okapi:
  Instruction-tuned large language models in multiple languages with
  reinforcement learning from human feedback}.
\newblock In \emph{Proceedings of the 2023 Conference on Empirical Methods in
  Natural Language Processing, {EMNLP} 2023 - System Demonstrations, Singapore,
  December 6-10, 2023}, pages 318--327. Association for Computational
  Linguistics.

\bibitem[{Liang et~al.(2023)Liang, Bommasani, Lee, Tsipras, Soylu, Yasunaga,
  Zhang, Narayanan, Wu, Kumar, Newman, Yuan, Yan, Zhang, Cosgrove, Manning,
  R{\'e}, Acosta-Navas, Hudson, Zelikman, Durmus, Ladhak, Rong, Ren, Yao, Wang,
  Santhanam, Orr, Zheng, Yuksekgonul, Suzgun, Kim, Guha, Chatterji, Khattab,
  Henderson, Huang, Chi, Xie, Santurkar, Ganguli, Hashimoto, Icard, Zhang,
  Chaudhary, Wang, Li, Mai, Zhang, and Koreeda}]{helm2023}
Percy Liang, Rishi Bommasani, Tony Lee, Dimitris Tsipras, Dilara Soylu,
  Michihiro Yasunaga, Yian Zhang, Deepak Narayanan, Yuhuai Wu, Ananya Kumar,
  Benjamin Newman, Binhang Yuan, Bobby Yan, Ce~Zhang, Christian Cosgrove,
  Christopher~D. Manning, Christopher R{\'e}, Diana Acosta-Navas, Drew~A.
  Hudson, and 31 others. 2023.
\newblock \href {https://openreview.net/forum?id=iO4LZibEqW} {Holistic
  evaluation of language models}.
\newblock \emph{Transactions on Machine Learning Research}.
\newblock Featured Certification, Expert Certification, Outstanding
  Certification.

\bibitem[{Ligeti{-}Nagy et~al.(2024)Ligeti{-}Nagy, Ferenczi, H{\'{e}}ja, Laki,
  Vad{\'{a}}sz, Yang, and V{\'{a}}radi}]{hulu2024}
No{\'{e}}mi Ligeti{-}Nagy, Gerg{\H{o}} Ferenczi, Enik{\H{o}} H{\'{e}}ja,
  L{\'{a}}szl{\'{o}}~J{\'{a}}nos Laki, No{\'{e}}mi Vad{\'{a}}sz,
  Zijian~Gy{\H{o}}z{\H{o}} Yang, and Tam{\'{a}}s V{\'{a}}radi. 2024.
\newblock \href {https://aclanthology.org/2024.lrec-main.733} {{HuLU:}
  {H}ungarian language understanding benchmark kit}.
\newblock In \emph{Proceedings of the 2024 Joint International Conference on
  Computational Linguistics, Language Resources and Evaluation, {LREC/COLING}
  2024, 20-25 May, 2024, Torino, Italy}, pages 8360--8371. {ELRA} and {ICCL}.

\bibitem[{Lin(2004)}]{rouge2004}
Chin-Yew Lin. 2004.
\newblock \href {https://aclanthology.org/W04-1013/} {{ROUGE}: A package for
  automatic evaluation of summaries}.
\newblock In \emph{Text Summarization Branches Out}, pages 74--81, Barcelona,
  Spain. Association for Computational Linguistics.

\bibitem[{Lin et~al.(2022)Lin, Hilton, and Evans}]{truthfulqa2022}
Stephanie Lin, Jacob Hilton, and Owain Evans. 2022.
\newblock \href {https://doi.org/10.18653/V1/2022.ACL-LONG.229} {{TruthfulQA:}
  measuring how models mimic human falsehoods}.
\newblock In \emph{Proceedings of the 60th Annual Meeting of the Association
  for Computational Linguistics (Volume 1: Long Papers), {ACL} 2022, Dublin,
  Ireland, May 22-27, 2022}, pages 3214--3252. Association for Computational
  Linguistics.

\bibitem[{Luo et~al.(2023)Luo, Yang, Meng, Li, Zhou, and
  Zhang}]{luo2023forgetting}
Yun Luo, Zhen Yang, Fandong Meng, Yafu Li, Jie Zhou, and Yue Zhang. 2023.
\newblock \href {https://arxiv.org/abs/2308.08747} {An empirical study of
  catastrophic forgetting in large language models during continual
  fine-tuning}.
\newblock \emph{Preprint}, arXiv:2308.08747.

\bibitem[{Martins et~al.(2024)Martins, Fernandes, Alves, Guerreiro, Rei, Alves,
  Pombal, Farajian, Faysse, Klimaszewski, Colombo, Haddow, de~Souza, Birch, and
  Martins}]{eurollm2024}
Pedro~Henrique Martins, Patrick Fernandes, João Alves, Nuno~M. Guerreiro,
  Ricardo Rei, Duarte~M. Alves, José Pombal, Amin Farajian, Manuel Faysse,
  Mateusz Klimaszewski, Pierre Colombo, Barry Haddow, José G.~C. de~Souza,
  Alexandra Birch, and André F.~T. Martins. 2024.
\newblock \href {https://arxiv.org/abs/2409.16235} {{EuroLLM}: Multilingual
  language models for {E}urope}.
\newblock \emph{Preprint}, arXiv:2409.16235.

\bibitem[{MiniMax et~al.(2025)MiniMax, Li, Gong, Yang, Shan, Liu, Zhu, Zhang,
  Guo, Chen, Li, Jiao, Li, Zhang, Sun, Dong, Zhu, Zhuang, Song, Zhu, Han, Li,
  Xie, Xu, Yan, Zhang, Xiao, Kang, Han, Wang, Yu, Feng, Zheng, Chai, Xing, Ju,
  Chi, Zhang, Huang, Niu, Li, Zhao, Yang, Xu, Wang, Wang, Li, Leng, Shi, Yu,
  Li, Zhu, Huang, Liang, Sun, Sun, Cheng, Li, Song, Su, Han, Zhang, Hou, Min,
  Zou, Shen, Gong, Zhu, Zhou, Zhong, Hu, Fan, Yu, Yang, Li, Huang, Li, Huang,
  Xu, Mao, Li, Li, Tao, Ying, Cong, Qin, Fan, Yu, Jiang, and Wu}]{minimax01}
MiniMax, Aonian Li, Bangwei Gong, Bo~Yang, Boji Shan, Chang Liu, Cheng Zhu,
  Chunhao Zhang, Congchao Guo, Da~Chen, Dong Li, Enwei Jiao, Gengxin Li, Guojun
  Zhang, Haohai Sun, Houze Dong, Jiadai Zhu, Jiaqi Zhuang, Jiayuan Song, and 71
  others. 2025.
\newblock \href {https://arxiv.org/abs/2501.08313} {Minimax-01: Scaling
  foundation models with lightning attention}.
\newblock \emph{Preprint}, arXiv:2501.08313.

\bibitem[{Ociepa et~al.(2025)Ociepa, Łukasz Flis, Wróbel, Gwoździej, and
  Kinas}]{bielik}
Krzysztof Ociepa, Łukasz Flis, Krzysztof Wróbel, Adrian Gwoździej, and
  Remigiusz Kinas. 2025.
\newblock \href {https://arxiv.org/abs/2505.02410} {Bielik 11b v2 technical
  report}.
\newblock \emph{Preprint}, arXiv:2505.02410.

\bibitem[{Ondrejov{\'a} and {\v{S}}uppa(2024)}]{slovakcopa}
Vikt{\'o}ria Ondrejov{\'a} and Marek {\v{S}}uppa. 2024.
\newblock \href {https://doi.org/10.18653/v1/2024.vardial-1.11} {Can {LLM}s
  handle low-resource dialects? a case study on translation and common sense
  reasoning in {\v{s}}ari{\v{s}}}.
\newblock In \emph{Proceedings of the Eleventh Workshop on NLP for Similar
  Languages, Varieties, and Dialects (VarDial 2024)}, pages 130--139, Mexico
  City, Mexico. Association for Computational Linguistics.

\bibitem[{Ondrejov{\'{a}} and Suppa(2024)}]{slovaksum2024}
Viktoria Ondrejov{\'{a}} and Marek Suppa. 2024.
\newblock \href {https://aclanthology.org/2024.lrec-main.1298} {{SlovakSum:}
  {A} large scale {S}lovak summarization dataset}.
\newblock In \emph{Proceedings of the 2024 Joint International Conference on
  Computational Linguistics, Language Resources and Evaluation, {LREC/COLING}
  2024, 20-25 May, 2024, Torino, Italy}, pages 14916--14922. {ELRA} and {ICCL}.

\bibitem[{OpenAI et~al.(2025)OpenAI, :, Agarwal, Ahmad, Ai, Altman, Applebaum,
  Arbus, Arora, Bai, Baker, Bao, Barak, Bennett, Bertao, Brett, Brevdo,
  Brockman, Bubeck, Chang, Chen, Chen, Cheung, Clark, Cook, Dukhan, Dvorak,
  Fives, Fomenko, Garipov, Georgiev, Glaese, Gogineni, Goucher, Gross, Guzman,
  Hallman, Hehir, Heidecke, Helyar, Hu, Huet, Huh, Jain, Johnson, Koch, Kofman,
  Kundel, Kwon, Kyrylov, Le, Leclerc, Lennon, Lessans, Lezcano-Casado, Li, Li,
  Lin, Liss, Lily, Liu, Liu, Lu, Lu, Martinovic, McCallum, McGrath, McKinney,
  McLaughlin, Mei, Mostovoy, Mu, Myles, Neitz, Nichol, Pachocki, Paino, Palmie,
  Pantuliano, Parascandolo, Park, Pathak, Paz, Peran, Pimenov, Pokrass, Proehl,
  Qiu, Raila, Raso, Ren, Richardson, Robinson, Rotsted, Salman, Sanjeev,
  Schwarzer, Sculley, Sikchi, Simon, Singhal, Song, Stuckey, Sun, Tillet,
  Toizer, Tsimpourlas, Vyas, Wallace, Wang, Wang, Watkins, Weil, Wendling,
  Whinnery, Whitney, Wong, Yang, Yang, Yasunaga, Ying, Zaremba, Zhan, Zhang,
  Zhang, Zhang, and Zhao}]{gptoss}
OpenAI, :, Sandhini Agarwal, Lama Ahmad, Jason Ai, Sam Altman, Andy Applebaum,
  Edwin Arbus, Rahul~K. Arora, Yu~Bai, Bowen Baker, Haiming Bao, Boaz Barak,
  Ally Bennett, Tyler Bertao, Nivedita Brett, Eugene Brevdo, Greg Brockman,
  Sebastien Bubeck, and 108 others. 2025.
\newblock \href {https://arxiv.org/abs/2508.10925} {gpt-oss-120b \& gpt-oss-20b
  model card}.
\newblock \emph{Preprint}, arXiv:2508.10925.

\bibitem[{{OpenAI}(2026)}]{gpt5}
{OpenAI}. 2026.
\newblock {GPT-5.5 System Card}.
\newblock \url{https://deploymentsafety.openai.com/gpt-5-5}.
\newblock Accessed: 2026-05-10.

\bibitem[{Pecar et~al.(2019)Pecar, Simko, and
  Bielikov{\'{a}}}]{pecar-simko-2019-sentiment}
Samuel Pecar, Mari{\'{a}}n Simko, and M{\'{a}}ria Bielikov{\'{a}}. 2019.
\newblock \href {https://doi.org/10.18653/V1/W19-3716} {Improving sentiment
  classification in {S}lovak language}.
\newblock In \emph{Proceedings of the 7th Workshop on Balto-Slavic Natural
  Language Processing, BSNLP@ACL 2019, Florence, Italy, August 2, 2019}, pages
  114--119. Association for Computational Linguistics.

\bibitem[{Pikuliak et~al.(2022)Pikuliak, Grivalsk{\'y}, Kon{\^{o}}pka,
  Bl{\v{s}}t{\'a}k, Tamajka, Bachrat{\'{y}}, Simko, Bal{\'{a}}{\v{z}}ik, Trnka,
  and Uhl{\'{a}}rik}]{slovakbert2022}
Mat{\'{u}}{\v{s}} Pikuliak, {\v{S}}tefan Grivalsk{\'y}, Martin Kon{\^{o}}pka,
  Miroslav Bl{\v{s}}t{\'a}k, Martin Tamajka, Viktor Bachrat{\'{y}},
  Mari{\'{a}}n Simko, Pavol Bal{\'{a}}{\v{z}}ik, Michal Trnka, and Filip
  Uhl{\'{a}}rik. 2022.
\newblock \href {https://doi.org/10.18653/V1/2022.FINDINGS-EMNLP.530}
  {{SlovakBERT}: {S}lovak masked language model}.
\newblock In \emph{Findings of the Association for Computational Linguistics:
  {EMNLP} 2022, Abu Dhabi, United Arab Emirates, December 7-11, 2022}, pages
  7156--7168. Association for Computational Linguistics.

\bibitem[{Ponti et~al.(2020)Ponti, Glava{\v{s}}, Majewska, Liu, Vuli{\'c}, and
  Korhonen}]{xcopa2020}
Edoardo~Maria Ponti, Goran Glava{\v{s}}, Olga Majewska, Qianchu Liu, Ivan
  Vuli{\'c}, and Anna Korhonen. 2020.
\newblock \href {https://doi.org/10.18653/v1/2020.emnlp-main.185} {{XCOPA}: A
  multilingual dataset for causal commonsense reasoning}.
\newblock In \emph{Proceedings of the 2020 Conference on Empirical Methods in
  Natural Language Processing (EMNLP)}, pages 2362--2376, Online. Association
  for Computational Linguistics.

\bibitem[{Popovi{\'c}(2015)}]{chrf2015}
Maja Popovi{\'c}. 2015.
\newblock \href {https://doi.org/10.18653/v1/W15-3049} {chr{F}: character
  n-gram {F}-score for automatic {MT} evaluation}.
\newblock In \emph{Proceedings of the Tenth Workshop on Statistical Machine
  Translation}, pages 392--395, Lisbon, Portugal. Association for Computational
  Linguistics.

\bibitem[{Popovi{\'c}(2017)}]{chrfpp2017}
Maja Popovi{\'c}. 2017.
\newblock \href {https://doi.org/10.18653/v1/W17-4770} {chr{F}++: words helping
  character n-grams}.
\newblock In \emph{Proceedings of the Second Conference on Machine
  Translation}, pages 612--618, Copenhagen, Denmark. Association for
  Computational Linguistics.

\bibitem[{Romanou et~al.(2025)Romanou, Foroutan, Sotnikova, Nelaturu, Singh,
  Maheshwary, Altomare, Chen, Haggag, A., Amayuelas, Amirudin, Boiko, Chang,
  Chim, Cohen, Dalmia, Diress, Duwal, Dzenhaliou, Florez, Farestam, Imperial,
  Islam, Isotalo, Jabbarishiviari, Karlsson, Khalilov, Klamm, Koto,
  Krzemi{\'{n}}ski, de~Melo, Montariol, Nan, Niklaus, Novikova, Obando{-}Ceron,
  Paul, Ploeger, Purbey, Rajwal, Ravi, Rydell, Santhosh, Sharma, Skenduli,
  Moakhar, Moakhar, Tarun, Wasi, Weerasinghe, Yilmaz, Zhang, Schlag, Fadaee,
  Hooker, and Bosselut}]{include2024}
Angelika Romanou, Negar Foroutan, Anna Sotnikova, Sree~Harsha Nelaturu,
  Shivalika Singh, Rishabh Maheshwary, Micol Altomare, Zeming Chen, Mohamed~A.
  Haggag, Snegha A., Alfonso Amayuelas, Azril~Hafizi Amirudin, Danylo Boiko,
  Michael Chang, Jenny Chim, Gal Cohen, Aditya~Kumar Dalmia, Abraham Diress,
  Sharad Duwal, and 38 others. 2025.
\newblock \href
  {https://proceedings.iclr.cc/paper_files/paper/2025/hash/ced46a50befedcb884ccf0cbe8c3ad23-Abstract-Conference.html}
  {{INCLUDE:} evaluating multilingual language understanding with regional
  knowledge}.
\newblock In \emph{The Thirteenth International Conference on Learning
  Representations, {ICLR} 2025, Singapore, April 24-28, 2025}. OpenReview.net.

\bibitem[{Rybak et~al.(2020)Rybak, Mroczkowski, Tracz, and Gawlik}]{klej2020}
Piotr Rybak, Robert Mroczkowski, Janusz Tracz, and Ireneusz Gawlik. 2020.
\newblock \href {https://doi.org/10.18653/V1/2020.ACL-MAIN.111} {{KLEJ:}
  comprehensive benchmark for {P}olish language understanding}.
\newblock In \emph{Proceedings of the 58th Annual Meeting of the Association
  for Computational Linguistics, {ACL} 2020, Online, July 5-10, 2020}, pages
  1191--1201. Association for Computational Linguistics.

\bibitem[{Sainz et~al.(2024)Sainz, Garc{\'i}a-Ferrero, Jacovi, Ander~Campos,
  Elazar, Agirre, Goldberg et~al.}]{conda2024}
Oscar Sainz, Iker Garc{\'i}a-Ferrero, Alon Jacovi, Jon Ander~Campos, Yanai
  Elazar, Eneko Agirre, Yoav Goldberg, and 1 others. 2024.
\newblock \href {https://doi.org/10.18653/v1/2024.conda-1.4} {Data
  contamination report from the 2024 {CONDA} shared task}.
\newblock In \emph{Proceedings of the 1st Workshop on Data Contamination
  (CONDA)}, pages 41--56, Bangkok, Thailand. Association for Computational
  Linguistics.

\bibitem[{Shavrina et~al.(2020)Shavrina, Fenogenova, Emelyanov, Shevelev,
  Artemova, Malykh, Mikhailov, Tikhonova, Chertok, and Evlampiev}]{rsglue2020}
Tatiana Shavrina, Alena Fenogenova, Anton~A. Emelyanov, Denis Shevelev,
  Ekaterina Artemova, Valentin Malykh, Vladislav Mikhailov, Maria Tikhonova,
  Andrey Chertok, and Andrey Evlampiev. 2020.
\newblock \href {https://doi.org/10.18653/V1/2020.EMNLP-MAIN.381}
  {{RussianSuperGLUE:} {A} {R}ussian language understanding evaluation
  benchmark}.
\newblock In \emph{Proceedings of the 2020 Conference on Empirical Methods in
  Natural Language Processing, {EMNLP} 2020, Online, November 16-20, 2020},
  pages 4717--4726. Association for Computational Linguistics.

\bibitem[{Singh et~al.(2025)Singh, Romanou, Fourrier, Adelani, Ngui,
  Vila-Suero, Limkonchotiwat, Marchisio, Leong, Susanto, Ng, Longpre, Ruder,
  Ko, Bosselut, Oh, Martins, Choshen, Ippolito, Ferrante, Fadaee, Ermis, and
  Hooker}]{globalmmlu2024}
Shivalika Singh, Angelika Romanou, Cl{\'e}mentine Fourrier, David~Ifeoluwa
  Adelani, Jian~Gang Ngui, Daniel Vila-Suero, Peerat Limkonchotiwat, Kelly
  Marchisio, Wei~Qi Leong, Yosephine Susanto, Raymond Ng, Shayne Longpre,
  Sebastian Ruder, Wei-Yin Ko, Antoine Bosselut, Alice Oh, Andre Martins,
  Leshem Choshen, Daphne Ippolito, and 4 others. 2025.
\newblock \href {https://doi.org/10.18653/v1/2025.acl-long.919} {Global {MMLU}:
  Understanding and addressing cultural and linguistic biases in multilingual
  evaluation}.
\newblock In \emph{Proceedings of the 63rd Annual Meeting of the Association
  for Computational Linguistics (Volume 1: Long Papers)}, pages 18761--18799,
  Vienna, Austria. Association for Computational Linguistics.

\bibitem[{Smart(2025)}]{multiwikiqa2025}
Dan~Saattrup Smart. 2025.
\newblock \href {https://arxiv.org/abs/2509.04111} {Multiwikiqa: A reading
  comprehension benchmark in 300+ languages}.
\newblock \emph{Preprint}, arXiv:2509.04111.

\bibitem[{{SNK}(2025)}]{snk2025}
{SNK}. 2025.
\newblock {Slovak National Corpus} ({prim-11.0}).
\newblock \url{https://korpus.sk}.
\newblock Accessed: 2025-04-11. \emph{Ľ. Štúr Institute of Linguistics,
  Slovak Academy of Sciences}.

\bibitem[{Sokolov\'{a} et~al.(2025)Sokolov\'{a}, Harahus, Hl\'{a}dek, and
  Sta\v{s}}]{toxic_sk}
Zuzana Sokolov\'{a}, Maro\v{s} Harahus, Daniel Hl\'{a}dek, and J\'{a}n
  Sta\v{s}. 2025.
\newblock \href {https://doi.org/10.2478/jazcas-2025-0025} {Annotated slovak
  datasets for toxicity, hate speech, and sentiment analysis}.
\newblock \emph{Journal of Linguistics/Jazykovedn\'{y} \v{c}asopis},
  76(1):279--289.

\bibitem[{Srivastava et~al.(2023)Srivastava, Rastogi, Rao, Shoeb, Abid, Fisch,
  Brown, Santoro, Gupta, Garriga{-}Alonso et~al.}]{bigbench2023}
Aarohi Srivastava, Abhinav Rastogi, Abhishek Rao, Abu Awal~Md Shoeb, Abubakar
  Abid, Adam Fisch, Adam~R. Brown, Adam Santoro, Aditya Gupta, Adri{\`{a}}
  Garriga{-}Alonso, and 1 others. 2023.
\newblock \href {https://openreview.net/forum?id=uyTL5Bvosj} {Beyond the
  imitation game: Quantifying and extrapolating the capabilities of language
  models}.
\newblock \emph{Trans. Mach. Learn. Res.}, 2023.

\bibitem[{{\v{S}}tef{\'a}nik et~al.(2023){\v{S}}tef{\'a}nik, Kadl{\v{c}}{\'i}k,
  Gramacki, and Sojka}]{csfd}
Michal {\v{S}}tef{\'a}nik, Marek Kadl{\v{c}}{\'i}k, Piotr Gramacki, and Petr
  Sojka. 2023.
\newblock \href {https://doi.org/10.18653/v1/2023.bsnlp-1.12} {Resources and
  few-shot learners for in-context learning in {S}lavic languages}.
\newblock In \emph{Proceedings of the 9th Workshop on Slavic Natural Language
  Processing 2023 (SlavicNLP 2023)}, pages 94--105, Dubrovnik, Croatia.
  Association for Computational Linguistics.

\bibitem[{Suba et~al.(2023)Suba, Suppa, Kubik, Hamerlik, and
  Takac}]{wikigoldsk2023}
David Suba, Marek Suppa, Jozef Kubik, Endre Hamerlik, and Martin Takac. 2023.
\newblock \href {https://doi.org/10.18653/v1/2023.bsnlp-1.16}
  {{W}iki{G}old{SK}: Annotated dataset, baselines and few-shot learning
  experiments for {S}lovak named entity recognition}.
\newblock In \emph{Proceedings of the 9th Workshop on Slavic Natural Language
  Processing 2023 (SlavicNLP 2023)}, pages 138--145, Dubrovnik, Croatia.
  Association for Computational Linguistics.

\bibitem[{Suppa and Adamec(2020)}]{smesum2020}
Marek Suppa and Jergus Adamec. 2020.
\newblock \href {https://aclanthology.org/2020.lrec-1.830/} {A summarization
  dataset of {S}lovak news articles}.
\newblock In \emph{Proceedings of The 12th Language Resources and Evaluation
  Conference, {LREC} 2020, Marseille, France, May 11-16, 2020}, pages
  6725--6730. European Language Resources Association.

\bibitem[{Suppa et~al.(2025)Suppa, Ridzik, Hl{\'{a}}dek, Jav{\r{u}}rek,
  Ondrejov{\'{a}}, S{\'{a}}sikov{\'{a}}, Tamajka, and {\v{S}}imko}]{sklep2025}
Marek Suppa, Andrej Ridzik, Daniel Hl{\'{a}}dek, Tom{\'{a}}{\v{s}}
  Jav{\r{u}}rek, Viktoria Ondrejov{\'{a}}, Krist{\'{\i}}na
  S{\'{a}}sikov{\'{a}}, Martin Tamajka, and Mari{\'{a}}n {\v{S}}imko. 2025.
\newblock \href {https://doi.org/10.18653/v1/2025.findings-acl.1371} {{skLEP:}
  {A} {S}lovak general language understanding benchmark}.
\newblock In \emph{Findings of the Association for Computational Linguistics,
  {ACL} 2025, Vienna, Austria, July 27 - August 1, 2025}, pages 26716--26743.
  Association for Computational Linguistics.

\bibitem[{{\v S}uppa et~al.(2026){\v S}uppa, Ridzik, Hl{\'a}dek, K{\v n}a{\v
  z}ekov{\'a}, and Ondrejov{\'a}}]{skmteb2026}
Marek {\v S}uppa, Andrej Ridzik, Daniel Hl{\'a}dek, Nat{\'a}lia K{\v n}a{\v
  z}ekov{\'a}, and Vikt{\'o}ria Ondrejov{\'a}. 2026.
\newblock \href {https://doi.org/10.18653/v1/2026.acl-long.2114} {{SkMTEB}:
  {S}lovak massive text embedding benchmark and model adaptation}.
\newblock In \emph{Proceedings of the 64th Annual Meeting of the Association
  for Computational Linguistics (Volume 1: Long Papers)}, pages 45597--45628.
  Association for Computational Linguistics.

\bibitem[{Taori et~al.(2023)Taori, Gulrajani, Zhang, Dubois, Li, Guestrin,
  Liang, and Hashimoto}]{alpaca2023}
Rohan Taori, Ishaan Gulrajani, Tianyi Zhang, Yann Dubois, Xuechen Li, Carlos
  Guestrin, Percy Liang, and Tatsunori~B. Hashimoto. 2023.
\newblock Stanford alpaca: An instruction-following {LL}a{MA} model.
\newblock \url{https://github.com/tatsu-lab/stanford_alpaca}.

\bibitem[{Team et~al.(2025{\natexlab{a}})Team, Kamath, Ferret, Pathak,
  Vieillard, Merhej, Perrin, Matejovicova, Ramé, Rivière, Rouillard, Mesnard,
  Cideron, bastien Grill, Ramos, Yvinec, Casbon, Pot, Penchev, Liu, Visin,
  Kenealy, Beyer, Zhai, Tsitsulin, Busa-Fekete, Feng, Sachdeva, Coleman, Gao,
  Mustafa, Barr, Parisotto, Tian, Eyal, Cherry, Peter, Sinopalnikov,
  Bhupatiraju, Agarwal, Kazemi, Malkin, Kumar, Vilar, Brusilovsky, Luo,
  Steiner, Friesen, Sharma, Sharma, Gilady, Goedeckemeyer, Saade, Feng,
  Kolesnikov, Bendebury, Abdagic, Vadi, György, Pinto, Das, Bapna, Miech,
  Yang, Paterson, Shenoy, Chakrabarti, Piot, Wu, Shahriari, Petrini, Chen, Lan,
  Choquette-Choo, Carey, Brick, Deutsch, Eisenbud, Cattle, Cheng, Paparas,
  Sreepathihalli, Reid, Tran, Zelle, Noland, Huizenga, Kharitonov, Liu,
  Amirkhanyan, Cameron, Hashemi, Klimczak-Plucińska, Singh, Mehta, Lehri,
  Hazimeh, Ballantyne, Szpektor, Nardini, Pouget-Abadie, Chan, Stanton,
  Wieting, Lai, Orbay, Fernandez, Newlan, yeong Ji, Singh, Black, Yu, Hui,
  Vodrahalli, Greff, Qiu, Valentine, Coelho, Ritter, Hoffman, Watson,
  Chaturvedi, Moynihan, Ma, Babar, Noy, Byrd, Roy, Momchev, Chauhan, Sachdeva,
  Bunyan, Botarda, Caron, Rubenstein, Culliton, Schmid, Sessa, Xu, Stanczyk,
  Tafti, Shivanna, Wu, Pan, Rokni, Willoughby, Vallu, Mullins, Jerome, Smoot,
  Girgin, Iqbal, Reddy, Sheth, Põder, Bhatnagar, Panyam, Eiger, Zhang, Liu,
  Yacovone, Liechty, Kalra, Evci, Misra, Roseberry, Feinberg, Kolesnikov, Han,
  Kwon, Chen, Chow, Zhu, Wei, Egyed, Cotruta, Giang, Kirk, Rao, Black, Babar,
  Lo, Moreira, Martins, Sanseviero, Gonzalez, Gleicher, Warkentin, Mirrokni,
  Senter, Collins, Barral, Ghahramani, Hadsell, Matias, Sculley, Petrov,
  Fiedel, Shazeer, Vinyals, Dean, Hassabis, Kavukcuoglu, Farabet, Buchatskaya,
  Alayrac, Anil, Dmitry, Lepikhin, Borgeaud, Bachem, Joulin, Andreev, Hardin,
  Dadashi, and Hussenot}]{gemma3}
Gemma Team, Aishwarya Kamath, Johan Ferret, Shreya Pathak, Nino Vieillard,
  Ramona Merhej, Sarah Perrin, Tatiana Matejovicova, Alexandre Ramé, Morgane
  Rivière, Louis Rouillard, Thomas Mesnard, Geoffrey Cideron, Jean bastien
  Grill, Sabela Ramos, Edouard Yvinec, Michelle Casbon, Etienne Pot, Ivo
  Penchev, and 197 others. 2025{\natexlab{a}}.
\newblock \href {https://arxiv.org/abs/2503.19786} {Gemma 3 technical report}.
\newblock \emph{Preprint}, arXiv:2503.19786.

\bibitem[{Team et~al.(2025{\natexlab{b}})Team, Zeng, Lv, Zheng, Hou, Chen, Xie,
  Wang, Yin, Zeng, Zhang, Wang, Zhong, Liu, Lu, Cao, Zhang, Huang, Wei, Cheng,
  An, Niu, Wen, Bai, Du, Wang, Zhu, Zhang, Wen, Wu, Xu, Huang, Zhao, Cai, Yu,
  Li, Ge, Huang, Zhang, Xu, Zhu, Li, Yin, Lin, Yang, Jiang, Ai, Zhu, Wang, Pan,
  Wang, Sun, Li, Li, Hu, Zhang, Peng, Tai, Zhang, Wang, Yang, Liu, Zhao, Liu,
  Yan, Liu, Chen, Li, Zhao, Ren, Jiao, Zhao, Yan, Wang, Gui, Zhao, Liu, Li, Li,
  Lu, Wang, Yuan, Li, Du, Du, Liu, Zhi, Gao, Wang, Yang, Xu, Fan, Wu, Ding,
  Wang, Zhang, Li, Xu, Zhao, Zhai, Du, Dong, Lei, Tu, Yang, Lu, Li, Li,
  Shuang-Li, Yang, Yi, Yu, Tian, Wang, Yu, Tam, Liang, Liu, Wang, Jia, Gu,
  Ling, Wang, Fan, Pan, Zhang, Zhang, Fu, Zhang, Xu, Wu, Lu, Wang, Zhou, Pan,
  Zhang, Wang, Li, Su, Geng, Zhu, Yang, Li, Wu, Li, Liu, Wang, Li, Zhang, Liu,
  Yang, Zhou, Qiao, Feng, Liu, Zhang, Wang, Yao, Wang, Liu, Chai, Li, Zhao,
  Chen, Zhai, Xu, Huang, Wang, Li, Dong, and Tang}]{glm}
GLM-4.5 Team, Aohan Zeng, Xin Lv, Qinkai Zheng, Zhenyu Hou, Bin Chen, Chengxing
  Xie, Cunxiang Wang, Da~Yin, Hao Zeng, Jiajie Zhang, Kedong Wang, Lucen Zhong,
  Mingdao Liu, Rui Lu, Shulin Cao, Xiaohan Zhang, Xuancheng Huang, Yao Wei, and
  152 others. 2025{\natexlab{b}}.
\newblock \href {https://arxiv.org/abs/2508.06471} {Glm-4.5: Agentic,
  reasoning, and coding (arc) foundation models}.
\newblock \emph{Preprint}, arXiv:2508.06471.

\bibitem[{Team et~al.(2025{\natexlab{c}})Team, Bai, Bao, Charles, Chen, Chen,
  Chen, Chen, Chen, Chen, Chen, Chen, Chen, Chen, Chen, Cui, Ding, Dong, Du,
  Du, Du, Du, Fan, Feng, Fu, Gao, Gao, Gao, Gao, Gao, Ge, Geng, Gu, Gu, Guan,
  Guo, Guo, Hao, He, He, He, He, Hong, Hu, Hu, Hu, Huang, Huang, Huang, Jiang,
  Jiang, Jin, Kang, Lai, Li, Li, Li, Li, Li, Li, Li, Li, Li, Li, Lin, Lin, Lin,
  Liu, Liu, Liu, Liu, Liu, Liu, Liu, Liu, Liu, Liu, Liu, Liu, Liu, Liu, Liu,
  Lu, Lu, Lu, Luo, Ma, Ma, Ma, Mao, Mei, Men, Miao, Pan, Peng, Qin, Qin, Qu,
  Shang, Shi, Shi, Song, Su, Su, Sui, Sun, Sung, Tai, Tang, Tao, Teng, Tian,
  Wang, Wang, Wang, Wang, Wang, Wang, Wang, Wang, Wang, Wang, Wang, Wang, Wang,
  Wang, Wang, Wang, Wang, Wang, Wang, Wang, Wang, Wei, Wei, Wu, Wu, Wu, Wu,
  Xiao, Xie, Xie, Xiong, Xu, Xu, Xu, Xu, Xu, Xu, Xu, Xu, Xu, Xu, Xu, Yan, Yan,
  Yang, Yang, Yang, Yang, Yang, Yang, Yang, Yao, Yao, Ye, Ye, Yin, Yu, Yuan,
  Yuan, Yuan, Yuan, Zhan, Zhang, Zhang, Zhang, Zhang, Zhang, Zhang, Zhang,
  Zhang, Zhang, Zhang, Zhang, Zhang, Zhang, Zhao, Zhao, Zhao, Zheng, Zheng,
  Zhong, Zhou, Zhou, Zhou, Zhu, Zhu, Zhuang, and Zu}]{kimik2}
Kimi Team, Yifan Bai, Yiping Bao, Y.~Charles, Cheng Chen, Guanduo Chen, Haiting
  Chen, Huarong Chen, Jiahao Chen, Ningxin Chen, Ruijue Chen, Yanru Chen,
  Yuankun Chen, Yutian Chen, Zhuofu Chen, Jialei Cui, Hao Ding, Mengnan Dong,
  Angang Du, and 181 others. 2025{\natexlab{c}}.
\newblock \href {https://arxiv.org/abs/2507.20534} {Kimi k2: Open agentic
  intelligence}.
\newblock \emph{Preprint}, arXiv:2507.20534.

\bibitem[{Upadhayay and Behzadan(2023)}]{taco2024}
Bibek Upadhayay and Vahid Behzadan. 2023.
\newblock \href {https://arxiv.org/abs/2311.10797} {Taco: Enhancing
  cross-lingual transfer for low-resource languages in llms through
  translation-assisted chain-of-thought processes}.
\newblock \emph{Preprint}, arXiv:2311.10797.

\bibitem[{Wang et~al.(2019{\natexlab{a}})Wang, Pruksachatkun, Nangia, Singh,
  Michael, Hill, Levy, and Bowman}]{superglue2019}
Alex Wang, Yada Pruksachatkun, Nikita Nangia, Amanpreet Singh, Julian Michael,
  Felix Hill, Omer Levy, and Samuel~R. Bowman. 2019{\natexlab{a}}.
\newblock \href
  {https://proceedings.neurips.cc/paper/2019/hash/4496bf24afe7fab6f046bf4923da8de6-Abstract.html}
  {Superglue: {A} stickier benchmark for general-purpose language understanding
  systems}.
\newblock In \emph{Advances in Neural Information Processing Systems 32: Annual
  Conference on Neural Information Processing Systems 2019, NeurIPS 2019,
  December 8-14, 2019, Vancouver, BC, Canada}, pages 3261--3275.

\bibitem[{Wang et~al.(2019{\natexlab{b}})Wang, Singh, Michael, Hill, Levy, and
  Bowman}]{glue2018}
Alex Wang, Amanpreet Singh, Julian Michael, Felix Hill, Omer Levy, and
  Samuel~R. Bowman. 2019{\natexlab{b}}.
\newblock \href {https://openreview.net/forum?id=rJ4km2R5t7} {{GLUE:} {A}
  multi-task benchmark and analysis platform for natural language
  understanding}.
\newblock In \emph{7th International Conference on Learning Representations,
  {ICLR} 2019, New Orleans, LA, USA, May 6-9, 2019}. OpenReview.net.

\bibitem[{Yang et~al.(2025)Yang, Li, Yang, Zhang, Hui, Zheng, Yu, Gao, Huang,
  Lv, Zheng, Liu, Zhou, Huang, Hu, Ge, Wei, Lin, Tang, Yang, Tu, Zhang, Yang,
  Yang, Zhou, Zhou, Lin, Dang, Bao, Yang, Yu, Deng, Li, Xue, Li, Zhang, Wang,
  Zhu, Men, Gao, Liu, Luo, Li, Tang, Yin, Ren, Wang, Zhang, Ren, Fan, Su,
  Zhang, Zhang, Wan, Liu, Wang, Cui, Zhang, Zhou, and Qiu}]{qwen3}
An~Yang, Anfeng Li, Baosong Yang, Beichen Zhang, Binyuan Hui, Bo~Zheng, Bowen
  Yu, Chang Gao, Chengen Huang, Chenxu Lv, Chujie Zheng, Dayiheng Liu, Fan
  Zhou, Fei Huang, Feng Hu, Hao Ge, Haoran Wei, Huan Lin, Jialong Tang, and 41
  others. 2025.
\newblock \href {https://arxiv.org/abs/2505.09388} {Qwen3 technical report}.
\newblock \emph{Preprint}, arXiv:2505.09388.

\bibitem[{Yong et~al.(2023)Yong, Schoelkopf, Muennighoff, Aji, Adelani,
  Almubarak, Bari, Sutawika, Kasai, Baruwa, Winata, Biderman, Raff, Radev, and
  Nikoulina}]{bloomplus2023}
Zheng~Xin Yong, Hailey Schoelkopf, Niklas Muennighoff, Alham~Fikri Aji,
  David~Ifeoluwa Adelani, Khalid Almubarak, M~Saiful Bari, Lintang Sutawika,
  Jungo Kasai, Ahmed Baruwa, Genta Winata, Stella Biderman, Edward Raff,
  Dragomir Radev, and Vassilina Nikoulina. 2023.
\newblock \href {https://doi.org/10.18653/v1/2023.acl-long.653} {{BLOOM}+1:
  Adding language support to {BLOOM} for zero-shot prompting}.
\newblock In \emph{Proceedings of the 61st Annual Meeting of the Association
  for Computational Linguistics (Volume 1: Long Papers)}, pages 11682--11703,
  Toronto, Canada. Association for Computational Linguistics.

\bibitem[{Zahradn{\'i}k and Suppa(2025)}]{trojsten2025}
Adam Zahradn{\'i}k and Marek Suppa. 2025.
\newblock \href {https://doi.org/10.18653/v1/2025.emnlp-main.1779} {Trojsten
  benchmark: Evaluating {LLM} problem-solving in {S}lovak {STEM} competition
  problems}.
\newblock In \emph{Proceedings of the 2025 Conference on Empirical Methods in
  Natural Language Processing}, pages 35106--35121, Suzhou, China. Association
  for Computational Linguistics.

\bibitem[{Zhou et~al.(2023)Zhou, Lu, Mishra, Brahma, Basu, Luan, Zhou, and
  Hou}]{ifeval2023}
Jeffrey Zhou, Tianjian Lu, Swaroop Mishra, Siddhartha Brahma, Sujoy Basu,
  Yi~Luan, Denny Zhou, and Le~Hou. 2023.
\newblock \href {https://arxiv.org/abs/2311.07911} {Instruction-following
  evaluation for large language models}.
\newblock \emph{Preprint}, arXiv:2311.07911.

\end{thebibliography}

\appendix

\section{Lessons for Under-Resourced-Language Benchmark Builders}
\label{app:lessons}

This appendix expands the four lessons of \S\ref{sec:conclusion}, each
with the Slovak result and per-model table behind it, and ends with two
recommendations for benchmark builders.

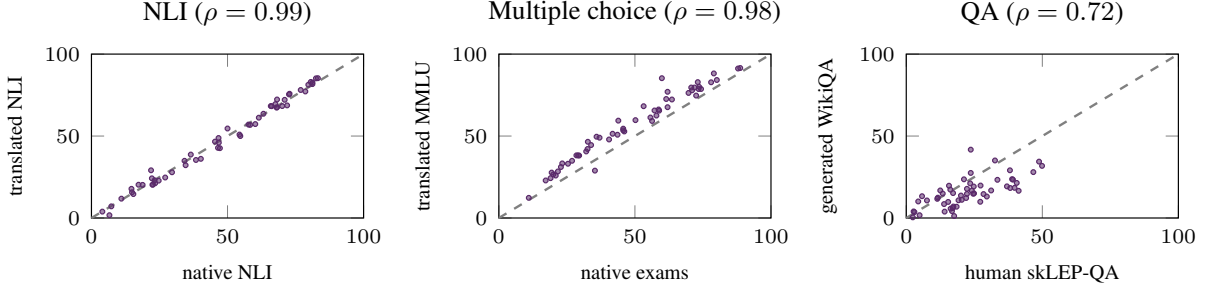
\begin{figure*}[t]
\centering
\resizebox{\textwidth}{!}{%
\begin{tikzpicture}
\begin{groupplot}[
  group style={group size=3 by 1, horizontal sep=1.6cm},
  width=0.30\textwidth, height=0.22\textwidth,
  xmin=0, xmax=100, ymin=0, ymax=100,
  xtick={0,50,100}, ytick={0,50,100},
  tick label style={font=\scriptsize},
  title style={font=\footnotesize}, label style={font=\scriptsize},
  scatter/use mapped color={draw=promptframe, fill=promptframe},
]
\nextgroupplot[title={NLI ($\rho=0.99$)}, xlabel={native NLI}, ylabel={translated NLI}]
\addplot[domain=0:100, dashed, gray, thick, forget plot] {x};
\addplot[only marks, mark=*, mark size=0.8pt, draw=promptframe, fill=promptframe, fill opacity=0.55] coordinates {
  (67.6,69.5) (36.5,38.7) (17.3,20.3) (61.5,61.2) (63.2,63.6) (72.6,75.4) (23.4,21.2) (14.9,15.7) (50.0,54.6) (24.7,22.8) (70.1,68.3) (45.4,46.6) (78.6,77.2) (65.9,68.3) (29.6,27.9) (40.2,36.1) (46.5,42.8) (6.6,1.7) (83.2,85.3) (80.6,83.0) (81.2,81.7) (82.5,85.2) (22.2,24.2) (81.2,82.6) (11.0,11.8) (34.2,34.9) (7.3,7.1) (27.1,24.7) (4.0,3.9) (46.7,48.8) (34.6,32.2) (23.2,23.0) (68.2,72.3) (58.3,56.8) (15.4,14.6) (54.7,50.0) (46.8,46.0) (71.9,68.7) (60.2,57.3) (18.8,20.2) (68.0,67.4) (38.4,35.4) (22.4,20.3) (68.2,67.5) (79.7,81.3) (72.8,75.8) (71.0,72.1) (21.9,29.1) (22.6,20.4) (66.2,68.2) (47.2,42.6) (14.7,17.8) (54.4,50.9) (57.7,57.0) (76.9,78.1)
};
\nextgroupplot[title={Multiple choice ($\rho=0.98$)}, xlabel={native exams}, ylabel={translated MMLU}]
\addplot[domain=0:100, dashed, gray, thick, forget plot] {x};
\addplot[only marks, mark=*, mark size=0.8pt, draw=promptframe, fill=promptframe, fill opacity=0.55] coordinates {
  (43.9,59.4) (32.7,46.5) (53.2,68.2) (37.0,49.0) (35.9,49.6) (73.1,82.9) (20.9,25.9) (22.8,31.1) (58.9,65.5) (25.4,33.1) (73.6,80.0) (33.9,44.5) (78.0,82.8) (62.0,67.6) (71.9,79.6) (69.7,76.3) (41.8,51.4) (32.6,42.1) (79.0,88.2) (59.9,85.3) (80.1,84.2) (88.7,91.5) (21.7,28.4) (70.7,79.6) (29.1,38.2) (28.5,38.2) (19.4,27.7) (32.1,40.6) (20.0,26.6) (57.2,65.6) (40.5,47.9) (29.5,38.1) (63.6,72.3) (57.9,62.5) (35.3,28.9) (50.2,59.7) (46.1,52.8) (62.0,77.0) (73.4,78.7) (11.0,12.3) (58.7,66.2) (43.6,50.8) (26.9,34.9) (72.5,74.7) (87.9,91.2) (71.0,77.9) (61.5,72.6) (23.3,33.3) (17.2,22.9) (55.7,61.3) (45.7,53.5) (19.0,24.3) (56.3,59.2) (45.7,54.5) (74.3,78.7)
};
\nextgroupplot[title={QA ($\rho=0.72$)}, xlabel={human skLEP-QA}, ylabel={generated WikiQA}]
\addplot[domain=0:100, dashed, gray, thick, forget plot] {x};
\addplot[only marks, mark=*, mark size=0.8pt, draw=promptframe, fill=promptframe, fill opacity=0.55] coordinates {
  (22.3,17.8) (12.5,16.8) (2.7,4.0) (11.6,12.4) (11.4,11.7) (40.6,21.4) (17.6,1.2) (17.2,15.2) (33.5,23.3) (13.4,14.9) (41.3,16.6) (23.5,21.1) (27.2,9.9) (39.8,18.4) (48.8,34.4) (38.9,23.7) (23.7,41.7) (4.4,10.1) (31.2,16.8) (22.3,14.7) (20.4,11.1) (49.9,31.8) (20.1,13.7) (29.6,13.1) (14.1,3.9) (13.9,8.5) (3.0,3.6) (17.1,6.1) (4.9,1.7) (24.9,15.0) (32.6,35.1) (21.2,23.6) (38.1,29.1) (16.3,5.7) (2.4,0.4) (15.4,9.8) (19.2,10.8) (24.1,16.7) (39.2,23.4) (5.8,13.3) (7.6,10.7) (18.5,6.9) (16.8,4.2) (37.0,19.3) (38.2,18.3) (28.4,14.6) (24.6,14.8) (23.7,27.5) (15.7,19.7) (17.3,7.0) (24.8,19.2) (16.2,17.3) (22.0,12.2) (27.3,19.8) (46.7,28.0)
};
\end{groupplot}
\end{tikzpicture}%
}
\caption{RQ1 data-provenance contrasts: each point is one of the 55 models, scored on the native source ($x$) and its translated or LLM-generated counterpart ($y$). The dashed line is $y=x$. Points hug the diagonal for NLI and multiple choice (translation preserves the ranking) but scatter off it for QA (human-authored vs.\ generated questions disagree), concentrated among weaker models. Per-tier $\rho$ and top-10 overlap are in Table~\ref{tab:provenance-divergence}.}
\label{fig:provenance-scatter}
\end{figure*}

\paragraph{L1: Spend the native-data budget where translation fails.}
On closed-form skills translation preserves the \emph{coarse} ordering
of models but separates them less well at the frontier, where a
leaderboard matters most. Native and
translated NLI agree on the entire top-10, whereas native and
translated multiple choice agree on only 7 of 10, so translated MMLU
separates the strongest models less reliably than a native exam set
does (\S\ref{sec:rq1}, Table~\ref{tab:provenance-divergence}). Pooled
over all 55 models the two contrasts look equally safe ($\rho=0.99$ for
NLI, $\rho=0.98$ for multiple choice, per-model scatter in
Figure~\ref{fig:provenance-scatter}), but a single correlation over a
pool spanning 350\,M to 1.1\,T parameters is dominated by between-tier
ordering and so overstates how interchangeable the two leaderboards
are. Our frontier bands hold at most six or seven models, so the divergence within them is noisy and only qualitative ($\dagger$ in
Table~\ref{tab:provenance-divergence}). Native data pays off most where
translation is weak or impossible: on
question answering, where human-authored and LLM-generated questions
rank models very differently ($\rho=0.72$), and on grammar and
morphology, which translation cannot produce at all. A native paired
contrast is how a new language finds out \emph{which} skills those are.

\paragraph{L2: Budget an instruction-repair step after language
adaptation.} Continued training is known to degrade previously learned behaviour
\citep{french1999forgetting, luo2023forgetting}, and the cost of
adapting LLMs to new languages is an active research topic
\citep{bloomplus2023, sambalingo2024}. We measure how large this effect
is for a low-resource language and what it costs to repair. Continued pretraining on Slovak
lowers an instruction-tuned model's overall score from $47.4$ to
$33.5$ (Qwen3-14B, \S\ref{sec:rq2}), and the loss is concentrated in
classification ($-30.6$), reading ($-25.2$) and knowledge ($-23.7$)
against a grammar gain ($+6.8$): adaptation helps Slovak grammar but
hurts content and comprehension. Instruction following falls by
only $-6.4$. A
small Slovak instruction set then recovers roughly three quarters of
the loss ($\to\!43.8$, per-base scores in
Table~\ref{tab:alpaca-delta}), but only for bases that retain enough
capability. A repair step should therefore be planned from the start.
\begin{table}[t]
\centering\small
\setlength{\tabcolsep}{4pt}
\resizebox{\columnwidth}{!}{%
\begin{tabular}{lrrr@{\hskip 7pt}rrrr}
\toprule
& \multicolumn{3}{c}{\textbf{Overall}} & \multicolumn{4}{c}{\textbf{$\Delta$ by group}} \\
\cmidrule(lr){2-4}\cmidrule(l){5-8}
\textbf{Lineage} & Base & +Alp. & $\Delta$ & Know. & Class. & Instr. & Gram. \\
\midrule
Mistral-SK 7B & 19.0 & 14.9 & \textbf{-4.0} & -1.1 & -16.9 & -0.4 & +4.9 \\
Qwen3-SK 14B & 33.5 & 43.8 & +10.3 & +26.6 & +16.5 & +13.2 & -19.8 \\
TildeOpen-SK 30B & 31.9 & 32.8 & +0.8 & -0.6 & -29.2 & +11.5 & +30.7 \\
\bottomrule
\end{tabular}}
\caption{Effect of Slovak Alpaca-style instruction-tuning on three base models (Slovak, 0-shot), sorted by size. \textbf{Base} and \textbf{+Alp.} are the overall macro-average of the base Slovak LM and its instruction-tuned counterpart. The $\Delta$ columns are tuned $-$ base for the overall score and the four most affected reporting groups (Knowledge \& reasoning, Classification, Instruction following, Grammar \& morphology). The net effect rises with base capability---negative for the weakest base (in bold), flat for the mid one, clearly positive for the strongest---while instruction following (the tuned skill) improves or holds for all three and the other groups swing in both directions.}
\label{tab:alpaca-delta}
\end{table}

\begin{figure}[t]
\centering
\includegraphics[width=\columnwidth]{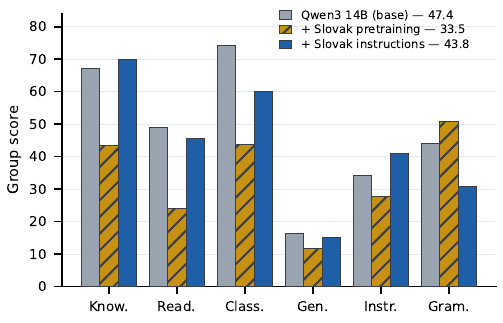}
\caption{The Qwen3-14B lineage: base, after Slovak continued
pretraining, and after the subsequent Slovak instruction-tuning step,
per reporting group (overall score beside each label). Adaptation
trades broad capability for a single grammar gain. Instruction tuning
recovers knowledge, reading, and instruction following, and reverses
the grammar gain. Scores are the six group averages from
Table~\ref{tab:results-full}.}
\label{fig:adaptation-trajectory}
\end{figure}

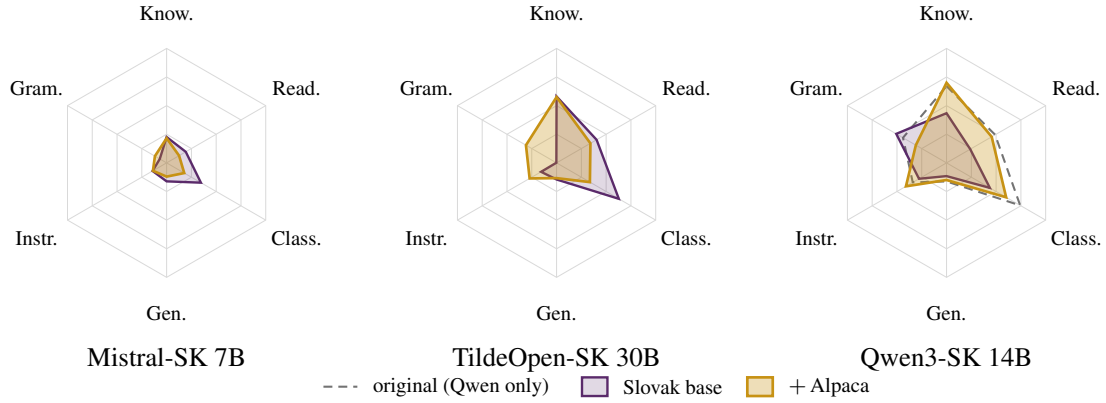
\begin{figure*}[t]
\centering
\resizebox{0.92\textwidth}{!}{%
\begin{tikzpicture}
\begin{scope}[shift={(0.0cm,0)}]
  \draw[gray!30,line width=0.3pt] (90:0.338cm) -- (30:0.338cm) -- (-30:0.338cm) -- (-90:0.338cm) -- (-150:0.338cm) -- (150:0.338cm) -- cycle;
  \draw[gray!30,line width=0.3pt] (90:0.675cm) -- (30:0.675cm) -- (-30:0.675cm) -- (-90:0.675cm) -- (-150:0.675cm) -- (150:0.675cm) -- cycle;
  \draw[gray!30,line width=0.3pt] (90:1.013cm) -- (30:1.013cm) -- (-30:1.013cm) -- (-90:1.013cm) -- (-150:1.013cm) -- (150:1.013cm) -- cycle;
  \draw[gray!30,line width=0.3pt] (90:1.350cm) -- (30:1.350cm) -- (-30:1.350cm) -- (-90:1.350cm) -- (-150:1.350cm) -- (150:1.350cm) -- cycle;
  \draw[gray!30,line width=0.3pt] (0,0) -- (90:1.35cm);
  \draw[gray!30,line width=0.3pt] (0,0) -- (30:1.35cm);
  \draw[gray!30,line width=0.3pt] (0,0) -- (-30:1.35cm);
  \draw[gray!30,line width=0.3pt] (0,0) -- (-90:1.35cm);
  \draw[gray!30,line width=0.3pt] (0,0) -- (-150:1.35cm);
  \draw[gray!30,line width=0.3pt] (0,0) -- (150:1.35cm);
  \node[font=\scriptsize] at (90:1.77cm) {Know.};
  \node[font=\scriptsize] at (30:1.77cm) {Read.};
  \node[font=\scriptsize] at (-30:1.77cm) {Class.};
  \node[font=\scriptsize] at (-90:1.77cm) {Gen.};
  \node[font=\scriptsize] at (-150:1.77cm) {Instr.};
  \node[font=\scriptsize] at (150:1.77cm) {Gram.};
  \draw[draw=promptframe,line width=0.8pt,fill=promptframe,fill opacity=0.18] (90:0.305cm) -- (30:0.261cm) -- (-30:0.467cm) -- (-90:0.218cm) -- (-150:0.191cm) -- (150:0.093cm) -- cycle;
  \draw[draw=newgold,line width=0.9pt,fill=newgold,fill opacity=0.30] (90:0.291cm) -- (30:0.170cm) -- (-30:0.240cm) -- (-90:0.162cm) -- (-150:0.186cm) -- (150:0.159cm) -- cycle;
  \node[font=\small,below] at (0,-2.05cm) {Mistral-SK 7B};
\end{scope}
\begin{scope}[shift={(4.6cm,0)}]
  \draw[gray!30,line width=0.3pt] (90:0.338cm) -- (30:0.338cm) -- (-30:0.338cm) -- (-90:0.338cm) -- (-150:0.338cm) -- (150:0.338cm) -- cycle;
  \draw[gray!30,line width=0.3pt] (90:0.675cm) -- (30:0.675cm) -- (-30:0.675cm) -- (-90:0.675cm) -- (-150:0.675cm) -- (150:0.675cm) -- cycle;
  \draw[gray!30,line width=0.3pt] (90:1.013cm) -- (30:1.013cm) -- (-30:1.013cm) -- (-90:1.013cm) -- (-150:1.013cm) -- (150:1.013cm) -- cycle;
  \draw[gray!30,line width=0.3pt] (90:1.350cm) -- (30:1.350cm) -- (-30:1.350cm) -- (-90:1.350cm) -- (-150:1.350cm) -- (150:1.350cm) -- cycle;
  \draw[gray!30,line width=0.3pt] (0,0) -- (90:1.35cm);
  \draw[gray!30,line width=0.3pt] (0,0) -- (30:1.35cm);
  \draw[gray!30,line width=0.3pt] (0,0) -- (-30:1.35cm);
  \draw[gray!30,line width=0.3pt] (0,0) -- (-90:1.35cm);
  \draw[gray!30,line width=0.3pt] (0,0) -- (-150:1.35cm);
  \draw[gray!30,line width=0.3pt] (0,0) -- (150:1.35cm);
  \node[font=\scriptsize] at (90:1.77cm) {Know.};
  \node[font=\scriptsize] at (30:1.77cm) {Read.};
  \node[font=\scriptsize] at (-30:1.77cm) {Class.};
  \node[font=\scriptsize] at (-90:1.77cm) {Gen.};
  \node[font=\scriptsize] at (-150:1.77cm) {Instr.};
  \node[font=\scriptsize] at (150:1.77cm) {Gram.};
  \draw[draw=promptframe,line width=0.8pt,fill=promptframe,fill opacity=0.18] (90:0.783cm) -- (30:0.545cm) -- (-30:0.848cm) -- (-90:0.196cm) -- (-150:0.211cm) -- (150:0.003cm) -- cycle;
  \draw[draw=newgold,line width=0.9pt,fill=newgold,fill opacity=0.30] (90:0.776cm) -- (30:0.461cm) -- (-30:0.454cm) -- (-90:0.180cm) -- (-150:0.367cm) -- (150:0.417cm) -- cycle;
  \node[font=\small,below] at (0,-2.05cm) {TildeOpen-SK 30B};
\end{scope}
\begin{scope}[shift={(9.2cm,0)}]
  \draw[gray!30,line width=0.3pt] (90:0.338cm) -- (30:0.338cm) -- (-30:0.338cm) -- (-90:0.338cm) -- (-150:0.338cm) -- (150:0.338cm) -- cycle;
  \draw[gray!30,line width=0.3pt] (90:0.675cm) -- (30:0.675cm) -- (-30:0.675cm) -- (-90:0.675cm) -- (-150:0.675cm) -- (150:0.675cm) -- cycle;
  \draw[gray!30,line width=0.3pt] (90:1.013cm) -- (30:1.013cm) -- (-30:1.013cm) -- (-90:1.013cm) -- (-150:1.013cm) -- (150:1.013cm) -- cycle;
  \draw[gray!30,line width=0.3pt] (90:1.350cm) -- (30:1.350cm) -- (-30:1.350cm) -- (-90:1.350cm) -- (-150:1.350cm) -- (150:1.350cm) -- cycle;
  \draw[gray!30,line width=0.3pt] (0,0) -- (90:1.35cm);
  \draw[gray!30,line width=0.3pt] (0,0) -- (30:1.35cm);
  \draw[gray!30,line width=0.3pt] (0,0) -- (-30:1.35cm);
  \draw[gray!30,line width=0.3pt] (0,0) -- (-90:1.35cm);
  \draw[gray!30,line width=0.3pt] (0,0) -- (-150:1.35cm);
  \draw[gray!30,line width=0.3pt] (0,0) -- (150:1.35cm);
  \node[font=\scriptsize] at (90:1.77cm) {Know.};
  \node[font=\scriptsize] at (30:1.77cm) {Read.};
  \node[font=\scriptsize] at (-30:1.77cm) {Class.};
  \node[font=\scriptsize] at (-90:1.77cm) {Gen.};
  \node[font=\scriptsize] at (-150:1.77cm) {Instr.};
  \node[font=\scriptsize] at (150:1.77cm) {Gram.};
  \draw[draw=black!55,densely dashed,line width=0.7pt] (90:0.905cm) -- (30:0.662cm) -- (-30:1.003cm) -- (-90:0.219cm) -- (-150:0.460cm) -- (150:0.593cm) -- cycle;
  \draw[draw=promptframe,line width=0.8pt,fill=promptframe,fill opacity=0.18] (90:0.586cm) -- (30:0.323cm) -- (-30:0.590cm) -- (-90:0.156cm) -- (-150:0.375cm) -- (150:0.685cm) -- cycle;
  \draw[draw=newgold,line width=0.9pt,fill=newgold,fill opacity=0.30] (90:0.945cm) -- (30:0.615cm) -- (-30:0.812cm) -- (-90:0.202cm) -- (-150:0.553cm) -- (150:0.418cm) -- cycle;
  \node[font=\small,below] at (0,-2.05cm) {Qwen3-SK 14B};
\end{scope}
\begin{scope}[shift={(4.6cm,-2.75cm)}]
  \draw[draw=black!55,densely dashed,line width=0.7pt] (-2.75,0.11) -- (-2.35,0.11);
  \node[font=\scriptsize,anchor=west] at (-2.30,0.11) {original (Qwen only)};
  \draw[draw=promptframe,line width=0.8pt,fill=promptframe,fill opacity=0.18] (0.30,0) rectangle (0.60,0.22);
  \node[font=\scriptsize,anchor=west] at (0.65,0.11) {Slovak base};
  \draw[draw=newgold,line width=0.9pt,fill=newgold,fill opacity=0.30] (2.25,0) rectangle (2.55,0.22);
  \node[font=\scriptsize,anchor=west] at (2.60,0.11) {$+$\,Alpaca};
\end{scope}
\end{tikzpicture}%
}
\caption{RQ2: per-reporting-group scores (Slovak, 0-shot, axes $0$--$100$) of three Slovak base checkpoints before (Slovak base) and after ($+$Alpaca) a small Slovak instruction-tuning step. For Qwen3-SK, whose upstream multilingual model (Qwen3-14B) was also evaluated, the dashed polygon shows that released model before Slovak adaptation. The upstream bases of Mistral-SK and TildeOpen-SK are not in the sweep. The effect varies by base and by group: Alpaca broadly lifts Qwen3-SK (knowledge, reading, classification, instruction-following) at the cost of grammar, is mixed for TildeOpen-SK, and mostly shrinks the weakest (Mistral-SK), and Slovak adaptation itself drops Qwen3-14B before Alpaca partly restores it (\S\ref{sec:rq2}, with per-group deltas in Table~\ref{tab:alpaca-delta}).}
\label{fig:adaptation}
\end{figure*}

\paragraph{L3: Turn on test-time reasoning before increasing model size.}
Reasoning mode yields large, consistent gains at $\geq$9B ($+8.5$ to
$+12.5$ overall, \S\ref{sec:rq3}, per-model deltas in
Table~\ref{tab:thinking-delta}), mostly in instruction following and
grammaticality. The gains hold in Slovak, but only above a capacity
floor: below it, small models exhaust their generation budget before
emitting an answer.
\begin{table}[t]
\centering\small
\setlength{\tabcolsep}{4pt}
\resizebox{\columnwidth}{!}{%
\begin{tabular}{lrrr@{\hskip 7pt}rrr}
\toprule
& \multicolumn{3}{c}{\textbf{Overall}} & \multicolumn{3}{c}{\textbf{$\Delta$ by group}} \\
\cmidrule(lr){2-4}\cmidrule(l){5-7}
\textbf{Model} & Std & Think & $\Delta$ & Instr. & Gram. & Gen. \\
\midrule
Qwen3 0.6B & 21.5 & 27.2 & +5.7 & +19.6 & +1.1 & -0.3 \\
Qwen3.5 0.8B\dag & 24.0 & 7.2 & \textbf{-16.8} & +10.1 & +3.6 & -9.4 \\
Qwen3 1.7B & 27.8 & 40.0 & +12.2 & +23.8 & +10.2 & -1.0 \\
Qwen3.5 2B\dag & 29.4 & 19.8 & \textbf{-9.7} & +7.1 & +12.5 & -10.8 \\
Qwen3.5 9B & 47.9 & 56.4 & +8.5 & +21.2 & +28.8 & -3.5 \\
Qwen3.5 27B & 55.3 & 64.3 & +8.9 & +35.8 & +11.6 & -0.9 \\
Qwen3.6 27B & 54.2 & 65.6 & +11.4 & +43.2 & +8.6 & -0.4 \\
Gemma 4 31B & 54.3 & 66.0 & +11.7 & +53.8 & +8.2 & +0.2 \\
Qwen3.6 35B-A3B & 52.2 & 64.7 & +12.5 & +44.1 & +13.8 & -0.5 \\
\bottomrule
\end{tabular}}
\caption{Effect of test-time reasoning (``thinking'' mode), Slovak, 0-shot, for the models evaluated in both modes (sorted by size). \textbf{Std} and \textbf{Think} are the overall macro-average. The $\Delta$ columns are thinking $-$ standard for the overall score and for the three most affected reporting groups (Instruction following, Grammar \& morphology, Generation). Reasoning runs under an 8{,}192-token generation cap. The two smallest models (\dag) exhaust it inside the reasoning trace without emitting a parseable answer, so their answer-extraction groups collapse while instruction and grammar still improve (negative overall $\Delta$ in bold). Full per-task thinking scores are in Table~\ref{tab:skbench-thinking}.}
\label{tab:thinking-delta}
\end{table}

\paragraph{Why small models regress under reasoning.} Under the token
cap the two smallest Qwen3.5 models exhaust their budget inside the
reasoning trace without emitting a parseable answer, so their
multiple-choice accuracy falls to $\sim$1\% (vs.\ 42\% in standard
mode), well below random. This pulls down every group scored by answer
extraction, while instruction following and
grammar, scored over the whole generation, still rise
(\S\ref{sec:rq3}). Thinking mode also raises the generation budget
(8{,}192 tokens vs.\ the per-task standard caps of 32--4{,}096), but
those caps already exceed each task's answer length, so the extra budget
goes to the reasoning trace, not the answer. The gains also land on
short-answer skills (grammar, knowledge), where a longer budget alone
cannot help. A budget-matched
control is left to future work.

\paragraph{L4: Do not over-invest in target-language prompts for
scoring.} Re-running the sweep with English prompts moves the overall
score by only $+0.4$ (Table~\ref{tab:lang-comparison}). The prompt language matters for grammar tasks and little elsewhere. This
concerns the instructions only: the test items themselves must still be
Slovak.
\begin{table}[t]
\centering\small
\setlength{\tabcolsep}{6pt}
\begin{tabular}{lrrr}
\toprule
\textbf{Reporting group} & \textbf{SK} & \textbf{EN} & \textbf{$\Delta$} \\
\midrule
Knowledge \& Reasoning & 58.1 & 58.3 & +0.1 \\
Reading \& QA & 39.4 & 41.0 & +1.6 \\
Classification & 61.3 & 61.1 & -0.2 \\
Generation & 13.2 & 13.1 & -0.1 \\
Grammar \& Morphology & 33.5 & 30.9 & -2.6 \\
\midrule
\textbf{Overall} & 49.9 & 50.3 & +0.4 \\
\bottomrule
\end{tabular}
\caption{Prompt-language effect: mean Slovak- vs.\ English-prompt score per reporting group, macro-averaged across all evaluated models over the tasks available in both languages (IFEval-SK and Slohry are Slovak-only and excluded). $\Delta = \text{EN} - \text{SK}$. A positive value means the English prompt scores higher. Full per-task English scores are in Table~\ref{tab:skbench-comprehensive-en}.}
\label{tab:lang-comparison}
\end{table}

\begin{table}[t]
\centering\small
\setlength{\tabcolsep}{4pt}
\resizebox{\columnwidth}{!}{%
\begin{tabular}{lcccccc}
\toprule
 & \textbf{skLEP} & \textbf{BCM} & \textbf{MERA} & \textbf{HuLU} & \textbf{INCL} & \textbf{sk-bench}\\
\midrule
Language                    & sk & cs & ru & hu & multi & sk\\
Generative LLM eval         &        & \yes & \yes &        & \yes & \yes\\
\midrule
Native paired contrast      &        &        &        &        &        & \yes\\
Open + closed, one harness  &        &        &        &        & $\sim$ & \yes\\
Grammar \& morphology       & $\sim$ &        &        &        &        & \yes\\
Instruction-following       &        &        & $\sim$ &        &        & \yes\\
\bottomrule
\end{tabular}}
\caption{How sk-bench differs from the closest Slavic, Central-European,
and multilingual LLM benchmarks. The lower block lists the design choices that set sk-bench apart. Cells
mark what each benchmark actually provides ($\sim$~=~partial).
BenCzechMark (BCM) and MERA score by \emph{log-likelihood},
which closed APIs do not expose, so neither ranks open- and
closed-weights models on one leaderboard. sk-bench's
generate-and-match scoring does. INCL~=~INCLUDE. Of sk-bench's 30
tasks, 11 are resources introduced or first packaged here
(Table~\ref{tab:dataset-catalogue}). We do not tabulate the same count
for the other suites, whose new-versus-integrated split we cannot
establish uniformly from their papers.}
\label{tab:benchmark-comparison}
\end{table}

\paragraph{Recommendations for benchmark builders.} (i)~Ship at least
one \emph{native paired contrast} per skill, so you can measure whether translation is safe for your language and
skill (Table~\ref{tab:benchmark-comparison}).
(ii)~Re-implement instruction and verifier checkers natively rather
than translating them: Slovak word boundaries, casing, and diacritics
break the original IFEval checker on its own examples
(Appendix~\ref{app:ifevalsk}).

\paragraph{Long-context robustness.}
Table~\ref{tab:oneruler-length} reports the OneRuler-SK results by
context length referenced in \S\ref{sec:oneruler}. Because the Slovak
and English haystacks are matched in length and run on the same models,
the Slovak$-$English gap at a fixed length separates a \emph{language} penalty from the model's context-window
limit. Two things follow. First, there is no Slovak-specific length cliff up to 16k:
macro-averaged over all 55 models, Slovak retrieval declines only mildly
from 4k to 16k ($58.1\!\to\!55.6\!\to\!54.4$), and the gap to English
does not widen with length; it narrows and slightly reverses ($-2.4 /
-0.7 / +1.1$). Long-context retrieval over Slovak text thus depends on
the model, not the language, consistent with L4. Second, scores vary
across models, not lengths: the six proprietary
APIs stay above $99\%$ at every length, a few models do degrade
with length (e.g.\ Kimi~K2.6, $63.7\!\to\!23.1$ from 4k to 16k), and
several base models---including the Slovak-adapted ones---score near
zero already at 4k. That floor is the same answer-not-emitted format
failure seen for the smallest reasoning models (\S\ref{sec:rq3}), not a
length effect. Because 16k is the longest length we evaluate, we cannot
locate a breaking point for the capable models, which stay flat across
this range. Finding where Slovak long-context retrieval breaks down would need
32k--128k haystacks, which we leave to future work.
\begin{table}[t]
\centering\small
\setlength{\tabcolsep}{6pt}
\begin{tabular}{lrrr}
\toprule
\textbf{Context} & \textbf{SK} & \textbf{EN} & \textbf{$\Delta$} \\
\midrule
4{,}096 & 58.1 & 60.5 & -2.4 \\
8{,}192 & 55.6 & 56.4 & -0.7 \\
16{,}384 & 54.4 & 53.4 & +1.1 \\
\bottomrule
\end{tabular}
\caption{OneRuler-SK by context length: mean retrieval accuracy ($\times100$, 0-shot) over the RULER-style subtasks at 4{,}096 / 8{,}192 / 16{,}384 tokens, macro-averaged across all evaluated models, for the matched Slovak and English haystacks. Because SK and EN share the same model and the same length, $\Delta=\text{SK}-\text{EN}$ isolates the language effect from the context-window limit. The gap does not widen with length---it closes and slightly reverses by 16k---so up to 16k there is no Slovak-specific long-context penalty. 16k is the longest length evaluated.}
\label{tab:oneruler-length}
\end{table}

\paragraph{Few-shot vs.\ zero-shot.} Table~\ref{tab:fewshot} gives the
full 3-shot vs.\ zero-shot breakdown by reporting group and prompt
language behind the summary in \S\ref{sec:oneruler}. The effect is small and preserves the ranking: the four-group overall
rises by only $+2.0$ (Slovak) and $+1.6$ (English), so zero-shot scores
do not noticeably understate any model. The gains come from
classification and reading \& QA, where a few examples fix the
\emph{output format} and label set; knowledge \& reasoning drops
slightly, since examples add no facts and only use up context. Slovak
and English prompts behave almost identically, except on grammar \&
morphology, where Slovak examples help slightly ($+1.1$) and English
ones hurt ($-1.6$). (The table lists all six reporting groups. Generation and
instruction following were run zero-shot only, so they appear at their
0-shot value with no 3-shot entry, and the ``Overall'' row spans the
four paired groups, so it is not directly comparable to the six-group
overall of the main leaderboard.) As with Alpaca tuning
(\S\ref{sec:alpaca}) and reasoning (\S\ref{sec:rq3}), examples teach a
model \emph{how to answer}, not \emph{what to know}. Zero-shot is
therefore a fair default, and few-shot is worth its context cost mainly
on fixed-label tasks with strict output formats.
\begin{table}[t]
\centering\small
\resizebox{\columnwidth}{!}{%
\setlength{\tabcolsep}{5pt}
\begin{tabular}{lrrrr}
\toprule
 & \multicolumn{2}{c}{\textbf{Slovak}} & \multicolumn{2}{c}{\textbf{English}}\\
\cmidrule(lr){2-3}\cmidrule(l){4-5}
\textbf{Reporting group} & 0-shot & 3-shot & 0-shot & 3-shot\\
\midrule
Knowledge \& reasoning & 60.6 & 59.4 & 61.2 & 60.4\\
Reading \& QA          & 29.4 & 32.6 & 30.7 & 33.6\\
Classification         & 61.9 & 66.8 & 61.3 & 67.0\\
Grammar \& morphology  & 35.7 & 36.8 & 28.6 & 27.0\\
\midrule
Overall & 46.9 & 48.9 & 45.4 & 47.0\\
\bottomrule
\end{tabular}}
\caption{Zero-shot vs.\ 3-shot, macro-averaged over the 52 models with few-shot results, for the four reporting groups of the main leaderboard (Table~\ref{tab:results-main}) that have a few-shot counterpart. Generation (summarisation) and instruction following were run zero-shot only and are therefore omitted. The \textbf{Overall} row is the macro-average over these four groups. Few-shot helps most on classification and reading \& QA, slightly hurts knowledge \& reasoning, and moves the overall little. The pattern is the same under Slovak and English prompts.}
\label{tab:fewshot}
\end{table}

\section{Costs}
\label{app:cost}

As shown in Table~\ref{tab:computational-costs}, our evaluation consumed 5,354 GPU hours (1,454 on the cluster with A100 40\,GB GPUs and 3,900 on the cluster with H200 141\,GB GPUs) and 15,500 CPU hours across 764,079 prompts across all open-source models. Proprietary API costs ranged from \$102.40 (Gemini 3.1 Flash Lite) to \$3,833.67 (Gemini 3.1 Pro), with GPT-5.4 Mini at \$286.96, GPT-5.5 at \$2,761.67, and Claude Sonnet 4.6 and Claude Opus 4.6 at \$1,560.83 and \$1,353.67, respectively. The Claude Sonnet 4.6 figure aggregates both its zero-shot and few-shot runs, whereas Claude Opus 4.6, like Gemini 3.1 Pro and GPT-5.5, was evaluated in the zero-shot setting only. Inference times for open-source models ranged from 1.84 to 80.83 hours, scaling with model size.

\begin{table*}
\centering
\tiny
\resizebox{\textwidth}{!}{%
\begin{tabular}{lrrrrr}
\toprule
\textbf{Model} & \makecell{\textbf{Input Tokens}\\\textbf{[M]}} & \makecell{\textbf{Completion Tokens}\\\textbf{[M]}} &  \makecell{\textbf{Total Tokens}\\\textbf{[M]}} & \textbf{Cost} & \makecell{\textbf{Evaluation Time}\\\textbf{[hours]}} \\
\midrule
\textit{Open-source Models} \\
\midrule
ibm-granite/granite-4.0-h-350m & 457.03 & 7.27 & 464.29 & N/A & 4.17 \\
ibm-granite/granite-4.0-350m & 457.03 & 10.83 & 467.85 & N/A & 1.97 \\
Qwen/Qwen3-0.6B & 439.36 & 14.13 & 453.49 & N/A & 2.49 \\
Qwen/Qwen3.5-0.8B & 369.88 & 13.04 & 382.92 & N/A & 3.16 \\
google/gemma-3-1b-it & 361.46 & 8.33 & 369.79 & N/A & 2.22 \\
ibm-granite/granite-4.0-h-1b & 458.30 & 7.39 & 465.69 & N/A & 5.04 \\
ibm-granite/granite-4.0-1b & 458.30 & 8.31 & 466.62 & N/A & 3.09 \\
Qwen/Qwen3-1.7B & 439.36 & 17.38 & 456.74 & N/A & 3.04 \\
utter-project/EuroLLM-1.7B-Instruct & 332.54 & 12.50 & 345.05 & N/A & 2.03 \\
ibm-granite/granite-3.2-2b-instruct & 506.76 & 20.90 & 527.67 & N/A & 5.12 \\
Qwen/Qwen3.5-2B & 369.88 & 12.82 & 382.70 & N/A & 3.96 \\
meta-llama/Llama-3.2-3B-Instruct & 405.58 & 10.80 & 416.39 & N/A & 3.36 \\
CohereLabs/tiny-aya-global & 602.03 & 13.82 & 615.85 & N/A & 3.60 \\
CohereLabs/tiny-aya-water & 602.03 & 13.74 & 615.77 & N/A & 3.65 \\
openGPT-X/Teuken-7B-instruct-v0.6 & 341.66 & 18.52 & 360.17 & N/A & 9.75 \\
slovak-nlp/mistral-sk-7b & 449.54 & 12.29 & 461.82 & N/A & 7.87 \\
meta-llama/Llama-3.1-8B-Instruct & 405.58 & 11.07 & 416.65 & N/A & 5.91 \\
mistralai/Ministral-3-8B-Instruct-2512 & 372.43 & 15.44 & 387.87 & N/A & 2.07 \\
mistralai/Ministral-3-8B-Reasoning-2512 & 372.43 & 46.72 & 419.15 & N/A & 2.34 \\
CohereLabs/aya-expanse-8b & 359.57 & 22.26 & 381.83 & N/A & 6.25 \\
utter-project/EuroLLM-9B-Instruct-2512 & 340.57 & 12.83 & 353.40 & N/A & 6.52 \\
Qwen/Qwen3.5-9B & 369.88 & 9.41 & 379.29 & N/A & 9.73 \\
speakleash/Bielik-11B-v3.0-Instruct & 456.55 & 15.94 & 472.50 & N/A & 12.11 \\
mistralai/Ministral-3-14B-Instruct-2512 & 372.43 & 14.39 & 386.82 & N/A & 1.84 \\
mistralai/Ministral-3-14B-Reasoning-2512 & 372.43 & 530.61 & 903.03 & N/A & 12.65 \\
Qwen/Qwen3-14B & 439.36 & 9.81 & 449.17 & N/A & 2.73 \\
slovak-nlp/Qwen3-14B-sk & 439.36 & 42.09 & 481.45 & N/A & 16.33 \\
openai/gpt-oss-20b & 402.44 & 85.76 & 488.20 & N/A & 3.32 \\
utter-project/EuroLLM-22B-Instruct-2512 & 340.57 & 9.84 & 350.41 & N/A & 4.92 \\
Qwen/Qwen3.5-27B & 369.89 & 9.85 & 379.74 & N/A & 12.55 \\
Qwen/Qwen3.6-27B & 369.89 & 8.85 & 378.73 & N/A & 11.38 \\
TildeAI/TildeOpen-30b & 446.01 & 9.37 & 455.38 & N/A & 7.13 \\
TildeAI/TildeOpen-30b-64k & 306.24 & 7.96 & 314.20 & N/A & 5.91 \\
google/gemma-4-31B-it & 364.16 & 7.97 & 372.13 & N/A & 8.39 \\
CohereLabs/aya-expanse-32b & 359.57 & 16.99 & 376.56 & N/A & 5.80 \\
ibm-granite/granite-4.0-h-small & 462.76 & 9.20 & 471.96 & N/A & 4.14 \\
Qwen/Qwen3.6-35B-A3B & 369.89 & 9.40 & 379.29 & N/A & 7.01 \\
meta-llama/Llama-3.3-70B-Instruct & 405.58 & 8.89 & 414.47 & N/A & 7.52 \\
swiss-ai/Apertus-70B-Instruct-2509 & 411.40 & 24.64 & 436.03 & N/A & 9.87 \\
CohereLabs/c4ai-command-a-03-2025 & 742.28 & 19.38 & 761.66 & N/A & 4.67 \\
openai/gpt-oss-120b & 402.44 & 67.54 & 469.98 & N/A & 6.52 \\
Qwen/Qwen3.5-397B-A17B & 369.89 & 9.90 & 379.79 & N/A & 10.54 \\
mistralai/Mistral-Large-3-675B-Instruct-2512 & 372.43 & 19.59 & 392.02 & N/A & 7.65 \\
MiniMaxAI/MiniMax-M2.7 & 438.59 & 836.75 & 1275.34 & N/A & 80.83 \\
zai-org/GLM-5.1-FP8 & 399.11 & 8.14 & 407.25 & N/A & 9.05 \\
moonshotai/Kimi-K2.6 & 414.52 & 14.44 & 428.96 & N/A & 6.66 \\
\hline
Mistral-SK 7B (Alpaca) & 456.42 & 28.10 & 484.52 & N/A & 3.01 \\
Qwen3-SK 14B (Alpaca) & 439.36 & 12.88 & 452.24 & N/A & 4.37 \\
TildeOpen-SK 30B (Alpaca) & 314.65 & 13.29 & 327.94 & N/A & 6.39 \\
\midrule
\textit{Thinking Versions} \\
\midrule
Qwen/Qwen3-0.6B & 435.63 & 602.11 & 1037.74 & N/A & 18.76 \\
Qwen/Qwen3.5-0.8B & 368.36 & 6119.13 & 6487.49 & N/A & 213.20 \\
Qwen/Qwen3-1.7B & 435.63 & 793.35 & 1228.98 & N/A & 67.53 \\
Qwen/Qwen3.5-2B & 368.96 & 3019.48 & 3388.44 & N/A & 310.53 \\
Qwen/Qwen3.5-9B & 368.35 & 2380.40 & 2748.75 & N/A & 393.67 \\
Qwen/Qwen3.5-27B & 368.36 & 1764.76 & 2133.12 & N/A & 136.75 \\
Qwen/Qwen3.6-27B & 368.36 & 808.09 & 1176.45 & N/A & 87.62 \\
google/gemma-4-31B-it & 366.46 & 456.12 & 822.58 & N/A & 89.84 \\
Qwen/Qwen3.6-35B-A3B & 368.36 & 865.91 & 1234.27 & N/A & 57.95 \\
\midrule
\textit{Proprietary Models} \\
\midrule
Gemini 3.1 Flash Lite & 358.41 & 9.13 & 367.53 & \$102.40 & ---\\
Gemini 3.1 Pro* & 166.26 & 577.82 & 744.08 & \$3,833.67 & --- \\
GPT-5.4 Mini & 357.55 & 9.48 & 367.02 & \$286.96 & --- \\
GPT-5.5* & 167.70 & 128.53 & 296.23 & \$2,761.67 & ---\\
Claude Sonnet 4.6 & 387.62 & 26.53 & 414.15 & \$1,560.83 & --- \\
Claude Opus 4.6* & 184.45 & 17.26 & 201.71 & \$1,353.67 & --- \\
\bottomrule
\end{tabular}}
\caption{Resource consumption and evaluation time across models on the sk-bench benchmark. Token counts are shown in millions (M). * denotes models evaluated only in the zero-shot setting due to API constraints. Evaluation time is measured in hours. ``---'' indicates data not available or not applicable.}
\label{tab:computational-costs}
\end{table*}

\onecolumn
\section{Full Dataset Catalogue}
\label{app:dataset-catalogue}

Table~\ref{tab:dataset-catalogue} lists every dataset in sk-bench with its
evaluation size, provenance, target skill, licence, and a one-line
description. It expands on the overview in
Figure~\ref{fig:overview}.

\begin{longtable}{p{0.15\textwidth}p{0.05\textwidth}p{0.04\textwidth}p{0.13\textwidth}p{0.11\textwidth}p{0.36\textwidth}}
\caption{Dataset catalogue for sk-bench. $|\textrm{test}|$ is the
evaluation test-set size used by the released harness. \textbf{Prov.}:
N=Slovak-native, MT=translated from English, MT$^{\!*}$=translated
and adapted with a custom Slovak instruction registry, G=generated
over Slovak source documents. Resources marked~\newdsmark{} are the eleven
introduced or first packaged with sk-bench, at three levels of
contribution: \emph{built from scratch} (IFEval-SK, Chiby, Slohry,
Slovak Trivia, OneRuler-SK), \emph{newly
curated from public institutional sources} (Autoškola, e-testy-sk,
Slovak-tests, Scio), and \emph{existing expert resources first packaged
for generative-LLM evaluation} (SKJ1, Word Analogies). \textbf{Licence} is the licence declared on the source artefact (``Mixed'' = the bundled release carries per-task licences; ``---'' = none declared).}
\label{tab:dataset-catalogue}\\
\toprule
\textbf{Dataset} & \textbf{$|\textrm{test}|$} & \textbf{Prov.} & \textbf{Skill} & \textbf{Licence} & \textbf{Description} \\
\midrule
\endfirsthead
\multicolumn{6}{l}{\small\itshape Table~\ref{tab:dataset-catalogue}, continued from previous page}\\[2pt]
\toprule
\textbf{Dataset} & \textbf{$|\textrm{test}|$} & \textbf{Prov.} & \textbf{Skill} & \textbf{Licence} & \textbf{Description} \\
\midrule
\endhead
\midrule
\multicolumn{6}{r}{\small\itshape continued on next page}\\
\endfoot
\bottomrule
\endlastfoot
\hyperref[task:senti]{SentiSK} & 4{,}677 & N & sentiment & CC BY-SA 4.0 & Slovak user-text sentiment corpus used in binary and 3-class settings. Neutral examples are excluded from the binary score. \\
\hyperref[task:movierev]{MovieReview} & 2{,}008 & N & sentiment & --- & Slovak film- and series-review sentiment with labels derived from user star ratings~\citep{csfd}, included as a second native binary sentiment source. See Appendix~\ref{task:movierev}. \\
\hyperref[task:sklep-sent]{skLEP-sentiment} & 1{,}042 & N & sentiment & Mixed & The sentiment subtask from skLEP, retained as a small validation cross-check against the larger sentiment resources. \\
\hyperref[task:toxic]{ToxicSK} & 881 & N & safety classification & CC BY-SA 3.0 & Slovak toxicity-detection corpus scored as yes/no classification. \\
\hyperref[task:sklep-hate]{skLEP-hate} & 1{,}319 & N & safety classification & Mixed & skLEP hate-speech subtask, paired with ToxicSK to separate toxicity from explicit hate-speech labelling. \\
\hyperref[task:nli-sk]{NLI-SK-annotated} & 597 & N & NLI & --- & Human-annotated Slovak 3-way entailment dataset after filtering the non-NLI ``Other'' category. \\
\hyperref[task:sklep-nli]{skLEP-NLI} & 5{,}004 & MT & NLI & Mixed & Translated 3-way NLI data from skLEP, used as the translated counterpart to NLI-SK-annotated. \\
\hyperref[task:sklep-rte]{skLEP-RTE} & 1{,}660 & MT & RTE & Mixed & Separate translated binary textual-entailment dataset from GLUE RTE, included as a second entailment task alongside skLEP-NLI. \\
\hyperref[task:sklep-qa]{skLEP-QA} & 9{,}583 & N & extractive QA & Mixed & Slovak Wikipedia/news question-answering data, evaluated with normalised exact match and token F1. \\
\hyperref[task:wikiqa]{MultiWikiQA-SK} & 4{,}997 & G & open QA & CC BY-NC-SA 4.0 & LLM-generated Slovak Wikipedia QA pairs, included to contrast human-authored and generated prompts over the same broad source domain. \\
\hyperref[task:slovaksum]{SlovakSum} & 1{,}500 & N & summarisation & --- & Large Slovak news summarisation corpus. sk-bench uses an evaluation-sized down-sample from 20{,}662 items with the original human reference summaries. \\
\hyperref[task:smesum]{SME-Sum} & 1{,}500 & N & summarisation & --- & Slovak news summarisation dataset based on SME articles. sk-bench uses an evaluation-sized down-sample from 8{,}001 items. \\
\hyperref[task:belebele]{Belebele-SK} & 900 & MT & reading comprehension & CC BY-SA 4.0 & Human-translated Slovak slice of the parallel Belebele multiple-choice reading-comprehension benchmark. \\
\hyperref[task:arc-sk]{ARC-SK} & 1{,}169 & MT & science reasoning & CC BY-NC 4.0 & Okapi translation of ARC Challenge science questions, used as a compact translated reasoning benchmark. \\
\hyperref[task:mmlu-sk]{Okapi-MMLU-SK} & 13{,}062 & MT & knowledge reasoning & --- & Slovak translation of MMLU-style multi-subject questions across 57 subjects, the largest translated multi-choice component in the suite. \\
\hyperref[task:truthqa-sk]{TruthfulQA-SK} & 778 & MT & truthfulness & CC BY-NC 4.0 & Slovak translation of TruthfulQA multiple-choice questions designed to elicit common false beliefs. \\
\hyperref[task:copa]{SlovakCOPA} & 500 & MT & causal reasoning & CC BY-NC 4.0 & Slovak version of COPA, testing plausible causes and effects in short everyday scenarios. \\
\hyperref[task:etest]{e-testy-sk}\newds & 444 & N & school exams & Other & Slovak high-school examination item bank covering multiple school subjects. \\
\hyperref[task:sktest]{Slovak-tests}\newds & 343 & N & Slovak-language exams & --- & Subset of native school-exam questions that specifically test Slovak grammar, orthography, and vocabulary. \\
\hyperref[task:finexam]{Financial Exams} & 1{,}334 & N & professional exams & CC BY-SA 4.0 & Slovak financial-sector certification questions, testing domain-specific regulatory and procedural knowledge. \\
\hyperref[task:autoskola]{Autoškola}\newds & 701 & N & professional exams & CC BY 4.0 & Slovak driving-licence exam questions, included for law- and culture-specific practical knowledge. \\
\hyperref[task:chiby]{Chiby}\newds & 3{,}432 & N & grammar & CC BY-SA 4.0 & Grammaticality classification data derived from Slovak Wikipedia revisions and manually filtered for spelling, diacritics, and capitalisation errors. \\
\hyperref[task:skj1]{SKJ1}\newds & 100 & N & grammar & --- & Slovak grammar-correction multiple-choice set from TUKE-UUI. \\
\hyperref[task:scio]{Scio}\newds & 1{,}785 & N & admission tests & --- & Slovak Scio admission-test questions spanning biology, mathematics, chemistry, and general knowledge. \\
\hyperref[task:slohry]{Slohry}\newds & 4{,}004 & N & grammar & CC BY 4.0 & Slovak word-game items testing lexical and morphological competence. \\
\hyperref[task:trivia]{Slovak Trivia}\newds & 353 & N & world knowledge & --- & Native Slovak trivia questions on culture, geography, and general knowledge. \\
\hyperref[task:analogies]{Word Analogies}\newds & 2{,}651 & N & analogy & --- & Slovak word-analogy completion items testing lexical-semantic relations. \\
\hyperref[task:gmmlu]{Global-MMLU-Lite} & 400 & MT & knowledge reasoning & Apache 2.0 & Curated Slovak slice of the Global-MMLU-Lite multilingual knowledge benchmark. \\
\hyperref[task:ifeval]{IFEval-SK}\newds & 537 & MT$^{\!*}$ & instruction following & --- & Slovak adaptation of IFEval with translated prompts and programmatic checkers re-implemented for Slovak word boundaries, casing, and punctuation. \\
\hyperref[task:oneruler]{OneRuler-SK}\newds & 9{,}000 & N & long context & --- & Slovak long-context reading-comprehension suite probing retrieval and reasoning over extended Slovak text. \\
\end{longtable}
\twocolumn

\section{Task Descriptions}
\label{app:tasks-detail}

\begin{table}[t]
\centering\small
\setlength{\tabcolsep}{4pt}
\resizebox{\columnwidth}{!}{%
\begin{tabular}{lll}
\toprule
\textbf{Skill} & \textbf{Native side} & \textbf{Contrast side}\\
\midrule
NLI & NLI-SK-annotated & skLEP-NLI (MT)\\
Multiple-choice & e-testy-sk\newds, & Okapi-MMLU-SK (MT)\\
knowledge & Slovak-tests\newds & \\
Question answering & skLEP-QA (human) & MultiWikiQA-SK (G)\\
Grammar \& morph. & Chiby\newds, SKJ1\newds, & \emph{translation cannot}\\
                  & Slohry\newds              & \emph{produce these}\\
\bottomrule
\end{tabular}}
\caption{The paired provenance contrasts that structure the suite. Each
row holds a skill fixed and varies how the evaluation data was
produced. \newdsmark{} marks a resource introduced or first packaged with sk-bench.
MT~=~machine-translated, G~=~LLM-generated. The grammar row has no contrast side: translation cannot produce these
items, so they must be built natively. \S\ref{sec:rq1} measures how far
the rankings disagree in each row.}
\label{tab:paired-contrasts}
\end{table}

Table~\ref{tab:dataset-catalogue} lists each dataset's size, provenance,
and skill. This appendix describes, for every task in the leaderboard
(Table~\ref{tab:results-main}), what the model is asked to do, how it is
scored, where the data comes from, and why the task is in the suite. Resources marked~\newdsmark{} are introduced or first packaged with sk-bench. Several
have full curation cards in Appendix~\ref{app:downsampling}. Prompt
templates are in Appendix~\ref{app:prompts}, and throughout the paper a
dataset name is a hyperlink to its entry here.

\subsection{Knowledge \& Reasoning}
\paragraph{ARC-SK.}\label{task:arc-sk} Multiple-choice grade-school
science questions. The model selects one option, scored by accuracy.
Okapi machine translation of ARC-Challenge, a small translated reasoning set~\citep{arc2018,okapi2023}.

\paragraph{Okapi-MMLU-SK.}\label{task:mmlu-sk} Four-option multiple
choice spanning MMLU's 57 academic subjects, scored by accuracy. This is the largest
translated multi-choice component. With ARC-SK it is the
\emph{translated} side of the multi-choice provenance contrast (RQ1)
against the native e-testy-sk/Slovak-tests~\citep{mmlu2021,okapi2023}.

\paragraph{Global-MMLU-Lite.}\label{task:gmmlu} Four-option multiple
choice, scored by accuracy. A small curated slice of a multilingual knowledge benchmark, kept as an
independent check on Okapi-MMLU-SK~\citep{globalmmlu2024}.

\paragraph{TruthfulQA-SK.}\label{task:truthqa-sk} Multiple choice in the
MC1 setting (single-true-answer multiple choice)---one true answer among
several plausible-sounding false ones, scored by accuracy. Probes susceptibility to common misconceptions rather
than recall~\citep{truthfulqa2022,okapi2023}.

\paragraph{Scio\newds.}\label{task:scio} Single-answer multiple choice,
scored by accuracy. Items parsed from Scio admission-test papers covering general
study prerequisites alongside biology, chemistry, and mathematics.
Curated by the authors: the parsing, cleaning, and answer-key
alignment are ours, and no external annotators were involved
(Appendix~\ref{app:card-scio}).

\paragraph{Financial Exams.}\label{task:finexam} Five-option
single-answer multiple choice, scored by accuracy. Parsed from the publicly
released financial-advisor certification exams of the National Bank of
Slovakia and tagged by topic (insurance, capital markets, deposits,
loans, pensions) and difficulty~\citep{financial_exams}.

\paragraph{Autoškola\newds.}\label{task:autoskola} Multiple choice,
scored by accuracy. Questions from the theory part of the official Slovak
driving-licence test across 35 topic categories, with the official
answer key as gold. Assembled by the authors from that official
question bank with LLM assistance. The gold labels are the official
answer key, not model output (Appendix~\ref{app:card-autoskola}).

\paragraph{SlovakCOPA.}\label{task:copa} Binary forced choice between
two alternatives for the more plausible cause or effect of a premise,
scored by accuracy (chance~$=50\%$). A standard-Slovak translation of
COPA~\citep{slovakcopa}.

\paragraph{SKJ1\newds.}\label{task:skj1} Five-option (A--E) multiple
choice, scored by accuracy. Slovak grammar-correction items spanning nine error
types---gender agreement, inflection, reflexive \emph{sa}/\emph{si},
Bohemisms, Anglicisms, and others. The items are expert-authored by
the team behind the SKJ1 dataset~\citep{skj1} and adopted from that
work. We recruited no experts ourselves, and sk-bench's contribution
is the generative-LLM wrapping---prompts, parsing, and scoring
(Appendix~\ref{app:card-skj1}).

\paragraph{Slovak Trivia\newds.}\label{task:trivia}
Single-answer multiple choice, scored by accuracy. Native questions on Slovak
culture, geography, and general knowledge, derived from short articles.
Constructed by the authors for sk-bench. No external annotators were
involved.

\subsection{Reading \& QA}
\paragraph{Belebele-SK.}\label{task:belebele} Four-option
multiple-choice reading comprehension over a short passage, scored by accuracy.
The human-translated Slovak slice of the parallel Belebele set, so scores are comparable
across its 122 languages~\citep{belebele2024}.

\paragraph{Chiby\newds.}\label{task:chiby} Binary grammaticality
judgement: decide whether a sentence drawn from Slovak Wikipedia
revisions is correct or carries a spelling/diacritic/capitalisation
error, scored by accuracy. Annotated from Chiby~v0.4 by three native Slovak
speakers who are co-authors of this work, after cosine-similarity pairing and language filtering (full procedure
in Appendix~\ref{app:chiby})~\citep{Garabik2026}. Translation cannot
produce such pairs.

\paragraph{MultiWikiQA-SK.}\label{task:wikiqa} Free-text answer
generation grounded in a Wikipedia passage, scored by case- and
diacritic-normalised exact match and token~F1. Questions were generated
by an LLM (Gemini~1.5~Pro) over the 2023 Wikipedia dump~\citep{multiwikiqa2025}
and then \emph{rephrased by the same model in a second pass} to reduce surface word overlap with the passage. It is the
\emph{generated} side of the QA contrast (RQ1) against the
human-authored skLEP-QA~\citep{sklep2025} task (built on
SK-QuAD~\citep{skquad2023}).

\paragraph{e-testy-sk\newds.}\label{task:etest} High-school exam items
from the e-testy.sk portal across nine subjects, reported in two
formats: single-answer multiple choice (\emph{ETest-M}, accuracy) and
free-text questions (\emph{ETest-Q}, normalised exact match / token~F1).
The items were public, but not as a dataset: the parsing, cleaning, and
answer-key alignment are ours. It is the native multiple-choice
counterpart to Okapi-MMLU-SK.

\paragraph{Slovak-tests\newds.}\label{task:sktest} School and
standardised-exam items (maturita, entrance exams, Monitor~9) focused on
Slovak language and literature, reported as \emph{SKTest-M} (multiple
choice, accuracy) and \emph{SKTest-Q} (free-text, normalised exact
match / token~F1). As with e-testy-sk, the items were public, but not as a dataset: the
parsing, cleaning, and answer-key alignment are ours.

\paragraph{OneRuler-SK\newds.}\label{task:oneruler} Long-context
retrieval and reasoning, scored by exact match against a gold string (the
absent-needle case answers \emph{žiadne}). A Slovak adaptation of
ONERULER built over an 18th-century Slovak novel as the haystack with
needle words drawn from Wiktionary, spanning needle-in-a-haystack and
common-word-extraction tasks at 4k--16k tokens. The haystack, the
needle placement, and the gold strings were constructed by the authors.
The gold answer is determined by construction, so no annotation round
was required.

\paragraph{skLEP-QA.}\label{task:sklep-qa} Extractive free-text QA over
Slovak Wikipedia/news, scored by normalised exact match and token~F1. The
question-answering subtask of skLEP~\citep{sklep2025} is built on the
human-authored, crowd-annotated SK-QuAD~\citep{skquad2023}, and forms the
\emph{human} side of the QA contrast against generated MultiWikiQA-SK.
The main text and the per-task tables of Appendix~\ref{app:detailed}
refer to the evaluated task as \texttt{skLEP-QA}.

\subsection{Classification}
\paragraph{NLI-SK-annotated.}\label{task:nli-sk} Three-way
entailment (entailment / neutral / contradiction), scored by accuracy. Hypotheses
are hand-written. Premises are drawn from SlovakSum titles, hand-written
items on Slovak traditions and geography, and image captions. Labels
are assigned by three native Slovak speakers using unanimous or majority
agreement~\citep{nli_sk}. The evaluated subset includes both. The resource was
first announced in SkMTEB~\citep{skmteb2026}. It forms the
\emph{native} side of the NLI contrast (RQ1) against translated
skLEP-NLI~\citep{sklep2025}.

\paragraph{skLEP-NLI.}\label{task:sklep-nli} Three-way entailment,
scored by accuracy. The skLEP NLI subtask, machine-translated from XNLI with
native post-editing---the translated counterpart in the NLI provenance
contrast~\citep{sklep2025}.

\paragraph{skLEP-RTE.}\label{task:sklep-rte} Binary entailment,
scored by accuracy. A separate machine-translated dataset from the GLUE RTE
corpus, testing entailment versus non-entailment~\citep{sklep2025}.

\paragraph{SentiSK.}\label{task:senti} User-comment sentiment, reported
both 3-class (\emph{Senti-3}) and binary (\emph{Senti-2}, neutral
dropped), scored by accuracy. Slovak Facebook comments manually annotated at TU
Košice. The 3-class set skews negative~\citep{toxic_sk}.

\paragraph{skLEP-sentiment.}\label{task:sklep-sent} Binary sentiment,
scored by accuracy. The skLEP sentiment subtask (from the Reviews3 corpus), a
small validation cross-check against the larger sentiment
resources~\citep{sklep2025}.

\paragraph{MovieReview.}\label{task:movierev} Binary sentiment over
Slovak film and series reviews, scored by accuracy. Labels derive from user star
ratings, making it a second native binary sentiment set from a different
domain~\citep{csfd}.

\paragraph{ToxicSK.}\label{task:toxic} Binary toxicity detection
(toxic / non-toxic), scored by accuracy. Slovak Facebook comments manually
annotated at TU Košice, class-balanced between the two
labels~\citep{toxic_sk}.

\paragraph{skLEP-hate.}\label{task:sklep-hate} Binary hate-speech
detection, scored by accuracy. The skLEP hate-speech subtask (from the Slovak Hate
Speech Database), paired with ToxicSK to separate general toxicity from
explicit hate~\citep{sklep2025}.

\subsection{Generation}
\paragraph{SlovakSum.}\label{task:slovaksum} Abstractive news
summarisation, with a generated summary scored by ROUGE-L against a single
human reference. A length-stratified down-sample of a multi-source
Slovak news corpus that preserves the original length distribution
(Appendix~\ref{app:downsampling})~\citep{slovaksum2024}.

\paragraph{SME-Sum.}\label{task:smesum} Abstractive summarisation of
SME.sk news articles, with the article perex as the reference summary,
scored by ROUGE-L. A length-stratified down-sample of SME-Sum
(Appendix~\ref{app:downsampling})~\citep{smesum2020}.

\subsection{Grammar \& Morphology}
\paragraph{Word Analogies\newds.}\label{task:analogies} Open-ended
completion: given a labelled relation and three terms of an analogy,
the model produces the missing fourth, scored by exact match. Probes
lexical-semantic relations. The relation inventory and term pairs are
taken from the Slovak Conceptual
Dictionary~\citep{blstak2025slovakconceptualdictionary}. sk-bench's
contribution is the generative-LLM wrapping---prompts, parsing, and
scoring.

\paragraph{Slohry\newds.}\label{task:slohry} Inflectional-morphology
\emph{production} across four sub-tasks---verb conjugation,
adjective/adverb comparison, noun declension, and grammatical-form
identification. The model generates the requested form, scored by exact
match against a single gold answer. 4{,}004 items targeting
Slovak morphological competence, generated by the authors from the
standard Slovak paradigms, so the gold form follows from the paradigm
rather than from an annotation judgement.

\subsection{Instruction Following}
\paragraph{IFEval-SK\newds.}\label{task:ifeval} Verifiable instruction
following: each prompt carries machine-checkable constraints (length,
casing, keywords, format, response language) verified by programmatic
checkers. Scoring uses prompt-level strict accuracy. A Slovak adaptation of M-IFEval
whose 25 instruction types were re-implemented for Slovak word
boundaries, casing, and punctuation
(Appendix~\ref{app:ifevalsk})~\citep{ifeval2023, mifeval2025}. Both the Slovak
templates and the checker registry are ours. Verification is
programmatic, so no annotators were involved.

\section{Scoring and Parsing Details}
\label{app:scoring}

This appendix describes how raw model outputs are parsed and normalised
before scoring.

\paragraph{Output parsing.}
The free-text parser proceeds in this fixed order:
(i)~strip leading/trailing whitespace and chat-template artefacts
(``\texttt{The answer is}'', ``\texttt{Odpove\v{d}'}\,:'', \dots);
(ii)~apply Unicode NFC normalisation;
(iii)~for multi-choice tasks, scan for the first occurrence of an
isolated answer letter (\texttt{A}--\texttt{Z}) optionally followed
by \texttt{)}, \texttt{.}, or whitespace and return that letter;
(iv)~for fixed-vocabulary classification, lowercase under the Slovak
locale (preserving diacritics --- \v{c}/c, \'i/i, \v{s}/s pairs are
semantically distinct in Slovak) and match the first allowed label
substring. If neither rule fires, the example is counted as incorrect
for the main score.

\paragraph{Classification metric calculation.}
We use two classification metrics. The \emph{conservative} metric
counts an answer as correct only when exactly one valid class label
appears in the response, so an answer naming several labels scores zero
even if one of them is correct; this penalises hedging. The
\emph{multi-label} metric, for tasks with several correct labels,
extracts the set of labels in the response and requires it to equal the
gold set exactly.

\paragraph{Invalid outputs and refusals.}
Refusals (``Som AI a nem\^o\v{z}em\dots''), safety wrappers, and any
output that the parser cannot resolve are counted as incorrect for
the main score.

\paragraph{Slovak normalisation for free-form tasks.}
For QA we apply NFC normalisation, Slovak-locale case folding
(without diacritic stripping), and the QEM/QF1 token splitter from
LightEval. We do \emph{not} stem because Slovak inflection makes stemming
brittle. Quasi-exact match and token-level quasi-F1 then compare the normalised
answer with the gold answer.

\paragraph{Summarisation metrics.}
ROUGE-1/2/L is computed with the \texttt{rouge\_score} package
(default tokeniser, no stemming). chrF/chrF++ and METEOR are computed
through LightEval. Absolute scores are treated as ordinal.

\paragraph{Token budgets.}
A single per-task generation budget is applied identically across
providers (32 tokens for short-answer classification, 512 for QA
and summarisation, 4096 for IFEval). Because Slovak tokenisation
differs by model, this is not the same character or word budget
across checkpoints.

\paragraph{Output length and format failures.}
The harness's per-task token logs show that output length is not a problem for almost any model. On multiple-choice tasks the graded
answer is a median of 2 tokens, well within the 32-token
budget, and across standard (non-reasoning) models the mean generation
is only 8--37 tokens, none of which ever reaches the 8{,}192-token
ceiling. Long, budget-exhausting generations happen only in reasoning
mode: at least one generation reaches the ceiling on 88.6\% of
MiniMax~M2.7's evaluation runs and 46.3\% of
Ministral~3~14B-Reasoning's, against 0\% for every standard model.
Even in reasoning mode we grade only the answer and not the reasoning
trace --- gpt-oss~20B averages 112 tokens per generation, of which
only about 14 are the answer we score. Format failures that affect scores are thus concentrated in reasoning mode on small models (\S\ref{sec:rq3}): when they exhaust the budget, their
multiple-choice accuracy falls to about 1\% against 42\% in standard
mode, because no answer is emitted, not because their reasoning got
worse.

\section{Model Catalogue}
\label{app:models}
Table~\ref{tab:model-catalogue} lists the evaluated open-weights models
with their parameter count, licence, reported knowledge or pretraining
cutoff where published (``---'' otherwise), language coverage, the Hugging
Face commit SHA pinned at evaluation time, and a reference to its family's technical
report. References are at the family level (e.g.\ all Qwen3.x variants
cite the Qwen3 report). Our Slovak-adapted and fine-tuned models have no
separate report and are described in \S\ref{sec:setup}.

We record a reported knowledge or pretraining cutoff wherever the model
card or technical report states one. Nine of the catalogued checkpoints
do. The remaining entries are ``---'' because no cutoff is published. Cutoff-versus-dataset-date reasoning about
possible contamination is therefore available for a minority of our
models, which is why Appendix~\ref{app:contamination} documents
\emph{dataset} availability instead.

\begin{table*}
\scriptsize
\setlength{\tabcolsep}{3pt}
\resizebox{\textwidth}{!}{%
\begin{tabular}{lllllll}
\toprule
\textbf{Model} & \textbf{Params.} & \textbf{Licence} & \textbf{Reported cutoff} & \textbf{\# Langs} & \textbf{HF SHA} & \textbf{Ref.} \\
\midrule
\href{https://huggingface.co/ibm-granite/granite-4.0-h-350m}{ibm-granite/granite-4.0-h-350m} & 350M & Apache 2.0 & --- & 12 & 3b17b717b8f2f5d305b0a92c1491e239aeda19c8 & \citep{granite2024} \\
\href{https://huggingface.co/ibm-granite/granite-4.0-350m}{ibm-granite/granite-4.0-350m} & 350M & Apache 2.0 & --- & 12 & bd8a1497065c0d6ba1ef19af6b0d2b14bacf71c2 & \citep{granite2024} \\
\href{https://huggingface.co/Qwen/Qwen3-0.6B}{Qwen/Qwen3-0.6B} & 0.6B & Apache 2.0 & --- & 100+ & c1899de289a04d12100db370d81485cdf75e47ca & \citep{qwen3} \\
\href{https://huggingface.co/Qwen/Qwen3.5-0.8B}{Qwen/Qwen3.5-0.8B} & 0.8B & Apache 2.0 & --- & 201 & 2fc06364715b967f1860aea9cf38778875588b17 & \citep{qwen3} \\
\href{https://huggingface.co/google/gemma-3-1b-it}{google/gemma-3-1b-it} & 1B & gemma & --- & 140+ & dcc83ea841ab6100d6b47a070329e1ba4cf78752 & \citep{gemma3} \\
\href{https://huggingface.co/ibm-granite/granite-4.0-h-1b}{ibm-granite/granite-4.0-h-1b} & 1B & Apache 2.0 & --- & 12 & d18cca4c121edb87d022116d281ce212c9136f57 & \citep{granite2024} \\
\href{https://huggingface.co/ibm-granite/granite-4.0-1b}{ibm-granite/granite-4.0-1b} & 1B & Apache 2.0 & --- & 12 & 6a7381ba1f54d684ff508d991aeb7dc580157103 & \citep{granite2024} \\
\href{https://huggingface.co/Qwen/Qwen3-1.7B}{Qwen/Qwen3-1.7B} & 1.7B & Apache 2.0 & --- & 100+ & 70d244cc86ccca08cf5af4e1e306ecf908b1ad5e & \citep{qwen3} \\
\href{https://huggingface.co/utter-project/EuroLLM-1.7B-Instruct}{utter-project/EuroLLM-1.7B-Instruct} & 1.7B & Apache 2.0 & --- & 35 & a25c7fa65fc2a644e6270b8940dbe295b51da681 & \citep{eurollm2024} \\
\href{https://huggingface.co/ibm-granite/granite-3.2-2b-instruct}{ibm-granite/granite-3.2-2b-instruct} & 2B & Apache 2.0 & --- & 12 & 641593c3b25bec0b1efe9f0f7d7a67f7243f86a3 & \citep{granite2024} \\
\href{https://huggingface.co/Qwen/Qwen3.5-2B}{Qwen/Qwen3.5-2B} & 2B & Apache 2.0 & --- & 201 & 15852e8c16360a2fea060d615a32b45270f8a8fc & \citep{qwen3} \\
\href{https://huggingface.co/meta-llama/Llama-3.2-3B-Instruct}{meta-llama/Llama-3.2-3B-Instruct} & 3B & llama3.2 & --- & 8 & 0cb88a4f764b7a12671c53f0838cd831a0843b95 & \citep{llama3} \\
\href{https://huggingface.co/CohereLabs/tiny-aya-global}{CohereLabs/tiny-aya-global} & 3.35B & CC-BY-NC-4.0 & --- & 67 & b2d281bab0beaecb3418d0c0dc53eaba90574011 & \citep{ayaexpanse} \\
\href{https://huggingface.co/CohereLabs/tiny-aya-water}{CohereLabs/tiny-aya-water} & 3.35B & CC-BY-NC-4.0 & --- & 67 & 3696a27d9f455538e50887778122cc898b6370b8 & \citep{ayaexpanse} \\
\href{https://huggingface.co/openGPT-X/Teuken-7B-instruct-v0.6}{openGPT-X/Teuken-7B-instruct-v0.6} & 7B & CC-BY-NC-4.0 & 09/2023 & 24 & eb3cb8fdd396037c7029ee3b2087f54c415086ac & \citep{teuken2025} \\
\href{https://huggingface.co/slovak-nlp/mistral-sk-7b}{slovak-nlp/mistral-sk-7b} & 7B & Apache 2.0 & --- & 1 & 089497ae0d72018e9591895c54f774bf0b4a83a4 & --- \\
\href{https://huggingface.co/meta-llama/Llama-3.1-8B-Instruct}{meta-llama/Llama-3.1-8B-Instruct} & 8B & llama3.1 & 12/2023 & 8 & 0e9e39f249a16976918f6564b8830bc894c89659 & \citep{llama3} \\
\href{https://huggingface.co/mistralai/Ministral-3-8B-Instruct-2512}{mistralai/Ministral-3-8B-Instruct-2512} & 8B & Apache 2.0 & --- & 11 & dcfc440ed367bc62945a537e9e31473406535292 & \citep{mistral7b} \\
\href{https://huggingface.co/mistralai/Ministral-3-8B-Reasoning-2512}{mistralai/Ministral-3-8B-Reasoning-2512} & 8B & Apache 2.0 & --- & 11 & 6c2c4c82a395598a424f3a17555ae0cc19e3f5bb & \citep{mistral7b} \\
\href{https://huggingface.co/CohereLabs/aya-expanse-8b}{CohereLabs/aya-expanse-8b} & 8B & CC-BY-NC-4.0 & --- & 23 & 5062468bf9bc0c6035fd64e06274333ec127d980 & \citep{ayaexpanse} \\
\href{https://huggingface.co/utter-project/EuroLLM-9B-Instruct-2512}{utter-project/EuroLLM-9B-Instruct-2512} & 9B & Apache 2.0 & --- & 35 & 1acaaeb8d9e8d9b5e3bf0240aacd32fc11ee4213 & \citep{eurollm2024} \\
\href{https://huggingface.co/Qwen/Qwen3.5-9B}{Qwen/Qwen3.5-9B} & 9B & Apache 2.0 & 12/2025 & 201 & c202236235762e1c871ad0ccb60c8ee5ba337b9a & \citep{qwen3} \\
\href{https://huggingface.co/speakleash/Bielik-11B-v3.0-Instruct}{speakleash/Bielik-11B-v3.0-Instruct} & 11B & Apache 2.0 & --- & 33 & e0ee273ce65377e91790567115d9b841911f3864 & \citep{bielik} \\
\href{https://huggingface.co/mistralai/Ministral-3-14B-Instruct-2512}{mistralai/Ministral-3-14B-Instruct-2512} & 14B & Apache 2.0 & --- & 11 & 1e4bed9a74c1d8af713dd9e2545d69020ced05dc & \citep{mistral7b} \\
\href{https://huggingface.co/mistralai/Ministral-3-14B-Reasoning-2512}{mistralai/Ministral-3-14B-Reasoning-2512} & 14B & Apache 2.0 & --- & 11 & eaabe4241136f11758e111c5b6b8bae5dde0b031 & \citep{mistral7b} \\
\href{https://huggingface.co/Qwen/Qwen3-14B}{Qwen/Qwen3-14B} & 14B & Apache 2.0 & --- & 100+ & 40c069824f4251a91eefaf281ebe4c544efd3e18 & \citep{qwen3} \\
\href{https://huggingface.co/slovak-nlp/Qwen3-14B-sk}{slovak-nlp/Qwen3-14B-sk} & 14B & Apache 2.0 & --- & 1 & 61bc40572b2fa4720e75a7a9de457151aefce396 & --- \\
\href{https://huggingface.co/openai/gpt-oss-20b}{openai/gpt-oss-20b} & 20B & Apache 2.0 & 06/2024 & N/A & 6cee5e81ee83917806bbde320786a8fb61efebee & \citep{gptoss} \\
\href{https://huggingface.co/utter-project/EuroLLM-22B-Instruct-2512}{utter-project/EuroLLM-22B-Instruct-2512} & 22B & Apache 2.0 & --- & 35 & 38196362cf70aebbc19d1acb9ebc7c18ef81d1b2 & \citep{eurollm2024} \\
\href{https://huggingface.co/Qwen/Qwen3.5-27B}{Qwen/Qwen3.5-27B} & 27B & Apache 2.0 & 12/2025 & 201 & fc05daec18b0a78c049392ed2e771dde82bdf654 & \citep{qwen3} \\
\href{https://huggingface.co/Qwen/Qwen3.6-27B}{Qwen/Qwen3.6-27B} & 27B & Apache 2.0 & --- & 201 & 6a9e13bd6fc8f0983b9b99948120bc37f49c13e9 & \citep{qwen3} \\
\href{https://huggingface.co/TildeAI/TildeOpen-30b}{TildeAI/TildeOpen-30b} & 30B & CC-BY-4.0 & --- & 32 & faea2654cb208f13c0f84702b82c3113400bdd4a & \citep{tildeopen2025} \\
\href{https://huggingface.co/TildeAI/TildeOpen-30b-64k}{TildeAI/TildeOpen-30b-64k} & 30B & CC-BY-4.0 & --- & 32 & e4c6ad4d87756665566e8e3fd9a23d3a413f2489 & \citep{tildeopen2025} \\
\href{https://huggingface.co/google/gemma-4-31B-it}{google/gemma-4-31B-it} & 31B & Apache 2.0 & 01/2025 & 35+ & fcf2302760ae9c6e528a8dbba9dd636e56848237 & \citep{gemma3} \\
\href{https://huggingface.co/CohereLabs/aya-expanse-32b}{CohereLabs/aya-expanse-32b} & 32B & CC-BY-NC-4.0 & --- & 23 & b306ea27e360683b50c005d7fcbad6a242317910 & \citep{ayaexpanse} \\
\href{https://huggingface.co/ibm-granite/granite-4.0-h-small}{ibm-granite/granite-4.0-h-small} & 32B & Apache 2.0 & --- & 12 & b8c0982bab7fde4eb48110f5a069527c008fab39 & \citep{granite2024} \\
\href{https://huggingface.co/Qwen/Qwen3.6-35B-A3B}{Qwen/Qwen3.6-35B-A3B} & 35B & Apache 2.0 & --- & 201 & 995ad96eacd98c81ed38be0c5b274b04031597b0 & \citep{qwen3} \\
\href{https://huggingface.co/meta-llama/Llama-3.3-70B-Instruct}{meta-llama/Llama-3.3-70B-Instruct} & 70B & llama3.3 & 12/2023 & 8 & 6f6073b423013f6a7d4d9f39144961bfbfbc386b & \citep{llama3} \\
\href{https://huggingface.co/swiss-ai/Apertus-70B-Instruct-2509}{swiss-ai/Apertus-70B-Instruct-2509} & 70B & Apache 2.0 & --- & 1000 & ec382cbace709945d642a54f4448411df05f793b & \citep{apertus} \\
\href{https://huggingface.co/CohereLabs/c4ai-command-a-03-2025}{CohereLabs/c4ai-command-a-03-2025} & 111B & CC-BY-NC-4.0 & --- & 23 & 066daaa11d57404b07cc650d1068936d31431979 & \citep{commanda} \\
\href{https://huggingface.co/openai/gpt-oss-120b}{openai/gpt-oss-120b} & 120B & Apache 2.0 & 06/2024 & N/A & b5c939de8f754692c1647ca79fbf85e8c1e70f8a & \citep{gptoss} \\
\href{https://huggingface.co/Qwen/Qwen3.5-397B-A17B}{Qwen/Qwen3.5-397B-A17B} & 397B & Apache 2.0 & 12/2025 & 201 & 8472618112abcbd45acbcdc58436aff4233c23f7 & \citep{qwen3} \\
\href{https://huggingface.co/mistralai/Mistral-Large-3-675B-Instruct-2512}{mistralai/Mistral-Large-3-675B-Instruct-2512} & 675B & Apache 2.0 & --- & 11 & 5bb3d32a0a147594527f38cb136a8390eaa82222 & \citep{mistral7b} \\
\href{https://huggingface.co/MiniMaxAI/MiniMax-M2.7}{MiniMaxAI/MiniMax-M2.7} & 230B & other & --- & N/A  & d494266a4affc0d2995ba1fa35c8481cbd84294b & \citep{minimax01} \\
\href{https://huggingface.co/zai-org/GLM-5.1-FP8}{zai-org/GLM-5.1-FP8} & 754B & MIT & --- & 2 & f396cf805182f4ca10fa675e1a99815b3ca384db & \citep{glm} \\
\href{https://huggingface.co/moonshotai/Kimi-K2.6}{moonshotai/Kimi-K2.6} & 1.1T & other & --- & N/A & 7eb5002f6aadc958aed6a9177b7ed26bb94011bb & \citep{kimik2} \\
\midrule
 \textit{Fine-Tuned Models} \\
\midrule
TildeOpen-SK 30B (Alpaca) & 30B & CC-BY-4.0 & --- & 2 & N/A & --- \\
Qwen3-SK 14B (Alpaca) & 14B & Apache 2.0 & --- & 2 & N/A & --- \\
Mistral-SK 7B (Alpaca) & 7B & Apache 2.0 & --- & 2 & N/A & --- \\
\bottomrule
\end{tabular}%
}
\caption{Summary of benchmarked models. The \textbf{Ref.}\ column cites
each model's family technical report. ``---'' indicates models for which no technical report is available, including our own Slovak-adapted/fine-tuned models as well as community Slovak models sourced from Hugging Face.}
\label{tab:model-catalogue}
\end{table*}

\section{Data Availability and Contamination}
\label{app:contamination}

As noted in the Limitations, leakage can happen at three levels: \emph{benchmark-format} leakage (the exact prompt and gold
answer in sk-bench form), \emph{answer-key} leakage (the
question--answer pairs of the underlying item bank), and
\emph{source-document} leakage (the raw text behind the task). The
latter two correspond to \emph{annotation} and \emph{raw text}
contamination in the sense of \citet{conda2024}, while
benchmark-format leakage is specific to newly packaged suites. No
contamination measurement was performed for this paper, and none of
the numbers we report should be read as contamination-free. Instead, we document availability: when the source material was
published, whether an evaluation split over it had been released before,
and whether the underlying source is public. Comparing dataset dates
with model cutoffs is less useful: the model catalogue
(Table~\ref{tab:model-catalogue}) records checkpoint revisions and
reported cutoffs, but most models have no published cutoff, so the
comparison is possible for only a minority of the 55 models.

\begin{table*}
\small
\centering
\begin{tabular}{p{0.25\textwidth}p{0.19\textwidth}p{0.21\textwidth}p{0.23\textwidth}}
\toprule
\textbf{Group (examples)} & \textbf{Source published} & \textbf{Split previously released?} & \textbf{Underlying source public?} \\
\midrule
Newly packaged here (Chiby, Slohry, OneRuler-SK, Scio, Autoškola,
e-testy-sk, Slovak-tests, Word Analogies, Slovak Trivia)
  & 2024--2026
  & No --- evaluation split first assembled for sk-bench
  & Often yes (public exams, Wikipedia, dictionary) \\
\addlinespace
Previously released native (NLI-SK-annotated, SK-QuAD, skLEP,
SlovakSum, SME-Sum)
  & 2020--2026
  & Yes
  & Yes \\
\addlinespace
Translated proxies (Okapi-MMLU-SK, ARC-SK, TruthfulQA-SK,
Belebele-SK, SlovakCOPA)
  & English 2018--2022, Slovak 2023--2024
  & Yes
  & Yes (English originals ubiquitous) \\
\addlinespace
LLM-generated (MultiWikiQA-SK)
  & 2025, over a 2023 Wikipedia dump
  & Yes
  & Yes (Wikipedia) \\
\bottomrule
\end{tabular}
\caption{Data availability by group. ``Source published'' is when the
underlying material became available, not when sk-bench packaged it.
``Split previously released'' asks whether an evaluation split over
that material was already circulating before this work. No
contamination measurement was performed. The table documents
exposure opportunity, not exposure.}
\label{tab:contamination}
\end{table*}

We draw only a narrow conclusion from Table~\ref{tab:contamination}. The newly packaged tasks are the only ones whose exact split was not
previously circulating, and even for those the source material may
still have appeared in training. The previously released native group
and the translated proxies warrant the most caution, particularly for closed
models whose training data is undisclosed and whose knowledge cutoff
is not published. Readers who want to compare a specific model
against a specific cutoff should use the released per-task breakdowns
(Appendix~\ref{app:detailed}) to restrict the leaderboard to the
first row of the table, which is the subset for which
benchmark-format leakage can be excluded.

\section{Per-Task Detailed Results}
\label{app:detailed}

This appendix gives the complete results behind the main leaderboard (Table~\ref{tab:results-main}, which shows the top two open
models per size tier and top three proprietary APIs):
Table~\ref{tab:results-full} is the full reporting-group
leaderboard for all 55 models, and the remaining tables give the
per-task scores. All cells are the task's primary metric ($\times100$)
with Slovak prompts, 0-shot. \textbf{Best} and \underline{second-best}
per column are marked.
Table~\ref{tab:skbench-comprehensive} gives every model on every task and
Table~\ref{tab:skbench-thinking} repeats it for thinking-mode models. The
per-reporting-group averages used in the body leaderboard
(Table~\ref{tab:results-main}) are computed directly from these per-task
scores.

\begin{figure*}[t]
  \centering
  \includegraphics[width=\linewidth]{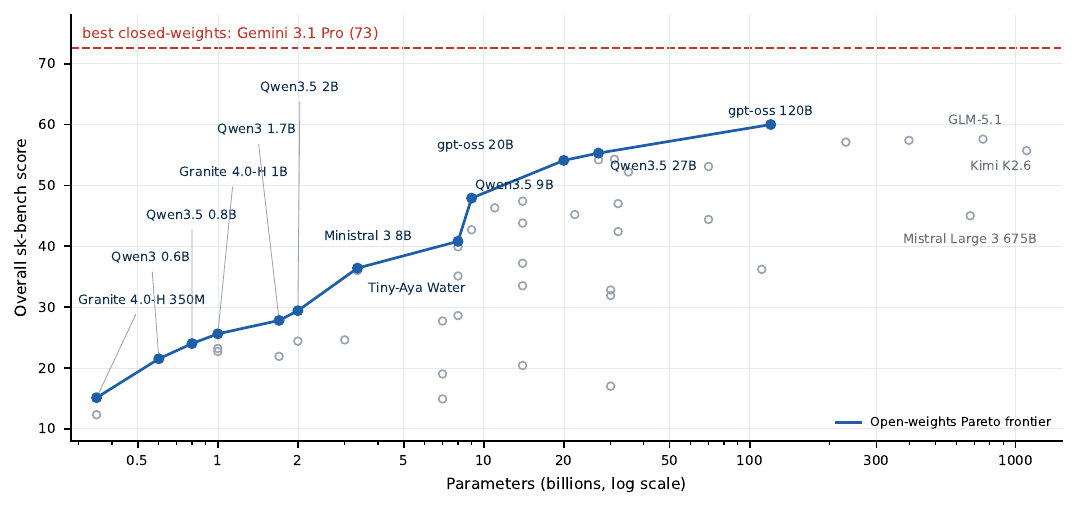}
  \caption{Open-weights size--score Pareto frontier, with \emph{every}
  frontier model labelled. Overall sk-bench score against parameter
  count (log scale). The blue line is the frontier, the best score
  reachable at or below each size. Gains are steep up to $\sim$9B, then
  the frontier flattens: past gpt-oss~20B and Qwen3.5~27B no open model
  beats gpt-oss~120B (60.0), and the 675B--1.1T MoEs sit below it. The
  dashed line marks the best closed-weights model (Gemini~3.1~Pro,
  72.6).}
  \label{fig:pareto-full}
\end{figure*}

\onecolumn
\begingroup
\small
\setlength{\tabcolsep}{2pt}
\begin{longtable}{lrrrrrr|r}
\caption{Full sk-bench leaderboard: all 55 models by reporting group, grouped by size tier (Slovak, 0-shot). Each cell is the unweighted mean of that group's per-task primary metrics ($\times100$), with \textbf{best} and \underline{second-best} per column. \textbf{Overall} is the unweighted mean of the six group averages. The small grey $\pm$ value is the standard deviation across the three prompt templates. The body leaderboard (Table~\ref{tab:results-main}) shows the top two open models per size tier and top three proprietary APIs. Per-task scores are in Table~\ref{tab:skbench-comprehensive}.}\label{tab:results-full}\\
\toprule
\textbf{Model} & \makecell{\textbf{Knowledge \&}\\\textbf{reasoning}} & \makecell{\textbf{Reading \&}\\\textbf{QA}} & \textbf{Classification} & \textbf{Generation} & \makecell{\textbf{Instruction}\\\textbf{following}} & \makecell{\textbf{Grammar \&}\\\textbf{morphology}} & \textbf{Overall} \\
\midrule
\endfirsthead
\multicolumn{8}{l}{\small\itshape continued from previous page}\\[2pt]
\toprule
\textbf{Model} & \makecell{\textbf{Knowledge \&}\\\textbf{reasoning}} & \makecell{\textbf{Reading \&}\\\textbf{QA}} & \textbf{Classification} & \textbf{Generation} & \makecell{\textbf{Instruction}\\\textbf{following}} & \makecell{\textbf{Grammar \&}\\\textbf{morphology}} & \textbf{Overall} \\
\midrule
\endhead
\midrule
\multicolumn{8}{r}{\small\itshape continued on next page}\\
\endfoot
\bottomrule
\endlastfoot
\multicolumn{8}{l}{\textit{$<$2B}} \\
Granite 4.0 350M & 27.5\,{\tiny\color{gray}$\pm$3.1} & 10.1\,{\tiny\color{gray}$\pm$0.6} & 13.8\,{\tiny\color{gray}$\pm$10.3} & 9.7\,{\tiny\color{gray}$\pm$3.4} & 10.8 & 1.8\,{\tiny\color{gray}$\pm$1.3} & 12.3\,{\tiny\color{gray}$\pm$1.9} \\
Granite 4.0-H 350M & 25.5\,{\tiny\color{gray}$\pm$5.7} & 9.4\,{\tiny\color{gray}$\pm$0.9} & 27.3\,{\tiny\color{gray}$\pm$10.1} & 11.7\,{\tiny\color{gray}$\pm$2.0} & 15.3 & 1.3\,{\tiny\color{gray}$\pm$0.0} & 15.1\,{\tiny\color{gray}$\pm$3.2} \\
Qwen3 0.6B & 33.1\,{\tiny\color{gray}$\pm$5.2} & 20.9\,{\tiny\color{gray}$\pm$3.5} & 43.3\,{\tiny\color{gray}$\pm$7.5} & 10.6\,{\tiny\color{gray}$\pm$0.2} & 16.8 & 4.2\,{\tiny\color{gray}$\pm$1.3} & 21.5\,{\tiny\color{gray}$\pm$1.6} \\
Qwen3.5 0.8B & 33.4\,{\tiny\color{gray}$\pm$2.8} & 26.0\,{\tiny\color{gray}$\pm$0.3} & 53.5\,{\tiny\color{gray}$\pm$7.8} & 9.7\,{\tiny\color{gray}$\pm$0.2} & 14.3 & 7.0\,{\tiny\color{gray}$\pm$2.5} & 24.0\,{\tiny\color{gray}$\pm$1.3} \\
Gemma 3 1B & 35.1\,{\tiny\color{gray}$\pm$2.0} & 19.9\,{\tiny\color{gray}$\pm$3.0} & 54.4\,{\tiny\color{gray}$\pm$4.1} & 11.8\,{\tiny\color{gray}$\pm$1.0} & 12.3 & 3.0\,{\tiny\color{gray}$\pm$0.8} & 22.7\,{\tiny\color{gray}$\pm$0.6} \\
Granite 4.0 1B & 34.5\,{\tiny\color{gray}$\pm$0.3} & 27.6\,{\tiny\color{gray}$\pm$1.0} & 41.8\,{\tiny\color{gray}$\pm$10.9} & 13.2\,{\tiny\color{gray}$\pm$0.2} & 17.1 & 5.2\,{\tiny\color{gray}$\pm$2.4} & 23.2\,{\tiny\color{gray}$\pm$1.9} \\
Granite 4.0-H 1B & 39.8\,{\tiny\color{gray}$\pm$0.6} & 28.2\,{\tiny\color{gray}$\pm$0.6} & 50.1\,{\tiny\color{gray}$\pm$7.8} & 14.8\,{\tiny\color{gray}$\pm$0.9} & 14.9 & 6.0\,{\tiny\color{gray}$\pm$0.8} & 25.6\,{\tiny\color{gray}$\pm$1.8} \\
EuroLLM 1.7B & 35.2\,{\tiny\color{gray}$\pm$1.9} & 19.2\,{\tiny\color{gray}$\pm$0.6} & 40.3\,{\tiny\color{gray}$\pm$6.9} & 13.4\,{\tiny\color{gray}$\pm$0.2} & 14.7 & 8.4\,{\tiny\color{gray}$\pm$1.4} & 21.9\,{\tiny\color{gray}$\pm$1.4} \\
Qwen3 1.7B & 35.0\,{\tiny\color{gray}$\pm$5.5} & 30.9\,{\tiny\color{gray}$\pm$2.8} & 52.4\,{\tiny\color{gray}$\pm$7.8} & 12.8\,{\tiny\color{gray}$\pm$0.9} & 23.5 & 12.3\,{\tiny\color{gray}$\pm$1.9} & 27.8\,{\tiny\color{gray}$\pm$2.4} \\
\midrule
\multicolumn{8}{l}{\textit{2--9B}} \\
Granite 3.2 2B & 38.2\,{\tiny\color{gray}$\pm$0.2} & 22.5\,{\tiny\color{gray}$\pm$2.1} & 51.7\,{\tiny\color{gray}$\pm$2.5} & 8.7\,{\tiny\color{gray}$\pm$0.1} & 13.8 & 11.6\,{\tiny\color{gray}$\pm$0.9} & 24.4\,{\tiny\color{gray}$\pm$0.2} \\
Qwen3.5 2B & 43.4\,{\tiny\color{gray}$\pm$1.9} & 34.2\,{\tiny\color{gray}$\pm$4.2} & 56.9\,{\tiny\color{gray}$\pm$6.9} & 11.2\,{\tiny\color{gray}$\pm$0.8} & 18.8 & 12.2\,{\tiny\color{gray}$\pm$1.0} & 29.4\,{\tiny\color{gray}$\pm$1.6} \\
Llama 3.2 3B & 34.7\,{\tiny\color{gray}$\pm$0.8} & 25.8\,{\tiny\color{gray}$\pm$2.7} & 46.2\,{\tiny\color{gray}$\pm$2.4} & 14.8\,{\tiny\color{gray}$\pm$1.1} & 14.0 & 12.5\,{\tiny\color{gray}$\pm$2.4} & 24.6\,{\tiny\color{gray}$\pm$0.7} \\
Tiny-Aya Global & 50.7\,{\tiny\color{gray}$\pm$1.1} & 34.6\,{\tiny\color{gray}$\pm$5.5} & 73.9\,{\tiny\color{gray}$\pm$2.3} & 10.6\,{\tiny\color{gray}$\pm$0.3} & 26.8 & 19.6\,{\tiny\color{gray}$\pm$0.4} & 36.0\,{\tiny\color{gray}$\pm$1.6} \\
Tiny-Aya Water & 50.5\,{\tiny\color{gray}$\pm$1.1} & 34.9\,{\tiny\color{gray}$\pm$5.5} & 74.3\,{\tiny\color{gray}$\pm$1.7} & 10.8\,{\tiny\color{gray}$\pm$0.1} & 27.6 & 20.4\,{\tiny\color{gray}$\pm$1.0} & 36.4\,{\tiny\color{gray}$\pm$1.5} \\
Mistral-SK 7B & 22.6\,{\tiny\color{gray}$\pm$4.1} & 19.3\,{\tiny\color{gray}$\pm$1.0} & 34.6\,{\tiny\color{gray}$\pm$11.4} & 16.2\,{\tiny\color{gray}$\pm$0.8} & 14.2 & 6.9\,{\tiny\color{gray}$\pm$2.3} & 19.0\,{\tiny\color{gray}$\pm$2.0} \\
Mistral-SK 7B (Alpaca) & 21.5\,{\tiny\color{gray}$\pm$1.6} & 12.6\,{\tiny\color{gray}$\pm$1.3} & 17.7\,{\tiny\color{gray}$\pm$7.6} & 12.0\,{\tiny\color{gray}$\pm$0.8} & 13.8 & 11.8\,{\tiny\color{gray}$\pm$2.6} & 14.9\,{\tiny\color{gray}$\pm$1.2} \\
Teuken 7B & 39.8\,{\tiny\color{gray}$\pm$2.4} & 26.1\,{\tiny\color{gray}$\pm$1.6} & 59.1\,{\tiny\color{gray}$\pm$3.7} & 10.3\,{\tiny\color{gray}$\pm$0.3} & 15.3 & 15.4\,{\tiny\color{gray}$\pm$2.9} & 27.7\,{\tiny\color{gray}$\pm$2.0} \\
Aya Expanse 8B & 44.5\,{\tiny\color{gray}$\pm$3.1} & 25.8\,{\tiny\color{gray}$\pm$4.9} & 59.0\,{\tiny\color{gray}$\pm$8.7} & 12.3\,{\tiny\color{gray}$\pm$0.8} & 15.1 & 14.8\,{\tiny\color{gray}$\pm$2.2} & 28.6\,{\tiny\color{gray}$\pm$3.0} \\
Llama 3.1 8B & 47.8\,{\tiny\color{gray}$\pm$2.0} & 41.4\,{\tiny\color{gray}$\pm$4.3} & 60.9\,{\tiny\color{gray}$\pm$4.9} & 15.3\,{\tiny\color{gray}$\pm$0.6} & 19.6 & 25.8\,{\tiny\color{gray}$\pm$0.6} & 35.1\,{\tiny\color{gray}$\pm$2.2} \\
Ministral 3 8B & 60.4\,{\tiny\color{gray}$\pm$1.3} & 39.2\,{\tiny\color{gray}$\pm$6.2} & 70.6\,{\tiny\color{gray}$\pm$4.4} & 13.5\,{\tiny\color{gray}$\pm$0.8} & 24.0 & 37.1\,{\tiny\color{gray}$\pm$1.0} & 40.8\,{\tiny\color{gray}$\pm$2.3} \\
Ministral 3 8B (R) & 54.9\,{\tiny\color{gray}$\pm$1.9} & 37.8\,{\tiny\color{gray}$\pm$4.0} & 61.9\,{\tiny\color{gray}$\pm$7.2} & 13.1\,{\tiny\color{gray}$\pm$1.1} & 39.9 & 32.1\,{\tiny\color{gray}$\pm$2.8} & 39.9\,{\tiny\color{gray}$\pm$2.4} \\
\midrule
\multicolumn{8}{l}{\textit{9--32B}} \\
EuroLLM 9B & 55.8\,{\tiny\color{gray}$\pm$1.3} & 42.7\,{\tiny\color{gray}$\pm$8.4} & 74.6\,{\tiny\color{gray}$\pm$1.4} & 14.9\,{\tiny\color{gray}$\pm$0.7} & 22.7 & 45.2\,{\tiny\color{gray}$\pm$0.4} & 42.7\,{\tiny\color{gray}$\pm$2.0} \\
Qwen3.5 9B & 69.7\,{\tiny\color{gray}$\pm$0.7} & 50.4\,{\tiny\color{gray}$\pm$5.2} & 77.0\,{\tiny\color{gray}$\pm$2.1} & 15.0\,{\tiny\color{gray}$\pm$0.2} & 31.1 & 44.4\,{\tiny\color{gray}$\pm$2.4} & 47.9\,{\tiny\color{gray}$\pm$1.4} \\
Bielik 11B & 66.5\,{\tiny\color{gray}$\pm$1.0} & 47.1\,{\tiny\color{gray}$\pm$3.9} & 74.6\,{\tiny\color{gray}$\pm$7.4} & 12.6\,{\tiny\color{gray}$\pm$0.6} & 30.4 & 46.9\,{\tiny\color{gray}$\pm$0.5} & 46.3\,{\tiny\color{gray}$\pm$2.4} \\
Ministral 3 14B & 63.8\,{\tiny\color{gray}$\pm$0.9} & 34.6\,{\tiny\color{gray}$\pm$4.8} & 55.9\,{\tiny\color{gray}$\pm$17.1} & 14.2\,{\tiny\color{gray}$\pm$1.3} & 18.4 & 36.5\,{\tiny\color{gray}$\pm$0.8} & 37.2\,{\tiny\color{gray}$\pm$3.1} \\
Ministral 3 14B (R) & 29.8\,{\tiny\color{gray}$\pm$13.4} & 16.2\,{\tiny\color{gray}$\pm$6.4} & 16.8\,{\tiny\color{gray}$\pm$19.7} & 0.2\,{\tiny\color{gray}$\pm$0.2} & 34.3 & 25.2\,{\tiny\color{gray}$\pm$2.5} & 20.4\,{\tiny\color{gray}$\pm$2.4} \\
Qwen3 14B & 67.1\,{\tiny\color{gray}$\pm$1.1} & 49.1\,{\tiny\color{gray}$\pm$6.5} & 74.3\,{\tiny\color{gray}$\pm$8.7} & 16.2\,{\tiny\color{gray}$\pm$0.4} & 34.1 & 44.0\,{\tiny\color{gray}$\pm$1.3} & 47.4\,{\tiny\color{gray}$\pm$2.4} \\
Qwen3-SK 14B & 43.4\,{\tiny\color{gray}$\pm$21.7} & 23.9\,{\tiny\color{gray}$\pm$8.4} & 43.7\,{\tiny\color{gray}$\pm$17.8} & 11.6\,{\tiny\color{gray}$\pm$0.6} & 27.7 & 50.8\,{\tiny\color{gray}$\pm$5.1} & 33.5\,{\tiny\color{gray}$\pm$10.2} \\
Qwen3-SK 14B (Alpaca) & 70.0\,{\tiny\color{gray}$\pm$0.7} & 45.6\,{\tiny\color{gray}$\pm$3.5} & 60.2\,{\tiny\color{gray}$\pm$8.3} & 15.0\,{\tiny\color{gray}$\pm$0.5} & 41.0 & 30.9\,{\tiny\color{gray}$\pm$1.1} & 43.8\,{\tiny\color{gray}$\pm$2.5} \\
gpt-oss 20B & 67.4\,{\tiny\color{gray}$\pm$0.5} & 47.6\,{\tiny\color{gray}$\pm$4.8} & 76.9\,{\tiny\color{gray}$\pm$1.2} & 13.0\,{\tiny\color{gray}$\pm$0.5} & 62.4 & 57.2\,{\tiny\color{gray}$\pm$0.8} & 54.1\,{\tiny\color{gray}$\pm$0.9} \\
EuroLLM 22B & 63.9\,{\tiny\color{gray}$\pm$1.6} & 41.6\,{\tiny\color{gray}$\pm$6.5} & 75.0\,{\tiny\color{gray}$\pm$5.7} & 15.2\,{\tiny\color{gray}$\pm$0.7} & 26.4 & 49.2\,{\tiny\color{gray}$\pm$3.1} & 45.2\,{\tiny\color{gray}$\pm$2.0} \\
Qwen3.5 27B & 81.2\,{\tiny\color{gray}$\pm$0.2} & 55.7\,{\tiny\color{gray}$\pm$6.3} & 79.5\,{\tiny\color{gray}$\pm$3.4} & 14.9\,{\tiny\color{gray}$\pm$0.4} & 36.1 & 64.6\,{\tiny\color{gray}$\pm$0.8} & 55.3\,{\tiny\color{gray}$\pm$1.8} \\
Qwen3.6 27B & 79.8\,{\tiny\color{gray}$\pm$0.5} & 59.7\,{\tiny\color{gray}$\pm$3.7} & 70.7\,{\tiny\color{gray}$\pm$10.7} & 14.9\,{\tiny\color{gray}$\pm$0.4} & 35.0 & 65.0\,{\tiny\color{gray}$\pm$1.8} & 54.2\,{\tiny\color{gray}$\pm$2.8} \\
TildeOpen 30B & 45.1\,{\tiny\color{gray}$\pm$2.6} & 21.5\,{\tiny\color{gray}$\pm$1.6} & 8.8\,{\tiny\color{gray}$\pm$0.5} & 3.4\,{\tiny\color{gray}$\pm$2.1} & 16.8 & 6.3\,{\tiny\color{gray}$\pm$4.3} & 17.0\,{\tiny\color{gray}$\pm$0.5} \\
TildeOpen 30B-64k & 58.0\,{\tiny\color{gray}$\pm$0.6} & 40.4\,{\tiny\color{gray}$\pm$3.3} & 62.8\,{\tiny\color{gray}$\pm$8.9} & 14.5\,{\tiny\color{gray}$\pm$1.9} & 15.6 & 0.2\,{\tiny\color{gray}$\pm$0.4} & 31.9\,{\tiny\color{gray}$\pm$2.3} \\
TildeOpen-SK 30B (Alpaca) & 57.4\,{\tiny\color{gray}$\pm$2.1} & 34.2\,{\tiny\color{gray}$\pm$3.2} & 33.6\,{\tiny\color{gray}$\pm$2.7} & 13.3\,{\tiny\color{gray}$\pm$0.9} & 27.2 & 30.9\,{\tiny\color{gray}$\pm$1.9} & 32.8\,{\tiny\color{gray}$\pm$0.8} \\
Gemma 4 31B & 78.8\,{\tiny\color{gray}$\pm$0.7} & 53.1\,{\tiny\color{gray}$\pm$8.7} & 82.4\,{\tiny\color{gray}$\pm$0.7} & 14.4\,{\tiny\color{gray}$\pm$0.5} & 30.2 & 66.7\,{\tiny\color{gray}$\pm$2.1} & 54.3\,{\tiny\color{gray}$\pm$1.3} \\
Aya Expanse 32B & 61.5\,{\tiny\color{gray}$\pm$0.9} & 41.9\,{\tiny\color{gray}$\pm$8.5} & 76.9\,{\tiny\color{gray}$\pm$2.5} & 14.2\,{\tiny\color{gray}$\pm$0.2} & 24.2 & 35.7\,{\tiny\color{gray}$\pm$1.6} & 42.4\,{\tiny\color{gray}$\pm$2.4} \\
Granite 4.0-H Small & 67.3\,{\tiny\color{gray}$\pm$0.3} & 42.4\,{\tiny\color{gray}$\pm$6.3} & 71.9\,{\tiny\color{gray}$\pm$6.2} & \underline{16.4}\,{\tiny\color{gray}$\pm$0.3} & 33.9 & 50.0\,{\tiny\color{gray}$\pm$1.3} & 47.0\,{\tiny\color{gray}$\pm$2.4} \\
\midrule
\multicolumn{8}{l}{\textit{32--200B}} \\
Qwen3.6 35B-A3B & 77.2\,{\tiny\color{gray}$\pm$0.4} & 55.4\,{\tiny\color{gray}$\pm$7.0} & 71.8\,{\tiny\color{gray}$\pm$9.9} & 14.5\,{\tiny\color{gray}$\pm$0.5} & 32.8 & 61.4\,{\tiny\color{gray}$\pm$2.5} & 52.2\,{\tiny\color{gray}$\pm$2.8} \\
Apertus 70B & 57.5\,{\tiny\color{gray}$\pm$1.2} & 43.3\,{\tiny\color{gray}$\pm$6.1} & 69.7\,{\tiny\color{gray}$\pm$3.8} & 13.2\,{\tiny\color{gray}$\pm$0.5} & 27.9 & 54.8\,{\tiny\color{gray}$\pm$3.6} & 44.4\,{\tiny\color{gray}$\pm$1.6} \\
Llama 3.3 70B & 72.1\,{\tiny\color{gray}$\pm$0.5} & 53.9\,{\tiny\color{gray}$\pm$4.0} & 79.4\,{\tiny\color{gray}$\pm$0.2} & \textbf{21.4}\,{\tiny\color{gray}$\pm$0.5} & 35.4 & 56.4\,{\tiny\color{gray}$\pm$1.0} & 53.1\,{\tiny\color{gray}$\pm$1.0} \\
Command A & 64.9\,{\tiny\color{gray}$\pm$3.0} & 32.4\,{\tiny\color{gray}$\pm$3.7} & 38.4\,{\tiny\color{gray}$\pm$5.2} & 12.4\,{\tiny\color{gray}$\pm$0.5} & 26.4 & 42.8\,{\tiny\color{gray}$\pm$6.5} & 36.2\,{\tiny\color{gray}$\pm$1.9} \\
gpt-oss 120B & 73.7\,{\tiny\color{gray}$\pm$1.0} & 53.2\,{\tiny\color{gray}$\pm$6.3} & 79.0\,{\tiny\color{gray}$\pm$0.5} & 13.5\,{\tiny\color{gray}$\pm$0.4} & 74.9 & 65.3\,{\tiny\color{gray}$\pm$1.0} & 60.0\,{\tiny\color{gray}$\pm$1.1} \\
\midrule
\multicolumn{8}{l}{\textit{$>$200B}} \\
MiniMax M2.7 & 78.0\,{\tiny\color{gray}$\pm$1.1} & 51.5\,{\tiny\color{gray}$\pm$4.6} & 78.4\,{\tiny\color{gray}$\pm$1.1} & 13.5\,{\tiny\color{gray}$\pm$0.1} & 44.7 & 76.6\,{\tiny\color{gray}$\pm$0.3} & 57.1\,{\tiny\color{gray}$\pm$1.2} \\
Qwen3.5 397B-A17B & 83.4\,{\tiny\color{gray}$\pm$0.5} & 55.5\,{\tiny\color{gray}$\pm$8.2} & 80.6\,{\tiny\color{gray}$\pm$0.8} & 14.6\,{\tiny\color{gray}$\pm$0.2} & 37.8 & 72.3\,{\tiny\color{gray}$\pm$3.0} & 57.4\,{\tiny\color{gray}$\pm$1.7} \\
Mistral Large 3 675B & 77.3\,{\tiny\color{gray}$\pm$0.2} & 49.3\,{\tiny\color{gray}$\pm$10.6} & 65.3\,{\tiny\color{gray}$\pm$5.6} & 11.8\,{\tiny\color{gray}$\pm$0.8} & 20.9 & 45.3\,{\tiny\color{gray}$\pm$5.7} & 45.0\,{\tiny\color{gray}$\pm$0.9} \\
GLM-5.1 & 80.6\,{\tiny\color{gray}$\pm$0.4} & 61.2\,{\tiny\color{gray}$\pm$6.9} & 82.6\,{\tiny\color{gray}$\pm$1.0} & 14.2\,{\tiny\color{gray}$\pm$0.2} & 35.9 & 70.8\,{\tiny\color{gray}$\pm$1.3} & 57.6\,{\tiny\color{gray}$\pm$1.2} \\
Kimi K2.6 & 80.6\,{\tiny\color{gray}$\pm$0.3} & 52.7\,{\tiny\color{gray}$\pm$7.9} & 78.0\,{\tiny\color{gray}$\pm$7.9} & 14.7\,{\tiny\color{gray}$\pm$0.4} & 31.8 & 76.2\,{\tiny\color{gray}$\pm$0.3} & 55.7\,{\tiny\color{gray}$\pm$3.0} \\
\midrule
\multicolumn{8}{l}{\textit{Proprietary (size undisclosed)}} \\
Gemini 3.1 Pro & \textbf{92.4}\,{\tiny\color{gray}$\pm$0.2} & \textbf{71.3}\,{\tiny\color{gray}$\pm$4.5} & \underline{82.6}\,{\tiny\color{gray}$\pm$0.4} & 14.8\,{\tiny\color{gray}$\pm$0.0} & \textbf{87.2} & \textbf{87.6}\,{\tiny\color{gray}$\pm$0.4} & \textbf{72.6}\,{\tiny\color{gray}$\pm$0.9} \\
GPT-5.5 & \underline{89.2}\,{\tiny\color{gray}$\pm$0.3} & \underline{67.3}\,{\tiny\color{gray}$\pm$6.8} & 81.6\,{\tiny\color{gray}$\pm$0.3} & 15.5\,{\tiny\color{gray}$\pm$0.3} & \underline{86.4} & \underline{86.1}\,{\tiny\color{gray}$\pm$0.3} & \underline{71.0}\,{\tiny\color{gray}$\pm$1.2} \\
Claude Opus 4.6 & 87.2\,{\tiny\color{gray}$\pm$0.5} & 58.6\,{\tiny\color{gray}$\pm$10.2} & \textbf{83.4}\,{\tiny\color{gray}$\pm$0.4} & 15.4\,{\tiny\color{gray}$\pm$0.2} & 80.6 & 83.6\,{\tiny\color{gray}$\pm$0.5} & 68.1\,{\tiny\color{gray}$\pm$2.0} \\
Gemini 3.1 Flash-Lite & 85.7\,{\tiny\color{gray}$\pm$0.6} & 54.8\,{\tiny\color{gray}$\pm$11.2} & 82.2\,{\tiny\color{gray}$\pm$0.7} & 14.0\,{\tiny\color{gray}$\pm$0.3} & 82.7 & 80.8\,{\tiny\color{gray}$\pm$0.5} & 66.7\,{\tiny\color{gray}$\pm$2.4} \\
Claude Sonnet 4.6 & 84.7\,{\tiny\color{gray}$\pm$0.3} & 54.0\,{\tiny\color{gray}$\pm$9.6} & 81.6\,{\tiny\color{gray}$\pm$1.9} & 15.3\,{\tiny\color{gray}$\pm$0.3} & 78.0 & 75.8\,{\tiny\color{gray}$\pm$0.6} & 64.9\,{\tiny\color{gray}$\pm$1.9} \\
GPT-5.4 mini & 78.2\,{\tiny\color{gray}$\pm$0.5} & 58.0\,{\tiny\color{gray}$\pm$9.1} & 79.0\,{\tiny\color{gray}$\pm$1.1} & 15.2\,{\tiny\color{gray}$\pm$0.4} & 80.1 & 73.0\,{\tiny\color{gray}$\pm$2.2} & 63.9\,{\tiny\color{gray}$\pm$1.5} \\
\end{longtable}
\endgroup
\twocolumn

\begin{table*}[t]
\centering
\resizebox{\linewidth}{!}{%
\begin{tabular}{lrrrrrrrrrrrrrrrrrrrrrrrrrrrrrrrrr|r}
\toprule
\textbf{Model} & \rotatebox{90}{\hyperref[task:arc-sk]{ARC}} & \rotatebox{90}{\hyperref[task:mmlu-sk]{MMLU}} & \rotatebox{90}{\hyperref[task:gmmlu]{G-MMLU}} & \rotatebox{90}{\hyperref[task:truthqa-sk]{TruthQA}} & \rotatebox{90}{\hyperref[task:scio]{SCIO}} & \rotatebox{90}{\hyperref[task:finexam]{FinExam}} & \rotatebox{90}{\hyperref[task:autoskola]{Driving}} & \rotatebox{90}{\hyperref[task:copa]{COPA}} & \rotatebox{90}{\hyperref[task:skj1]{SKJ1}} & \rotatebox{90}{\hyperref[task:trivia]{Trivia}} & \rotatebox{90}{\hyperref[task:belebele]{Belebele}} & \rotatebox{90}{\hyperref[task:chiby]{Chiby}} & \rotatebox{90}{\hyperref[task:wikiqa]{WikiQA}} & \rotatebox{90}{\hyperref[task:etest]{ETest-M}} & \rotatebox{90}{\hyperref[task:etest]{ETest-Q}} & \rotatebox{90}{\hyperref[task:sktest]{SKTest-M}} & \rotatebox{90}{\hyperref[task:sktest]{SKTest-Q}} & \rotatebox{90}{\hyperref[task:oneruler]{OneRuler}} & \rotatebox{90}{\hyperref[task:sklep-qa]{skLEP-QA}} & \rotatebox{90}{\hyperref[task:nli-sk]{NLI}} & \rotatebox{90}{\hyperref[task:sklep-nli]{skLEP-NLI}} & \rotatebox{90}{\hyperref[task:sklep-rte]{skLEP-RTE}} & \rotatebox{90}{\hyperref[task:senti]{Senti-3}} & \rotatebox{90}{\hyperref[task:senti]{Senti-2}} & \rotatebox{90}{\hyperref[task:sklep-sent]{skLEP-Sent}} & \rotatebox{90}{\hyperref[task:movierev]{MovieRev}} & \rotatebox{90}{\hyperref[task:toxic]{Toxic}} & \rotatebox{90}{\hyperref[task:sklep-hate]{skLEP-Hate}} & \rotatebox{90}{\hyperref[task:slovaksum]{SlovakSum}} & \rotatebox{90}{\hyperref[task:smesum]{SME-Sum}} & \rotatebox{90}{\hyperref[task:ifeval]{IFEval}} & \rotatebox{90}{\hyperref[task:analogies]{Analogies}} & \rotatebox{90}{\hyperref[task:slohry]{Slohry}} & \rotatebox{90}{\textbf{Avg}} \\
\midrule
\multicolumn{35}{l}{\textit{$<$2B}} \\
Granite 4.0 350M & 26.5 & 27.7 & 26.7 & 21.9 & 25.9 & 25.9 & 33.2 & 48.7 & 12.7 & 26.1 & 30.0 & 0.9 & 3.6 & 19.4 & 0.2 & 25.4 & 1.3 & 7.2 & 3.0 & 7.3 & 7.1 & 5.2 & 28.5 & 22.0 & 23.8 & 25.6 & 3.1 & 1.4 & 9.8 & 9.6 & 10.8 & 0.3 & 3.2 & 15.9 \\
Granite 4.0-H 350M & 24.1 & 26.6 & 25.6 & 13.8 & 21.8 & 22.3 & 32.2 & 50.5 & 15.3 & 22.4 & 26.7 & 2.7 & 1.7 & 20.0 & 0.2 & 26.9 & 0.8 & 0.5 & 4.9 & 4.0 & 3.9 & 14.9 & 43.3 & 67.4 & 10.3 & 52.5 & 41.2 & 8.2 & 10.8 & 12.6 & 15.3 & 0.0 & 2.6 & 19.0 \\
Qwen3 0.6B & 25.9 & 25.9 & 26.2 & 65.3 & 21.4 & 31.0 & 61.1 & 49.4 & 3.3 & 21.1 & 34.2 & 26.8 & 1.2 & 20.9 & 0.5 & 29.0 & 1.5 & 56.0 & 17.6 & 23.4 & 21.2 & 36.1 & 28.3 & 75.8 & 33.8 & 58.6 & 56.3 & 56.6 & 11.6 & 9.7 & 16.8 & 2.6 & 5.9 & 28.9 \\
Qwen3.5 0.8B & 41.1 & 33.1 & 35.8 & 28.5 & 28.7 & 34.5 & 54.7 & 50.1 & 2.7 & 24.6 & 47.1 & 43.9 & 14.9 & 25.4 & 0.9 & 36.8 & 1.3 & 50.5 & 13.4 & 24.7 & 22.8 & 56.7 & \underline{67.5} & 85.2 & 57.6 & 69.0 & 57.7 & 40.2 & 10.9 & 8.5 & 14.3 & 7.3 & 6.7 & 33.2 \\
Gemma 3 1B & 27.3 & 28.4 & 32.8 & 52.8 & 23.5 & 35.0 & 70.0 & 51.8 & 2.7 & 26.9 & 34.0 & 34.7 & 13.7 & 21.7 & 2.6 & 35.9 & 2.1 & 13.9 & 20.1 & 22.2 & 24.2 & 51.7 & 65.2 & 81.5 & 82.0 & 73.1 & 56.1 & 33.3 & 12.0 & 11.7 & 12.3 & 2.0 & 4.0 & 32.2 \\
Granite 4.0 1B & 41.5 & 38.2 & 39.4 & 19.2 & 35.2 & 36.1 & 47.3 & 51.8 & 2.3 & 33.6 & 61.4 & 51.0 & 8.5 & 28.5 & 1.6 & 40.2 & 2.1 & 41.2 & 13.9 & 34.2 & 34.9 & 60.8 & 30.8 & 59.2 & 30.9 & 47.9 & 27.4 & 50.6 & 11.9 & 14.5 & 17.1 & 4.0 & 6.5 & 31.0 \\
Granite 4.0-H 1B & 49.1 & 40.6 & 44.8 & 18.6 & 35.4 & 42.3 & 43.1 & 51.1 & 41.3 & 31.6 & 54.6 & 49.8 & 6.1 & 32.1 & 1.3 & 40.9 & 2.7 & 49.4 & 17.1 & 27.1 & 24.7 & 63.5 & 38.4 & 55.5 & 90.3 & 70.4 & 40.7 & 40.0 & 13.3 & 16.3 & 14.9 & 3.4 & 8.6 & 35.1 \\
EuroLLM 1.7B & 25.9 & 24.3 & 24.4 & 45.6 & 22.4 & 18.8 & 57.1 & 50.9 & 57.3 & 24.8 & 27.9 & 50.0 & 17.3 & 19.0 & 5.5 & 27.5 & 3.8 & 5.6 & 16.2 & 14.7 & 17.8 & 50.2 & 19.3 & 24.4 & 74.8 & 51.1 & 46.5 & 63.9 & 13.1 & 13.8 & 14.7 & 6.9 & 9.9 & 28.7 \\
Qwen3 1.7B & 40.2 & 31.1 & 36.1 & 37.7 & 19.6 & 40.0 & 62.0 & 51.1 & 14.0 & 18.2 & 52.5 & 50.4 & 15.2 & 22.8 & 2.7 & 34.8 & 2.4 & 80.2 & 17.2 & 14.9 & 15.7 & 47.0 & 55.3 & 70.6 & 71.5 & 70.6 & 61.3 & 65.0 & 12.7 & 13.0 & 23.5 & 6.9 & 17.8 & 35.6 \\
\midrule
\multicolumn{35}{l}{\textit{2--9B}} \\
Granite 3.2 2B & 41.1 & 38.2 & 39.7 & 14.4 & 30.1 & 47.5 & 49.4 & 55.7 & 38.3 & 28.0 & 55.1 & 24.2 & 3.9 & 29.1 & 1.4 & 39.7 & 1.5 & 33.2 & 14.1 & 11.0 & 11.8 & 44.7 & 57.6 & 76.4 & 93.3 & 85.1 & 41.4 & 44.1 & 9.5 & 7.8 & 13.8 & 9.5 & 13.7 & 33.5 \\
Qwen3.5 2B & 60.6 & 44.5 & 51.6 & 22.9 & 41.1 & 47.6 & 53.1 & 55.6 & 20.7 & 36.2 & 71.1 & 56.9 & 21.1 & 33.9 & 4.4 & 44.8 & 3.9 & 47.8 & 23.5 & 45.4 & 46.6 & 72.5 & 56.3 & 75.2 & 44.8 & 54.2 & 63.9 & 53.3 & 11.8 & 10.5 & 18.8 & 7.1 & 17.2 & 40.0 \\
Llama 3.2 3B & 40.1 & 38.1 & 43.2 & 11.0 & 33.9 & 45.6 & 37.9 & 49.1 & 15.7 & 32.9 & 52.4 & 18.3 & 23.6 & 29.5 & 9.3 & 40.5 & 3.2 & 34.1 & 21.2 & 23.2 & 23.0 & 49.7 & 64.7 & 85.7 & 46.1 & 69.3 & 32.1 & 21.7 & 13.9 & 15.6 & 14.0 & 12.0 & 12.9 & 32.2 \\
Tiny-Aya Global & 64.9 & 49.0 & 54.2 & 46.2 & 40.9 & 58.5 & 67.8 & 63.7 & 20.3 & 41.4 & 78.4 & 50.6 & 12.4 & 37.0 & 8.8 & 53.2 & 4.9 & 54.9 & 11.6 & 61.5 & 61.2 & 74.7 & 55.6 & 83.4 & 91.7 & 83.4 & 77.3 & 76.0 & 11.4 & 9.9 & 26.8 & 11.5 & 27.7 & 47.6 \\
Tiny-Aya Water & 66.1 & 49.6 & 54.6 & 43.4 & 41.3 & 59.7 & 68.0 & 62.6 & 17.7 & 41.9 & 79.0 & 50.7 & 11.7 & 35.9 & 9.5 & 53.8 & 4.3 & 57.4 & 11.4 & 63.2 & 63.6 & 74.6 & 55.4 & 83.2 & 92.7 & 83.0 & 75.4 & 77.5 & 11.5 & 10.1 & 27.6 & 13.8 & 26.9 & 47.8 \\
Mistral-SK 7B & 23.8 & 22.9 & 26.6 & 25.3 & 20.6 & 24.0 & 31.1 & 34.6 & 3.0 & 14.4 & 32.1 & 43.3 & 19.7 & 17.2 & 17.1 & 24.8 & 3.8 & 0.0 & 15.7 & 22.6 & 20.4 & 46.2 & 25.0 & 23.4 & 50.8 & 35.1 & 50.7 & 37.5 & 13.5 & \underline{18.8} & 14.2 & 13.8 & 0.0 & 23.4 \\
Mistral-SK 7B (Alpaca) & 14.3 & 12.3 & 16.2 & 30.2 & 10.8 & 15.6 & 46.9 & 57.3 & 4.7 & 7.0 & 22.7 & 5.5 & 13.3 & 11.0 & 0.7 & 17.0 & 2.0 & 35.6 & 5.8 & 18.8 & 20.2 & 30.0 & 11.1 & 15.2 & 12.2 & 22.6 & 8.5 & 21.3 & 12.8 & 11.1 & 13.8 & 13.8 & 9.8 & 16.7 \\
Teuken 7B & 49.9 & 34.9 & 39.7 & 32.6 & 23.4 & 54.4 & 66.0 & 57.5 & 6.7 & 33.5 & 66.6 & 53.8 & 4.2 & 26.9 & 4.3 & 39.7 & 4.2 & 17.9 & 16.8 & 22.4 & 20.3 & 60.2 & 59.7 & 79.8 & 90.2 & 80.6 & 46.7 & 71.9 & 10.6 & 10.0 & 15.3 & 9.6 & 21.1 & 37.3 \\
Aya Expanse 8B & 61.4 & 46.5 & 51.2 & 36.5 & 39.3 & 54.6 & 58.0 & 33.5 & 29.0 & 35.2 & 78.7 & 15.5 & 16.8 & 32.7 & 6.5 & 52.6 & 3.6 & 13.2 & 12.5 & 36.5 & 38.7 & 59.2 & 47.8 & 65.9 & 88.3 & 87.7 & 48.5 & 57.9 & 12.2 & 12.5 & 15.1 & 14.8 & 14.7 & 38.7 \\
Llama 3.1 8B & 56.5 & 47.9 & 53.9 & 36.4 & 41.2 & 58.3 & 53.3 & 58.1 & 36.3 & 35.9 & 74.1 & 50.1 & \underline{35.1} & 40.5 & 13.8 & 49.7 & 8.5 & 68.3 & 32.6 & 34.6 & 32.2 & 57.5 & 54.7 & 75.9 & 90.8 & 86.7 & 53.0 & 63.1 & 14.7 & 16.0 & 19.6 & 24.5 & 27.1 & 45.5 \\
Ministral 3 8B & 78.6 & 59.7 & 64.9 & 45.4 & 53.4 & 68.7 & 71.5 & 67.3 & 47.0 & 47.0 & 85.2 & 54.2 & 9.8 & 50.2 & 12.7 & 61.2 & 6.2 & 58.0 & 15.4 & 54.7 & 50.0 & 74.7 & 51.6 & 72.4 & 96.2 & 89.5 & 81.5 & 65.0 & 13.1 & 14.0 & 24.0 & 32.3 & 41.9 & 52.0 \\
Ministral 3 8B (R) & 70.6 & 52.8 & 57.9 & 45.2 & 43.7 & 59.8 & 63.8 & 51.1 & 67.0 & 36.5 & 78.7 & 38.5 & 10.8 & 46.1 & 13.1 & 54.5 & 5.6 & 74.1 & 19.2 & 46.8 & 46.0 & 65.4 & 47.4 & 67.4 & 87.8 & 77.9 & 46.9 & 71.7 & 12.9 & 13.2 & 39.9 & 28.0 & 36.2 & 47.8 \\
\midrule
\multicolumn{35}{l}{\textit{9--32B}} \\
EuroLLM 9B & 75.7 & 54.5 & 62.5 & 40.2 & 45.4 & 63.2 & 69.1 & 70.9 & 30.3 & 46.1 & 82.5 & 58.4 & 19.8 & 45.7 & 16.7 & 63.0 & 8.1 & 63.1 & 27.3 & 57.7 & 57.0 & 78.3 & \textbf{68.4} & \textbf{88.8} & 87.1 & 81.6 & 81.7 & 70.6 & 13.9 & 16.0 & 22.7 & 40.4 & 50.0 & 53.2 \\
Qwen3.5 9B & 88.7 & 67.6 & 72.2 & 64.8 & 61.3 & 73.9 & 78.3 & 81.0 & 61.0 & 47.9 & 89.3 & 61.6 & 18.4 & 62.0 & 20.2 & 65.8 & 12.9 & 83.2 & 39.8 & 65.9 & 68.3 & 83.5 & 60.1 & 82.5 & 95.4 & 90.9 & 69.1 & 77.1 & 13.6 & 16.3 & 31.1 & 42.1 & 46.7 & 60.4 \\
Bielik 11B & 83.7 & 61.3 & 68.4 & 66.7 & 53.5 & 73.5 & 75.4 & 85.9 & 45.3 & 51.7 & 89.9 & 59.3 & 7.0 & 55.7 & 25.4 & 66.2 & 16.4 & 86.7 & 17.3 & 66.2 & 68.2 & 80.4 & 50.8 & 72.3 & 81.0 & 86.2 & \textbf{90.4} & 76.3 & 12.9 & 12.3 & 30.4 & 41.9 & 51.8 & 57.9 \\
Ministral 3 14B & 82.3 & 62.5 & 66.2 & 51.0 & 51.0 & 72.9 & 73.0 & 68.7 & 61.7 & 48.9 & 87.3 & 58.4 & 5.7 & 57.9 & 13.1 & 63.5 & 4.8 & 4.2 & 16.3 & 58.3 & 56.8 & 77.6 & 23.8 & 40.7 & 49.6 & 46.3 & 78.6 & 71.7 & 13.4 & 15.0 & 18.4 & 36.7 & 36.2 & 47.6 \\
Ministral 3 14B (R) & 48.0 & 28.9 & 33.4 & 25.0 & 22.6 & 12.6 & 39.3 & 35.3 & 24.7 & 28.2 & 52.5 & 11.7 & 0.4 & 35.3 & 3.2 & 40.0 & 0.6 & 0.0 & 2.4 & 15.4 & 14.6 & 26.2 & 11.6 & 15.0 & 17.8 & 8.2 & 28.8 & 13.6 & 0.2 & 0.2 & 34.3 & 11.3 & 39.2 & 20.6 \\
Qwen3 14B & 85.7 & 65.5 & 65.1 & 63.1 & 59.4 & 70.9 & 79.6 & 76.5 & 59.7 & 45.2 & 88.7 & 63.4 & 23.3 & 58.9 & 16.0 & 62.5 & 10.1 & 85.1 & 33.5 & 50.0 & 54.6 & 83.4 & 58.1 & 76.6 & 94.3 & 90.0 & 84.3 & 77.4 & 14.2 & 18.2 & 34.1 & 41.4 & 46.5 & 58.6 \\
Qwen3-SK 14B & 39.1 & 33.3 & 35.0 & 52.3 & 18.5 & 69.9 & 31.3 & 79.0 & 55.0 & 20.8 & 88.7 & 0.1 & 27.5 & 23.3 & 4.6 & 40.6 & 5.5 & 1.0 & 23.7 & 21.9 & 29.1 & 54.1 & 39.9 & 54.7 & 38.5 & 66.4 & 25.8 & 62.7 & 11.9 & 11.2 & 27.7 & 49.2 & 52.3 & 36.2 \\
Qwen3-SK 14B (Alpaca) & 86.7 & 66.2 & 68.9 & 70.7 & 59.3 & 71.0 & 81.9 & 80.5 & 65.0 & 50.1 & 89.7 & 63.5 & 10.7 & 58.7 & 18.4 & 67.6 & 9.9 & 83.9 & 7.6 & 68.0 & 67.4 & 82.8 & 32.6 & 51.0 & 15.3 & 59.0 & 87.3 & 78.0 & 14.5 & 15.5 & 41.0 & 29.3 & 32.5 & 54.1 \\
gpt-oss 20B & 88.1 & 72.6 & 74.8 & 58.7 & 67.1 & 65.5 & 72.8 & 83.3 & 56.7 & 34.4 & 89.6 & 62.6 & 14.8 & 61.5 & 15.6 & 65.0 & 9.4 & 85.1 & 24.6 & 71.0 & 72.1 & 80.7 & 56.1 & 72.5 & 96.1 & 87.3 & 82.1 & 73.9 & 12.5 & 13.5 & 62.4 & 42.5 & 71.9 & 60.5 \\
EuroLLM 22B & 80.6 & 59.2 & 68.0 & 58.2 & 48.4 & 69.2 & 76.9 & 77.5 & 47.3 & 53.4 & 83.4 & 53.9 & 12.2 & 56.3 & 16.0 & 66.5 & 9.9 & 54.0 & 22.0 & 54.4 & 50.9 & 78.8 & 61.4 & 85.8 & 94.6 & 88.6 & 86.2 & 74.3 & 14.2 & 16.2 & 26.4 & 44.9 & 53.6 & 55.9 \\
Qwen3.5 27B & 94.5 & 80.0 & 84.9 & 76.8 & 69.1 & 83.2 & 87.8 & 90.5 & 82.7 & 62.1 & 93.7 & 69.0 & 16.6 & 73.6 & 27.8 & 77.4 & 18.2 & 83.3 & 41.3 & 70.1 & 68.3 & 79.9 & 61.3 & 80.0 & 97.3 & 93.6 & 86.5 & \underline{78.9} & 13.7 & 16.2 & 36.1 & 54.7 & 74.5 & 67.4 \\
Qwen3.6 27B & 93.7 & 79.6 & 83.3 & 77.8 & 69.3 & 83.6 & 86.9 & 88.5 & 76.7 & 58.1 & 94.0 & 67.4 & 34.4 & 71.9 & 39.7 & 76.1 & 21.6 & 83.3 & \underline{48.8} & 29.6 & 27.9 & \textbf{85.9} & 62.5 & 83.5 & 97.1 & 93.6 & 81.2 & 75.1 & 13.7 & 16.2 & 35.0 & 54.6 & 75.4 & 65.6 \\
TildeOpen 30B & 57.7 & 42.1 & 47.5 & 45.2 & 27.4 & 46.5 & 74.4 & 32.1 & 51.3 & 26.4 & 66.0 & 17.4 & 10.1 & 32.6 & 9.6 & 51.3 & 2.4 & 0.0 & 4.4 & 6.6 & 1.7 & 13.3 & 1.4 & 0.2 & 0.2 & 1.1 & 39.7 & 14.7 & 4.2 & 2.6 & 16.8 & 12.6 & 0.0 & 23.0 \\
TildeOpen 30B-64k & 69.4 & 51.4 & 59.9 & 42.4 & 41.3 & 60.6 & 81.0 & 58.5 & 62.7 & 52.9 & 74.3 & 52.4 & \textbf{41.7} & 41.8 & 43.5 & 64.0 & 22.1 & 0.0 & 23.7 & 46.5 & 42.8 & 66.4 & 60.6 & 70.4 & 89.1 & 59.2 & 60.8 & 69.1 & 13.2 & 15.8 & 15.6 & 0.5 & 0.0 & 47.1 \\
TildeOpen-SK 30B (Alpaca) & 72.6 & 50.8 & 59.1 & 43.5 & 44.2 & 58.5 & 76.7 & 71.3 & 49.0 & 48.7 & 77.4 & 24.8 & 6.9 & 43.6 & 19.4 & 64.1 & 9.5 & 43.5 & 18.5 & 38.4 & 35.4 & 63.4 & 8.3 & 10.5 & 3.7 & 20.7 & 48.7 & 73.4 & 13.4 & 13.2 & 27.2 & 20.1 & 41.7 & 39.4 \\
Gemma 4 31B & 94.7 & 79.6 & 86.9 & 78.7 & 63.4 & 84.5 & 87.9 & 85.9 & 66.3 & 59.7 & 94.1 & 66.9 & 13.1 & 70.7 & 27.9 & 77.7 & 14.8 & 83.3 & 29.6 & 81.2 & 82.6 & 84.5 & 60.3 & 75.9 & 97.1 & 93.3 & 88.3 & 78.1 & 13.8 & 15.0 & 30.2 & 54.0 & 79.4 & 66.7 \\
Aya Expanse 32B & 78.0 & 59.4 & 63.9 & 55.3 & 48.2 & 68.9 & 71.0 & 81.1 & 43.3 & 45.7 & 88.7 & 57.8 & 17.8 & 43.9 & 13.3 & 63.3 & 4.9 & 65.4 & 22.3 & 67.6 & 69.5 & 75.1 & 57.6 & 77.5 & 94.7 & 91.0 & 87.6 & 71.8 & 13.7 & 14.7 & 24.2 & 32.8 & 38.7 & 54.8 \\
Granite 4.0-H Small & 84.2 & 65.6 & 73.8 & 63.4 & 56.5 & 69.3 & 74.3 & 83.5 & 54.3 & 47.7 & 86.1 & 53.5 & 15.0 & 57.2 & 13.8 & 66.3 & 8.1 & 56.9 & 24.9 & 46.7 & 48.8 & 82.7 & 55.1 & 76.4 & 95.7 & 89.1 & 79.7 & 72.8 & \underline{14.7} & 18.1 & 33.9 & 44.2 & 55.9 & 56.6 \\
\midrule
\multicolumn{35}{l}{\textit{32--200B}} \\
Qwen3.6 35B-A3B & 92.3 & 76.3 & 81.5 & 76.0 & 65.8 & 80.7 & 83.5 & 85.5 & 76.3 & 53.7 & 92.5 & 69.8 & 23.7 & 69.7 & 31.0 & 75.6 & 14.4 & 83.4 & 38.9 & 40.2 & 36.1 & 85.5 & 55.6 & 77.8 & 97.3 & 93.0 & 84.6 & 76.4 & 13.6 & 15.3 & 32.8 & 51.2 & 71.6 & 63.7 \\
Apertus 70B & 71.3 & 53.5 & 62.8 & 43.2 & 45.8 & 67.8 & 67.1 & 73.0 & 43.0 & 47.6 & 78.8 & 50.7 & 19.2 & 45.7 & 17.2 & 60.2 & 12.5 & 80.5 & 24.8 & 47.2 & 42.6 & 70.4 & 54.8 & 74.8 & 95.9 & 89.2 & 79.6 & 72.9 & 13.2 & 13.2 & 27.9 & 48.7 & 60.8 & 53.2 \\
Llama 3.3 70B & 87.5 & 72.3 & 80.6 & 66.0 & 59.7 & 76.2 & 83.8 & 87.3 & 57.0 & 50.2 & 91.6 & 62.2 & 29.1 & 63.6 & 33.3 & 69.4 & 16.2 & 81.2 & 38.1 & 68.2 & 72.3 & 81.2 & 57.6 & 82.0 & 96.9 & 92.7 & 85.1 & \textbf{79.0} & \textbf{16.1} & \textbf{26.6} & 35.4 & 47.2 & 65.6 & 64.0 \\
Command A & 88.0 & 68.2 & 72.6 & 65.3 & 57.2 & 78.1 & 79.4 & 19.7 & 61.7 & 59.0 & 92.3 & 7.0 & 4.0 & 53.2 & 10.4 & 67.7 & 3.6 & 50.5 & 2.7 & 17.3 & 20.3 & 59.0 & 31.0 & 29.7 & 36.3 & 35.9 & 47.7 & 68.8 & 11.6 & 13.2 & 26.4 & 49.6 & 36.1 & 43.1 \\
gpt-oss 120B & 90.6 & 77.9 & 81.0 & 65.9 & 71.6 & 74.4 & 76.0 & 83.9 & 64.7 & 51.4 & 92.2 & 65.0 & 14.6 & 71.0 & 27.9 & 71.7 & 17.5 & 90.9 & 28.4 & 72.8 & 75.8 & 83.2 & 57.8 & 75.6 & 96.4 & 89.8 & 84.8 & 75.3 & 12.9 & 14.2 & 74.9 & 50.5 & 80.2 & 65.5 \\
\midrule
\multicolumn{35}{l}{\textit{$>$200B}} \\
MiniMax M2.7 & 93.6 & 82.9 & 86.9 & 73.1 & 71.7 & 81.6 & 83.5 & 83.1 & 63.3 & 59.9 & 93.0 & 59.5 & 21.4 & 73.1 & 42.4 & 76.0 & 22.8 & 34.7 & 40.6 & 72.6 & 75.4 & 81.7 & 57.2 & 71.7 & 96.8 & 92.4 & 83.5 & 74.4 & 13.0 & 14.1 & 44.7 & 63.0 & 90.2 & 65.9 \\
Qwen3.5 397B-A17B & 94.9 & 82.8 & 90.1 & 80.8 & 73.6 & 87.8 & 86.3 & 90.1 & 79.0 & 68.7 & 94.6 & 72.2 & 9.9 & 78.0 & 33.9 & 80.5 & 19.9 & 83.4 & 27.2 & 78.6 & 77.2 & 82.5 & 61.7 & 81.3 & 96.3 & 94.0 & 78.2 & 75.2 & 13.5 & 15.8 & 37.8 & 61.7 & 82.8 & 68.8 \\
Mistral Large 3 675B & 90.8 & 77.0 & 83.7 & 72.7 & 65.4 & 82.2 & 83.9 & 89.6 & 64.3 & 63.2 & 93.0 & 61.8 & 16.7 & 62.0 & 24.8 & 76.8 & 11.1 & 73.2 & 24.1 & 71.9 & 68.7 & 83.0 & 28.8 & 35.8 & 85.3 & 47.4 & 88.2 & 78.2 & 11.2 & 12.3 & 20.9 & 58.0 & 32.6 & 58.7 \\
GLM-5.1 & 92.3 & 78.7 & 86.2 & 77.3 & 68.6 & 84.8 & 87.7 & 89.8 & 76.3 & 63.9 & 94.0 & 66.5 & 28.0 & 74.3 & 40.8 & 78.7 & 22.0 & 99.3 & 46.7 & 76.9 & 78.1 & 85.0 & 61.9 & \underline{85.9} & 96.6 & 92.5 & 87.9 & 78.6 & 13.5 & 15.0 & 35.9 & 59.7 & 82.0 & 69.9 \\
Kimi K2.6 & 93.1 & 78.7 & 84.8 & 78.6 & 64.7 & 87.0 & 85.7 & 90.2 & 81.0 & 62.4 & 94.0 & 65.2 & 23.4 & 73.4 & 38.1 & 77.6 & 18.3 & 45.4 & 39.2 & 60.2 & 57.3 & 84.1 & 60.8 & 82.8 & 97.0 & 93.1 & 88.6 & 77.9 & 13.8 & 15.7 & 31.8 & 62.6 & 89.8 & 66.6 \\
\midrule
\multicolumn{35}{l}{\textit{Proprietary (size undisclosed)}} \\
Gemini 3.1 Pro & \textbf{97.0} & \textbf{91.5} & \textbf{94.0} & \textbf{94.4} & \textbf{82.0} & \textbf{96.3} & \textbf{98.3} & \textbf{91.5} & \underline{84.3} & \textbf{94.3} & \textbf{95.8} & 74.4 & 31.8 & \textbf{88.7} & \textbf{73.7} & \textbf{89.6} & \textbf{37.5} & \textbf{100.0} & \textbf{49.9} & \underline{82.5} & \underline{85.2} & \underline{85.8} & 62.2 & 77.3 & \underline{97.8} & \textbf{95.7} & 83.6 & 73.8 & 13.9 & 15.7 & \textbf{87.2} & \textbf{76.8} & \textbf{98.5} & \textbf{78.8} \\
GPT-5.5 & \underline{96.5} & \underline{91.2} & \underline{92.9} & 84.8 & \underline{81.1} & \underline{96.0} & \underline{96.1} & 77.8 & 83.3 & \underline{92.4} & \underline{95.0} & \underline{74.5} & 18.3 & \underline{87.9} & \underline{67.6} & \underline{88.8} & \underline{35.6} & 99.9 & 38.2 & 79.7 & 81.3 & 83.7 & 61.3 & 76.0 & \textbf{98.3} & \underline{95.2} & 85.1 & 74.0 & 14.1 & 16.9 & \underline{86.4} & \underline{74.1} & \underline{98.0} & \underline{76.4} \\
Claude Opus 4.6 & 94.8 & 88.2 & 91.5 & \underline{91.4} & 74.3 & 93.2 & 91.6 & 90.8 & 80.3 & 76.2 & 94.8 & 73.2 & 16.8 & 79.0 & 33.9 & 83.9 & 14.8 & 99.8 & 31.2 & \textbf{83.2} & \textbf{85.3} & 80.9 & 59.9 & 79.7 & 97.4 & 94.8 & \underline{90.3} & 78.8 & 14.6 & 16.1 & 80.6 & 71.4 & 95.7 & 73.6 \\
Gemini 3.1 Flash-Lite & 94.8 & 84.2 & 89.9 & 81.2 & 71.1 & 89.7 & 91.5 & 90.7 & \textbf{84.7} & 79.7 & 93.9 & \textbf{75.0} & 11.1 & 80.1 & 17.8 & 85.5 & 9.9 & 99.8 & 20.4 & 81.2 & 81.7 & 82.8 & 61.3 & 78.5 & 97.4 & 94.1 & 85.9 & 76.5 & 13.5 & 14.5 & 82.7 & 67.9 & 93.7 & 71.6 \\
Claude Sonnet 4.6 & 94.0 & 85.3 & 89.6 & 80.6 & 73.2 & 89.6 & 88.5 & \underline{90.9} & 82.0 & 73.4 & 93.3 & 68.8 & 14.7 & 59.9 & 32.6 & 81.9 & 12.3 & \underline{100.0} & 22.3 & 80.6 & 83.0 & 80.1 & 60.3 & 81.3 & 95.4 & 91.2 & 85.1 & 77.7 & 14.4 & 16.1 & 78.0 & 63.1 & 88.5 & 70.5 \\
GPT-5.4 mini & 91.3 & 74.7 & 82.2 & 70.6 & 64.0 & 84.3 & 87.0 & 90.6 & 78.0 & 59.3 & 92.0 & 71.0 & 19.3 & 72.5 & 32.4 & 80.4 & 17.9 & 99.9 & 37.0 & 68.2 & 67.5 & 81.4 & 58.8 & 82.5 & 96.7 & 91.5 & 86.4 & 77.7 & 13.9 & 16.5 & 80.1 & 60.4 & 85.5 & 68.8 \\
\bottomrule
\end{tabular}%
}
\caption{sk-bench results across all tasks. Slovak (SK), 0-shot. Each cell is the task's primary metric ($\times100$): accuracy for multiple-choice and classification, quasi-exact-match (qem) for open QA, ROUGE-L for summarisation, and prompt-level strict accuracy for IFEval. \textbf{Best} and \underline{second-best} per column. Avg is the unweighted mean of the columns shown.}
\label{tab:skbench-comprehensive}
\end{table*}

\begin{table*}[t]
\centering
\resizebox{\linewidth}{!}{%
\begin{tabular}{lrrrrrrrrrrrrrrrrrrrrrrrrrrrrrrrrr|r}
\toprule
\textbf{Model} & \rotatebox{90}{\hyperref[task:arc-sk]{ARC}} & \rotatebox{90}{\hyperref[task:mmlu-sk]{MMLU}} & \rotatebox{90}{\hyperref[task:gmmlu]{G-MMLU}} & \rotatebox{90}{\hyperref[task:truthqa-sk]{TruthQA}} & \rotatebox{90}{\hyperref[task:scio]{SCIO}} & \rotatebox{90}{\hyperref[task:finexam]{FinExam}} & \rotatebox{90}{\hyperref[task:autoskola]{Driving}} & \rotatebox{90}{\hyperref[task:copa]{COPA}} & \rotatebox{90}{\hyperref[task:skj1]{SKJ1}} & \rotatebox{90}{\hyperref[task:trivia]{Trivia}} & \rotatebox{90}{\hyperref[task:belebele]{Belebele}} & \rotatebox{90}{\hyperref[task:chiby]{Chiby}} & \rotatebox{90}{\hyperref[task:wikiqa]{WikiQA}} & \rotatebox{90}{\hyperref[task:etest]{ETest-M}} & \rotatebox{90}{\hyperref[task:etest]{ETest-Q}} & \rotatebox{90}{\hyperref[task:sktest]{SKTest-M}} & \rotatebox{90}{\hyperref[task:sktest]{SKTest-Q}} & \rotatebox{90}{\hyperref[task:oneruler]{OneRuler}} & \rotatebox{90}{\hyperref[task:sklep-qa]{skLEP-QA}} & \rotatebox{90}{\hyperref[task:nli-sk]{NLI}} & \rotatebox{90}{\hyperref[task:sklep-nli]{skLEP-NLI}} & \rotatebox{90}{\hyperref[task:sklep-rte]{skLEP-RTE}} & \rotatebox{90}{\hyperref[task:senti]{Senti-3}} & \rotatebox{90}{\hyperref[task:senti]{Senti-2}} & \rotatebox{90}{\hyperref[task:sklep-sent]{skLEP-Sent}} & \rotatebox{90}{\hyperref[task:movierev]{MovieRev}} & \rotatebox{90}{\hyperref[task:toxic]{Toxic}} & \rotatebox{90}{\hyperref[task:sklep-hate]{skLEP-Hate}} & \rotatebox{90}{\hyperref[task:slovaksum]{SlovakSum}} & \rotatebox{90}{\hyperref[task:smesum]{SME-Sum}} & \rotatebox{90}{\hyperref[task:ifeval]{IFEval}} & \rotatebox{90}{\hyperref[task:analogies]{Analogies}} & \rotatebox{90}{\hyperref[task:slohry]{Slohry}} & \rotatebox{90}{\textbf{Avg}} \\
\midrule
\multicolumn{35}{l}{\textit{$<$2B}} \\
Qwen3 0.6B & 39.1 & 34.5 & 33.2 & 29.3 & 30.6 & 40.4 & 53.7 & 53.1 & 7.0 & 26.7 & 46.7 & 46.1 & 10.3 & 26.8 & 2.9 & 31.1 & 2.4 & 49.3 & 18.7 & 36.9 & 38.6 & 64.9 & 35.1 & 47.2 & 79.5 & 65.5 & 54.8 & 31.8 & 11.2 & 9.6 & 36.3 & 4.5 & 6.1 & 33.5 \\
Qwen3.5 0.8B & 7.1 & 2.8 & 2.3 & 0.6 & 2.5 & 1.7 & 3.6 & 1.9 & 0.0 & 0.3 & 1.2 & 0.1 & 0.0 & 4.1 & 0.1 & 2.3 & 0.8 & 0.5 & 0.0 & 2.5 & 3.2 & 0.3 & 2.9 & 4.5 & 15.2 & 6.1 & 3.9 & 2.5 & 0.3 & 0.2 & 24.4 & 2.3 & 18.9 & 3.6 \\
Qwen3 1.7B & 67.5 & 51.6 & 53.9 & 36.9 & 49.5 & 49.6 & 59.4 & 68.8 & 22.0 & 31.4 & 72.0 & 51.1 & 25.1 & 42.4 & 14.8 & 43.7 & 5.6 & 93.2 & 23.5 & 54.8 & 57.2 & 76.1 & 49.3 & 63.4 & 92.0 & 81.1 & 71.2 & 68.7 & 12.0 & 11.7 & 47.3 & 22.2 & 22.8 & 48.2 \\
\midrule
\multicolumn{35}{l}{\textit{2--9B}} \\
Qwen3.5 2B & 42.9 & 31.2 & 25.8 & 10.9 & 23.9 & 24.7 & 31.9 & 13.5 & 1.3 & 4.9 & 29.9 & 18.8 & 21.3 & 18.5 & 1.4 & 13.5 & 1.4 & 26.9 & 23.5 & 20.6 & 18.2 & 23.8 & 20.3 & 26.2 & 62.6 & 41.7 & 26.4 & 24.3 & 0.5 & 0.3 & 25.9 & 23.7 & 25.7 & 21.4 \\
\midrule
\multicolumn{35}{l}{\textit{9--32B}} \\
Qwen3.5 9B & 93.5 & 80.9 & 84.6 & 68.8 & 74.9 & 77.1 & 82.1 & 82.0 & 63.7 & 50.8 & 92.6 & 46.0 & 26.8 & 69.6 & 25.0 & 69.0 & 11.5 & 80.4 & 39.8 & 71.8 & 72.0 & 75.2 & 53.8 & 66.4 & 95.5 & 90.3 & 73.0 & 71.2 & 10.5 & 12.4 & 52.3 & 59.8 & 86.5 & 63.9 \\
Qwen3.5 27B & \underline{95.8} & 86.6 & \underline{89.4} & 79.9 & 77.9 & \underline{86.1} & \underline{88.8} & \textbf{84.3} & \textbf{77.0} & \textbf{66.4} & 94.7 & 64.7 & 25.6 & 75.6 & 32.9 & 79.0 & 19.5 & 96.0 & \textbf{44.9} & \textbf{80.3} & \underline{79.8} & \underline{83.9} & 60.0 & 73.9 & \underline{97.1} & \textbf{94.2} & 85.5 & \textbf{75.7} & 12.9 & 15.0 & 71.9 & \underline{63.0} & \textbf{89.5} & \underline{71.1} \\
Qwen3.6 27B & 95.7 & \underline{86.7} & 89.2 & \underline{80.0} & \underline{78.3} & 84.1 & 84.1 & \underline{83.6} & \underline{74.7} & \underline{64.4} & \textbf{95.0} & \underline{69.2} & \textbf{38.4} & \underline{76.1} & \textbf{50.3} & \underline{79.8} & \underline{25.5} & \underline{98.4} & \underline{42.2} & 78.1 & 78.7 & 83.7 & \underline{60.1} & \underline{76.9} & \textbf{97.3} & 93.9 & \underline{86.1} & 75.3 & \underline{13.4} & \textbf{15.7} & \underline{78.2} & 61.3 & 85.9 & \textbf{72.1} \\
Gemma 4 31B & \textbf{96.1} & \textbf{86.9} & \textbf{90.8} & \textbf{80.1} & \textbf{78.8} & \textbf{87.0} & \textbf{89.9} & 82.3 & 74.7 & 64.1 & \underline{94.8} & 66.9 & 15.9 & \textbf{76.7} & 30.9 & \textbf{82.1} & 17.2 & \textbf{100.0} & 33.5 & \underline{80.1} & \textbf{82.2} & 83.7 & \textbf{60.8} & 74.5 & 97.1 & \underline{94.0} & \textbf{87.6} & \underline{75.4} & \textbf{13.8} & \underline{15.5} & \textbf{84.0} & 62.1 & \underline{87.7} & 71.1 \\
\midrule
\multicolumn{35}{l}{\textit{32--200B}} \\
Qwen3.6 35B-A3B & 93.5 & 84.8 & 88.8 & 78.7 & 76.2 & 81.5 & 84.1 & 82.3 & 70.7 & 62.5 & 93.9 & \textbf{69.6} & \underline{31.6} & 72.7 & \underline{43.6} & 78.9 & \textbf{25.6} & 96.7 & 39.4 & 75.2 & 75.0 & \textbf{84.0} & 59.7 & \textbf{77.0} & 97.0 & 93.1 & 85.4 & 75.0 & 13.1 & 14.9 & 76.9 & \textbf{63.6} & 86.9 & 70.7 \\
\bottomrule
\end{tabular}%
}
\caption{sk-bench results for thinking-mode models. Slovak (SK), 0-shot. Each cell is the task's primary metric ($\times100$): accuracy for multiple-choice and classification, quasi-exact-match (qem) for open QA, ROUGE-L for summarisation, and prompt-level strict accuracy for IFEval. \textbf{Best} and \underline{second-best} per column. Avg is the unweighted mean of the columns shown.}
\label{tab:skbench-thinking}
\end{table*}

\subsection{English-prompt results}
\label{app:english}
The released leaderboard uses the Slovak instruction templates, but the
harness ships a parallel English template for every task. We re-ran the
full sweep with the English prompts. For the tasks that have an
English counterpart (IFEval-SK and Slohry are Slovak-only by
construction) Table~\ref{tab:skbench-comprehensive-en} gives the full
English per-task scores. The per-reporting-group Slovak-vs-English
comparison and its discussion are in the main text
(\S\ref{sec:oneruler}, Table~\ref{tab:lang-comparison}).

\begin{table*}[t]
\centering
\resizebox{\linewidth}{!}{%
\begin{tabular}{lrrrrrrrrrrrrrrrrrrrrrrrrrrrrrrr|r}
\toprule
\textbf{Model} & \rotatebox{90}{\hyperref[task:arc-sk]{ARC}} & \rotatebox{90}{\hyperref[task:mmlu-sk]{MMLU}} & \rotatebox{90}{\hyperref[task:gmmlu]{G-MMLU}} & \rotatebox{90}{\hyperref[task:truthqa-sk]{TruthQA}} & \rotatebox{90}{\hyperref[task:scio]{SCIO}} & \rotatebox{90}{\hyperref[task:finexam]{FinExam}} & \rotatebox{90}{\hyperref[task:autoskola]{Driving}} & \rotatebox{90}{\hyperref[task:copa]{COPA}} & \rotatebox{90}{\hyperref[task:skj1]{SKJ1}} & \rotatebox{90}{\hyperref[task:trivia]{Trivia}} & \rotatebox{90}{\hyperref[task:belebele]{Belebele}} & \rotatebox{90}{\hyperref[task:chiby]{Chiby}} & \rotatebox{90}{\hyperref[task:wikiqa]{WikiQA}} & \rotatebox{90}{\hyperref[task:etest]{ETest-M}} & \rotatebox{90}{\hyperref[task:etest]{ETest-Q}} & \rotatebox{90}{\hyperref[task:sktest]{SKTest-M}} & \rotatebox{90}{\hyperref[task:sktest]{SKTest-Q}} & \rotatebox{90}{\hyperref[task:oneruler]{OneRuler}} & \rotatebox{90}{\hyperref[task:sklep-qa]{skLEP-QA}} & \rotatebox{90}{\hyperref[task:nli-sk]{NLI}} & \rotatebox{90}{\hyperref[task:sklep-nli]{skLEP-NLI}} & \rotatebox{90}{\hyperref[task:sklep-rte]{skLEP-RTE}} & \rotatebox{90}{\hyperref[task:senti]{Senti-3}} & \rotatebox{90}{\hyperref[task:senti]{Senti-2}} & \rotatebox{90}{\hyperref[task:sklep-sent]{skLEP-Sent}} & \rotatebox{90}{\hyperref[task:movierev]{MovieRev}} & \rotatebox{90}{\hyperref[task:toxic]{Toxic}} & \rotatebox{90}{\hyperref[task:sklep-hate]{skLEP-Hate}} & \rotatebox{90}{\hyperref[task:slovaksum]{SlovakSum}} & \rotatebox{90}{\hyperref[task:smesum]{SME-Sum}} & \rotatebox{90}{\hyperref[task:analogies]{Analogies}} & \rotatebox{90}{\textbf{Avg}} \\
\midrule
\multicolumn{33}{l}{\textit{$<$2B}} \\
Granite 4.0 350M & 26.3 & 27.0 & 25.4 & 7.5 & 27.7 & 24.7 & 19.0 & 50.3 & 0.0 & 26.5 & 32.0 & 50.3 & 6.4 & 20.6 & 0.6 & 28.1 & 1.3 & 16.8 & 8.5 & 34.4 & 31.2 & 53.8 & 46.6 & 74.8 & 59.9 & 70.8 & 46.7 & 72.7 & 8.2 & 9.0 & 0.9 & 29.3 \\
Granite 4.0-H 350M & 25.5 & 27.8 & 25.9 & 7.6 & 24.3 & 25.3 & 25.7 & 50.7 & 48.3 & 22.9 & 28.7 & 50.2 & 6.3 & 22.4 & 0.6 & 20.2 & 0.8 & 24.7 & 14.3 & 25.2 & 23.7 & 44.5 & 62.5 & \underline{87.4} & 30.4 & 65.4 & 47.3 & 67.0 & 11.1 & 15.1 & 0.3 & 30.1 \\
Qwen3 0.6B & 29.9 & 29.5 & 29.0 & 30.2 & 23.8 & 32.6 & 65.5 & 48.5 & 7.7 & 24.1 & 39.6 & 50.7 & 8.1 & 24.7 & 0.3 & 30.5 & 1.1 & 52.3 & 18.6 & 32.3 & 29.6 & 52.5 & 65.9 & 66.8 & 46.2 & 55.7 & 53.3 & 28.0 & 12.4 & 12.5 & 2.0 & 32.4 \\
Qwen3.5 0.8B & 40.5 & 32.8 & 37.8 & 48.9 & 27.6 & 37.1 & 60.5 & 50.6 & 0.0 & 26.1 & 53.5 & 51.4 & 19.5 & 25.8 & 1.0 & 37.6 & 1.7 & 37.3 & 15.0 & 31.6 & 28.7 & 60.0 & 64.8 & 85.8 & 57.2 & 67.2 & 64.4 & 67.6 & 12.4 & 11.9 & 1.1 & 37.3 \\
Gemma 3 1B & 32.0 & 31.3 & 36.9 & 39.8 & 26.7 & 35.6 & 65.8 & 51.5 & 1.7 & 28.6 & 37.1 & 52.0 & 14.5 & 24.1 & 1.7 & 35.9 & 2.2 & 20.1 & 15.2 & 44.4 & 43.3 & 65.3 & \textbf{66.7} & 87.2 & 60.1 & 72.5 & 63.6 & 46.8 & 10.1 & 10.2 & 1.3 & 36.3 \\
Granite 4.0 1B & 40.9 & 36.5 & 39.3 & 14.2 & 32.6 & 38.6 & 33.4 & 52.7 & 10.3 & 35.1 & 62.3 & 52.0 & 16.9 & 27.3 & 1.4 & 39.1 & 2.0 & 65.2 & 20.3 & 47.6 & 45.3 & 72.9 & 38.6 & 73.8 & 75.8 & 78.5 & 51.1 & 72.7 & 12.2 & 14.1 & 5.6 & 39.0 \\
Granite 4.0-H 1B & 49.5 & 40.7 & 43.0 & 21.0 & 34.8 & 42.1 & 46.2 & 56.9 & 40.0 & 29.2 & 58.1 & 51.9 & 19.0 & 30.5 & 1.0 & 40.9 & 2.0 & 56.3 & 20.0 & 48.1 & 43.4 & 72.0 & 45.6 & 52.1 & 94.2 & 81.7 & 53.7 & 72.7 & 13.3 & 16.9 & 5.2 & 41.4 \\
EuroLLM 1.7B & 27.3 & 25.4 & 29.2 & 16.0 & 27.1 & 20.9 & 29.8 & 49.8 & 64.7 & 25.3 & 29.3 & 50.0 & 15.5 & 21.1 & 3.1 & 26.2 & 3.8 & 12.6 & 14.2 & 28.6 & 29.5 & 49.2 & \underline{66.2} & \textbf{88.0} & 43.6 & 64.7 & 44.2 & 67.5 & 8.4 & 7.0 & 3.9 & 32.0 \\
Qwen3 1.7B & 40.6 & 29.0 & 32.4 & 40.1 & 19.4 & 34.9 & 66.4 & 54.7 & 16.7 & 25.6 & 56.9 & 51.5 & 21.8 & 27.0 & 2.5 & 35.9 & 2.7 & 79.0 & 24.8 & 32.0 & 28.6 & 75.3 & 50.5 & 70.1 & 59.4 & 77.6 & 63.8 & 42.8 & 13.2 & 14.3 & 4.8 & 38.5 \\
\midrule
\multicolumn{33}{l}{\textit{2--9B}} \\
Granite 3.2 2B & 40.8 & 37.6 & 39.8 & 18.9 & 30.1 & 46.7 & 45.6 & 57.4 & 41.7 & 27.8 & 59.5 & 51.5 & 16.8 & 29.8 & 2.6 & 38.8 & 2.1 & 46.6 & 23.2 & 62.7 & 65.6 & 76.6 & 53.6 & 81.9 & 87.6 & 84.9 & 65.3 & 74.5 & 10.0 & 8.4 & 9.0 & 43.1 \\
Qwen3.5 2B & 60.3 & 44.5 & 52.8 & 29.3 & 41.8 & 45.3 & 59.5 & 53.1 & 11.3 & 34.7 & 73.5 & 56.5 & 17.5 & 35.8 & 4.1 & 46.8 & 5.5 & 38.9 & 25.6 & 40.0 & 37.1 & 79.1 & 48.4 & 63.5 & 71.2 & 68.0 & 65.0 & 72.5 & 13.5 & 16.2 & 2.5 & 42.4 \\
Llama 3.2 3B & 38.2 & 37.0 & 44.2 & 9.5 & 33.5 & 43.5 & 29.4 & 53.8 & 32.0 & 34.3 & 53.7 & 52.1 & 29.7 & 28.0 & 6.1 & 39.5 & 3.1 & 32.4 & 16.2 & 28.5 & 29.1 & 55.4 & 45.5 & 67.8 & 49.6 & 55.5 & 47.6 & 65.5 & 14.0 & 15.5 & 6.0 & 35.4 \\
Tiny-Aya Global & 64.7 & 48.8 & 53.9 & 45.9 & 39.8 & 57.5 & 67.8 & 65.3 & 22.3 & 41.1 & 78.4 & 52.8 & 15.5 & 37.3 & 8.1 & 52.6 & 5.2 & 49.2 & 14.8 & 42.7 & 38.2 & 58.2 & 47.7 & 75.3 & 86.0 & 80.5 & 60.3 & 75.8 & 12.9 & 12.7 & 10.7 & 45.9 \\
Tiny-Aya Water & 65.2 & 49.5 & 53.8 & 41.8 & 40.7 & 57.7 & 68.0 & 65.1 & 19.7 & 42.1 & 79.4 & 52.0 & 14.7 & 37.7 & 7.8 & 52.2 & 5.3 & 60.5 & 13.9 & 36.5 & 33.7 & 61.4 & 47.9 & 74.5 & 87.6 & 81.8 & 53.8 & 75.5 & 13.0 & 13.0 & 11.6 & 45.7 \\
Mistral-SK 7B & 21.3 & 18.7 & 20.0 & 9.3 & 16.2 & 21.1 & 40.8 & 25.9 & 1.7 & 13.1 & 22.9 & 27.0 & 17.7 & 12.8 & 13.0 & 20.3 & 4.3 & 0.0 & 16.4 & 7.4 & 5.9 & 14.8 & 17.8 & 28.5 & 10.1 & 36.8 & 26.1 & 9.4 & 8.5 & 10.6 & 4.0 & 16.2 \\
Mistral-SK 7B (Alpaca) & 18.9 & 13.6 & 18.2 & 32.6 & 10.7 & 12.4 & 54.1 & 47.0 & 4.7 & 5.8 & 26.5 & 0.4 & 14.2 & 10.3 & 0.6 & 19.2 & 1.8 & 40.5 & 5.8 & 0.4 & 0.7 & 0.0 & 0.2 & 0.2 & 0.0 & 0.3 & 0.3 & 0.1 & 12.7 & 10.9 & 4.9 & 11.9 \\
Teuken 7B & 48.7 & 35.1 & 40.8 & 34.9 & 22.9 & 52.9 & 71.8 & 53.1 & 2.0 & 33.1 & 66.3 & 45.6 & 3.8 & 30.8 & 3.7 & 45.6 & 3.9 & 37.9 & 9.1 & 24.0 & 24.1 & 59.0 & 40.7 & 54.5 & 86.3 & 84.4 & 46.9 & 72.7 & 10.2 & 9.3 & 4.4 & 37.4 \\
Aya Expanse 8B & 62.0 & 48.0 & 51.5 & 38.5 & 40.7 & 53.1 & 58.3 & 72.3 & 28.0 & 34.3 & 80.6 & 24.0 & \underline{42.7} & 33.6 & 8.8 & 52.1 & 3.9 & 49.3 & 24.6 & 47.4 & 55.2 & 75.4 & 42.0 & 60.6 & 82.1 & 76.9 & 79.9 & 77.5 & 12.5 & 13.7 & 10.9 & 46.5 \\
Llama 3.1 8B & 55.5 & 47.7 & 55.1 & 32.5 & 40.7 & 55.6 & 52.7 & 56.8 & 39.7 & 31.3 & 75.7 & 42.0 & 38.0 & 37.6 & 9.6 & 49.5 & 8.4 & 77.9 & 23.8 & 20.7 & 18.2 & 51.1 & 32.6 & 49.1 & 77.2 & 61.1 & 63.9 & 62.9 & 14.7 & 15.7 & 8.1 & 42.1 \\
Ministral 3 8B & 78.6 & 60.3 & 66.0 & 43.5 & 52.9 & 59.3 & 72.8 & 65.8 & 48.3 & 46.1 & 85.6 & 54.1 & 13.7 & 50.4 & 12.3 & 62.1 & 6.3 & 35.4 & 15.9 & 43.0 & 42.2 & 77.4 & 53.4 & 75.2 & 92.0 & 87.6 & 61.2 & 73.6 & 13.2 & 14.7 & 31.5 & 51.4 \\
Ministral 3 8B (R) & 71.5 & 52.0 & 59.7 & 60.2 & 48.5 & 39.1 & 69.9 & 57.7 & 28.7 & 36.5 & 80.5 & 52.9 & 13.1 & 43.3 & 11.8 & 49.7 & 4.8 & 90.0 & 18.4 & 49.1 & 49.4 & 76.3 & 55.9 & 80.6 & 78.9 & 76.6 & 49.1 & 72.7 & 12.6 & 13.4 & 21.2 & 49.2 \\
\midrule
\multicolumn{33}{l}{\textit{9--32B}} \\
EuroLLM 9B & 73.8 & 52.9 & 59.6 & 41.3 & 43.5 & 61.2 & 73.0 & 68.1 & 36.3 & 42.1 & 79.1 & 55.5 & 19.6 & 45.1 & 20.5 & 62.1 & 12.9 & 47.5 & 24.5 & 54.7 & 52.5 & 81.5 & 60.1 & 84.7 & 86.7 & 82.3 & 79.2 & 74.2 & 13.5 & 14.9 & 30.9 & 52.7 \\
Qwen3.5 9B & 88.5 & 67.6 & 71.8 & 60.9 & 60.7 & 73.1 & 78.9 & 82.4 & 60.7 & 47.6 & 88.9 & 63.9 & 22.2 & 62.5 & 26.6 & 66.9 & 15.3 & 84.5 & 38.5 & 74.5 & 71.5 & 85.1 & 46.6 & 60.0 & 89.4 & 88.8 & 74.2 & 76.5 & 13.5 & 16.0 & 39.2 & 61.2 \\
Bielik 11B & 82.2 & 61.3 & 69.2 & 58.4 & 54.7 & 72.5 & 73.9 & 82.2 & 52.0 & 52.9 & 89.6 & 58.7 & 22.8 & 54.9 & 30.8 & 65.4 & 18.9 & 86.5 & 31.8 & 76.3 & 78.4 & 81.9 & 58.5 & 76.0 & 95.9 & 92.4 & 84.6 & 77.4 & 13.0 & 12.6 & 38.0 & 61.4 \\
Ministral 3 14B & 82.1 & 63.4 & 67.3 & 53.3 & 55.8 & 70.3 & 75.9 & 71.8 & 67.7 & 49.1 & 87.6 & 52.9 & 11.8 & 59.7 & 14.2 & 63.8 & 5.5 & 0.0 & 16.7 & 49.2 & 45.1 & 79.8 & 19.7 & 31.3 & 36.0 & 35.7 & 78.8 & 74.0 & 13.4 & 15.1 & 37.6 & 47.9 \\
Ministral 3 14B (R) & 57.2 & 39.0 & 43.3 & 9.6 & 31.5 & 42.2 & 20.6 & 52.8 & 37.7 & 26.9 & 43.7 & 22.1 & 3.2 & 30.8 & 8.4 & 35.7 & 2.0 & 0.6 & 5.3 & 42.0 & 41.9 & 66.8 & 16.5 & 10.4 & 22.1 & 20.3 & 57.1 & 63.3 & 0.9 & 0.5 & 21.5 & 28.3 \\
Qwen3 14B & 83.8 & 64.7 & 66.2 & 58.5 & 58.7 & 67.9 & 79.6 & 75.7 & 63.3 & 46.8 & 88.7 & 60.5 & 28.3 & 57.8 & 21.6 & 62.0 & 11.8 & 85.8 & 37.8 & 47.2 & 52.8 & 83.7 & 47.4 & 71.9 & 89.1 & 88.7 & 76.5 & 76.8 & 13.9 & 16.6 & 37.0 & 58.7 \\
Qwen3-SK 14B & 85.4 & 64.3 & 68.2 & 72.2 & 55.2 & 73.6 & 86.2 & 69.3 & 53.7 & 52.5 & 71.9 & 61.7 & 40.1 & 61.6 & 23.4 & 64.5 & 12.7 & 32.4 & 24.1 & 56.1 & 51.1 & 27.8 & 26.5 & 49.1 & 65.2 & 63.2 & 64.7 & 74.6 & 12.4 & 10.9 & 48.8 & 52.4 \\
Qwen3-SK 14B (Alpaca) & 86.1 & 65.6 & 69.5 & 68.9 & 59.1 & 70.1 & 83.1 & 83.9 & 66.7 & 49.6 & 89.6 & 20.6 & 11.7 & 58.3 & 17.9 & 67.5 & 11.1 & 76.1 & 7.7 & 46.2 & 45.7 & 50.0 & 12.8 & 4.3 & 28.3 & 9.3 & 3.7 & 5.5 & 14.2 & 14.4 & 32.0 & 42.9 \\
gpt-oss 20B & 88.5 & 73.6 & 74.8 & 53.2 & 67.6 & 64.5 & 73.4 & 83.3 & 59.0 & 34.6 & 90.0 & 65.1 & 28.2 & 62.3 & 24.2 & 65.6 & 12.9 & 85.7 & 33.2 & 74.0 & 76.4 & 83.6 & 53.9 & 70.9 & 95.7 & 87.9 & 75.6 & 74.1 & 12.8 & 13.7 & 40.9 & 61.3 \\
EuroLLM 22B & 78.9 & 56.9 & 65.6 & 58.4 & 46.9 & 66.9 & 81.0 & 76.1 & 50.7 & 49.8 & 82.3 & 62.3 & 14.9 & 54.8 & 19.9 & 66.1 & 12.0 & 64.9 & 28.6 & 68.0 & 72.1 & 83.4 & 57.1 & 82.0 & 94.6 & 88.3 & 81.1 & 78.8 & 14.4 & 16.3 & 38.6 & 58.4 \\
Qwen3.5 27B & 94.3 & 80.0 & 84.8 & 74.9 & 69.1 & 82.3 & 86.1 & 87.5 & 80.0 & 61.0 & 93.4 & 68.9 & 21.6 & 72.2 & 33.9 & 76.7 & 19.0 & 83.3 & 42.6 & 29.7 & 29.3 & 85.0 & 44.4 & 64.3 & 91.1 & 89.6 & 81.0 & \underline{79.0} & 13.9 & 16.4 & 51.6 & 64.1 \\
Qwen3.6 27B & 94.0 & 79.5 & 84.4 & 77.1 & 69.0 & 82.5 & 87.2 & 86.5 & 73.3 & 57.1 & 93.8 & 68.6 & 33.0 & 70.6 & 39.6 & 75.9 & 21.0 & 83.3 & \underline{49.4} & 27.6 & 26.4 & 85.3 & 41.5 & 56.6 & 85.1 & 77.3 & 79.1 & 76.4 & 13.7 & 16.3 & 51.2 & 63.3 \\
TildeOpen 30B & 65.2 & 46.7 & 54.9 & 47.9 & 41.2 & 53.0 & 75.1 & 46.8 & 18.0 & 43.4 & 68.4 & 0.0 & 17.3 & 36.5 & 25.5 & 56.4 & 8.5 & 0.0 & 10.3 & 16.2 & 17.1 & 28.9 & 28.6 & 31.1 & 30.9 & 27.3 & 49.0 & 45.2 & 1.2 & 0.8 & 11.1 & 32.4 \\
TildeOpen 30B-64k & 71.5 & 52.0 & 61.3 & 46.5 & 43.8 & 64.4 & 82.6 & 63.1 & 62.7 & 54.1 & 76.6 & 51.8 & \textbf{43.9} & 40.3 & 52.6 & 66.0 & 24.4 & 0.0 & 24.2 & 40.8 & 38.2 & 55.3 & 46.1 & 56.8 & 91.9 & 67.2 & 48.1 & 69.1 & 7.7 & 7.5 & 20.0 & 49.4 \\
TildeOpen-SK 30B (Alpaca) & 68.4 & 45.7 & 51.7 & 37.7 & 41.0 & 58.6 & 70.1 & 72.2 & 53.0 & 48.7 & 73.5 & 0.2 & 10.8 & 40.0 & 22.6 & 61.4 & 10.5 & 38.0 & 16.3 & 5.0 & 1.8 & 9.1 & 0.5 & 0.1 & 0.0 & 0.0 & 0.0 & 1.9 & 13.2 & 12.5 & 20.2 & 28.5 \\
Gemma 4 31B & 94.6 & 78.5 & 85.4 & 74.9 & 62.4 & 83.4 & 86.9 & 83.3 & 68.7 & 56.6 & 93.9 & 63.6 & 11.8 & 66.7 & 32.1 & 77.3 & 19.5 & 83.3 & 33.8 & \underline{83.8} & \underline{84.3} & 85.4 & 58.9 & 77.7 & 96.4 & 93.3 & 81.2 & 76.9 & 13.9 & 15.3 & 50.0 & 66.9 \\
Aya Expanse 32B & 78.0 & 60.0 & 64.5 & 57.1 & 49.0 & 69.4 & 70.7 & 79.9 & 41.0 & 45.0 & 90.7 & 44.8 & 22.2 & 44.3 & 14.3 & 63.7 & 6.7 & 49.6 & 27.1 & 58.4 & 58.2 & 76.4 & 40.1 & 59.0 & 84.9 & 89.1 & 82.0 & 76.2 & 14.5 & 16.5 & 30.4 & 53.7 \\
Granite 4.0-H Small & 83.6 & 64.5 & 72.5 & 62.2 & 55.5 & 66.9 & 69.9 & 77.9 & 59.3 & 46.4 & 86.7 & 58.5 & 25.2 & 56.5 & 16.5 & 66.7 & 9.2 & 75.0 & 26.4 & 72.4 & 72.4 & 84.2 & 47.5 & 62.8 & 94.1 & 88.6 & 73.7 & 73.6 & 14.6 & \underline{18.1} & 43.6 & 58.9 \\
\midrule
\multicolumn{33}{l}{\textit{32--200B}} \\
Qwen3.6 35B-A3B & 92.6 & 75.8 & 80.9 & 74.6 & 66.6 & 79.4 & 81.7 & 82.9 & 74.0 & 50.9 & 92.3 & 68.3 & 34.6 & 69.2 & 36.1 & 75.5 & 15.5 & 83.3 & 43.3 & 27.6 & 27.0 & 83.4 & 39.8 & 58.0 & 91.9 & 90.0 & 81.5 & 76.5 & 13.9 & 15.9 & 46.4 & 62.2 \\
Apertus 70B & 71.3 & 53.5 & 63.2 & 44.2 & 45.9 & 66.4 & 70.1 & 65.6 & 51.0 & 48.8 & 83.0 & 49.9 & 22.5 & 45.7 & 14.6 & 61.2 & 12.2 & 80.1 & 21.2 & 63.0 & 60.6 & 76.6 & 49.0 & 69.3 & 93.3 & 88.8 & 51.7 & 71.8 & 13.5 & 13.8 & 37.9 & 53.5 \\
Llama 3.3 70B & 86.7 & 71.5 & 79.2 & 63.6 & 59.4 & 73.8 & 81.6 & 83.6 & 48.7 & 53.4 & 91.3 & 58.5 & 33.9 & 62.3 & 33.1 & 69.2 & 18.2 & 82.1 & 41.0 & 69.0 & 73.4 & 79.7 & 51.7 & 72.1 & 93.9 & 91.0 & 83.1 & 77.3 & \textbf{16.2} & \textbf{24.3} & 45.4 & 63.5 \\
Command A & 81.7 & 57.7 & 64.2 & 60.4 & 42.6 & 76.1 & 74.5 & 24.9 & 59.7 & 54.1 & 92.0 & 0.1 & 8.3 & 47.3 & 5.0 & 60.4 & 3.9 & 72.5 & 8.4 & 50.5 & 38.4 & 64.4 & 6.1 & 7.8 & 26.5 & 12.5 & 17.8 & 30.0 & 10.2 & 10.9 & 43.8 & 39.1 \\
gpt-oss 120B & 91.6 & 78.8 & 80.6 & 63.0 & 72.2 & 73.3 & 77.8 & 80.1 & 65.7 & 52.6 & 91.4 & 66.6 & 20.3 & 70.6 & 36.0 & 72.4 & 21.0 & 91.0 & 36.0 & 76.8 & 79.1 & \underline{85.8} & 55.2 & 71.1 & 96.1 & 89.9 & 79.3 & 74.8 & 12.9 & 14.4 & 47.9 & 65.3 \\
\midrule
\multicolumn{33}{l}{\textit{$>$200B}} \\
MiniMax M2.7 & 93.6 & 82.2 & 86.7 & 71.1 & 71.8 & 80.8 & 82.6 & 82.9 & 60.7 & 60.3 & 92.9 & 61.0 & 24.9 & 73.2 & 46.9 & 75.7 & 24.1 & 34.3 & 42.5 & 74.7 & 77.8 & 82.1 & 55.8 & 69.5 & 96.4 & 92.5 & 78.7 & 74.5 & 13.1 & 14.5 & 62.5 & 65.8 \\
Qwen3.5 397B-A17B & 95.0 & 82.8 & 89.6 & 79.8 & 73.4 & 87.6 & 86.3 & 90.1 & 82.3 & 68.5 & \underline{94.7} & 70.0 & 14.0 & 77.0 & 45.0 & 80.6 & 25.8 & 83.3 & 31.8 & 64.8 & 67.4 & 82.6 & 45.2 & 57.2 & 85.2 & 84.0 & 72.9 & 75.4 & 13.7 & 16.1 & 57.7 & 67.1 \\
Mistral Large 3 675B & 90.6 & 76.7 & 83.5 & 71.9 & 64.8 & 79.7 & 84.2 & 85.7 & 62.0 & 61.9 & 92.4 & 59.7 & 15.0 & 61.2 & 22.3 & 75.7 & 12.6 & 10.4 & 21.6 & 70.0 & 69.0 & 67.5 & 28.4 & 42.8 & 73.5 & 47.1 & 78.5 & 76.4 & 12.9 & 15.0 & 54.6 & 57.0 \\
GLM-5.1 & 91.7 & 76.3 & 84.4 & 75.5 & 61.5 & 82.8 & 86.4 & 79.6 & 74.7 & 61.1 & 92.6 & 65.7 & 34.8 & 71.4 & 39.6 & 77.0 & 18.3 & 83.2 & 45.7 & 77.3 & 73.8 & 85.1 & 61.0 & 80.7 & 96.0 & 92.1 & \textbf{86.5} & 77.4 & 13.4 & 15.2 & 55.4 & 68.3 \\
Kimi K2.6 & 92.6 & 79.1 & 85.7 & 78.1 & 67.7 & 85.8 & 85.7 & \underline{90.9} & 79.3 & 64.3 & 94.7 & 63.3 & 37.1 & 74.4 & 44.1 & 78.8 & 23.1 & 0.7 & 44.1 & 65.3 & 61.4 & 84.2 & 58.2 & 77.8 & 96.4 & 92.6 & 83.3 & 77.8 & 14.2 & 16.6 & 57.5 & 66.3 \\
\midrule
\multicolumn{33}{l}{\textit{Proprietary (size undisclosed)}} \\
Gemini 3.1 Pro & \textbf{96.9} & \textbf{91.3} & \textbf{94.8} & \textbf{93.9} & \textbf{81.5} & \underline{96.2} & \textbf{98.0} & \textbf{91.6} & 82.3 & \textbf{93.6} & \textbf{96.0} & \textbf{75.1} & 35.8 & \textbf{88.3} & \underline{69.7} & \textbf{89.7} & \textbf{38.1} & \underline{100.0} & \textbf{52.3} & \textbf{84.1} & \textbf{85.6} & \textbf{86.4} & 62.2 & 77.3 & 97.6 & \textbf{96.1} & 82.1 & 73.3 & 14.1 & 16.0 & \textbf{75.7} & \textbf{77.9} \\
GPT-5.5 & \underline{96.4} & \underline{91.3} & \underline{92.7} & 84.0 & \underline{81.1} & \textbf{96.4} & \underline{96.1} & 82.5 & 82.0 & \underline{93.5} & 94.2 & \underline{74.1} & 22.7 & \underline{87.5} & \textbf{70.5} & \underline{88.8} & \underline{35.6} & 99.9 & 43.0 & 80.4 & 80.9 & 83.7 & 60.7 & 75.3 & \textbf{97.9} & \underline{95.2} & 80.7 & 73.8 & 14.3 & 17.0 & \underline{71.8} & \underline{75.6} \\
Gemini 3.1 Flash-Lite & 94.7 & 83.8 & 88.9 & 79.3 & 70.7 & 89.5 & 92.4 & 90.3 & \textbf{84.7} & 77.6 & 94.2 & 71.8 & 17.6 & 80.8 & 26.6 & 85.5 & 18.3 & 99.8 & 31.0 & 83.4 & 82.7 & 83.9 & 60.5 & 80.8 & 97.5 & 94.4 & 81.0 & 75.4 & 13.6 & 15.1 & 64.1 & 71.3 \\
Claude Opus 4.6 & 93.8 & 86.8 & 90.5 & \underline{89.8} & 71.6 & 92.6 & 91.3 & 87.9 & \underline{83.0} & 78.5 & 94.4 & 71.5 & 14.1 & 70.6 & 23.7 & 81.3 & 13.4 & \textbf{100.0} & 33.3 & 81.6 & 83.0 & 80.2 & 57.8 & 72.7 & \underline{97.8} & 94.3 & \underline{85.5} & \textbf{79.3} & \underline{14.9} & 16.8 & 66.8 & 70.9 \\
GPT-5.4 mini & 91.1 & 75.0 & 82.4 & 69.9 & 63.4 & 84.2 & 87.1 & 90.1 & 79.3 & 58.5 & 91.7 & 69.9 & 26.0 & 72.7 & 34.7 & 80.2 & 20.2 & 99.7 & 38.9 & 74.4 & 68.8 & 81.2 & 57.7 & 83.2 & 95.3 & 91.5 & 78.9 & 76.3 & 13.8 & 16.9 & 56.7 & 68.1 \\
Claude Sonnet 4.6 & 93.0 & 84.0 & 87.6 & 78.6 & 71.5 & 88.5 & 88.2 & 90.1 & 80.3 & 73.0 & 91.2 & 66.2 & 14.1 & 58.9 & 28.2 & 79.4 & 14.3 & 100.0 & 19.8 & 72.5 & 74.1 & 81.3 & 54.1 & 73.8 & 95.3 & 87.0 & 78.1 & 76.9 & 14.6 & 16.4 & 62.9 & 67.5 \\
\bottomrule
\end{tabular}%
}
\caption{sk-bench results across all tasks. English (EN) prompts, 0-shot --- the English-template counterpart of Table~\ref{tab:skbench-comprehensive}. IFEval-SK and Slohry are Slovak-only by construction and omitted. Each cell is the task's primary metric ($\times100$), with \textbf{best} and \underline{second-best} per column. Avg is the unweighted mean of the columns shown.}
\label{tab:skbench-comprehensive-en}
\end{table*}

\subsection{Few-shot (3-shot) results}
\label{app:fewshot-tables}
The main results are zero-shot. We also ran the applicable tasks 3-shot
under both prompt languages (\S\ref{sec:oneruler},
Table~\ref{tab:fewshot}); summarisation and instruction following were
run zero-shot only. For the 52 models run 3-shot,
Table~\ref{tab:skbench-comprehensive-3shot} gives the Slovak 3-shot
per-task scores and
Table~\ref{tab:skbench-comprehensive-en-3shot} their English-prompt
counterpart, parallel to the 0-shot
Tables~\ref{tab:skbench-comprehensive} and~\ref{tab:skbench-comprehensive-en}.

\begin{table*}[t]
\centering
\resizebox{\linewidth}{!}{%
\begin{tabular}{lrrrrrrrrrrrrrrrrrrr|r}
\toprule
\textbf{Model} & \rotatebox{90}{\hyperref[task:arc-sk]{ARC}} & \rotatebox{90}{\hyperref[task:mmlu-sk]{MMLU}} & \rotatebox{90}{\hyperref[task:gmmlu]{G-MMLU}} & \rotatebox{90}{\hyperref[task:scio]{Scio}} & \rotatebox{90}{\hyperref[task:autoskola]{Driving}} & \rotatebox{90}{\hyperref[task:copa]{COPA}} & \rotatebox{90}{\hyperref[task:sktest]{SKTest-M}} & \rotatebox{90}{\hyperref[task:sktest]{SKTest-Q}} & \rotatebox{90}{\hyperref[task:sklep-qa]{skLEP-QA}} & \rotatebox{90}{\hyperref[task:sklep-nli]{skLEP-NLI}} & \rotatebox{90}{\hyperref[task:sklep-rte]{skLEP-RTE}} & \rotatebox{90}{\hyperref[task:senti]{Senti-3}} & \rotatebox{90}{\hyperref[task:senti]{Senti-2}} & \rotatebox{90}{\hyperref[task:sklep-sent]{skLEP-Sent}} & \rotatebox{90}{\hyperref[task:movierev]{MovieRev}} & \rotatebox{90}{\hyperref[task:toxic]{Toxic}} & \rotatebox{90}{\hyperref[task:sklep-hate]{skLEP-Hate}} & \rotatebox{90}{\hyperref[task:analogies]{Analogies}} & \rotatebox{90}{\hyperref[task:slohry]{Slohry}} & \rotatebox{90}{\textbf{Avg}} \\
\midrule
\multicolumn{21}{l}{\textit{$<$2B}} \\
Granite 4.0 350M & 24.6 & 27.8 & 27.1 & 23.9 & 24.9 & 49.1 & 26.2 & 1.0 & 10.1 & 24.9 & 42.2 & 16.1 & 21.2 & 36.8 & 48.0 & 41.8 & 62.9 & 0.5 & 1.5 & 26.9 \\
Granite 4.0-H 350M & 24.3 & 25.4 & 26.0 & 24.4 & 59.0 & 49.9 & 30.7 & 1.5 & 9.4 & 26.9 & 49.0 & \textbf{67.9} & 85.6 & 22.8 & 58.8 & 31.1 & 44.6 & 0.1 & 0.4 & 33.6 \\
Qwen3 0.6B & 35.5 & 31.8 & 32.7 & 26.5 & 36.5 & 43.5 & 33.9 & 1.1 & 15.1 & 35.7 & 45.2 & 47.0 & 84.6 & 51.8 & 59.1 & 54.9 & 46.9 & 0.7 & 8.8 & 36.4 \\
Qwen3.5 0.8B & 40.1 & 33.6 & 34.2 & 26.1 & 32.3 & 50.5 & 29.8 & 2.1 & 11.6 & 36.6 & 57.2 & 51.6 & 78.1 & 76.2 & 73.7 & 54.6 & 33.6 & 4.1 & 9.0 & 38.7 \\
Gemma 3 1B & 28.2 & 30.0 & 33.0 & 25.0 & 65.0 & 51.5 & 33.1 & 1.8 & 17.0 & 35.4 & 58.6 & 49.9 & 61.5 & 91.5 & 80.1 & 54.1 & 47.8 & 0.6 & 3.4 & 40.4 \\
Granite 4.0 1B & 44.2 & 39.8 & 42.9 & 35.1 & 59.1 & 56.8 & 42.1 & 3.2 & 22.8 & 47.5 & 65.8 & 55.1 & 71.5 & 67.8 & 65.8 & 50.6 & 57.0 & 1.6 & 12.3 & 44.3 \\
Granite 4.0-H 1B & 53.9 & 43.6 & 44.6 & 36.3 & 59.3 & 51.8 & 42.1 & 0.6 & 21.3 & 45.2 & 69.7 & 37.1 & 67.3 & 93.8 & 81.2 & 65.2 & 57.8 & 0.8 & 9.6 & 46.4 \\
EuroLLM 1.7B & 26.2 & 26.6 & 26.7 & 25.8 & 49.6 & 50.2 & 26.9 & 4.8 & 20.4 & 30.1 & 50.2 & 50.3 & \underline{87.7} & 22.6 & 58.7 & 45.0 & 70.2 & 5.9 & 24.4 & 37.0 \\
Qwen3 1.7B & 56.4 & 41.1 & 42.0 & 30.1 & 65.9 & 63.1 & 41.1 & 2.8 & 20.1 & 47.9 & 68.7 & 28.6 & 63.3 & 92.0 & 80.0 & 66.6 & 54.1 & 2.7 & 12.4 & 46.3 \\
\midrule
\multicolumn{21}{l}{\textit{2--9B}} \\
Granite 3.2 2B & 42.4 & 39.6 & 38.7 & 29.5 & 51.0 & 60.1 & 37.2 & 2.9 & 21.0 & 37.5 & 67.0 & 58.6 & 72.6 & 91.7 & 86.5 & 79.5 & 47.1 & 4.2 & 7.5 & 46.0 \\
Qwen3.5 2B & 60.6 & 43.4 & 45.8 & 36.0 & 57.2 & 53.2 & 41.5 & 3.6 & 24.8 & 42.7 & 68.7 & 45.0 & 66.9 & 40.3 & 67.6 & 62.8 & 60.5 & 2.8 & 18.2 & 44.3 \\
Llama 3.2 3B & 13.0 & 9.1 & 7.3 & 10.6 & 12.8 & 41.0 & 18.3 & 2.5 & 17.2 & 29.2 & 45.1 & 45.7 & 62.2 & 29.0 & 50.1 & 52.4 & 48.7 & 5.9 & 4.8 & 26.6 \\
Tiny-Aya Global & 67.3 & 50.7 & 54.2 & 41.1 & 65.4 & 69.8 & 54.4 & 4.6 & 20.4 & 65.7 & 74.5 & 49.4 & 81.0 & 94.4 & 89.7 & 87.4 & 75.9 & 7.5 & 19.5 & 56.5 \\
Tiny-Aya Water & 68.3 & 51.4 & 54.8 & 41.4 & 64.5 & 71.1 & 54.3 & 4.1 & 20.1 & 66.2 & 76.0 & 45.3 & 80.0 & 92.4 & 89.7 & 88.0 & 77.6 & 8.1 & 22.6 & 56.6 \\
Mistral-SK 7B & 37.2 & 29.3 & 37.9 & 30.6 & 56.2 & 57.4 & 31.7 & 11.9 & 28.0 & 35.2 & 58.1 & 25.6 & 50.3 & 94.8 & 91.3 & 47.9 & 71.9 & 14.4 & 47.6 & 45.1 \\
Mistral-SK 7B (Alpaca) & 19.2 & 7.2 & 12.2 & 5.0 & 5.4 & 54.6 & 4.9 & 1.5 & 6.9 & 8.8 & 10.2 & 0.3 & 50.4 & 17.6 & 9.2 & 37.1 & 40.5 & 0.2 & 18.6 & 16.3 \\
Teuken 7B & 53.6 & 42.8 & 49.8 & 34.8 & 68.0 & 60.0 & 47.0 & 5.5 & 25.2 & 37.2 & 58.2 & 46.6 & 58.5 & 75.2 & 69.4 & 46.7 & 73.5 & 0.1 & 1.4 & 44.9 \\
Aya Expanse 8B & 53.3 & 27.4 & 34.0 & 20.7 & 42.8 & 28.5 & 29.2 & 2.4 & 25.9 & 28.3 & 68.0 & 32.8 & 50.0 & 67.9 & 77.9 & 77.9 & 45.0 & 1.8 & 1.7 & 37.6 \\
Llama 3.1 8B & 51.0 & 39.9 & 50.8 & 34.2 & 60.7 & 64.8 & 49.6 & 9.0 & 30.6 & 44.9 & 67.2 & 55.8 & 70.0 & 87.3 & 80.0 & 73.5 & 66.9 & 20.9 & 23.7 & 51.6 \\
Ministral 3 8B & 70.8 & 54.3 & 54.7 & 47.6 & 41.4 & 34.2 & 51.4 & 14.1 & 27.1 & 46.1 & 73.9 & 43.1 & 54.0 & 91.6 & 86.8 & 79.5 & 59.6 & 31.4 & 64.8 & 54.0 \\
Ministral 3 8B (R) & 22.9 & 12.7 & 5.8 & 8.8 & 35.0 & 0.5 & 26.6 & 6.0 & 21.5 & 16.0 & 13.2 & 36.4 & 24.2 & 81.0 & 80.9 & 34.8 & 40.1 & 22.5 & 45.2 & 28.1 \\
\midrule
\multicolumn{21}{l}{\textit{9--32B}} \\
EuroLLM 9B & 75.2 & 54.3 & 60.5 & 41.4 & 79.9 & 73.2 & 60.1 & 9.1 & 42.2 & 66.2 & 75.8 & 58.1 & 83.8 & 95.9 & 89.9 & 90.5 & 73.0 & 30.5 & 63.6 & 64.4 \\
Qwen3.5 9B & 88.4 & 69.7 & 70.2 & 59.5 & 80.5 & 87.3 & 66.6 & 15.1 & 47.5 & 75.6 & 83.9 & 59.1 & 82.3 & 96.9 & 91.2 & 86.9 & 77.2 & 30.3 & 70.9 & 70.5 \\
Bielik 11B & 84.5 & 62.8 & 67.9 & 53.2 & 78.6 & 89.4 & 67.1 & 13.3 & 25.9 & 75.4 & 82.2 & 63.8 & 77.5 & 96.2 & 92.6 & 90.7 & 76.9 & 27.6 & 38.7 & 66.5 \\
Ministral 3 14B & 84.1 & 62.2 & 66.2 & 49.2 & 70.8 & 83.3 & 61.0 & 14.1 & 33.5 & 57.8 & 78.4 & 54.2 & 69.9 & 94.3 & 85.4 & 85.4 & 67.1 & 28.6 & 75.4 & 64.2 \\
Ministral 3 14B (R) & 68.3 & 39.3 & 28.8 & 20.0 & 39.6 & 70.3 & 38.3 & 1.8 & 14.7 & 24.2 & 64.6 & 10.4 & 1.9 & 17.1 & 16.4 & 19.1 & 37.9 & 1.6 & 53.2 & 29.9 \\
Qwen3 14B & 86.5 & 67.8 & 69.1 & 59.3 & 84.4 & 83.7 & 65.6 & 16.0 & 45.9 & 77.2 & 75.9 & 52.9 & 71.9 & 97.2 & 91.9 & 87.2 & 75.9 & 39.9 & 67.7 & 69.3 \\
Qwen3-SK 14B & 73.5 & 41.7 & 44.0 & 24.0 & 3.9 & 18.9 & 46.7 & 16.2 & 24.0 & 52.9 & 71.2 & 5.6 & 7.6 & 12.3 & 29.3 & 1.2 & 21.1 & 48.5 & 0.2 & 28.6 \\
Qwen3-SK 14B (Alpaca) & 88.0 & 69.7 & 68.4 & 59.4 & 86.9 & 86.9 & 68.9 & 8.3 & 11.5 & 78.1 & 81.0 & 57.7 & 74.4 & 96.1 & 91.7 & 90.0 & 77.6 & 18.9 & 59.9 & 67.0 \\
gpt-oss 20B & 86.3 & 71.8 & 71.9 & 64.1 & 72.2 & 84.0 & 62.4 & 7.3 & 25.5 & 75.3 & 82.4 & 57.1 & 77.1 & 96.3 & 88.3 & 85.4 & 75.1 & 25.2 & 73.5 & 67.4 \\
EuroLLM 22B & 81.0 & 61.3 & 64.1 & 48.4 & 81.8 & 88.6 & 66.3 & 13.3 & 42.0 & 72.9 & 75.8 & 59.5 & 84.2 & 95.9 & 91.3 & 73.2 & 62.9 & 33.5 & 64.8 & 66.4 \\
Qwen3.5 27B & 94.9 & 82.1 & 85.7 & 69.6 & 88.6 & 90.6 & 78.2 & 23.4 & 51.8 & 82.3 & 82.9 & 62.9 & 83.3 & 97.6 & 94.4 & 90.4 & \textbf{80.0} & 50.7 & 86.0 & 77.6 \\
Qwen3.6 27B & \underline{95.1} & 81.7 & 84.4 & 67.8 & 88.7 & 89.9 & 77.5 & 25.6 & \underline{53.0} & 80.4 & 85.3 & 63.2 & 87.6 & 97.4 & 94.4 & 87.6 & 75.8 & 51.1 & 85.4 & 77.5 \\
TildeOpen 30B & 36.4 & 31.4 & 21.2 & 23.3 & 63.9 & 66.9 & 41.8 & 2.5 & 1.8 & 18.9 & 31.5 & 66.3 & 45.5 & 72.8 & 62.1 & 54.8 & 35.3 & 13.2 & 46.1 & 38.7 \\
TildeOpen 30B-64k & 77.2 & 57.6 & 65.6 & 47.6 & 82.1 & 89.2 & 71.7 & 28.9 & 35.7 & 54.9 & 74.7 & \underline{66.6} & 80.5 & \textbf{97.8} & \textbf{94.8} & 80.1 & 71.2 & 44.4 & 85.4 & 68.7 \\
TildeOpen-SK 30B (Alpaca) & 75.3 & 50.6 & 52.7 & 39.2 & 80.8 & 89.4 & 66.9 & 15.0 & 26.5 & 46.5 & 65.5 & 56.0 & 76.0 & 46.5 & 76.6 & 84.6 & 70.8 & 6.8 & 54.1 & 56.8 \\
Gemma 4 31B & 94.7 & 81.9 & 86.7 & 72.1 & 89.0 & 84.6 & 77.7 & 27.6 & 50.0 & \textbf{85.0} & 85.6 & 62.5 & 83.1 & 97.4 & 94.5 & \textbf{91.3} & 78.8 & 51.8 & 86.3 & 77.9 \\
Aya Expanse 32B & 77.8 & 61.4 & 65.6 & 50.6 & 78.0 & 84.3 & 64.3 & 7.6 & 32.6 & 74.7 & 78.7 & 49.6 & 75.0 & 95.8 & 92.6 & 89.4 & 72.6 & 15.2 & 6.4 & 61.7 \\
Granite 4.0-H Small & 85.8 & 67.7 & 75.8 & 56.6 & 80.0 & 88.1 & 68.1 & 14.1 & 37.9 & 77.7 & 82.4 & 59.9 & 73.9 & 96.0 & 91.8 & 89.1 & 74.7 & 30.4 & 65.6 & 69.2 \\
\midrule
\multicolumn{21}{l}{\textit{32--200B}} \\
Qwen3.6 35B-A3B & 93.1 & 78.6 & 83.1 & 66.5 & 85.8 & 88.7 & 75.5 & 21.4 & 52.7 & 78.9 & 82.8 & 61.1 & 85.7 & 96.9 & 93.6 & 83.7 & 76.4 & 47.6 & 84.0 & 75.6 \\
Apertus 70B & 77.1 & 58.5 & 67.4 & 47.8 & 83.0 & 90.3 & 63.0 & 17.8 & 32.5 & 61.8 & 69.3 & 53.8 & 71.9 & 95.2 & 91.7 & 84.7 & 66.9 & 40.3 & 75.3 & 65.7 \\
Llama 3.3 70B & 88.3 & 73.1 & 77.0 & 58.4 & 83.5 & 89.1 & 69.3 & 21.0 & 44.2 & 78.0 & 82.2 & 59.4 & 84.6 & 97.4 & 94.0 & 89.9 & 77.6 & 41.7 & 74.2 & 72.8 \\
Command A & 37.6 & 24.0 & 14.8 & 22.2 & 1.9 & 18.9 & 19.9 & 2.5 & 2.3 & 37.6 & 23.3 & 8.5 & 8.1 & 23.9 & 33.3 & 26.2 & 29.1 & 42.4 & 5.0 & 20.1 \\
gpt-oss 120B & 91.2 & 78.0 & 79.4 & 68.3 & 78.4 & 84.5 & 71.7 & 20.3 & 34.0 & 78.5 & 84.6 & 58.5 & 80.8 & 96.3 & 78.9 & 86.2 & 72.1 & 44.4 & 83.3 & 72.1 \\
\midrule
\multicolumn{21}{l}{\textit{$>$200B}} \\
MiniMax M2.7 & 93.2 & 83.1 & 87.2 & \underline{72.5} & 84.5 & 86.1 & 76.0 & 27.6 & 44.7 & 79.6 & 83.5 & 60.0 & 79.4 & 97.1 & 93.6 & 86.3 & 75.3 & 57.0 & 92.1 & 76.8 \\
Qwen3.5 397B-A17B & \textbf{95.4} & \underline{84.9} & \underline{89.2} & \textbf{72.9} & 88.1 & 91.7 & 80.3 & 29.4 & 45.7 & 83.2 & \underline{85.6} & 61.2 & 81.7 & 97.5 & 94.5 & 85.8 & 75.8 & 60.4 & \underline{92.8} & 78.7 \\
Mistral Large 3 675B & 91.6 & 78.9 & 84.2 & 63.2 & 87.8 & 92.3 & 76.3 & 26.6 & 38.8 & 76.4 & 84.0 & 44.8 & 49.2 & 87.5 & 59.8 & \underline{91.3} & \underline{79.1} & 52.0 & 86.4 & 71.1 \\
GLM-5.1 & 92.6 & 81.2 & 86.5 & 68.0 & 86.5 & 92.1 & 79.6 & \textbf{32.6} & 50.8 & 80.1 & 85.5 & 60.6 & \textbf{89.8} & 96.7 & 93.7 & 90.4 & 77.9 & 57.2 & 91.1 & 78.6 \\
Kimi K2.6 & 93.8 & 82.4 & 88.8 & 72.5 & 88.4 & \underline{93.3} & 78.8 & 29.7 & \textbf{53.6} & 83.8 & 83.4 & 62.0 & 86.9 & 97.3 & 94.1 & 91.1 & 77.2 & 60.6 & 92.1 & 79.5 \\
\midrule
\multicolumn{21}{l}{\textit{Proprietary (size undisclosed)}} \\
Claude Sonnet 4.6 & 94.2 & 82.7 & -- & -- & 88.8 & 92.7 & \underline{82.2} & -- & 42.0 & 84.3 & 85.5 & 59.5 & 83.1 & 97.7 & \underline{94.7} & 88.8 & 77.4 & \underline{63.5} & -- & \textbf{81.1} \\
Gemini 3.1 Flash-Lite & 94.9 & \textbf{85.7} & \textbf{90.7} & 71.9 & \textbf{92.0} & \textbf{93.7} & \textbf{85.2} & 21.1 & 46.5 & \underline{84.4} & \textbf{86.4} & 63.0 & 80.2 & \underline{97.8} & 94.7 & 91.0 & 78.0 & \textbf{66.7} & \textbf{94.1} & \underline{79.9} \\
GPT-5.4 mini & 91.1 & 75.3 & 80.8 & 61.8 & \underline{89.7} & 90.5 & 79.4 & \underline{30.3} & 44.6 & 75.0 & 80.6 & 59.9 & 86.1 & 97.5 & 91.6 & 90.7 & 72.5 & 46.5 & 91.0 & 75.5 \\
\bottomrule
\end{tabular}%
}
\caption{sk-bench results across all tasks. Slovak (SK), 3-shot --- the few-shot counterpart of Table~\ref{tab:skbench-comprehensive}, restricted to the tasks and models that were run with 3-shot exemplars (random exemplars from each task's train/validation split). Summarisation, instruction following, and the other tasks run zero-shot only are omitted, as are the models without few-shot results. Each cell is the task's primary metric ($\times100$), with \textbf{best} and \underline{second-best} per column. Avg is the unweighted mean of the columns shown.}
\label{tab:skbench-comprehensive-3shot}
\end{table*}

\begin{table*}[t]
\centering
\resizebox{\linewidth}{!}{%
\begin{tabular}{lrrrrrrrrrrrrrrrrrr|r}
\toprule
\textbf{Model} & \rotatebox{90}{\hyperref[task:arc-sk]{ARC}} & \rotatebox{90}{\hyperref[task:mmlu-sk]{MMLU}} & \rotatebox{90}{\hyperref[task:gmmlu]{G-MMLU}} & \rotatebox{90}{\hyperref[task:scio]{Scio}} & \rotatebox{90}{\hyperref[task:autoskola]{Driving}} & \rotatebox{90}{\hyperref[task:copa]{COPA}} & \rotatebox{90}{\hyperref[task:sktest]{SKTest-M}} & \rotatebox{90}{\hyperref[task:sktest]{SKTest-Q}} & \rotatebox{90}{\hyperref[task:sklep-qa]{skLEP-QA}} & \rotatebox{90}{\hyperref[task:sklep-nli]{skLEP-NLI}} & \rotatebox{90}{\hyperref[task:sklep-rte]{skLEP-RTE}} & \rotatebox{90}{\hyperref[task:senti]{Senti-3}} & \rotatebox{90}{\hyperref[task:senti]{Senti-2}} & \rotatebox{90}{\hyperref[task:sklep-sent]{skLEP-Sent}} & \rotatebox{90}{\hyperref[task:movierev]{MovieRev}} & \rotatebox{90}{\hyperref[task:toxic]{Toxic}} & \rotatebox{90}{\hyperref[task:sklep-hate]{skLEP-Hate}} & \rotatebox{90}{\textbf{Avg}} \\
\midrule
\multicolumn{20}{l}{\textit{$<$2B}} \\
Granite 4.0 350M & 26.3 & 28.0 & 29.2 & 25.3 & 22.2 & 49.9 & 28.6 & 1.3 & 11.4 & 34.2 & 50.3 & 12.7 & 70.0 & 90.1 & 70.3 & 46.4 & 72.7 & 0.2 & 37.2 \\
Granite 4.0-H 350M & 27.1 & 27.6 & 28.1 & 28.9 & 31.3 & 50.6 & 30.7 & 1.1 & 11.1 & 31.0 & 55.1 & 55.1 & \underline{87.3} & 64.1 & 60.0 & 47.0 & 72.5 & 0.2 & 39.4 \\
Qwen3 0.6B & 36.8 & 34.1 & 34.3 & 29.6 & 30.7 & 53.3 & 31.8 & 2.1 & 15.5 & 36.7 & 55.2 & 44.1 & 81.0 & 83.0 & 67.6 & 55.7 & 63.7 & 0.4 & 42.0 \\
Qwen3.5 0.8B & 43.3 & 35.0 & 37.6 & 28.6 & 48.7 & 51.9 & 31.6 & 1.7 & 14.6 & 36.7 & 60.4 & 48.7 & 76.6 & 89.6 & 74.7 & 57.9 & 71.5 & 0.6 & 45.0 \\
Gemma 3 1B & 30.5 & 29.9 & 33.6 & 24.0 & 67.9 & 52.4 & 35.8 & 2.2 & 18.1 & 40.3 & 61.5 & 50.6 & 64.4 & 91.5 & 78.0 & 72.7 & 67.4 & 0.9 & 45.7 \\
Granite 4.0 1B & 45.5 & 40.2 & 43.3 & 34.5 & 48.9 & 58.9 & 39.8 & 3.5 & 23.0 & 57.5 & 71.6 & 51.2 & 61.8 & 92.6 & 83.8 & 72.7 & 72.8 & 1.9 & 50.2 \\
Granite 4.0-H 1B & 53.5 & 43.2 & 46.2 & 36.2 & 68.5 & 53.7 & 43.9 & 1.4 & 21.0 & 56.0 & 72.9 & 45.8 & 65.1 & 93.6 & 83.0 & 55.4 & 72.6 & 0.3 & 50.7 \\
EuroLLM 1.7B & 27.5 & 27.1 & 27.8 & 27.4 & 17.2 & 50.2 & 26.7 & 4.3 & 19.5 & 35.8 & 52.5 & 58.0 & 50.3 & 94.1 & 82.5 & 38.7 & 71.7 & 5.1 & 39.8 \\
Qwen3 1.7B & 55.8 & 41.7 & 44.1 & 37.7 & 63.5 & 62.6 & 39.4 & 4.2 & 27.1 & 56.1 & 70.1 & 39.1 & 71.3 & 90.7 & 83.1 & 68.0 & 58.0 & 3.3 & 50.9 \\
\midrule
\multicolumn{20}{l}{\textit{2--9B}} \\
Granite 3.2 2B & 42.5 & 40.7 & 42.7 & 31.7 & 50.6 & 59.8 & 37.9 & 2.2 & 29.0 & 62.5 & 75.4 & 60.7 & 73.9 & 93.3 & 86.6 & 74.0 & 76.0 & 3.5 & 52.4 \\
Qwen3.5 2B & 64.2 & 46.0 & 47.2 & 40.2 & 56.9 & 56.1 & 44.7 & 4.6 & 31.3 & 47.2 & 74.2 & 46.2 & 71.6 & 83.5 & 83.3 & 77.0 & 74.0 & 1.5 & 52.8 \\
Llama 3.2 3B & 11.1 & 14.5 & 9.3 & 9.6 & 14.6 & 50.3 & 18.0 & 2.2 & 15.7 & 35.4 & 48.9 & 38.5 & 58.1 & 40.3 & 53.9 & 58.4 & 61.1 & 5.2 & 30.3 \\
Tiny-Aya Global & 66.6 & 51.3 & 55.3 & 39.3 & 71.1 & 61.1 & 53.8 & 6.0 & 23.9 & 66.8 & 73.3 & 49.8 & 66.5 & 91.1 & 86.6 & 81.5 & 76.3 & 5.6 & 57.0 \\
Tiny-Aya Water & 67.9 & 52.0 & 55.7 & 40.8 & 70.5 & 63.5 & 54.4 & 6.0 & 23.1 & 66.6 & 72.3 & 51.1 & 72.0 & 92.4 & 89.1 & 82.6 & 77.1 & 5.9 & 57.9 \\
Mistral-SK 7B & 36.4 & 28.9 & 38.0 & 29.0 & 52.9 & 52.8 & 31.1 & 12.3 & 26.8 & 35.6 & 53.5 & 23.1 & 73.2 & 88.0 & 90.5 & 43.5 & 68.2 & 13.5 & 44.3 \\
Mistral-SK 7B (Alpaca) & 20.0 & 5.1 & 9.3 & 4.3 & 1.5 & 50.6 & 2.8 & 2.2 & 6.7 & 0.8 & 0.5 & 0.4 & 16.1 & 15.1 & 0.2 & 1.2 & 3.8 & 0.2 & 7.8 \\
Teuken 7B & 54.9 & 43.3 & 51.3 & 34.2 & 70.0 & 72.7 & 47.8 & 6.6 & 22.8 & 52.8 & 66.3 & 33.4 & 51.3 & 63.8 & 77.0 & 71.5 & 73.6 & 1.0 & 49.7 \\
Aya Expanse 8B & 64.2 & 50.5 & 54.6 & 40.7 & 60.5 & 77.1 & 54.5 & 6.2 & 33.2 & 64.4 & 76.7 & 26.1 & 44.4 & 63.2 & 55.5 & 83.6 & 78.0 & 3.5 & 52.0 \\
Llama 3.1 8B & 43.5 & 32.7 & 42.5 & 28.8 & 25.4 & 65.6 & 32.6 & 6.4 & 26.2 & 46.8 & 67.5 & 41.9 & 41.0 & 65.0 & 56.6 & 60.0 & 50.7 & 10.1 & 41.3 \\
Ministral 3 8B & 79.9 & 59.5 & 63.7 & 47.3 & 57.6 & 72.9 & 57.5 & 13.2 & 25.7 & 56.4 & 70.9 & 54.3 & 67.1 & 93.9 & 88.9 & 86.3 & 71.0 & 33.2 & 61.1 \\
Ministral 3 8B (R) & 33.7 & 18.2 & 16.6 & 13.3 & 60.9 & 0.3 & 30.4 & 1.8 & 21.1 & 37.8 & 52.2 & 61.7 & 74.8 & 59.8 & 72.9 & 54.4 & 43.0 & 30.0 & 37.9 \\
\midrule
\multicolumn{20}{l}{\textit{9--32B}} \\
EuroLLM 9B & 78.0 & 55.3 & 61.1 & 45.0 & 77.6 & 77.7 & 63.0 & 14.8 & 36.2 & 73.5 & 81.2 & 56.4 & 71.0 & 97.1 & 91.2 & 90.0 & 76.5 & 23.6 & 65.0 \\
Qwen3.5 9B & 89.1 & 69.6 & 70.4 & 60.2 & 81.4 & 87.3 & 67.7 & 16.7 & 46.1 & 78.6 & 83.0 & 54.8 & 57.2 & 95.2 & 90.8 & 85.4 & 77.2 & 31.8 & 69.0 \\
Bielik 11B & 84.9 & 64.3 & 69.7 & 55.2 & 79.0 & 88.4 & 68.1 & 19.9 & 43.0 & 80.0 & 81.5 & 62.7 & 75.8 & 96.7 & 93.2 & 89.7 & 78.3 & 34.2 & 70.3 \\
Ministral 3 14B & 81.5 & 62.6 & 65.8 & 51.1 & 74.4 & 85.1 & 60.0 & 12.2 & 31.3 & 62.9 & 80.5 & 53.9 & 59.9 & 95.1 & 85.7 & 82.7 & 75.5 & 39.1 & 64.4 \\
Ministral 3 14B (R) & 73.8 & 43.5 & 26.8 & 32.8 & 60.3 & 82.2 & 37.1 & 1.4 & 4.2 & 20.1 & 54.8 & 14.0 & 3.9 & 29.4 & 42.7 & 17.2 & 48.0 & 8.8 & 33.4 \\
Qwen3 14B & 87.1 & 67.5 & 68.8 & 59.1 & 85.9 & 83.8 & 63.9 & 16.5 & 45.5 & 81.8 & 82.7 & 49.2 & 76.9 & 96.5 & 91.1 & 85.2 & 77.7 & 38.3 & 69.9 \\
Qwen3-SK 14B & 83.7 & 54.5 & 62.7 & 31.4 & 66.4 & 3.5 & 64.1 & 17.5 & 34.1 & 62.9 & 57.7 & 11.4 & 12.6 & 15.4 & 29.0 & 60.4 & 60.7 & 47.1 & 43.1 \\
Qwen3-SK 14B (Alpaca) & 88.1 & 69.6 & 67.9 & 60.9 & 87.1 & 87.6 & 69.1 & 13.7 & 13.0 & 57.5 & 49.5 & 21.7 & 53.9 & 14.7 & 12.2 & 43.1 & 45.5 & 18.3 & 48.5 \\
gpt-oss 20B & 88.4 & 72.6 & 73.8 & 64.7 & 71.2 & 79.3 & 65.3 & 12.5 & 31.0 & 77.2 & 84.1 & 55.3 & 76.7 & 97.0 & 88.3 & 81.5 & 75.0 & 31.6 & 68.1 \\
EuroLLM 22B & 80.1 & 59.7 & 64.3 & 50.1 & 79.6 & 84.7 & 67.7 & 19.2 & 39.2 & 73.9 & 82.6 & 58.5 & 77.8 & 97.1 & 92.8 & 88.3 & 76.9 & 33.6 & 68.1 \\
Qwen3.5 27B & \underline{95.2} & 82.4 & 86.2 & 68.9 & 88.1 & 89.7 & 77.7 & 25.4 & 52.2 & 81.0 & 85.7 & 57.1 & 82.8 & 97.8 & 94.6 & 89.1 & 78.1 & 50.5 & 76.8 \\
Qwen3.6 27B & 94.7 & 81.9 & 84.2 & 68.5 & 88.1 & 87.8 & 76.2 & 25.8 & \textbf{53.2} & 77.9 & \underline{86.3} & 60.5 & 84.8 & \underline{98.0} & 94.7 & 86.7 & 75.5 & 50.6 & 76.4 \\
TildeOpen 30B & 56.1 & 33.9 & 38.5 & 30.1 & 76.1 & 51.9 & 44.6 & 2.4 & 1.8 & 13.5 & 21.2 & 34.5 & 31.8 & 56.2 & 33.9 & 49.7 & 21.8 & 16.2 & 34.1 \\
TildeOpen 30B-64k & 77.6 & 56.8 & 63.1 & 48.1 & 82.0 & 84.5 & 70.5 & 28.0 & 34.6 & 45.5 & 64.7 & \textbf{64.3} & 79.3 & \textbf{98.2} & 94.2 & 83.0 & 70.2 & 46.0 & 66.2 \\
TildeOpen-SK 30B (Alpaca) & 70.7 & 44.3 & 43.7 & 33.9 & 79.4 & 85.1 & 63.9 & 12.2 & 24.3 & 36.5 & 30.0 & 27.8 & 19.6 & 34.6 & 16.9 & 4.1 & 17.3 & 10.2 & 36.4 \\
Gemma 4 31B & 94.8 & 82.1 & 86.1 & 71.3 & 89.5 & 83.7 & 77.9 & 29.7 & 51.5 & \underline{85.2} & 85.6 & 61.8 & 84.3 & 97.4 & 94.9 & 90.4 & 77.4 & 51.8 & 77.5 \\
Aya Expanse 32B & 79.6 & 63.2 & 67.8 & 51.6 & 78.4 & 86.1 & 65.4 & 12.2 & 43.6 & 79.8 & 82.0 & 41.9 & 73.7 & 94.3 & 92.5 & \underline{90.8} & \underline{79.2} & 16.9 & 66.6 \\
Granite 4.0-H Small & 86.2 & 67.5 & 75.2 & 56.1 & 77.3 & 88.1 & 67.7 & 13.7 & 38.1 & 80.4 & 80.1 & 54.9 & 69.0 & 96.4 & 91.2 & 88.6 & 75.8 & 32.3 & 68.8 \\
\midrule
\multicolumn{20}{l}{\textit{32--200B}} \\
Qwen3.6 35B-A3B & 93.2 & 78.7 & 82.9 & 66.7 & 87.6 & 88.0 & 75.4 & 21.7 & 52.6 & 80.4 & 83.5 & 53.9 & 77.4 & 97.4 & 93.4 & 85.7 & 76.7 & 49.8 & 74.7 \\
Apertus 70B & 77.0 & 59.3 & 67.8 & 50.3 & 82.9 & 88.2 & 66.0 & 15.8 & 27.5 & 66.8 & 76.5 & 39.2 & 69.1 & 94.2 & 91.2 & 60.2 & 62.8 & 39.9 & 63.0 \\
Llama 3.3 70B & 88.1 & 73.5 & 78.8 & 59.2 & 83.4 & 88.7 & 70.4 & 21.3 & 44.6 & 78.5 & 80.8 & 50.0 & 62.4 & 95.6 & 93.4 & 90.0 & 78.1 & 43.8 & 71.2 \\
Command A & 1.2 & 11.0 & 7.2 & 6.8 & 0.5 & 5.2 & 14.6 & 4.2 & 10.1 & 20.8 & 35.6 & 22.9 & 36.9 & 32.6 & 30.8 & 35.7 & 27.5 & 45.9 & 19.4 \\
gpt-oss 120B & 91.5 & 77.7 & 80.8 & 67.2 & 77.0 & 82.7 & 67.6 & 20.6 & 36.2 & 80.4 & 85.5 & 57.7 & 79.9 & 96.4 & 83.6 & 83.6 & 74.4 & 47.2 & 71.7 \\
\midrule
\multicolumn{20}{l}{\textit{$>$200B}} \\
MiniMax M2.7 & 93.3 & 82.9 & 86.2 & \textbf{73.3} & 85.8 & 83.5 & 76.5 & 28.9 & 46.3 & 81.1 & 82.3 & 60.9 & 79.5 & 97.2 & 93.8 & 84.4 & 75.3 & 59.7 & 76.2 \\
Qwen3.5 397B-A17B & \textbf{95.4} & \underline{85.1} & \textbf{90.5} & \underline{73.1} & 89.1 & \textbf{93.5} & 81.0 & 32.2 & 47.8 & 85.1 & 84.7 & 62.5 & 83.5 & 97.6 & \underline{95.0} & 83.7 & 76.0 & 60.5 & 78.7 \\
Mistral Large 3 675B & 92.0 & 74.2 & 82.7 & 59.9 & 84.6 & 91.5 & 74.5 & 26.8 & 39.1 & 77.6 & 83.5 & 49.2 & 57.6 & 89.5 & 68.9 & 89.0 & \textbf{79.9} & 53.4 & 70.8 \\
GLM-5.1 & 92.5 & 81.4 & 85.5 & 68.4 & 86.5 & 92.7 & 79.4 & 32.2 & 49.6 & 82.5 & 85.3 & 62.4 & \textbf{89.6} & 97.2 & 93.5 & \textbf{91.1} & 78.1 & 57.7 & 78.1 \\
Kimi K2.6 & 94.0 & 82.6 & 88.8 & 72.6 & 87.4 & \underline{92.9} & 78.1 & \underline{32.5} & \underline{52.7} & \textbf{85.8} & 84.0 & 59.5 & 78.9 & 97.4 & 93.8 & 89.1 & 78.3 & 61.9 & 78.4 \\
\midrule
\multicolumn{20}{l}{\textit{Proprietary (size undisclosed)}} \\
Claude Sonnet 4.6 & 93.9 & 81.4 & -- & -- & \underline{90.1} & 92.4 & \underline{83.1} & -- & 45.5 & 81.8 & 85.0 & 58.2 & 84.4 & 97.6 & 94.6 & 87.4 & 77.5 & \underline{64.4} & \textbf{81.1} \\
Gemini 3.1 Flash-Lite & 94.6 & \textbf{85.6} & \underline{90.4} & 72.0 & \textbf{91.7} & 92.8 & \textbf{85.1} & \textbf{33.5} & 50.3 & 84.3 & \textbf{86.5} & \underline{63.1} & 83.8 & 98.0 & \textbf{95.4} & 90.0 & 77.2 & \textbf{66.3} & \underline{80.0} \\
GPT-5.4 mini & 90.7 & 75.3 & 83.0 & 61.1 & 89.7 & 90.3 & 79.4 & 29.1 & 44.6 & 74.4 & 79.7 & 60.9 & 86.9 & 97.6 & 91.3 & 90.2 & 76.4 & 46.7 & 74.9 \\
\bottomrule
\end{tabular}%
}
\caption{sk-bench results across all tasks. English (EN) prompts, 3-shot --- the English-template, few-shot counterpart of Table~\ref{tab:skbench-comprehensive}, restricted to the tasks and models run with 3-shot exemplars. Summarisation, instruction following, the Slovak-only tasks (IFEval-SK, Slohry), and the other tasks run zero-shot only are omitted, as are the models without few-shot results. Each cell is the task's primary metric ($\times100$), with \textbf{best} and \underline{second-best} per column. Avg is the unweighted mean of the columns shown.}
\label{tab:skbench-comprehensive-en-3shot}
\end{table*}

\subsection{Generator self-preference check}
\label{app:self-preference}
One task in the suite has LLM-generated questions: MultiWikiQA-SK's
questions were generated by Gemini~1.5~Pro over the November~2023
Slovak Wikipedia dump, then rephrased in a second pass by the same model to reduce surface word
overlap with the passage. This could favour the generator's own model
family. We check this against the human-authored task in the same
reporting group, skLEP-QA: if the generator family were favoured, Gemini models
would rank higher on the generated set than on human-authored skLEP-QA.
They do not. Across the 55 models, Gemini~3.1~Pro ranks 1st on
human-authored skLEP-QA and 4th on the Gemini-generated set, and
Gemini~3.1~Flash-Lite drops from 30th to 38th---both rank \emph{lower}
on the generated set, the opposite of self-preference. The models that gain most are small and older ones (e.g.\ Mistral-SK~7B,
$+26$ places), which suggests that weaker models rely on surface word
matching, not that the generator is favoured. The low agreement between
human-authored and LLM-generated QA in RQ1 ($\rho=0.72$,
\S\ref{sec:rq1}) thus reflects how the generated questions can be
answered, not which model generated them.

\section{Collection and Curation of New Datasets}
\label{app:collection}

For each dataset introduced with sk-bench whose provenance warrants
more than the one-line summary in Appendix~\ref{app:tasks-detail},
this section records how the items were obtained, who curated them,
and whether human annotators were involved. Chiby has its own curation
card in Appendix~\ref{app:chiby} and IFEval-SK's instruction registry
is described in Appendix~\ref{app:ifevalsk}.

\subsection{Scio}
\label{app:card-scio}
\textbf{Collection.}~Parsed by the authors from Scio admission-test
papers, covering general study prerequisites alongside biology,
chemistry, and mathematics.
\textbf{Curation.}~The parsing, cleaning, and answer-key alignment are
ours. No external annotators were involved.
\textbf{Licence.}~No licence is declared on the source artefact, so
the catalogue (Table~\ref{tab:dataset-catalogue}) records ``---''.

\subsection{Autoškola}
\label{app:card-autoskola}
\textbf{Collection.}~The official Slovak driving-licence theory
questions, spanning 35 topic categories, with the official answer key
as gold.
\textbf{Curation.}~An LLM assisted in constructing the dataset. The
gold labels themselves are the official answer key, not model output.
This use of an LLM inside the benchmark is also flagged in the
Ethical Considerations section and carries a provenance label in the
catalogue.

\subsection{SKJ1}
\label{app:card-skj1}
\textbf{Collection.}~Adopted from the SKJ1 dataset~\citep{skj1}: the
items are expert-authored by the team behind that dataset, and we did
not recruit experts ourselves.
\textbf{Curation.}~sk-bench's contribution is the generative-LLM
wrapping---the prompts, the answer parsing, and the scoring.
\textbf{Licence.}~No licence is declared on the source artefact, so
the catalogue (Table~\ref{tab:dataset-catalogue}) records ``---''.

\section{Summarisation Datasets Downsampling}
\label{app:downsampling}

To keep evaluation affordable while preserving the length distribution,
we applied a stratified downsampling procedure based on text length. For each dataset (SlovakSum and SME-Sum), we divided the full range of document lengths into 10 equal-width bins and allocated the target sample size of 1,500 examples per split proportionally according to each bin's representation in the original data. This keeps rare long documents in the sample, which uniform random
sampling would tend to miss. Any rounding remainder from the proportional allocation was distributed to the largest bins, with final samples drawn without replacement within each bin. The resulting subsets preserve the distributional statistics (median, interquartile range, and mean) while reducing computational costs.

\section{Chiby-classification: Dataset Curation}
\label{app:chiby}
To construct the dataset chiby-classification, we preprocessed and annotated the dataset Chiby v0.4 \cite{Garabik2026}, which is based on revisions of Slovak Wikipedia.
\newline

The preprocessing was carried out in several steps. First, the data were decompressed and cleaned by removing tags from sentences, and errors were extracted into new columns for annotation purposes. The error tags belong to one of seven categories: unclassified, spelling, typographical, punct, lexicosemantic, diacritics, and capitalisation.

Of the seven revision types present in the original dataset, three were selected: spelling, diacritics, and capitalisation. Sentence pairs not containing any of the selected error types were removed. The following sentence pairs were then excluded:

\begin{itemize}
    \item Pairs containing more than one type of error.
    \item Pairs in which the sentence had fewer than 55 characters or more than 300 characters.
    \item Pairs containing a URL (\texttt{http}, \texttt{www}).
    \item Pairs that were parsed incorrectly, identified as sentences not beginning with a capital letter or not ending with a full stop, exclamation mark, or question mark.
\end{itemize}

Sentence similarity was then checked with cosine similarity to confirm correct pairing between the original and revised sentences. The threshold was selected empirically through manual inspection of pairs produced by several candidate thresholds, including 0.90 and 0.95, and 0.99 was found to best preserve correctly paired sentences while filtering out mismatched pairs. Only pairs with a cosine similarity score above 0.99 were retained. 

Finally, the \texttt{langdetect} library\footnote{\url{https://pypi.org/project/langdetect/}} was used to verify that the sentences were in Slovak, those identified as being in another language were removed.

After preprocessing, 67\% of pairs contained spelling errors, 23\% diacritics errors, and 10\% capitalisation errors.

A total of 2,400 sentence pairs were then randomly sampled to match this distribution: 1,608 from the spelling category, 552 from the diacritics category, and 240 from the capitalisation category.
\newline

The dataset was annotated by three native Slovak speakers, all of whom hold a university-level degree and are co-authors of this work. Each sentence pair was annotated by a single annotator, who determined whether the first sentence (original) was correct, the second sentence (revised) was correct, or neither.

As each pair was reviewed by one annotator only, inter-annotator agreement was not computed. Where a grammatical error was minor, it was corrected in the revised sentence and the pair was retained. Pairs for which neither sentence was deemed correct and the error could not be straightforwardly corrected were discarded (see Section~\ref{app:chiby_annotations} for full annotation instructions). 
\newline
After annotation, 1,716 sentence pairs remained and were randomly shuffled. Sentences were labelled 0 if incorrect and 1 if correct, resulting in a total of 3,432 rows in the dataset.
\newline

This dataset is released under the Creative Commons Attribution-ShareAlike 4.0 International (CC BY-SA 4.0)\footnote{\url{https://creativecommons.org/licenses/by-sa/4.0/deed.en}} licence, consistent with the licence of the Chiby v0.4 dataset.

\subsection{Annotation Instructions}
\label{app:chiby_annotations}

You will be shown a series of sentence pairs. Each pair consists of an original sentence and a revised sentence. For each pair, read both sentences and assign \textbf{one} of the following labels:

\begin{itemize}
    \item \textit{original} – the original sentence is correct as written (the revision was unnecessary or introduces an error).
    \item \textit{revision} – the revised sentence is the correct one (it fixes an error present in the original).
    \item \textit{fixed} – both sentences contain an error, but you as the annotator have corrected the revised sentence and provided the correct version.
    \item \textit{neither} – both sentences are incorrect and the error cannot be straightforwardly corrected.
\end{itemize}

\textbf{When in doubt, ask yourself:}\textit{Is either sentence fully correct on its own?} If yes, label accordingly. If neither sentence is correct but you can see a clear correction, write the corrected version and label it \textit{fixed}. If the error is too complex to correct, label it \textit{neither}.

\textbf{\noindent Please note:}
\vspace{-\topsep}
\begin{itemize}
    \item You do not need to identify or describe what type of error is present – only whether the sentence is correct or not.
    \item When using the \textit{fixed} label, make sure to provide your corrected version of the sentence.
    \item If you label a pair as \textit{neither}, it will be discarded from the final dataset. Only use this label when the error is too complex or unclear to correct in a simple revision.
    \item Each sentence pair is reviewed by you alone, so take your time and read carefully.
\end{itemize}

\section{IFEval-SK: Slovak Instruction Registry}
\label{app:ifevalsk}
\textbf{Size and splits.}~537 prompts.
\textbf{Adaptation procedure.}~For each English instruction type in
\citet{ifeval2023}, we (i)~translated the instruction template, and
(ii)~re-implemented the programmatic checker against Slovak word
boundaries, casing rules, and punctuation.

\onecolumn
\section{Prompt Library}
\label{app:prompts}
Most tasks have three prompt variants in both Slovak and English to
measure prompt sensitivity (\S\ref{sec:setup}). IFEval-SK uses one
Slovak prompt per item, and Slohry is also Slovak-only.
Each box below pairs the Slovak (left)
and English (right) form of one variant. \ph{slots} mark the fields
filled in at run time. Typographic quotation marks are reproduced as in
the released templates.

\subsection{NLI --- binary}
\begin{tcolorbox}[promptbox,title={NLI --- binary \hfill \normalfont\itshape Prompt 1}]
\begin{minipage}[t]{0.475\linewidth}
\promptlang{Slovak}\\[2pt]
{\promptmono\raggedright\footnotesize Aký je vzťah medzi nasledujúcimi dvomi vetami? Vyplýva druhá veta z prvej alebo jej protirečí? Odpovedz iba slovom \quotedblbase{}vyplýva\textquotedblleft{} alebo \quotedblbase{}protirečí\textquotedblleft{}.

Premisa: \ph{sentence1} \\
Hypotéza: \ph{sentence2}

Odpoveď:\par}
\end{minipage}\hfill\promptdivider\hfill
\begin{minipage}[t]{0.475\linewidth}
\promptlang{English}\\[2pt]
{\promptmono\raggedright\footnotesize What is the relationship between the two sentences? Does the first sentence entail or contradict the second sentence? Answer only with the word "entailment" or "contradiction".

Premise: \ph{sentence1} \\
Hypothesis: \ph{sentence2}

Answer:\par}
\end{minipage}
\end{tcolorbox}
\begin{tcolorbox}[promptbox,title={NLI --- binary \hfill \normalfont\itshape Prompt 2}]
\begin{minipage}[t]{0.475\linewidth}
\promptlang{Slovak}\\[2pt]
{\promptmono\raggedright\footnotesize Urči, či hypotéza vyplýva z danej premisy. Odpovedz iba slovom \quotedblbase{}áno\textquotedblleft{} alebo \quotedblbase{}nie\textquotedblleft{}.

Premisa: \ph{sentence1} \\
Hypotéza: \ph{sentence2}

Odpoveď:\par}
\end{minipage}\hfill\promptdivider\hfill
\begin{minipage}[t]{0.475\linewidth}
\promptlang{English}\\[2pt]
{\promptmono\raggedright\footnotesize Determine whether the premise entails the hypothesis. Answer only "yes" or "no".

Premise: \ph{sentence1} \\
Hypothesis: \ph{sentence2}

Answer:\par}
\end{minipage}
\end{tcolorbox}
\begin{tcolorbox}[promptbox,title={NLI --- binary \hfill \normalfont\itshape Prompt 3}]
\begin{minipage}[t]{0.475\linewidth}
\promptlang{Slovak}\\[2pt]
{\promptmono\raggedright\footnotesize Napíšem dve vety, ktorým hovorím premisa a hypotéza. Hypotéza môže z premisy buď vyplývať alebo jej protirečiť. Vyhodnoť vzťah týchto viet a odpovedz buď slovom \quotedblbase{}vyplýva\textquotedblleft{} alebo \quotedblbase{}protirečí\textquotedblleft{}.

Premisa: \ph{sentence1} \\
Hypotéza: \ph{sentence2}

Odpoveď:\par}
\end{minipage}\hfill\promptdivider\hfill
\begin{minipage}[t]{0.475\linewidth}
\promptlang{English}\\[2pt]
{\promptmono\raggedright\footnotesize I will write two sentences, which I call the premise and the hypothesis. The premise may entail the hypothesis or contradict it. Determine the relationship between these sentences and answer with either "entailment" or "contradiction".

Premise: \ph{sentence1} \\
Hypothesis: \ph{sentence2}

Answer:\par}
\end{minipage}
\end{tcolorbox}

\subsection{NLI --- 3-class}
\begin{tcolorbox}[promptbox,title={NLI --- 3-class \hfill \normalfont\itshape Prompt 1}]
\begin{minipage}[t]{0.475\linewidth}
\promptlang{Slovak}\\[2pt]
{\promptmono\raggedright\footnotesize Aký je vzťah medzi dvomi vetami? Druhá veta z prvej vyplýva, protirečí jej alebo je neutrálna? Odpovedz iba slovom \quotedblbase{}vyplýva\textquotedblleft{}, \quotedblbase{}protirečí\textquotedblleft{} alebo \quotedblbase{}neutrálny\textquotedblleft{}.

Premisa: \ph{sentence1} \\
Hypotéza: \ph{sentence2}

Odpoveď:\par}
\end{minipage}\hfill\promptdivider\hfill
\begin{minipage}[t]{0.475\linewidth}
\promptlang{English}\\[2pt]
{\promptmono\raggedright\footnotesize What is the relationship between the two sentences? Does the first sentence entail, contradict, or is neutral towards the second sentence? Answer only "entailment", "contradiction", or "neutral".

Premise: \ph{sentence1} \\
Hypothesis: \ph{sentence2}

Answer:\par}
\end{minipage}
\end{tcolorbox}
\begin{tcolorbox}[promptbox,title={NLI --- 3-class \hfill \normalfont\itshape Prompt 2}]
\begin{minipage}[t]{0.475\linewidth}
\promptlang{Slovak}\\[2pt]
{\promptmono\raggedright\footnotesize Napíšem dve vety, ktorým hovorím premisa a hypotéza. Hypotéza môže z premisy vyplývať, alebo jej protirečiť. Pokiaľ ani nevyplýva, ani neprotirečí, tak je neutrálna. Vyhodnoť, či hypotéza z premisy \quotedblbase{}vyplýva\textquotedblleft{}, alebo jej \quotedblbase{}protirečí\textquotedblleft{}, alebo je vzťah \quotedblbase{}neutrálny\textquotedblleft{}.

Premisa: \ph{sentence1} \\
Hypotéza: \ph{sentence2}

Odpoveď:\par}
\end{minipage}\hfill\promptdivider\hfill
\begin{minipage}[t]{0.475\linewidth}
\promptlang{English}\\[2pt]
{\promptmono\raggedright\footnotesize I will write two sentences, which I call the premise and the hypothesis. The premise may entail the hypothesis or contradict it. If it neither entails nor contradicts the hypothesis, then it is neutral. Determine the relationship as "entailment", "contradiction" or "neutral".

Premise: \ph{sentence1} \\
Hypothesis: \ph{sentence2}

Answer:\par}
\end{minipage}
\end{tcolorbox}
\begin{tcolorbox}[promptbox,title={NLI --- 3-class \hfill \normalfont\itshape Prompt 3}]
\begin{minipage}[t]{0.475\linewidth}
\promptlang{Slovak}\\[2pt]
{\promptmono\raggedright\footnotesize Úloha: Urči, či hypotéza vyplýva z premisy (vyplýva), či je s ňou v spore (protirečí) alebo ani jeden z týchto vzťahov (neutrálny).

Premisa: \ph{sentence1} \\
Hypotéza: \ph{sentence2}

Odpoveď:\par}
\end{minipage}\hfill\promptdivider\hfill
\begin{minipage}[t]{0.475\linewidth}
\promptlang{English}\\[2pt]
{\promptmono\raggedright\footnotesize Task: Determine whether the premise entails the hypothesis (entailment), contradicts the hypothesis (contradiction), or neither (neutral).

Premise: \ph{sentence1} \\
Hypothesis: \ph{sentence2}

Answer:\par}
\end{minipage}
\end{tcolorbox}

\subsection{Sentiment --- binary}
\begin{tcolorbox}[promptbox,title={Sentiment --- binary \hfill \normalfont\itshape Prompt 1}]
\begin{minipage}[t]{0.475\linewidth}
\promptlang{Slovak}\\[2pt]
{\promptmono\raggedright\footnotesize Urči sentiment nasledujúceho textu. Odpovedz iba slovom \quotedblbase{}negatívny\textquotedblleft{} alebo \quotedblbase{}pozitívny\textquotedblleft{}.

Text: \ph{text} \\
Sentiment považujem za:\par}
\end{minipage}\hfill\promptdivider\hfill
\begin{minipage}[t]{0.475\linewidth}
\promptlang{English}\\[2pt]
{\promptmono\raggedright\footnotesize Determine the sentiment of the following text. Answer only "negative" or "positive".

Text: \ph{text} \\
I consider the sentiment to be:\par}
\end{minipage}
\end{tcolorbox}
\begin{tcolorbox}[promptbox,title={Sentiment --- binary \hfill \normalfont\itshape Prompt 2}]
\begin{minipage}[t]{0.475\linewidth}
\promptlang{Slovak}\\[2pt]
{\promptmono\raggedright\footnotesize Rozhodni, či je sentiment nasledujúceho textu negatívny alebo pozitívny. Odpovedz iba slovom \quotedblbase{}negatívny\textquotedblleft{} alebo \quotedblbase{}pozitívny\textquotedblleft{}.

Text: \ph{text} \\
Odpoveď:\par}
\end{minipage}\hfill\promptdivider\hfill
\begin{minipage}[t]{0.475\linewidth}
\promptlang{English}\\[2pt]
{\promptmono\raggedright\footnotesize Determine whether the sentiment of the following text is negative or positive. Answer only "negative" or "positive".

Text: \ph{text} \\
Answer:\par}
\end{minipage}
\end{tcolorbox}
\begin{tcolorbox}[promptbox,title={Sentiment --- binary \hfill \normalfont\itshape Prompt 3}]
\begin{minipage}[t]{0.475\linewidth}
\promptlang{Slovak}\\[2pt]
{\promptmono\raggedright\footnotesize Klasifikuj sentiment do kategórií: negatívny/pozitívny.

Text: \ph{text} \\
Sentiment:\par}
\end{minipage}\hfill\promptdivider\hfill
\begin{minipage}[t]{0.475\linewidth}
\promptlang{English}\\[2pt]
{\promptmono\raggedright\footnotesize Classify sentiment into the following categories: negative/positive.

Text: \ph{text} \\
Sentiment:\par}
\end{minipage}
\end{tcolorbox}

\subsection{Sentiment --- 3-class}
\begin{tcolorbox}[promptbox,title={Sentiment --- 3-class \hfill \normalfont\itshape Prompt 1}]
\begin{minipage}[t]{0.475\linewidth}
\promptlang{Slovak}\\[2pt]
{\promptmono\raggedright\footnotesize Urči sentiment nasledujúceho textu. Odpovedz iba slovom \quotedblbase{}negatívny\textquotedblleft{}, \quotedblbase{}neutrálny\textquotedblleft{} alebo \quotedblbase{}pozitívny\textquotedblleft{}.

Text: \ph{text} \\
Sentiment považujem za:\par}
\end{minipage}\hfill\promptdivider\hfill
\begin{minipage}[t]{0.475\linewidth}
\promptlang{English}\\[2pt]
{\promptmono\raggedright\footnotesize Determine the sentiment of the following text. Answer only "negative", "neutral", or "positive".

Text: \ph{text} \\
I consider the sentiment to be:\par}
\end{minipage}
\end{tcolorbox}
\begin{tcolorbox}[promptbox,title={Sentiment --- 3-class \hfill \normalfont\itshape Prompt 2}]
\begin{minipage}[t]{0.475\linewidth}
\promptlang{Slovak}\\[2pt]
{\promptmono\raggedright\footnotesize Rozhodni, či sentiment nasledujúceho textu je negatívny, neutrálny alebo pozitívny. Odpovedz iba slovom \quotedblbase{}negatívny\textquotedblleft{}, \quotedblbase{}neutrálny\textquotedblleft{} alebo \quotedblbase{}pozitívny\textquotedblleft{}.

Text: \ph{text} \\
Odpoveď:\par}
\end{minipage}\hfill\promptdivider\hfill
\begin{minipage}[t]{0.475\linewidth}
\promptlang{English}\\[2pt]
{\promptmono\raggedright\footnotesize Determine whether the sentiment of the following text is negative, neutral, or positive. Answer only with "negative", "neutral", or "positive".

Text: \ph{text} \\
Answer:\par}
\end{minipage}
\end{tcolorbox}
\begin{tcolorbox}[promptbox,title={Sentiment --- 3-class \hfill \normalfont\itshape Prompt 3}]
\begin{minipage}[t]{0.475\linewidth}
\promptlang{Slovak}\\[2pt]
{\promptmono\raggedright\footnotesize Klasifikuj sentiment do kategórií: negatívny/neutrálny/pozitívny.

Text: \ph{text} \\
Sentiment:\par}
\end{minipage}\hfill\promptdivider\hfill
\begin{minipage}[t]{0.475\linewidth}
\promptlang{English}\\[2pt]
{\promptmono\raggedright\footnotesize Classify sentiment into the following categories: negative/neutral/positive.

Text: \ph{text} \\
Sentiment:\par}
\end{minipage}
\end{tcolorbox}

\subsection{Hate speech --- binary}
\begin{tcolorbox}[promptbox,title={Hate speech --- binary \hfill \normalfont\itshape Prompt 1}]
\begin{minipage}[t]{0.475\linewidth}
\promptlang{Slovak}\\[2pt]
{\promptmono\raggedright\footnotesize Urči, či zadaný výrok obsahuje nenávistné prejavy. Odpovedz len slovom \quotedblbase{}áno\textquotedblleft{} alebo \quotedblbase{}nie\textquotedblleft{}.

Výrok: \ph{text} \\
Odpoveď:\par}
\end{minipage}\hfill\promptdivider\hfill
\begin{minipage}[t]{0.475\linewidth}
\promptlang{English}\\[2pt]
{\promptmono\raggedright\footnotesize Determine whether the given claim includes hate speech. Answer only "yes" or "no".

Claim: \ph{text} \\
Answer:\par}
\end{minipage}
\end{tcolorbox}
\begin{tcolorbox}[promptbox,title={Hate speech --- binary \hfill \normalfont\itshape Prompt 2}]
\begin{minipage}[t]{0.475\linewidth}
\promptlang{Slovak}\\[2pt]
{\promptmono\raggedright\footnotesize Obsahuje nasledujúci text nenávistné prejavy? Odpovedz áno/nie.

Text: \ph{text} \\
Odpoveď:\par}
\end{minipage}\hfill\promptdivider\hfill
\begin{minipage}[t]{0.475\linewidth}
\promptlang{English}\\[2pt]
{\promptmono\raggedright\footnotesize Does the following text contain hate speech? Answer yes/no.

Text: \ph{text} \\
Answer:\par}
\end{minipage}
\end{tcolorbox}
\begin{tcolorbox}[promptbox,title={Hate speech --- binary \hfill \normalfont\itshape Prompt 3}]
\begin{minipage}[t]{0.475\linewidth}
\promptlang{Slovak}\\[2pt]
{\promptmono\raggedright\footnotesize Rozhodni, či nasledujúci text obsahuje známky nenávistného prejavu (hate speech). Odpovedz iba slovom \quotedblbase{}áno\textquotedblleft{} alebo \quotedblbase{}nie\textquotedblleft{}.

Text: \ph{text} \\
Odpoveď:\par}
\end{minipage}\hfill\promptdivider\hfill
\begin{minipage}[t]{0.475\linewidth}
\promptlang{English}\\[2pt]
{\promptmono\raggedright\footnotesize Determine whether the following text contains signs of hate speech. Respond only with the word "yes" or "no".

Text: \ph{text} \\
Answer:\par}
\end{minipage}
\end{tcolorbox}

\subsection{Toxicity --- binary}
\begin{tcolorbox}[promptbox,title={Toxicity --- binary \hfill \normalfont\itshape Prompt 1}]
\begin{minipage}[t]{0.475\linewidth}
\promptlang{Slovak}\\[2pt]
{\promptmono\raggedright\footnotesize Urči, či daný text obsahuje toxické prejavy. Odpovedz len \quotedblbase{}áno\textquotedblleft{} alebo \quotedblbase{}nie\textquotedblleft{}.

Text: \ph{text} \\
Odpoveď:\par}
\end{minipage}\hfill\promptdivider\hfill
\begin{minipage}[t]{0.475\linewidth}
\promptlang{English}\\[2pt]
{\promptmono\raggedright\footnotesize Determine whether the given text includes toxic speech. Answer only "yes" or "no".

Text: \ph{text} \\
Answer:\par}
\end{minipage}
\end{tcolorbox}
\begin{tcolorbox}[promptbox,title={Toxicity --- binary \hfill \normalfont\itshape Prompt 2}]
\begin{minipage}[t]{0.475\linewidth}
\promptlang{Slovak}\\[2pt]
{\promptmono\raggedright\footnotesize Je nasledujúci text toxický? Odpovedz len \quotedblbase{}áno\textquotedblleft{} alebo \quotedblbase{}nie\textquotedblleft{}.

Text: \ph{text} \\
Odpoveď:\par}
\end{minipage}\hfill\promptdivider\hfill
\begin{minipage}[t]{0.475\linewidth}
\promptlang{English}\\[2pt]
{\promptmono\raggedright\footnotesize Is the following text toxic? Answer only "yes" or "no".

Text: \ph{text} \\
Answer:\par}
\end{minipage}
\end{tcolorbox}
\begin{tcolorbox}[promptbox,title={Toxicity --- binary \hfill \normalfont\itshape Prompt 3}]
\begin{minipage}[t]{0.475\linewidth}
\promptlang{Slovak}\\[2pt]
{\promptmono\raggedright\footnotesize Zisti, či v texte nižšie nájdeš akúkoľvek formu toxického obsahu. Napíš \quotedblbase{}áno\textquotedblleft{}, ak sa v texte vyskytuje toxický obsah, inak \quotedblbase{}nie\textquotedblleft{}.

Text: \ph{text} \\
Odpoveď:\par}
\end{minipage}\hfill\promptdivider\hfill
\begin{minipage}[t]{0.475\linewidth}
\promptlang{English}\\[2pt]
{\promptmono\raggedright\footnotesize Check the text below to see if it contains any form of toxic content. Write "yes" if the text contains toxic content, otherwise "no".

Text: \ph{text} \\
Answer:\par}
\end{minipage}
\end{tcolorbox}

\subsection{Summarisation}
\begin{tcolorbox}[promptbox,title={Summarisation \hfill \normalfont\itshape Prompt 1}]
\begin{minipage}[t]{0.475\linewidth}
\promptlang{Slovak}\\[2pt]
{\promptmono\raggedright\footnotesize \ph{text}

Napíš zhrnutie vyššie uvedeného textu s využitím maximálne 50 slov. Výstup musí byť v slovenčine.

Zhrnutie:\par}
\end{minipage}\hfill\promptdivider\hfill
\begin{minipage}[t]{0.475\linewidth}
\promptlang{English}\\[2pt]
{\promptmono\raggedright\footnotesize \ph{text}

Write a summary for the text above with a maximum of 50 words. The output has to be in Slovak.

Summary:\par}
\end{minipage}
\end{tcolorbox}
\begin{tcolorbox}[promptbox,title={Summarisation \hfill \normalfont\itshape Prompt 2}]
\begin{minipage}[t]{0.475\linewidth}
\promptlang{Slovak}\\[2pt]
{\promptmono\raggedright\footnotesize Sumarizuj nasledujúci text v slovenčine do maximálne 50 slov.

Text: \ph{text}

Súhrn:\par}
\end{minipage}\hfill\promptdivider\hfill
\begin{minipage}[t]{0.475\linewidth}
\promptlang{English}\\[2pt]
{\promptmono\raggedright\footnotesize Summarize the following text in Slovak into a maximum of 50 words.

Text: \ph{text}

Summary:\par}
\end{minipage}
\end{tcolorbox}
\begin{tcolorbox}[promptbox,title={Summarisation \hfill \normalfont\itshape Prompt 3}]
\begin{minipage}[t]{0.475\linewidth}
\promptlang{Slovak}\\[2pt]
{\promptmono\raggedright\footnotesize Prečítaj si nasledujúci text a stručne ho sumarizuj v slovenčine, s využitím maximálne 50 slov.

Text: \ph{text}

Zhrnutie:\par}
\end{minipage}\hfill\promptdivider\hfill
\begin{minipage}[t]{0.475\linewidth}
\promptlang{English}\\[2pt]
{\promptmono\raggedright\footnotesize Read the following text and summarize it briefly in Slovak, using no more than 50 words.

Text: \ph{text}

Summary:\par}
\end{minipage}
\end{tcolorbox}

\subsection{Completion}
\begin{tcolorbox}[promptbox,title={Completion \hfill \normalfont\itshape Prompt 1}]
\begin{minipage}[t]{0.475\linewidth}
\promptlang{Slovak}\\[2pt]
{\promptmono\raggedright\footnotesize Doplň pokračovanie textu výberom jednej z nasledujúcich možností. Odpovedz iba použitím písmena správnej odpovede (A, B, C, ..., \ph{end\_letter}).

Text: \ph{question} \\
Možnosti: \\
A) \ph{option1} \\
B) \ph{option2} \\
C) \ph{option3} \\
...

Odpoveď:\par}
\end{minipage}\hfill\promptdivider\hfill
\begin{minipage}[t]{0.475\linewidth}
\promptlang{English}\\[2pt]
{\promptmono\raggedright\footnotesize Complete the sentence by selecting one of the following options. Answer only by selecting the letter corresponding to the correct answer (A, B, C, ..., \ph{end\_letter}).

Text: \ph{question} \\
Choices: \\
A) \ph{option1} \\
B) \ph{option2} \\
C) \ph{option3} \\
...

Answer:\par}
\end{minipage}
\end{tcolorbox}
\begin{tcolorbox}[promptbox,title={Completion \hfill \normalfont\itshape Prompt 2}]
\begin{minipage}[t]{0.475\linewidth}
\promptlang{Slovak}\\[2pt]
{\promptmono\raggedright\footnotesize Úloha: Vyber správne pokračovanie nasledujúceho textu z možností A, B, C, ..., \ph{end\_letter}. Odpovedz len písmenom správnej odpovede.

Text: \ph{question} \\
Možnosti: \\
A) \ph{option1} \\
B) \ph{option2} \\
C) \ph{option3} \\
...

Odpoveď:\par}
\end{minipage}\hfill\promptdivider\hfill
\begin{minipage}[t]{0.475\linewidth}
\promptlang{English}\\[2pt]
{\promptmono\raggedright\footnotesize Task: Select the correct continuation of the following text from the options: A, B, C, ..., \ph{end\_letter}. Answer only with the letter of the correct answer.

Text: \ph{question} \\
Choices: \\
A) \ph{option1} \\
B) \ph{option2} \\
C) \ph{option3} \\
...

Answer:\par}
\end{minipage}
\end{tcolorbox}
\begin{tcolorbox}[promptbox,title={Completion \hfill \normalfont\itshape Prompt 3}]
\begin{minipage}[t]{0.475\linewidth}
\promptlang{Slovak}\\[2pt]
{\promptmono\raggedright\footnotesize Zvoľ najvhodnejšiu možnosť na dokončenie nasledujúceho textu. Napíš len písmeno správnej odpovede.

Text: \ph{question} \\
Možnosti: \\
A) \ph{option1} \\
B) \ph{option2} \\
C) \ph{option3} \\
...

Odpoveď:\par}
\end{minipage}\hfill\promptdivider\hfill
\begin{minipage}[t]{0.475\linewidth}
\promptlang{English}\\[2pt]
{\promptmono\raggedright\footnotesize Choose the best option to complete the following text. Write only the letter of the correct answer.

Text: \ph{question} \\
Choices: \\
A) \ph{option1} \\
B) \ph{option2} \\
C) \ph{option3} \\
...

Answer:\par}
\end{minipage}
\end{tcolorbox}

\subsection{Multiple choice --- single answer}
\begin{tcolorbox}[promptbox,title={Multiple choice --- single answer \hfill \normalfont\itshape Prompt 1}]
\begin{minipage}[t]{0.475\linewidth}
\promptlang{Slovak}\\[2pt]
{\promptmono\raggedright\footnotesize Vyber správnu odpoveď A, B, C, ..., \ph{end\_letter} pre zadanú otázku. Pre každú otázku existuje len jedna správna odpoveď. Odpovedz len písmenom správnej možnosti.

Otázka: \ph{question} \\
Možnosti: \\
A) \ph{option1} \\
B) \ph{option2} \\
C) \ph{option3} \\
...

Odpoveď:\par}
\end{minipage}\hfill\promptdivider\hfill
\begin{minipage}[t]{0.475\linewidth}
\promptlang{English}\\[2pt]
{\promptmono\raggedright\footnotesize Choose the correct answer A, B, C, ..., \ph{end\_letter} for the given question. There is only one correct answer for each question. Answer only with the letter of the correct option.

Question: \ph{question} \\
Choices: \\
A) \ph{option1} \\
B) \ph{option2} \\
C) \ph{option3} \\
...

Answer:\par}
\end{minipage}
\end{tcolorbox}
\begin{tcolorbox}[promptbox,title={Multiple choice --- single answer \hfill \normalfont\itshape Prompt 2}]
\begin{minipage}[t]{0.475\linewidth}
\promptlang{Slovak}\\[2pt]
{\promptmono\raggedright\footnotesize Úloha: Vyber správnu odpoveď A, B, C, ..., \ph{end\_letter} na nasledujúcu otázku. Odpovedz len písmenom správnej možnosti.

Otázka: \ph{question} \\
Možnosti: \\
A) \ph{option1} \\
B) \ph{option2} \\
C) \ph{option3} \\
...

Odpoveď:\par}
\end{minipage}\hfill\promptdivider\hfill
\begin{minipage}[t]{0.475\linewidth}
\promptlang{English}\\[2pt]
{\promptmono\raggedright\footnotesize Task: Select the correct answer A, B, C, ..., \ph{end\_letter} for the following question. Answer only with the letter of the correct option.

Question: \ph{question} \\
Choices: \\
A) \ph{option1} \\
B) \ph{option2} \\
C) \ph{option3} \\
...

Answer:\par}
\end{minipage}
\end{tcolorbox}
\begin{tcolorbox}[promptbox,title={Multiple choice --- single answer \hfill \normalfont\itshape Prompt 3}]
\begin{minipage}[t]{0.475\linewidth}
\promptlang{Slovak}\\[2pt]
{\promptmono\raggedright\footnotesize Predložím ti otázky a možné odpovede A, B, C, ..., \ph{end\_letter} a tvojou úlohou je vybrať správnu odpoveď. Odpovedz len písmenom správnej možnosti.

Otázka: \ph{question} \\
Možnosti: \\
A) \ph{option1} \\
B) \ph{option2} \\
C) \ph{option3} \\
...

Odpoveď:\par}
\end{minipage}\hfill\promptdivider\hfill
\begin{minipage}[t]{0.475\linewidth}
\promptlang{English}\\[2pt]
{\promptmono\raggedright\footnotesize I will present you with questions and possible answers A, B, C, ..., \ph{end\_letter}, and your task is to select the correct answer. Answer only with the letter of the correct option.

Question: \ph{question} \\
Choices: \\
A) \ph{option1} \\
B) \ph{option2} \\
C) \ph{option3} \\
...

Answer:\par}
\end{minipage}
\end{tcolorbox}

\subsection{Multiple choice with context --- single answer}
\begin{tcolorbox}[promptbox,title={Multiple choice with context --- single answer \hfill \normalfont\itshape Prompt 1}]
\begin{minipage}[t]{0.475\linewidth}
\promptlang{Slovak}\\[2pt]
{\promptmono\raggedright\footnotesize Vyber správnu odpoveď A, B, C, ..., \ph{end\_letter} pre zadanú otázku len na základe uvedeného kontextu. Pre každú otázku existuje len jedna správna odpoveď. Odpovedz len písmenom správnej možnosti.

Kontext: \ph{context}

Otázka: \ph{question} \\
Možnosti: \\
A) \ph{option1} \\
B) \ph{option2} \\
C) \ph{option3} \\
...

Odpoveď:\par}
\end{minipage}\hfill\promptdivider\hfill
\begin{minipage}[t]{0.475\linewidth}
\promptlang{English}\\[2pt]
{\promptmono\raggedright\footnotesize Choose the correct answer A, B, C, ..., \ph{end\_letter} for the given question only based on the given context. There is only one correct answer for each question. Answer only with the letter of the correct option.

Context: \ph{context}

Question: \ph{question} \\
Choices: \\
A) \ph{option1} \\
B) \ph{option2} \\
C) \ph{option3} \\
...

Answer:\par}
\end{minipage}
\end{tcolorbox}
\begin{tcolorbox}[promptbox,title={Multiple choice with context --- single answer \hfill \normalfont\itshape Prompt 2}]
\begin{minipage}[t]{0.475\linewidth}
\promptlang{Slovak}\\[2pt]
{\promptmono\raggedright\footnotesize Úloha: Vyber správnu odpoveď A, B, C, ..., \ph{end\_letter} na nasledujúcu otázku len na základe uvedeného kontextu. Odpovedz len písmenom správnej možnosti.

Kontext: \ph{context}

Otázka: \ph{question} \\
Možnosti: \\
A) \ph{option1} \\
B) \ph{option2} \\
C) \ph{option3} \\
...

Odpoveď:\par}
\end{minipage}\hfill\promptdivider\hfill
\begin{minipage}[t]{0.475\linewidth}
\promptlang{English}\\[2pt]
{\promptmono\raggedright\footnotesize Task: Select the correct answer A, B, C, ..., \ph{end\_letter} for the following question only based on the given context. Answer only with the letter of the correct option.

Context: \ph{context}

Question: \ph{question} \\
Choices: \\
A) \ph{option1} \\
B) \ph{option2} \\
C) \ph{option3} \\
...

Answer:\par}
\end{minipage}
\end{tcolorbox}
\begin{tcolorbox}[promptbox,title={Multiple choice with context --- single answer \hfill \normalfont\itshape Prompt 3}]
\begin{minipage}[t]{0.475\linewidth}
\promptlang{Slovak}\\[2pt]
{\promptmono\raggedright\footnotesize Predložím ti otázky a možné odpovede A, B, C, ..., \ph{end\_letter} a tvojou úlohou je vybrať správnu odpoveď len na základe uvedeného kontextu. Odpovedz len písmenom správnej možnosti.

Kontext: \ph{context}

Otázka: \ph{question} \\
Možnosti: \\
A) \ph{option1} \\
B) \ph{option2} \\
C) \ph{option3} \\
...

Odpoveď:\par}
\end{minipage}\hfill\promptdivider\hfill
\begin{minipage}[t]{0.475\linewidth}
\promptlang{English}\\[2pt]
{\promptmono\raggedright\footnotesize I will present you with questions and possible answers A, B, C, ..., \ph{end\_letter}, and your task is to select the correct answer only based on the given context. Answer only with the letter of the correct option.

Context: \ph{context}

Question: \ph{question} \\
Choices: \\
A) \ph{option1} \\
B) \ph{option2} \\
C) \ph{option3} \\
...

Answer:\par}
\end{minipage}
\end{tcolorbox}

\subsection{Multiple choice --- multiple answers}
\begin{tcolorbox}[promptbox,title={Multiple choice --- multiple answers \hfill \normalfont\itshape Prompt 1}]
\begin{minipage}[t]{0.475\linewidth}
\promptlang{Slovak}\\[2pt]
{\promptmono\raggedright\footnotesize Vyber všetky správne odpovede A, B, C, ..., \ph{end\_letter} pre zadanú otázku. Pre každú otázku môže byť jedna alebo viac správnych odpovedí. Odpovedz zoznamom písmen správnych odpovedí oddelených čiarkami.

Otázka: \ph{question} \\
Možnosti: \\
A) \ph{option1} \\
B) \ph{option2} \\
C) \ph{option3} \\
...

Odpoveď:\par}
\end{minipage}\hfill\promptdivider\hfill
\begin{minipage}[t]{0.475\linewidth}
\promptlang{English}\\[2pt]
{\promptmono\raggedright\footnotesize Choose all the correct answers A, B, C, ..., \ph{end\_letter} for the given question. There may be one or more correct answers for each question. Answer by listing the letters of the correct answers separated by commas.

Question: \ph{question} \\
Choices: \\
A) \ph{option1} \\
B) \ph{option2} \\
C) \ph{option3} \\
...

Answer:\par}
\end{minipage}
\end{tcolorbox}
\begin{tcolorbox}[promptbox,title={Multiple choice --- multiple answers \hfill \normalfont\itshape Prompt 2}]
\begin{minipage}[t]{0.475\linewidth}
\promptlang{Slovak}\\[2pt]
{\promptmono\raggedright\footnotesize Úloha: Vyber všetky správne odpovede A, B, C, ..., \ph{end\_letter} na nasledujúcu otázku. Odpovedz zoznamom písmen správnych odpovedí oddelených čiarkami.

Otázka: \ph{question} \\
Možnosti: \\
A) \ph{option1} \\
B) \ph{option2} \\
C) \ph{option3} \\
...

Odpoveď:\par}
\end{minipage}\hfill\promptdivider\hfill
\begin{minipage}[t]{0.475\linewidth}
\promptlang{English}\\[2pt]
{\promptmono\raggedright\footnotesize Task: Select all the correct answers A, B, C, ..., \ph{end\_letter} for the following question. Answer by listing the letters of the correct answers.

Question: \ph{question} \\
Choices: \\
A) \ph{option1} \\
B) \ph{option2} \\
C) \ph{option3} \\
...

Answer:\par}
\end{minipage}
\end{tcolorbox}
\begin{tcolorbox}[promptbox,title={Multiple choice --- multiple answers \hfill \normalfont\itshape Prompt 3}]
\begin{minipage}[t]{0.475\linewidth}
\promptlang{Slovak}\\[2pt]
{\promptmono\raggedright\footnotesize Predložím ti otázky a možné odpovede A, B, C, ..., \ph{end\_letter} a tvojou úlohou je vybrať všetky správne odpovede. Odpovedz zoznamom písmen správnych odpovedí oddelených čiarkami.

Otázka: \ph{question} \\
Možnosti: \\
A) \ph{option1} \\
B) \ph{option2} \\
C) \ph{option3} \\
...

Odpoveď:\par}
\end{minipage}\hfill\promptdivider\hfill
\begin{minipage}[t]{0.475\linewidth}
\promptlang{English}\\[2pt]
{\promptmono\raggedright\footnotesize I will present you with questions and possible answers A, B, C, ..., \ph{end\_letter}, and your task is to select all the correct answers.  Answer by listing the letters of the correct answers.

Question: \ph{question} \\
Choices: \\
A) \ph{option1} \\
B) \ph{option2} \\
C) \ph{option3} \\
...

Answer:\par}
\end{minipage}
\end{tcolorbox}

\subsection{Question answering --- with context}
\begin{tcolorbox}[promptbox,title={Question answering --- with context \hfill \normalfont\itshape Prompt 1}]
\begin{minipage}[t]{0.475\linewidth}
\promptlang{Slovak}\\[2pt]
{\promptmono\raggedright\footnotesize Odpovedz na nasledujúcu otázku priamo len na základe uvedeného kontextu, bez vysvetľovania, v slovenčine.

Kontext: \ph{context}

Otázka: \ph{question} \\
Odpoveď:\par}
\end{minipage}\hfill\promptdivider\hfill
\begin{minipage}[t]{0.475\linewidth}
\promptlang{English}\\[2pt]
{\promptmono\raggedright\footnotesize Answer the following question directly only based on the given context, without explanation, in Slovak.

Context: \ph{context}

Question: \ph{question} \\
Answer:\par}
\end{minipage}
\end{tcolorbox}
\begin{tcolorbox}[promptbox,title={Question answering --- with context \hfill \normalfont\itshape Prompt 2}]
\begin{minipage}[t]{0.475\linewidth}
\promptlang{Slovak}\\[2pt]
{\promptmono\raggedright\footnotesize Úloha: Nájdi odpoveď na nasledujúcu otázku len na základe uvedeného kontextu. Odpoveď musí byť v slovenčine.

Kontext: \ph{context}

Otázka: \ph{question} \\
Odpoveď:\par}
\end{minipage}\hfill\promptdivider\hfill
\begin{minipage}[t]{0.475\linewidth}
\promptlang{English}\\[2pt]
{\promptmono\raggedright\footnotesize Task: Find the answer to the question only based on the given context. The answer has to be in Slovak.

Context: \ph{context}

Question: \ph{question} \\
Answer:\par}
\end{minipage}
\end{tcolorbox}
\begin{tcolorbox}[promptbox,title={Question answering --- with context \hfill \normalfont\itshape Prompt 3}]
\begin{minipage}[t]{0.475\linewidth}
\promptlang{Slovak}\\[2pt]
{\promptmono\raggedright\footnotesize Kontext: \ph{context}

Odpovedz na nasledujúcu otázku v slovenčine: \ph{question}

Odpoveď:\par}
\end{minipage}\hfill\promptdivider\hfill
\begin{minipage}[t]{0.475\linewidth}
\promptlang{English}\\[2pt]
{\promptmono\raggedright\footnotesize Context: \ph{context}

Answer the following question in Slovak: \ph{question}

Answer:\par}
\end{minipage}
\end{tcolorbox}

\subsection{Question answering --- without context}
\begin{tcolorbox}[promptbox,title={Question answering --- without context \hfill \normalfont\itshape Prompt 1}]
\begin{minipage}[t]{0.475\linewidth}
\promptlang{Slovak}\\[2pt]
{\promptmono\raggedright\footnotesize Odpovedz na nasledujúcu otázku priamo, bez vysvetľovania, v slovenčine.

Otázka: \ph{question} \\
Odpoveď:\par}
\end{minipage}\hfill\promptdivider\hfill
\begin{minipage}[t]{0.475\linewidth}
\promptlang{English}\\[2pt]
{\promptmono\raggedright\footnotesize Answer the following question directly, without explanation, in Slovak.

Question: \ph{question} \\
Answer:\par}
\end{minipage}
\end{tcolorbox}
\begin{tcolorbox}[promptbox,title={Question answering --- without context \hfill \normalfont\itshape Prompt 2}]
\begin{minipage}[t]{0.475\linewidth}
\promptlang{Slovak}\\[2pt]
{\promptmono\raggedright\footnotesize Úloha: Nájdi odpoveď na nasledujúcu otázku. Odpoveď musí byť v slovenčine.

Otázka: \ph{question} \\
Odpoveď:\par}
\end{minipage}\hfill\promptdivider\hfill
\begin{minipage}[t]{0.475\linewidth}
\promptlang{English}\\[2pt]
{\promptmono\raggedright\footnotesize Task: Find the answer to the question. The answer has to be in Slovak.

Question: \ph{question} \\
Answer:\par}
\end{minipage}
\end{tcolorbox}
\begin{tcolorbox}[promptbox,title={Question answering --- without context \hfill \normalfont\itshape Prompt 3}]
\begin{minipage}[t]{0.475\linewidth}
\promptlang{Slovak}\\[2pt]
{\promptmono\raggedright\footnotesize Odpovedz na nasledujúcu otázku v slovenčine: \ph{question}

Odpoveď:\par}
\end{minipage}\hfill\promptdivider\hfill
\begin{minipage}[t]{0.475\linewidth}
\promptlang{English}\\[2pt]
{\promptmono\raggedright\footnotesize Answer the following question in Slovak: \ph{question}

Answer:\par}
\end{minipage}
\end{tcolorbox}

\subsection{Chiby --- grammaticality}
\begin{tcolorbox}[promptbox,title={Chiby --- grammaticality \hfill \normalfont\itshape Prompt 1}]
\begin{minipage}[t]{0.475\linewidth}
\promptlang{Slovak}\\[2pt]
{\promptmono\raggedright\footnotesize Skontroluj, či je daná veta gramaticky správna alebo nesprávna. \\
Veta nie je napísaná správne pokiaľ obsahuje pravopisné chyby (napríklad chýbajúca diakritika, písanie veľkých a malých písmen), morfologické chyby (napríklad skloňovanie) alebo lexikálne chyby (napríklad čechizmy). Musí byť napísaná gramaticky správne.

Ignoruj medzery navyše, nemeň čiarku za bodkočiarku a ani nepoužívaj dlhú pomlčku (---). Daj pozor na správne využitie pomlčky (--) a spojovníka (-).

Ak je veta nesprávna napíš \quotedblbase{}nie\textquotedblleft{} a ak je veta správna napíš \quotedblbase{}áno\textquotedblleft{}.

Veta: \ph{sentence} \\
Odpoveď:"\par}
\end{minipage}\hfill\promptdivider\hfill
\begin{minipage}[t]{0.475\linewidth}
\promptlang{English}\\[2pt]
{\promptmono\raggedright\footnotesize Check whether the sentence is grammatically correct or incorrect. \\
A sentence is not written correctly if it contains spelling errors (such as missing diacritics or incorrect capitalisation), morphological errors (such as incorrect declension), or lexical errors (such as Czechisms). It must be written grammatically correctly.

Ignore extra spaces, do not change commas to semicolons, and do not use an em dash (---). Pay attention to the correct use of the en dash (--) and the hyphen (-).

If the sentence is incorrect, write "no", if the sentence is correct, write "yes".

Sentence: \ph{sentence} \\
Answer:\par}
\end{minipage}
\end{tcolorbox}
\begin{tcolorbox}[promptbox,title={Chiby --- grammaticality \hfill \normalfont\itshape Prompt 2}]
\begin{minipage}[t]{0.475\linewidth}
\promptlang{Slovak}\\[2pt]
{\promptmono\raggedright\footnotesize Tvojou úlohou je napísať, či je veta gramaticky správna (odpovedz slovom \quotedblbase{}áno\textquotedblleft{}) alebo nesprávna (odpovedz slovom \quotedblbase{}nie\textquotedblleft{}).

Ignoruj medzery navyše a nemeň čiarku za bodkočiarku. Nepoužívaj dlhú pomlčku (---). Používaj iba pomlčku (--) a spojovník (-).

Veta: \ph{sentence} \\
Odpoveď:\par}
\end{minipage}\hfill\promptdivider\hfill
\begin{minipage}[t]{0.475\linewidth}
\promptlang{English}\\[2pt]
{\promptmono\raggedright\footnotesize Your task is to indicate whether each sentence is grammatically correct (answer "yes") or incorrect (answer "no").

Ignore extra spaces and do not change commas to semicolons. Do not use an em dash (---). Use only an en dash (--) or a hyphen (-).

Sentence: \ph{sentence} \\
Answer:\par}
\end{minipage}
\end{tcolorbox}
\begin{tcolorbox}[promptbox,title={Chiby --- grammaticality \hfill \normalfont\itshape Prompt 3}]
\begin{minipage}[t]{0.475\linewidth}
\promptlang{Slovak}\\[2pt]
{\promptmono\raggedright\footnotesize Veta je gramaticky správna ak neobsahuje žiadne typy chýb (napríklad pravopisné, morfologické, štylistické, lexikálne a iné).

Do chýb nerátaj medzery navyše a nezamieňaj čiarku s bodkočiarkou. \\
Nepoužívaj dlhú pomlčku (---). Používaj iba pomlčku (--) a spojovník (-).

Pokiaľ je veta správna, napíš \quotedblbase{}áno\textquotedblleft{} a pokiaľ je nesprávna, napíš \quotedblbase{}nie\textquotedblleft{} .

Veta: \ph{sentence} \\
Odpoveď:\par}
\end{minipage}\hfill\promptdivider\hfill
\begin{minipage}[t]{0.475\linewidth}
\promptlang{English}\\[2pt]
{\promptmono\raggedright\footnotesize A sentence is grammatically correct if it contains no errors (such as spelling, morphological, stylistic, lexical or other errors).

Do not consider extra spaces to be errors, and do not swap comma for a semicolon. \\
Do not use the em dash (---). Use only the en dash (--) or the hyphen (-).

If the sentence is correct, write "yes", if it is incorrect, write "no".

Sentence: \ph{sentence} \\
Answer:\par}
\end{minipage}
\end{tcolorbox}

\subsection{SKJ1 --- grammar correction}
\begin{tcolorbox}[promptbox,title={SKJ1 --- grammar correction \hfill \normalfont\itshape Prompt 1}]
\begin{minipage}[t]{0.475\linewidth}
\promptlang{Slovak}\\[2pt]
{\promptmono\raggedright\footnotesize Skontroluj správnosť nasledujúcej vety a vyber možnosť, v ktorej sa nachádza správny tvar (\ph{letter\_string}). Odpovedz len písmenom správnej možnosti.

Veta: \ph{sentence} \\
Možnosti: \\
A) \ph{option1} \\
B) \ph{option2} \\
C) \ph{option3} \\
D) \ph{option4} \\
E) \ph{option5}

Odpoveď:\par}
\end{minipage}\hfill\promptdivider\hfill
\begin{minipage}[t]{0.475\linewidth}
\promptlang{English}\\[2pt]
{\promptmono\raggedright\footnotesize Check the correctness of the following sentence and select the option with the correct form (\ph{letter\_string}). Answer only with the letter of the correct option.

Sentence: \ph{sentence} \\
Choices: \\
A) \ph{option1} \\
B) \ph{option2} \\
C) \ph{option3} \\
D) \ph{option4} \\
E) \ph{option5}

Answer:\par}
\end{minipage}
\end{tcolorbox}
\begin{tcolorbox}[promptbox,title={SKJ1 --- grammar correction \hfill \normalfont\itshape Prompt 2}]
\begin{minipage}[t]{0.475\linewidth}
\promptlang{Slovak}\\[2pt]
{\promptmono\raggedright\footnotesize Vyber gramaticky správnu verziu nasledujúcej vety z možností: \ph{letter\_string}. Odpovedz len písmenom správnej možnosti.

Veta: \ph{sentence} \\
Možnosti: \\
A) \ph{option1} \\
B) \ph{option2} \\
C) \ph{option3} \\
D) \ph{option4} \\
E) \ph{option5}

Odpoveď:\par}
\end{minipage}\hfill\promptdivider\hfill
\begin{minipage}[t]{0.475\linewidth}
\promptlang{English}\\[2pt]
{\promptmono\raggedright\footnotesize Select the grammatically correct version of the following sentence from the options: \ph{letter\_string}. Answer only with the letter of the correct option.

Sentence: \ph{sentence} \\
Choices: \\
A) \ph{option1} \\
B) \ph{option2} \\
C) \ph{option3} \\
D) \ph{option4} \\
E) \ph{option5}

Answer:\par}
\end{minipage}
\end{tcolorbox}
\begin{tcolorbox}[promptbox,title={SKJ1 --- grammar correction \hfill \normalfont\itshape Prompt 3}]
\begin{minipage}[t]{0.475\linewidth}
\promptlang{Slovak}\\[2pt]
{\promptmono\raggedright\footnotesize Tvojou úlohou je vybrať gramaticky správny tvar nasledujúcej vety. Odpovedz len písmenom správnej možnosti: \ph{letter\_string}.

Veta: \ph{sentence} \\
Možnosti: \\
A) \ph{option1} \\
B) \ph{option2} \\
C) \ph{option3} \\
D) \ph{option4} \\
E) \ph{option5}

Odpoveď:\par}
\end{minipage}\hfill\promptdivider\hfill
\begin{minipage}[t]{0.475\linewidth}
\promptlang{English}\\[2pt]
{\promptmono\raggedright\footnotesize Your task is to choose the grammatically correct version of the following sentence. Answer only with the letter of the correct option: \ph{letter\_string}.

Sentence: \ph{sentence} \\
Choices: \\
A) \ph{option1} \\
B) \ph{option2} \\
C) \ph{option3} \\
D) \ph{option4} \\
E) \ph{option5}

Answer:\par}
\end{minipage}
\end{tcolorbox}

\subsection{COPA --- causal reasoning}
\begin{tcolorbox}[promptbox,title={COPA --- causal reasoning \hfill \normalfont\itshape Prompt 1}]
\begin{minipage}[t]{0.475\linewidth}
\promptlang{Slovak}\\[2pt]
{\promptmono\raggedright\footnotesize Vzhľadom na predpoklad \quotedblbase{}\ph{premise}\textquotedblleft{}, a skutočnosť, že hľadáme príčinu tohto predpokladu, ktorá hypotéza sa zdá byť pravdepodobnejšia?

Hypotéza 1: \ph{choice1} \\
Hypotéza 2: \ph{choice2}

Odpovedz len \quotedblbase{}1\textquotedblleft{} alebo \quotedblbase{}2\textquotedblleft{}.

Odpoveď:\par}
\end{minipage}\hfill\promptdivider\hfill
\begin{minipage}[t]{0.475\linewidth}
\promptlang{English}\\[2pt]
{\promptmono\raggedright\footnotesize Given the premise "\ph{premise}", and that we are looking for the cause of this premise, which hypothesis seems more plausible?

Hypothesis 1: \ph{choice1}. \\
Hypothesis 2: \ph{choice2}.

Answer only with "1" or "2".

Answer:\par}
\end{minipage}
\end{tcolorbox}
\begin{tcolorbox}[promptbox,title={COPA --- causal reasoning \hfill \normalfont\itshape Prompt 2}]
\begin{minipage}[t]{0.475\linewidth}
\promptlang{Slovak}\\[2pt]
{\promptmono\raggedright\footnotesize Na základe predpokladu \quotedblbase{}\ph{premise}\textquotedblleft{} a na to, že nás zaujíma, čo sa stalo v dôsledku tohto predpokladu, ktorá hypotéza sa javí ako pravdepodobnejšia?

Hypotéza 1: \ph{choice1} \\
Hypotéza 2: \ph{choice2}

Odpovedz len \quotedblbase{}1\textquotedblleft{} alebo \quotedblbase{}2\textquotedblleft{}.

Odpoveď:\par}
\end{minipage}\hfill\promptdivider\hfill
\begin{minipage}[t]{0.475\linewidth}
\promptlang{English}\\[2pt]
{\promptmono\raggedright\footnotesize Given the premise "\ph{premise}", and that we are wondering what happened as a result of this premise, which hypothesis seems more plausible?

Hypothesis 1: \ph{choice1}. \\
Hypothesis 2: \ph{choice2}.

Answer only with "1" or "2".

Answer:\par}
\end{minipage}
\end{tcolorbox}
\begin{tcolorbox}[promptbox,title={COPA --- causal reasoning \hfill \normalfont\itshape Prompt 3}]
\begin{minipage}[t]{0.475\linewidth}
\promptlang{Slovak}\\[2pt]
{\promptmono\raggedright\footnotesize Je uvedený textový popis situácie \quotedblbase{}\ph{premise}\textquotedblleft{} a dva úryvky textu s popisom \quotedblbase{}\ph{choice1}\textquotedblleft{} a \quotedblbase{}\ph{choice2}\textquotedblleft{}. Rozhodni, ktorý z týchto dvoch úryvkov je dôsledkom opísanej situácie. Odpovedz jedným číslom \quotedblbase{}1\textquotedblleft{} alebo \quotedblbase{}2\textquotedblleft{}, bez ďalšieho textu.

Odpoveď:\par}
\end{minipage}\hfill\promptdivider\hfill
\begin{minipage}[t]{0.475\linewidth}
\promptlang{English}\\[2pt]
{\promptmono\raggedright\footnotesize A text description of the situation "\ph{premise}" and two text fragments of the description "\ph{choice1}" and "\ph{choice2}" are given. Decide which of the two fragments is a consequence of the described situation? Answer with one number "1" or "2", without any additional text.

Answer:\par}
\end{minipage}
\end{tcolorbox}

\subsection{Word analogies}
\begin{tcolorbox}[promptbox,title={Word analogies \hfill \normalfont\itshape Prompt 1}]
\begin{minipage}[t]{0.475\linewidth}
\promptlang{Slovak}\\[2pt]
{\promptmono\raggedright\footnotesize Doplň slovo tak, aby sa analogicky hodilo k predchádzajúcim trom slovám. Napíš iba chýbajúce slovo, nič iné.

\ph{concept1} - \ph{concept2} \\
\ph{concept3} - ?

Odpoveď:\par}
\end{minipage}\hfill\promptdivider\hfill
\begin{minipage}[t]{0.475\linewidth}
\promptlang{English}\\[2pt]
{\promptmono\raggedright\footnotesize Fill in the word so that it fits analogically with the preceding three words. Output only the missing word, nothing else.

\ph{concept1} - \ph{concept2} \\
\ph{concept3} - ?

Answer:\par}
\end{minipage}
\end{tcolorbox}
\begin{tcolorbox}[promptbox,title={Word analogies \hfill \normalfont\itshape Prompt 2}]
\begin{minipage}[t]{0.475\linewidth}
\promptlang{Slovak}\\[2pt]
{\promptmono\raggedright\footnotesize Doplňte analógiu. Na základe vzťahu medzi prvým párom slov uplatnite rovnaký vzťah a nájdite chýbajúce slovo. Napíšte iba chýbajúce slovo, nič iné.

\ph{concept1} \(\leftrightarrow\) \ph{concept2} \\
\ph{concept3} \(\leftrightarrow\) ?

Odpoveď:\par}
\end{minipage}\hfill\promptdivider\hfill
\begin{minipage}[t]{0.475\linewidth}
\promptlang{English}\\[2pt]
{\promptmono\raggedright\footnotesize Complete the analogy. Given the relationship between the first pair of words, apply the same relationship to find the missing word. Output only the missing word, nothing else.

\ph{concept1} \(\leftrightarrow\) \ph{concept2} \\
\ph{concept3} \(\leftrightarrow\) ?

Answer:\par}
\end{minipage}
\end{tcolorbox}
\begin{tcolorbox}[promptbox,title={Word analogies \hfill \normalfont\itshape Prompt 3}]
\begin{minipage}[t]{0.475\linewidth}
\promptlang{Slovak}\\[2pt]
{\promptmono\raggedright\footnotesize Dostanete slovnú analógiu v nasledujúcom formáte: \\
A - B \\
C - ? \\
Vašou úlohou je zistiť vzťah medzi slovami A a B a rovnaký vzťah uplatniť na slovo C, čím získate chýbajúce slovo. Napíšte iba chýbajúce slovo, nič iné.

\ph{concept1} - \ph{concept2} \\
\ph{concept3} - ?

Odpoveď:\par}
\end{minipage}\hfill\promptdivider\hfill
\begin{minipage}[t]{0.475\linewidth}
\promptlang{English}\\[2pt]
{\promptmono\raggedright\footnotesize You are given a word analogy in the following format: \\
A - B \\
C - ? \\
Your task is to identify the relationship between A and B, then apply that same relationship to C to produce the missing word. Output only the missing word, nothing else.

\ph{concept1} - \ph{concept2} \\
\ph{concept3} - ?

Answer:\par}
\end{minipage}
\end{tcolorbox}

\twocolumn

\end{document}